\documentclass[]{fairmeta}

\usepackage{amssymb}

\usepackage{algorithm}
\usepackage{algpseudocode}

\usepackage{tablefootnote}

\usepackage{tikz}      %
\usetikzlibrary{arrows.meta} 

\usepackage[toc,page,header]{appendix}
\usepackage{titletoc}

\newcommand{\eg}{\textit{e.g.},~}
\newcommand{\wrt}{\textit{w.r.t.}~}
\newcommand{\ie}{\textit{i.e.},~}

\newcommand{\xf}[1]{\textcolor{black}{#1}}

\title{The Intricate Dance of Prompt Complexity, Quality, Diversity, and Consistency in T2I Models}

\author[1,2,3]{Zhang Xiaofeng}
\author[2,3,5]{Aaron Courville}
\author[1]{Michal Drozdzal}
\author[1,2,4,5]{Adriana Romero-Soriano}

\affiliation[1]{FAIR at Meta - Montréal}
\affiliation[2]{Mila - Québec AI Institute}
\affiliation[3]{Université de Montréal}
\affiliation[4]{McGill University}
\affiliation[5]{Canada CIFAR AI Chair}

\abstract{Text-to-image (T2I) models offer great potential for creating virtually limitless synthetic data, a valuable resource compared to fixed and finite real datasets. Previous works evaluate the utility of synthetic data from T2I models on three key desiderata: quality, diversity, and consistency. While prompt engineering is the primary means of interacting with T2I models, the systematic impact of prompt complexity on these critical utility axes remains underexplored. In this paper, we first conduct synthetic experiments to motivate the difficulty of generalization \wrt prompt complexity and explain the observed difficulty with theoretical derivations. Then, we introduce a new evaluation framework that can compare the utility of real data and synthetic data, and present a comprehensive analysis of how prompt complexity influences the utility of synthetic data generated by commonly used T2I models. We conduct our study across diverse datasets, including CC12M, ImageNet-1k, and DCI, and evaluate different inference-time intervention methods. Our synthetic experiments show that generalizing to more general conditions is harder than the other way round, since the former needs an estimated likelihood that is not learned by diffusion models. Our large-scale empirical experiments reveal that increasing prompt complexity results in lower conditional diversity and prompt consistency, while reducing the synthetic-to-real distribution shift, which aligns with the synthetic experiments. Moreover, current inference-time interventions can augment the diversity of the generations at the expense of moving outside the support of real data. Among those interventions, prompt expansion, by deliberately using a pre-trained language model as a likelihood estimator, consistently achieves the highest performance in both image diversity and aesthetics, even higher than that of real data. Combining advanced guidance interventions with prompt expansion results in the most appealing utility trade-offs of synthetic data.\looseness-1
}

\date{\today}
\correspondence{Zhang Xiaofeng at \email{xiaofeng.zhang@mila.quebec}}

\metadata[Code]{\url{https://github.com/facebookresearch/synthetic_data_utility_prompt_complexity}}

\begin{document}

\maketitle

\section{Introduction}

Text-to-image (T2I) models have made significant progress in recent years, enabling the generation of high-quality images from textual descriptions~\citep{sd35model,flux2024}. These advancements have opened up new opportunities in several application domains, making it possible for the public to create images easily with their own words. For the research community, the success of T2I models has also unlocked the exploration of synthetic data generated by these models for downstream model training~\citep{feedbackguidance, askari2025improving, tian2023stablerep, fan2024scaling, dallasen2025} and model self-improvement~\citep{yoon2024model}. \looseness-1

Recent studies have yielded in-depth analyses on the utility of synthetic data from off-the-shelf T2I models~\citep{astolfi2024consistency,digin,lee2023holistic}. The utility of synthetic data from conditional image generative models, such as T2I, has been defined as a set of desiderata~\citep{astolfi2024consistency, devries2019evaluation}, including \emph{synthetic image quality} --\ie aesthetics and realism--, \emph{diversity}, and \emph{conditional consistency} --\ie alignment with the conditioning prompt. Analyses of these desiderata have revealed that the impressive progress in image quality has come at the expense of generation diversity. 
As a result, the community has devoted significant work to devise inference-time interventions --\eg prompt rewriting~\citep{promptexpansion} and advanced guidance approaches modifying standard classifier-free guidance (CFG)~\citep{cfg} -- that improve the diversity of synthetic data. Although these interventions focus on either changing the prompt conditioning or the flow of signals from the conditioning, \emph{the effect of the prompt content on the utility of synthetic data remains an open question}. Prompts may be characterized by their complexity, defined as the amount of details contained in the prompt or the specificity of its concepts, and state-of-the-art T2I models have been shown to benefit from synthetic and descriptive captions, generated by vision language models (VLMs), during training to improve their generation performance~\citep{bettercaptiondalle3}. It is now common practice to train high performance T2I models on a combination of real and synthetic captions~\citep{sd35model,pixart,qwenimage}. Yet, these models are known to produce poor results when sampled out of their training distribution~\citep{bettercaptiondalle3}. Considering the very descriptive and synthetic captions used to train T2I models, evaluating these models from the perspective of prompt complexity is crucial to better understand the models' performance.\looseness-1

Therefore, in this paper, we study the effect of prompt complexity on the different utility axes of synthetic image data. We start by conducting synthetic experiments over mixtures of Gaussians, conditioning the generation process on prompts of varying complexities.%
We show that generalization across prompt complexities is already hard in this toy synthetic setting, especially when attempting to generalize to more general prompts --\ie prompts that are shorter or less specific than the ones used for training. \xf{The results from the synthetic setting motivate} an in-depth evaluation of how prompt complexity influences different utility axes of synthetic data from T2I models. To do so, \xf{we propose a novel evaluation framework that constructs prompts with different complexities, enabling the evaluation from a prompt complexity perspecitve over a wide range of commonly used vision and vision-language datasets}
--\ie CC12M~\citep{datasetcc12m}, ImageNet-1k~\citep{imagenet}, and Densely Captioned Images (DCI)~\citep{dcidataset}.  We use the resulting prompts to condition the generation process of state-of-the-art T2I models --\ie LDMv1.5~\citep{ldmrombach}, LDM-XL~\citep{sdxl}, LDMv3.5M~\citep{sd35model}, LDMv3.5L~\citep{sd35model}, Flux-schnell~\citep{flux2024}, and Infinity~\citep{infinityautoregressive}--, and collect synthetic images with several inference-time intervention methods including CFG~\citep{cfg}, condition-annealing diffusion sampling (CADS)~\citep{cads}, interval guidance~\citep{intervalguide}, adapted projected guidance (APG)~\citep{apg}, and prompt expansion~\citep{promptexpansion}. 
Our analysis unlocks comparisons between the utility of real and synthetic data, which are beneficial to identify potential gaps between real and synthetic image distributions. To the best of our knowledge, \emph{we are the first to systematically evaluate the effect of prompt complexity on T2I generations}.\looseness-1

Our synthetic experiments show that generalizing to more general conditions is harder than generalizing to more fine-grained conditionings, since the former needs an estimated likelihood that is not learned by diffusion models. Our large-scale empirical evaluation \xf{enabled by the contributed framework further reveals interesting findings as follows:} 
\xf{1) The trend of utilities are non-linear, and especially the aesthetic score exhibits a slope steeper towards shorter prompt lengths and more gradual for longer ones (Fig.~\ref{fig:quality_SDexpansion_dci}), showing an asymmetry of prompt length generalization. 2) Diversity does not collapse but plateaus as prompt length increases, as shown in Fig.~\ref{fig:diversity_SDexpansion_dci}. This suggests an inherent “lower bound of diversity” in T2I models. 3) Optimizing for reference-free metrics harms distributional fidelity. As shown in Figs.~\ref{fig:marginal_dinov2_SDexpansion_cc12m} and \ref{fig:marginal_dinov2_SDexpansion_in1k}, prompt expansion degrades precision and density while newer models (e.g., LDMv3.5L) degrade in frechet distance with real data.
This suggests a cautious usage of synthetic data generated from T2I models for downstream applications.}
4) By combining advanced guidance methods (in particular APG) with prompt expansion, we can benefit from the advantages of both approaches and achieve the most interesting trade-offs. 
Overall, our study suggests that prompt complexity is a crucial axis to consider when prompting T2I models, and requires more investigation especially when generating from very general prompts. In addition, diversity is a key feature of real-world image distributions, which is still not properly captured by synthetic images from state-of-the-art T2I models when no explicit prompt expansion is conducted during inference.\looseness-1

\section{Generalizing to general conditions is hard}
\label{sec:synthetic}
In this section, we use a synthetic setting to build intuition of how T2I models may perform when evaluated on prompts of different complexities, and notably when evaluated on prompts that are outside of their training distribution. \xf{motivating the necessity of conducting systematic evaluations of T2I models from a prompt complexity perspective in the following sections.}\looseness-1

\begin{figure}
    \centering
    \begin{subfigure}[ht]{0.2\textwidth}
        \includegraphics[width=0.9\linewidth]{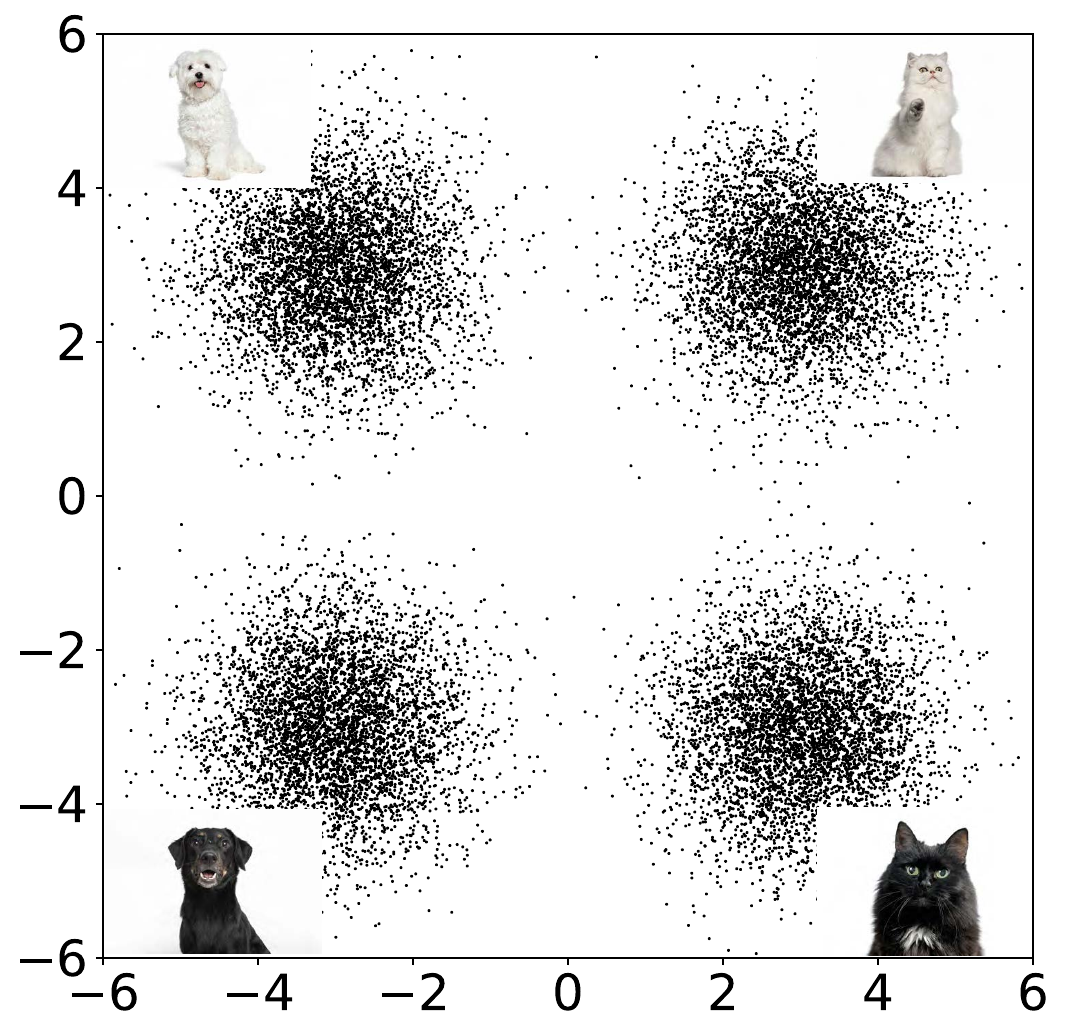}
        \caption{Training data \\ \textcolor{white}{spacing} \\ \textcolor{white}{spacing} }
        \label{fig:synthetic_training}
    \end{subfigure}
    \hspace{1em}
    \begin{subfigure}[ht]{0.2\textwidth}
        \includegraphics[width=0.9\linewidth]{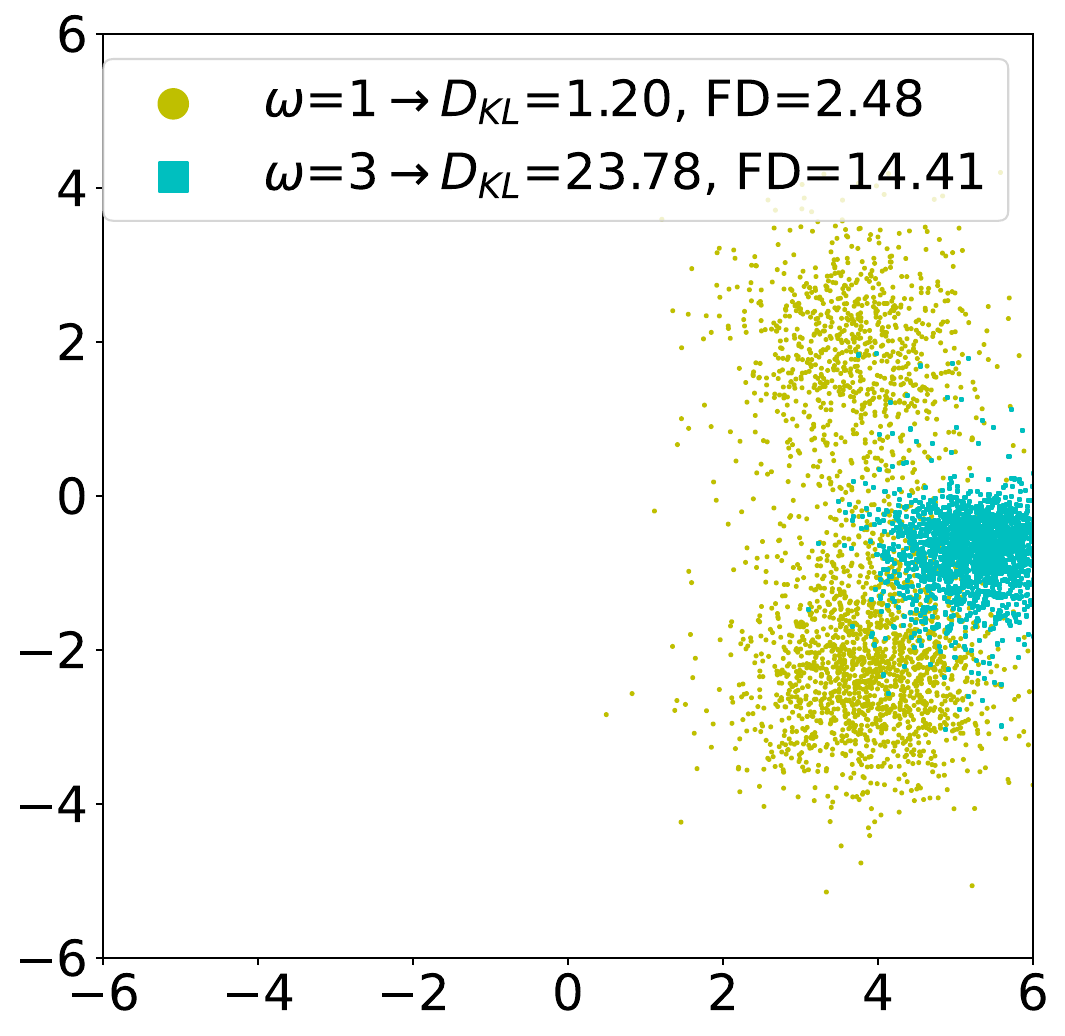}
        \caption{\texttt{cat} using model trained with fine-grained prompts}
        \label{fig:synthetic_general_cat}
    \end{subfigure}
    \hspace{1em}
    \begin{subfigure}[ht]{0.2\textwidth}
        \includegraphics[width=0.9\linewidth]{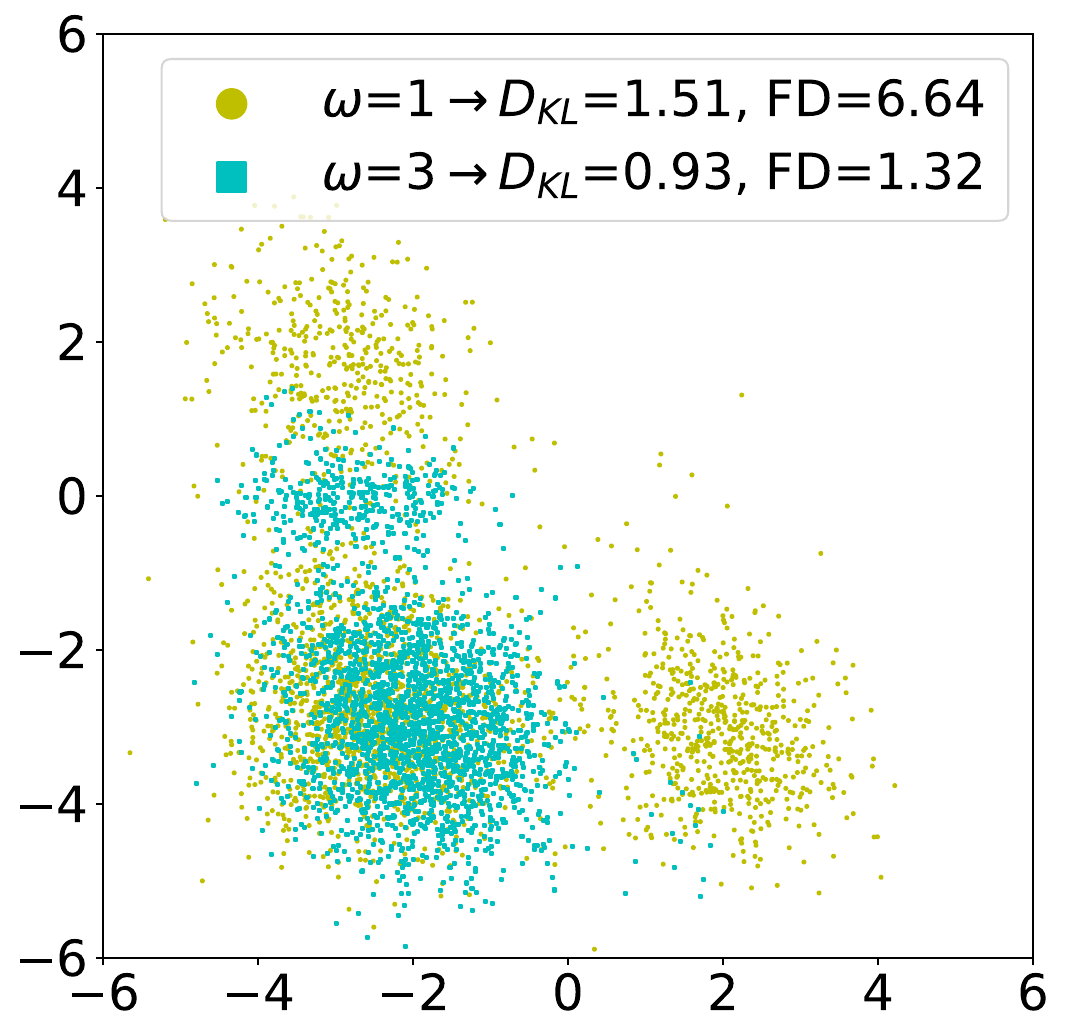}
        \caption{\texttt{black dog} using model trained with general prompts}
        \label{fig:synthetic_black_dog}
    \end{subfigure}
    \caption{\textbf{Generalization to prompts of different complexities during inference.} \ref{fig:synthetic_training} shows the training data distribution. \ref{fig:synthetic_general_cat} presents the generated samples using the general prompt \texttt{cat} with the model trained with fine-grained prompts. \ref{fig:synthetic_black_dog} shows the generated samples using the fine-grained prompt \texttt{black dog} with the model trained with general prompts. $\omega$ is the classifier-free guidance scale. With $\omega>1$, generalization towards more general prompts is harder in this synthetic setting.\looseness-1}
    \label{fig:synthetic_main_paper}
\end{figure}

\textbf{Dataset and model. \ \ } Let our data be a mixture of four Gaussians (Figure~\ref{fig:synthetic_training}), where each Gaussian represents a different category --\eg white dog, white cat, black dog, and black cat. We train two conditional U-Net models~\citep{unet,ldmrombach} using two sets of conditional prompts: one model is trained with fine-grained prompts (\texttt{white dog}, \texttt{white cat}, \texttt{black dog}, and \texttt{black cat}); the other model is trained with general prompts (\texttt{dog}, \texttt{cat}, \texttt{white} and \texttt{black}). We train both models, including their vocabulary dictionary, from scratch using DDPM schedule~\citep{ddpm}. We perform inference in a generalization setting, --\ie if the model is trained with general prompts, inference is done with fine-grained prompts, and vice versa. We also employ classifier-free guidance (CFG)~\citep{cfg} with a guidance scale $\omega$; $\omega=1$ refers to the conditional model, --\ie not using classifier-free guidance--, and $\omega>1$ refers to higher CFG guidance strengths. More details can be found in the Appendix~\ref{app:synthetic_exp_setting}.\looseness-1

\textbf{Mathematical derivations. \ \ } Conditioning on general prompts is analogous to applying an \texttt{OR} operator on fine-grained prompts --\eg \texttt{dog} would be \texttt{white dog OR black dog}--, while conditioning on fine-grained prompts may be seen as performing an \texttt{AND} operator among general prompts. Following \citet{compositional_diffusion,composition_reducereuserecycle}, we have the following derivations. Given a general prompt $c_{\text{g}}$ that can be decomposed into various independent fine-grained prompts ${c_{\text{f}}^i, i \in \{1,2,...,K\}}$, computing the score function $s_{\theta}(x_t|c_{\text{g}})$ at timestep $t$
requires both the score functions and the conditional likelihood $p_{\theta}(x_t|c_{\text{f}}^i)$ of fine-grained ones. The latter is not available from a diffusion model, as shown in Equation~\ref{eq:OR}. However, given a fine-grained prompt $c_{\text{f}}$ that is composed of several general concepts ${c_{\text{g}}^i, i\in \{1,2,...,M\}}$, the score function $s_{\theta}(x_t|c_{\text{f}})$ at timestep $t$ can be approximately estimated by the score functions of general prompts solely (\eg \texttt{white} and \texttt{dog}), as shown in Equation~\ref{eq:AND}. Derivations are in Appendix~\ref{app:synthetic_proof}.\looseness-1

\begin{equation}
\label{eq:OR}
    \text{\texttt{OR} operator: \ \ } s_{\theta}(x_t|c_{\text{g}}) = \sum_{i\in\{1,2,...,K\}} \left(\underbrace{\frac{p_{\theta}(x_t|c_{\text{f}}^i)}{{\sum_{j \in \{1,2,...,K\}}p(x_t|c_{\text{f}}^j)}}}_{\text{not learned by the diffusion model}} s_{\theta}(x_t|c_{\text{f}}^i)\right)
\end{equation}

\begin{equation}
\label{eq:AND}
    \text{\texttt{AND} operator: \ \ } s_{\theta}(x_t|c_{\text{f}}) = s_{\theta} (x_t) + \sum_{i \in \{1,2,...,M\}} (s_{\theta}(x_t|c_{\text{g}}^i) - s_{\theta}(x_t))
\end{equation}

\textbf{Generalizing to general prompts is hard. \ \ } 
We use forward KL-divergence ($D_{\text{KL}}$) and Fr\'{e}chet Distance (FD) to evaluate the generated samples. We also measure reference-free sample diversity using Vendi score~\citep{vendiscore} ($\text{VS}$)\footnote{For the synthetic setting, there is no pre-trained models to evaluate the reference-free quality and consistency. KL-divergence and FD cover both the quality and consistency.}, following the empirical evaluation in section~\ref{sec:exp}.
On the one hand, Figure~\ref{fig:synthetic_general_cat} shows the generated samples using the general prompt \texttt{cat} at inference time with the model trained with fine-grained prompts. When generalizing to general prompts (\texttt{OR} operator), the model tends to generate samples towards the mean of the referenced distributions. In particular, when $\omega>1$ (a common practice in T2I models), the generated samples cover a region where the training data density is very small, resulting in $D_{\text{KL}}=23.78$, $\text{FD}=14.41$, and $\text{VS}_{\text{gen}}=1.03$ compared to $\text{VS}_{\text{ref}}=1.82$. Yet, when $\omega=1$, this phenomenon is reduced, achieving $D_{\text{KL}}=1.20$, $\text{FD}=2.48$, and $\text{VS}_{\text{gen}}=1.43$ compared to $\text{VS}_{\text{ref}}=1.82$.
Since diffusion models only learn the score function (and not the likelihood weighting in Equation~\ref{eq:OR}, they may naively add up the score function values of fine-grained conditions together when generalizing to a general condition, hence disregarding the likelihood-based weighting, and %
leading to generated samples that correspond to the average of the fine-grained conditional training samples.
On the other hand, Figure~\ref{fig:synthetic_black_dog} shows the generated samples using fine-grained prompts at inference time with the model trained with general prompts only. In this case, we observe that the model can leverage both general prompts and generate compositional outcomes in a zero-shot manner. The generated samples achieve $D_{\text{KL}}=1.51$, $\text{FD}=6.64$, and $\text{VS}_{\text{gen}}=2.04$ compared to $\text{VS}_{\text{ref}}=1.10$, when using $\omega=1$. Note that Equation~\ref{eq:AND} is similar to CFG,  ${s_{\theta} (x_t) + \omega (s_{\theta}(x_t|c) - s_{\theta}(x_t))}$, when $\omega \approx M$. Empirically, using a larger $\omega$ pushes the distributions further towards the fine-grained conditional direction, reaching $D_{\text{KL}}=0.93$,  $\text{FD}=1.32$, and $\text{VS}_{\text{gen}}=1.33$ compared to $\text{VS}_{\text{ref}}=1.10$ when $\omega=3$. We do not observe severe distributional shift and diversity reduction in this simple synthetic setting.
\xf{We further discuss the relationship between synthetic settings and large-scale evaluations in Appendix~\ref{app:synthetic_relation}.}
\looseness-1

\section{Benchmarking Framework}
\label{sec:pipeline}
\xf{We propose a new evaluation framework designed to}
evaluate the utility (focusing on quality, diversity, and consistency) of synthetic data generated by T2I models as a function of prompt complexity, comparing with that of real data. 
Comparing the diversity of synthetic images conditioned on a prompt to that of real images in a dataset is challenging, as image-caption pairs in existing datasets are fixed, making it non-trivial to assemble a set of real images that matches a given prompt.\looseness-1

Given an image dataset ${\mathcal{X} \subseteq \mathcal{I} \times \mathcal{Y}}$, where ${\mathcal{I}}$ is the image set and ${\mathcal{Y}}$ is the label set. Each datapoint ${X_i \in \mathcal{X}, i \in \{1,2, \dots, |\mathcal{X}|\}}$ is ${(I_i, y_i)}$ where ${I_i}$ represents the image sample, and ${y_i}$ represents the associated label (either class or captions in our experiments).
We first synthesize the images using captions with $K$ different levels of complexities. The caption set associated with each complexity is noted as ${\mathcal{C}^k, k\in \{1, 2, \dots, K\}}$. We then use T2I models to generate synthetic images $\bar{\mathcal{I}}^k$ from the caption sets $\mathcal{C}^k$. Finally, we employ different evaluation functions ${f \in \mathcal{F}: \{\mathcal{I}, \bar{\mathcal{I}}^k\} \times \mathcal{C}^k \to \mathbb{R}}$ to evaluate the utilities of both synthetic and real data, where $\mathcal{F}$ is the set of evaluation functions considered in our analysis and $f$ is an evaluation function.
In the following, we describe our framework, which consists of captioning, pairing, alignment, sampling, and generation steps, in detail.\looseness-1

\textbf{Captioning. } This step creates captions of different complexities for each datapoint $X_i$. Given the complexity level $K$ that we want to consider in our study, we create captions ${c_i^k}, {i \in \{1,2,\dots,|\mathcal{X}|\}}, {k \in \{1, 2, \dots, K\}}$ of increasing complexities from each datapoint $(X_i, y_i)$. Thus, we transform the original dataset $\mathcal{X}$ into $K$ different datasets ${\tilde{\mathcal{X}}^k , k \in \{1, 2, \dots, K\}}$. The dataset $\tilde{\mathcal{X}}^k$ contains all the images in $\mathcal{I}$ and each image $I_i$ is paired with a caption of complexity level-$k$, $c_i^k$.\looseness-1 \\
\textbf{Pairing. } Given a certain complexity level ${k \in \{1, 2, \dots, K\}}$ and a caption ${c_i^k, i\in \{1,2,\dots,|\mathcal{X}|\}}$, we search for images that are semantically similar to the caption $c_i^k$. We note these images as a set $\bar{\mathcal{I}}^{ik}$ with elements ${I_j^{ik}, j\in \{1,2,\dots,|\mathcal{X}|\}}$. We only keep the image sets $\bar{\mathcal{I}}^{ik}$ with cardinality $\geq 20$. Thus, across complexity levels, the captions left after filtering are different. We denote the indices of the captions left in each complexity as sets ${\mathcal{N}_1^{\rm{p}}, \mathcal{N}_2^{\rm{p}}, ..., \mathcal{N}_k^{\rm{p}}}$.\looseness-1

\textbf{Alignment. } Across complexities, the images left in each complexity ${\bigcup_{i \in \mathcal{N}_k^{\rm{p}}} \bar{\mathcal{I}}^{ik}, k \in \{1, 2, \dots, K\}}$ could be very different due to the filtering performed in the pairing step. This step aligns the image modality across complexities to ensure comparability. Specifically, we iteratively remove images that are not shared across complexities, \ie not in the set ${\bigcap_{k\in \{1,2,\dots,K\}}\bigcup_{i\in \mathcal{N}_k^{\rm{p}}}\bar{\mathcal{I}}^{ik}}$. This process is repeated until the following criteria is satisfied, ${\bigcap_{k\in \{1,2,\dots,K\}}\bigcup_{i\in \mathcal{N}_k^{\rm{p}}}\bar{\mathcal{I}}^{ik} = \bigcup_{k\in \{1,2,\dots,K\}}\bigcup_{i\in \mathcal{N}_k^{\rm{p}}}\bar{\mathcal{I}}^{ik}}$. We denote the indices of the captions left in each complexity as sets ${\mathcal{N}_1^{\rm{a}}, \mathcal{N}_2^{\rm{a}}, ..., \mathcal{N}_k^{\rm{a}}}$.\looseness-1

\textbf{Sampling. } Following the alignment step, the number of remaining captions may still be too large for practical evaluation of T2I models. We randomly sample the same number captions per complexity in order to maximize semantic coverage in a consistent manner. This results in sets of indices of captions ${\mathcal{N}_1^{\rm{s}}, \mathcal{N}_2^{\rm{s}}, ..., \mathcal{N}_k^{\rm{s}}}$ where ${|\mathcal{N}_k^{\rm{s}}|}$ is the same ${\forall k \in \{1,2,\dots,K\}}$. The sampled captions are then used as prompts to collect synthetic data from T2I models. \looseness-1

\textbf{Generation. } For each prompt (caption) ${c_i^k, i\in \mathcal{N}_k^s, k\in \{1,2,\dots,K\}}$, we generate as many images as the cardinality of the smallest similar image set across different caption complexity ${N_{\rm{gen}} = \min_{i\in \mathcal{N}_k^{\rm{s}}, k\in \{1,2,\dots,K\}}|\bar{\mathcal{I}}^{ik}|}$, ensuring representativeness. The generated image set for prompt $c_i^k$ is noted as $\hat{\mathcal{I}}^{ik}$. 
With the prompt conditional image sets for both real images and synthetic images,
we comprehensively assess each axis of data utility using a suite of representative reference-free metrics, which do not constrain the analysis to any predefined target data distribution. 
Our framework can also be readily extended to incorporate reference-based evaluation metrics as needed, as shown in section~\ref{sec:exp}.\looseness-1

\section{Evaluation of Synthetic Data Generated from T2I Models}
\label{sec:exp}
\subsection{Experimental Setup}
\label{sec:exp_setup}

\begin{figure}[t]
    \centering
    \begin{subfigure}[ht]{0.20\textwidth}
        \includegraphics[width=0.9\linewidth]{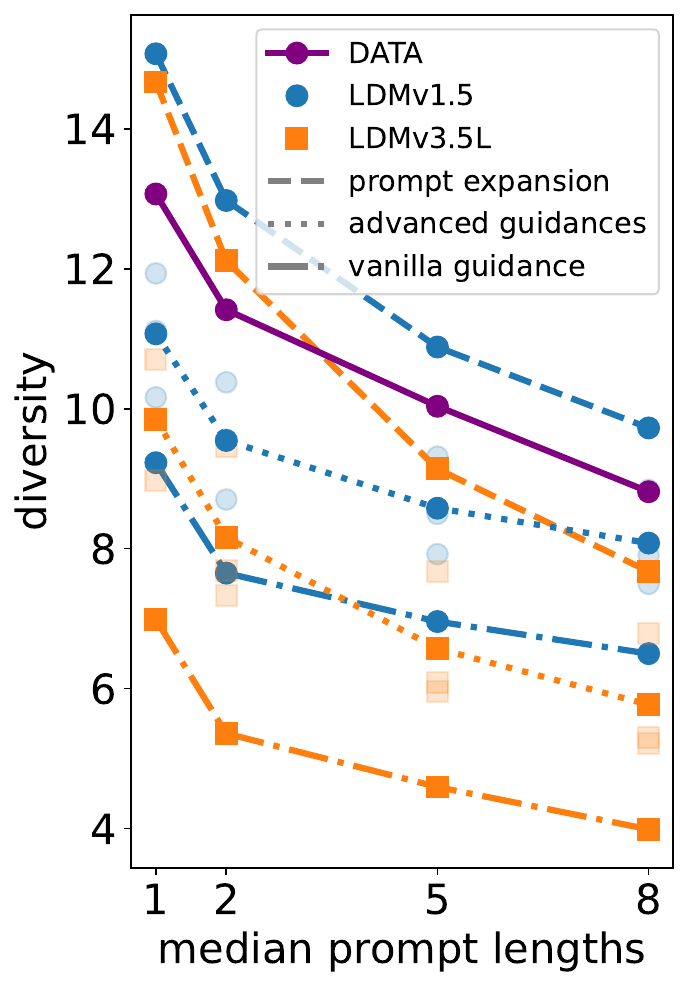}
        \caption{CC12M}
        \label{fig:diversity_SDexpansion_cc12m}
    \end{subfigure}
    \begin{subfigure}[ht]{0.20\textwidth}
        \includegraphics[width=0.9\textwidth]{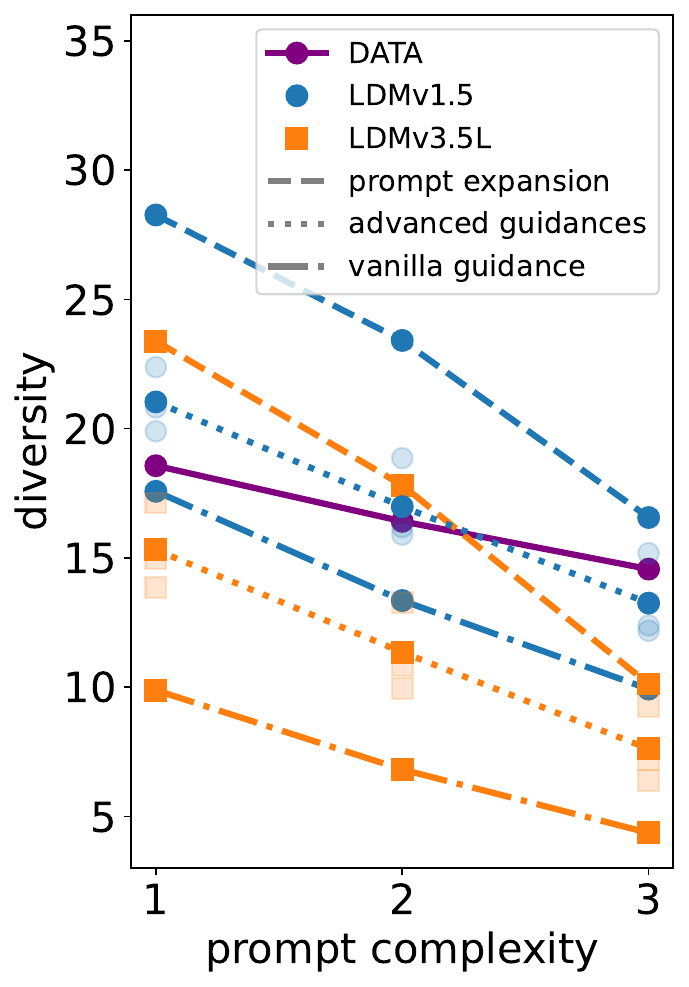}
        \caption{ImageNet-1k}
        \label{fig:diversity_SDexpansion_in1k}
    \end{subfigure}
    \begin{subfigure}[ht]{0.225\textwidth}
        \includegraphics[width=0.9\textwidth]{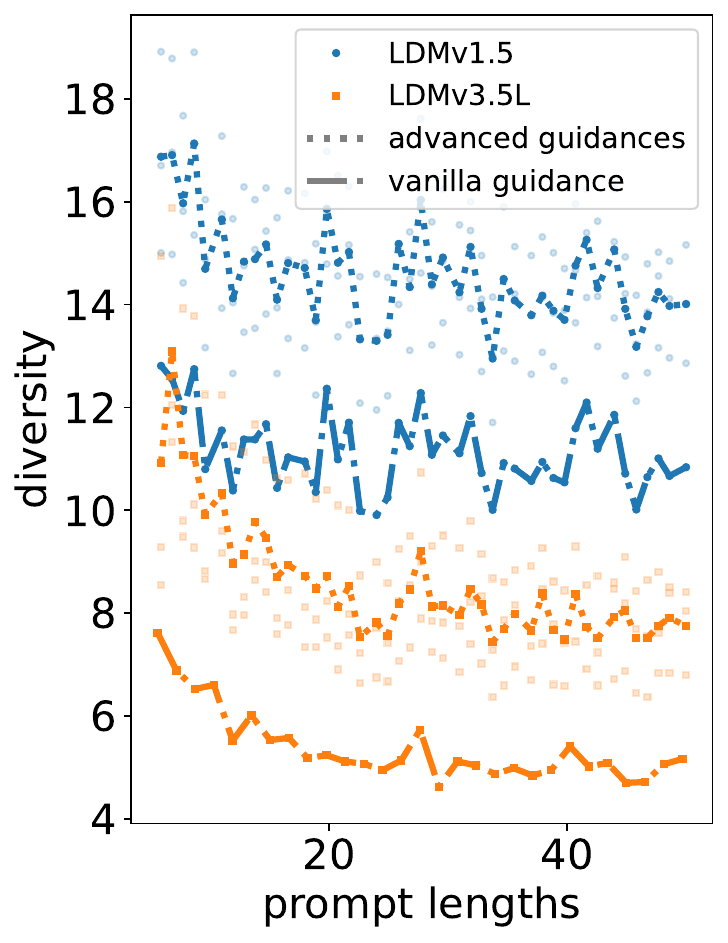}
        \caption{DCI~\footnotemark}
        \label{fig:diversity_SDexpansion_dci}
    \end{subfigure}
    \caption{\textbf{Reference-free diversity metric.} Diversity (Vendi) of LDMv1.5 and LDMv3.5L generations with CC12M and ImageNet-1k prompts when using: 1) vanilla guidance (CFG), 2) prompt expansion, and 3) advanced guidance methods, for which transparent markers correspond to different methods and the solid marker is the average over methods. 
    Both advanced guidance methods and prompt expansion lead to improved diversity over the vanilla guidance. Prompt expansion from shorter captions can surpass the real data diversity. We further extend to much longer DCI prompts. Diversity of all models first decreases then plateaus which is not observed within shorter prompt length ranges.\looseness-1} 
    \label{fig:diversity_SDexpansion}
\end{figure}
\footnotetext{We stop at prompt length of $50$, given the $77$ token limit of text encoders of LDMv1.5. For LDMv3.5 models, we push the prompt length to $100$, as shown in Figure~\ref{fig:DCI_SD}, Appendix~\ref{app:dci}.}

\textbf{Datasets. } We employ three datasets --CC12M, ImageNet-1k, and DCI-- to evaluate the utility of synthetic data generated from T2I models.
CC12M~\citep{datasetcc12m} is a large-scale, internet-crawled image-text dataset containing 12 million image-caption pairs, where we investigate the effect of increasing prompt detail and length on the utility axes of synthetic images.
ImageNet-1k~\citep{imagenet} is a widely used, object-centric, curated dataset with 1,000 classes, where we examine the effect of concept specificity on the utility axes. 
DCI~\citep{dcidataset} is a large-scale, human-curated image-caption dataset featuring detailed, long captions for each image, where we further explore prompt complexity using \emph{very long} prompts. More details are presented in Appendix~\ref{app:dataset_details}.\looseness-1 \\
\textbf{Models. } We select LDMv1.5~\citep{ldmrombach} and LDMv3.5L~\citep{sd35model} models for the presentation of results. We also include the evaluation results from LDMv3.5M~\citep{sd35model} and LDM-XL~\citep{sdxl} in Appendix~\ref{app:xl35m}.  LDMv1.5 is the early high-performing T2I model using the latent diffusion model (LDM) architecture~\citep{ldmrombach}, while LDMv3.5L is its latest successor, employing a rectified flow model~\citep{rectifiedflow}. These models are representative of current open-source T2I models and illustrate the trend of model improvement over time. We consider the following guidance techniques: 
CFG~\citep{cfg} (referred to as vanilla guidance), and advanced guidance approaches including CADS~\citep{cads}, interval guidance~\citep{intervalguide}, and APG~\citep{apg}. 
We also explore the effect of explicitly expanding prompts by adding information, using large language models to expand each caption to $N_{\text{gen}}$ different captions with maximum thirty words. 
Additionally, we report evaluation results of Flux-schnell~\citep{flux2024} and Infinity~\citep{infinityautoregressive} to cover more model types, to which advanced guidance methods are not applicable. Considering the limited inference-time interventions, we present the results for these two models in Appendix~\ref{app:additional_models}. More implementation details are presented in Appendix~\ref{app:exp_details}.\looseness-1 \\
\textbf{Metrics. } We use the aesthetic score~\citep{aesthetic_v2_5} to assess image quality, the Vendi score~\citep{vendiscore} to measure diversity, and the Davidsonian Scene Graph (DSG) score~\citep{dsg} to evaluate prompt consistency. \xf{Our human evaluations in Appendix~\ref{app:human_eval} confirm the validity of the automatic metrics.}
We also report widely used reference-based metrics, including 
FDD (Fr\'{e}chet distance using DINOv2~\citep{dinov2})~\citep{fddmetric}, precision~\citep{prmetric}, density and coverage~\citep{dcmetric}. \footnote{Unless otherwise specified, we use the DINOv2~\citep{dinov2} feature space to compute these metrics, since it has been found to correlate well with human judgement~\citep{hall2024towards,fddmetric}.}\looseness-1

\begin{figure}[t]
    \centering
    \begin{subfigure}[ht]{0.20\textwidth}
        \includegraphics[width=0.9\linewidth]{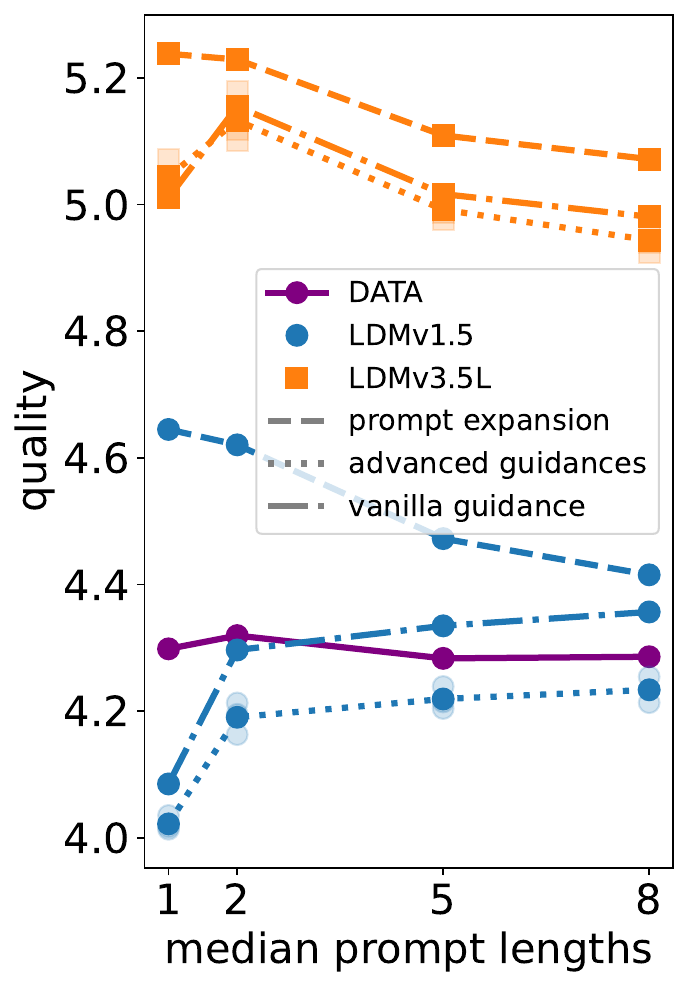}
        \caption{CC12M}
        \label{fig:quality_SDexpansion_cc12m}
    \end{subfigure}
    \begin{subfigure}[ht]{0.20\textwidth}
        \includegraphics[width=0.9\textwidth]{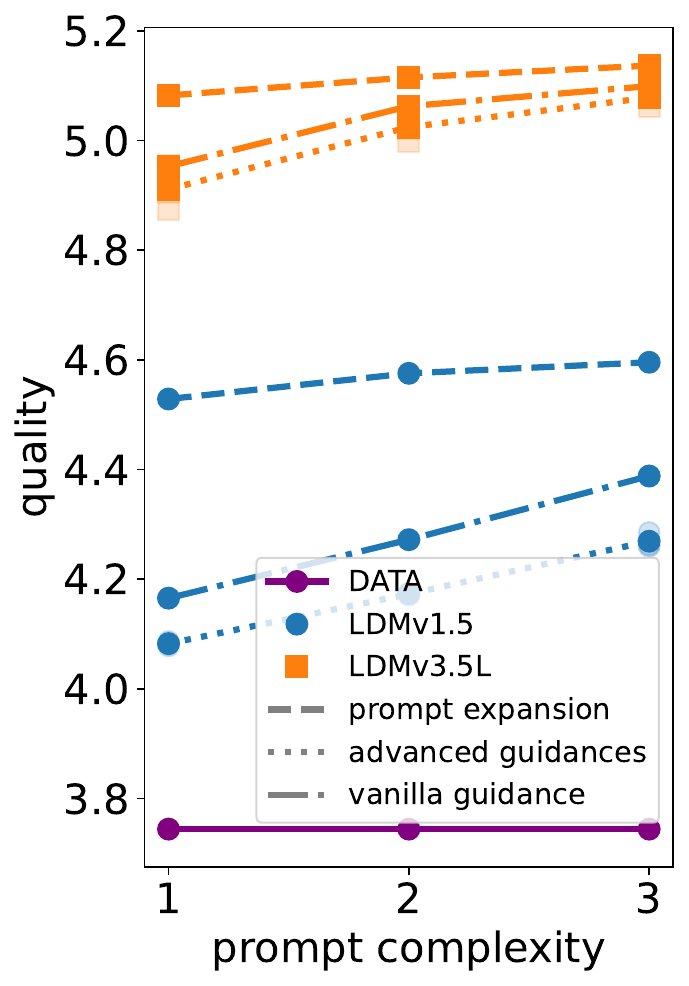}
        \caption{ImageNet-1k}
        \label{fig:quality_SDexpansion_in1k}
    \end{subfigure}
    \begin{subfigure}[ht]{0.225\textwidth}
        \includegraphics[width=0.9\textwidth]{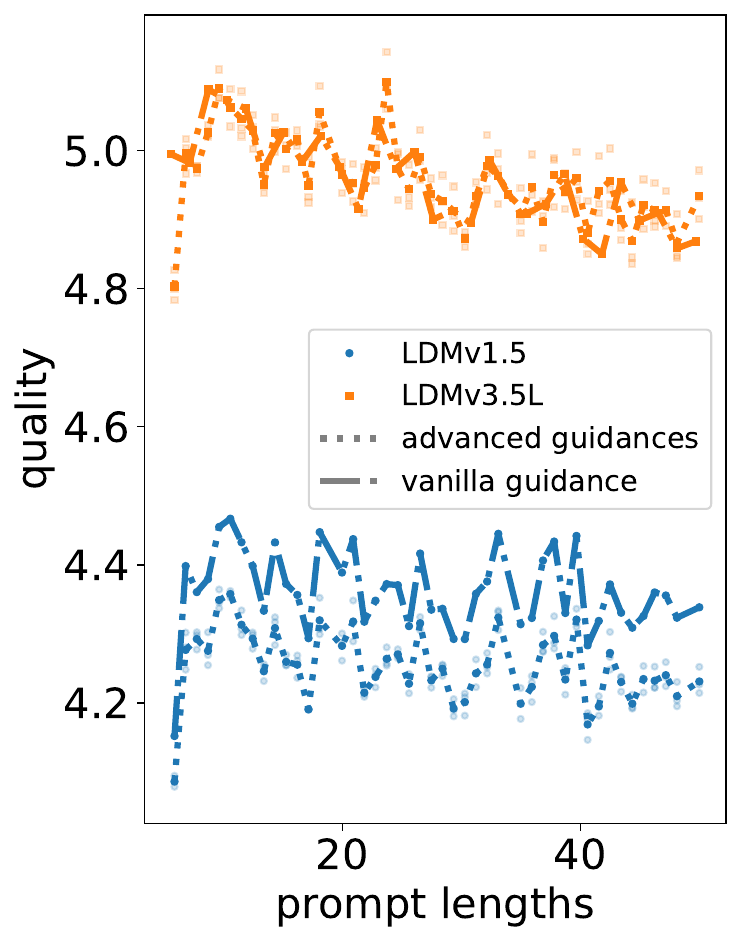}
        \caption{DCI}
        \label{fig:quality_SDexpansion_dci}
    \end{subfigure}
    \caption{\textbf{Reference-free quality metric.} Quality (aesthetic) of LDMv1.5 and LDMv3.5L generations with CC12M and ImageNet-1k prompts when using: 1) vanilla guidance (CFG), 2) prompt expansion, and 3) advanced guidance methods, for which transparent markers correspond to different methods and the solid marker is the average over methods. 
    Advanced guidance methods slightly impair the quality while prompt expansion increases it. We further extend to much longer DCI prompts. Quality of all models first increases then decreases which is not observed within shorter prompt ranges.\looseness-1} 
    \label{fig:quality_SDexpansion}
\end{figure}

\begin{figure}[t]
    \centering
    \begin{subfigure}[ht]{0.20\textwidth}
        \includegraphics[width=0.9\linewidth]{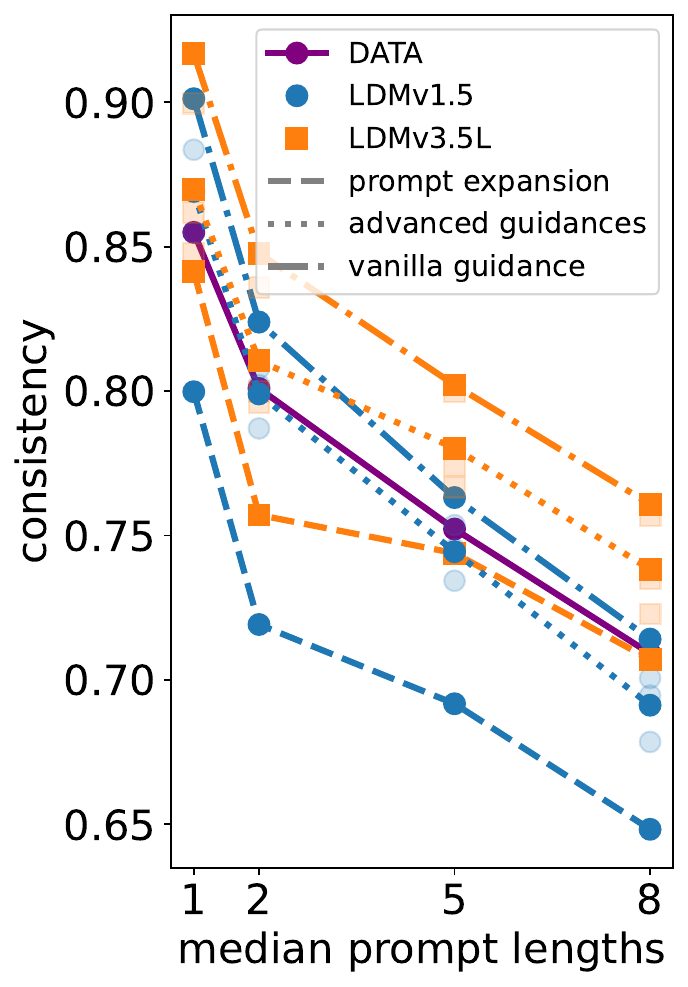}
        \caption{CC12M}
        \label{fig:consistency_SDexpansion_cc12m}
    \end{subfigure}
    \begin{subfigure}[ht]{0.20\textwidth}
        \includegraphics[width=0.9\textwidth]{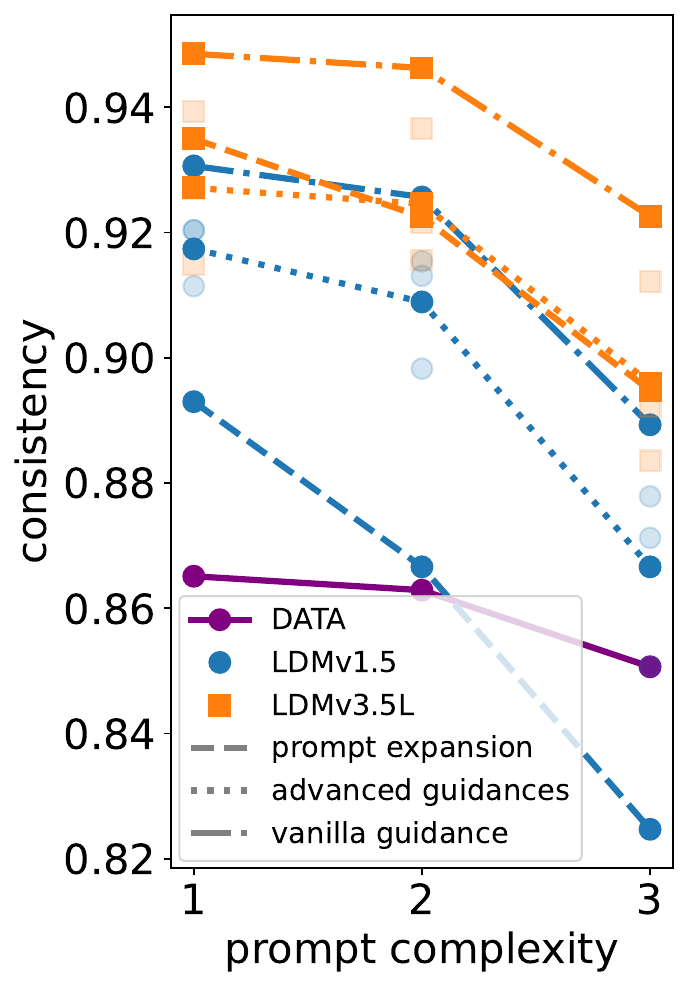}
        \caption{ImageNet-1k}
        \label{fig:consistency_SDexpansion_in1k}
    \end{subfigure}
    \begin{subfigure}[ht]{0.225\textwidth}
        \includegraphics[width=0.9\textwidth]{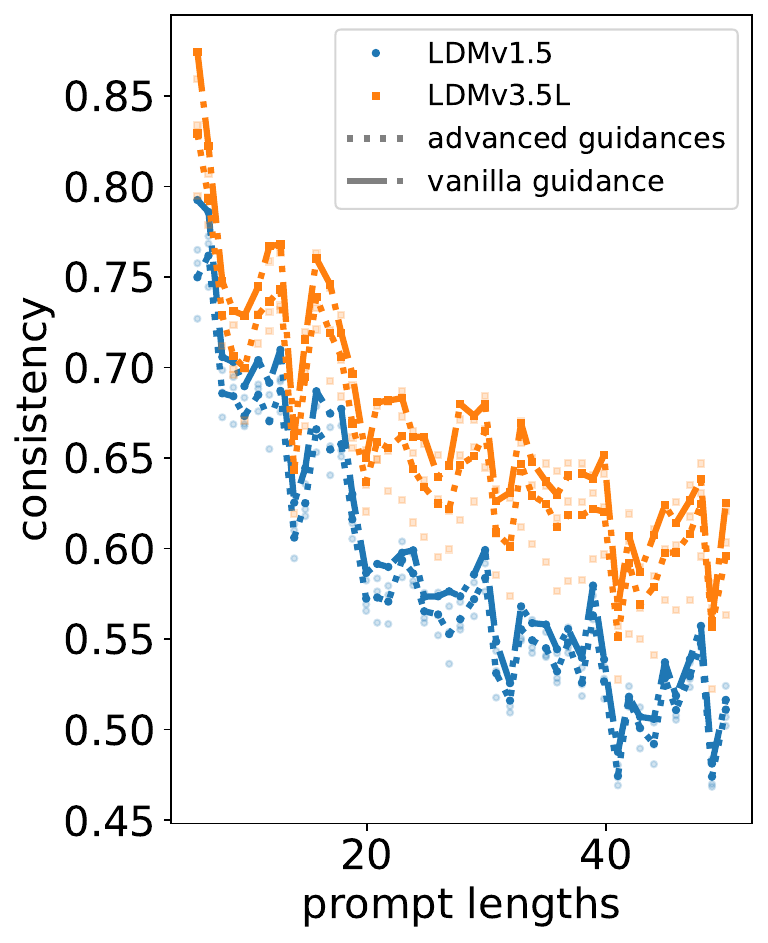}
        \caption{DCI}
        \label{fig:consistency_SDexpansion_dci}
    \end{subfigure}
    \caption{\textbf{Reference-free consistency metric.} Consistency (DSG) metrics of LDMv1.5 and LDMv3.5L generations with CC12M and ImageNet-1k prompts when using: 1) vanilla guidance (CFG), 2) prompt expansion, and 3) advanced guidance methods, for which transparent markers correspond to different methods and the solid marker is the average over methods. 
    Both advanced guidance methods and prompt expansion lead to lower consistency scores compared to vanilla guidance. We further extend to much longer DCI prompts. Consistency of all models decreases when the prompt lengths increases, which is the same as in the shorter prompt ranges.\looseness-1} 
    \label{fig:consistency_SDexpansion}
    \vspace{-1em}
\end{figure}

\subsection{Utility axes measured by reference-free metrics}
\label{sec:reference_free}
\textbf{Diversity. \ \ } We present results in Figure~\ref{fig:diversity_SDexpansion}. Our evaluation findings empirically verify that the diversity of synthetic data decreases as we increase prompt complexity for all models and interventions considered (Figures~\ref{fig:quality_SDexpansion_cc12m} and \ref{fig:quality_SDexpansion_in1k}). As increasing the complexity (either by adding details in CC12M or increasing the class specificity in ImageNet-1k) constrains the generation process, the degree of freedom of the model is reduced.
In terms of inference-time interventions, both advanced guidance methods and prompt expansion lead to higher diversity than vanilla guidance. Note that prompt expansion can be considered as an explicit way to sample the likelihood of more fine-grained prompts (Equation~\ref{eq:OR}) using a pre-trained language model~\citep{bettercaptiondalle3}, which effectively increases the diversity. It is possible to obtain higher diversity than the one captured by the real data when the prompt complexity is low, but this comes at the cost of consistency (as indicated in Figure~\ref{fig:consistency_SDexpansion}). 
In CC12M, the diversity gap between vanilla guidance and real data slightly decreases with the increase of prompt complexity, indicating that the diversity for more general prompts is harder to be captured, aligning with our synthetic experimental results (Figure~\ref{fig:synthetic_general_cat}). However, ImageNet-1k exhibits a different pattern. This may be due to the construction of the real data prompt-image pairs, as general prompts can cover more subcategories than the ones in ImageNet-1k. For example, the general prompt (complexity=1) \texttt{stringed instrument} can contain mandolin (covered by T2I models as shown in Figure~\ref{fig:header_visuals_pmt_complexity_in1k}, Appendix~\ref{app:qualitative_samples}), but this class is not part of the ImageNet-1k dataset, leading to an underestimation of the real data diversity.
We further push this evaluation towards much longer prompts using DCI dataset. We observe that the diversity decreases and then plateaus as we continuously increase the prompt length, indicating that the models may not be able to follow all the constraints added in the prompts (also reflected in the consistency measurements in Figure~\ref{fig:consistency_SDexpansion_dci}). For most LDM models, the plateau starts at a prompt length of $\sim30$ words. However, the plateau region of LDMv1.5 seems to occur earlier than its successor models.\looseness-1

\textbf{Quality. \ \ } Results across datasets and inference-time intervention methods are shown in Figure~\ref{fig:quality_SDexpansion}. In CC12M and ImageNet-1k datasets, where we use relatively shorter prompts, the aesthetics of the synthetic data remains relatively stable across prompt complexities, especially for the most recent model (LDMv3.5L). Further, advanced guidance methods lead to slightly lower performance than vanilla guidance. This is by contrast to prompt expansion, which consistently exhibits higher aesthetics than vanilla guidance. Compared to real data, synthetic data exhibits higher or competitive aesthetic quality. This is perhaps unsurprising given the aesthetics finetuning performed on the most recent models (LDMv3.5L). When we evaluate on the DCI dataset using much longer prompts, image aesthetics first increase and then start to gradually decrease as a function of prompt length for all models. The sharp slope observed towards more general prompts and the gradual decrease observed towards more fine-grained prompts appear aligned with the observation of our synthetic experiments that generalization to longer and fine-grained (higher complexity) prompts is easier than generalization to more general (lower complexity) ones, especially when leveraging CFG.\looseness-1 

\textbf{Consistency. \ \ } 
Figure~\ref{fig:consistency_SDexpansion} shows the results across datasets and inference-time intervention methods.
We observe that the prompt consistency decreases as we increase prompt complexity for all the cases considered. This suggests that T2I models struggle to incorporate the increasing amount of details (objects, attributes, and relations among objects) required by longer prompts, and to faithfully generate very specific concepts.
Real data also shows a decrease in consistency across complexities, which is due to the image-text pairing process.
The experiments on DCI dataset further consolidate this observation.
\xf{We present the 95\% confidence interval of these metrics in Appendix~\ref{app:95ci}, confirming the statistical significance of the observed trends.}
The findings connecting different utility axes and prompt complexity are visually captured in Figures~\ref{fig:header_visuals_pmt_complexity} and \ref{fig:header_visuals_pmt_complexity_in1k} in Appendix~\ref{app:qualitative_samples}.  \looseness-1

\subsection{Utility axes measured by reference-based metrics}
\label{para:reference_based}
\begin{figure}[t]
    \centering
    \begin{subfigure}[ht]{0.20\textwidth}
        \includegraphics[width=\textwidth]{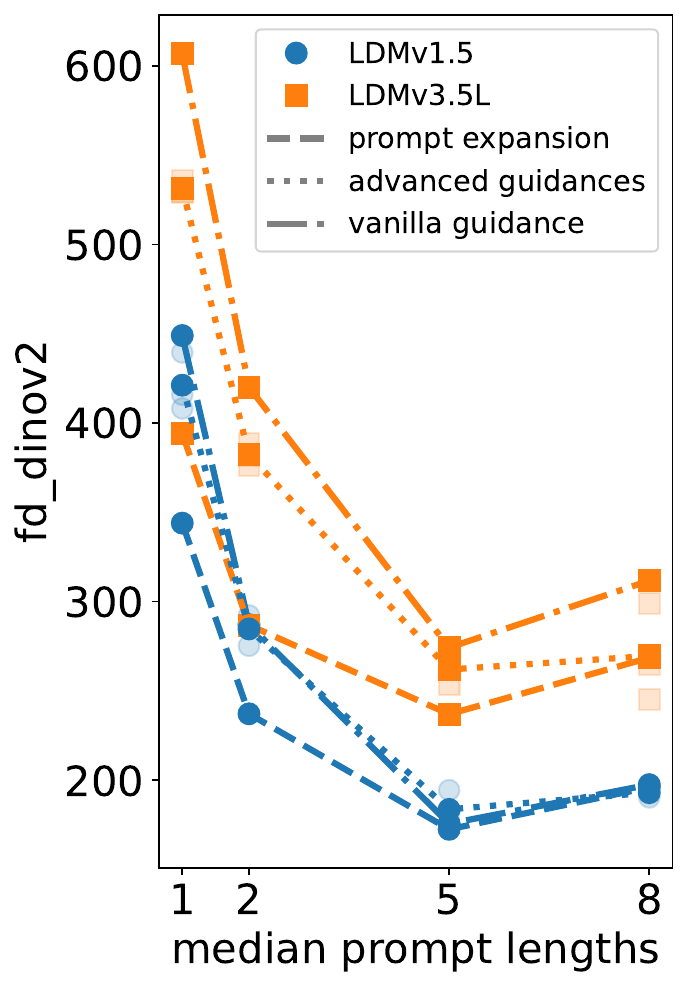}
        \caption{FDD}
        \label{fig:fd_dinov2_SDexpansion_cc12m}
    \end{subfigure}
    \begin{subfigure}[ht]{0.20\textwidth}
        \includegraphics[width=\textwidth]{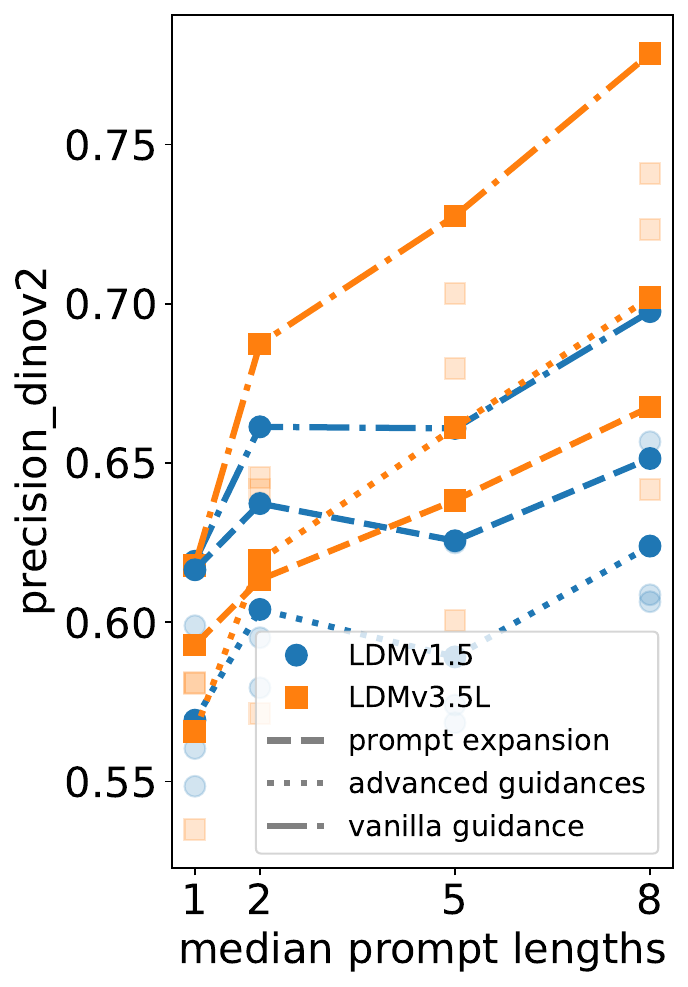}
        \caption{Precision}
        \label{fig:precision_dinov2_SDexpansion_cc12m}
    \end{subfigure}
    \begin{subfigure}[ht]{0.20\textwidth}
        \includegraphics[width=\linewidth]{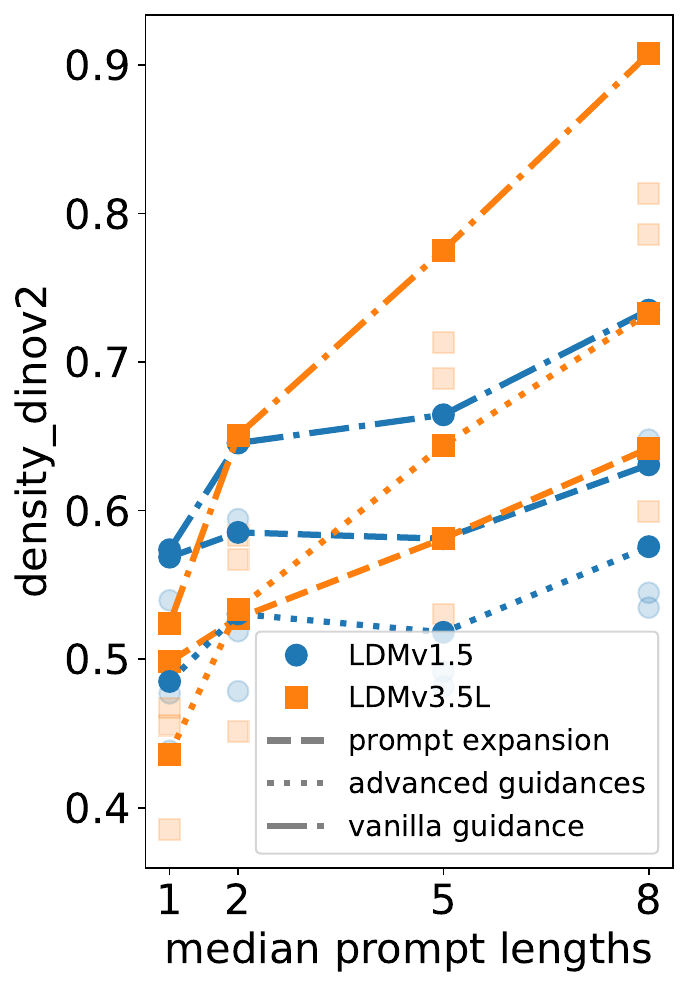}
        \caption{Density}
        \label{fig:density_dinov2_SDexpansion_cc12m}
    \end{subfigure}
    \begin{subfigure}[ht]{0.20\textwidth}
        \includegraphics[width=\textwidth]{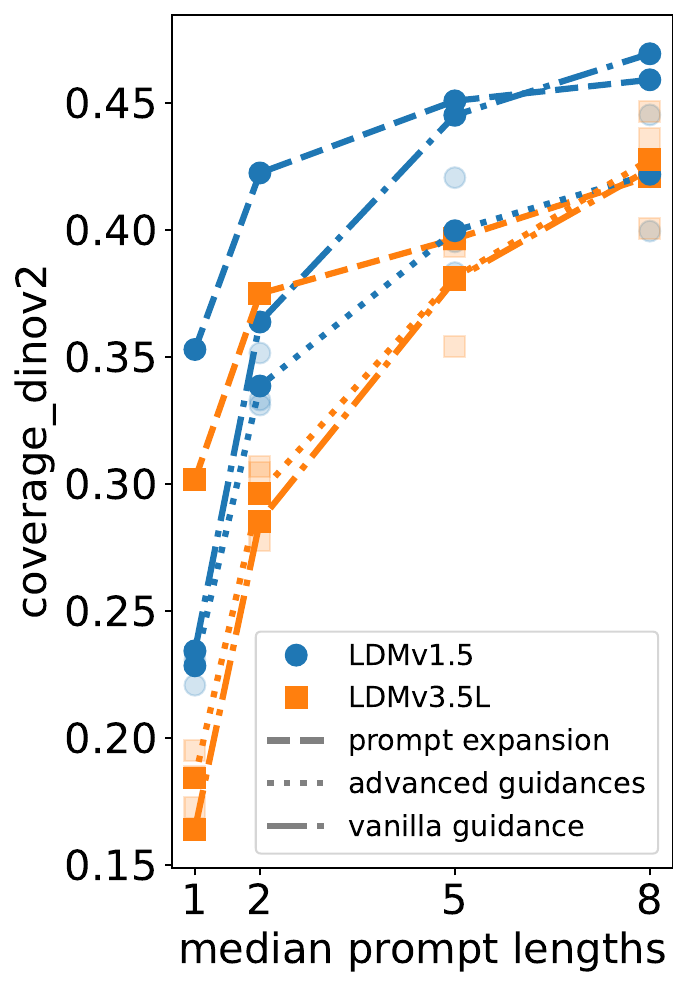}
        \caption{Coverage}
        \label{fig:coverage_dinov2_SDexpansion_cc12m}
    \end{subfigure}
    \caption{\textbf{Reference-based utility metrics of synthetic data using CC12M prompts.} FDD, precision, density and coverage for LDMv1.5 and LDMv3.5L generations with: (1) vanilla guidance (CFG), (2) prompt expansion, and (3) advanced guidance methods, for which transparent markers correspond to different methods and the solid marker is the average over methods. Both advanced guidance methods and prompt expansion lead to better FDD. Although prompt expansion improves coverage and advanced guidance methods match coverage for LDMv3.5, they both sacrifice precision and density. LDMv1.5 has thus better overall performance (lower FDD) than LDMv3.5L.\looseness-1}
\label{fig:marginal_dinov2_SDexpansion_cc12m}
\end{figure}

Reference-free metrics are suitable for evaluating the utility of synthetic data in the wild with prompts only. Yet, in our analysis, the prompts are extracted from existing datasets, \emph{containing both images and texts}, therefore enabling the use of reference-based metrics for evaluation too. 
The CC12M results are presented in Figures~\ref{fig:marginal_dinov2_SDexpansion_cc12m}. We leave the ImageNet-1k results in Figure~\ref{fig:marginal_dinov2_SDexpansion_in1k} in Appendix~\ref{app:reference_based_in1k}. These figures reveal that as prompt complexity increases, there is an overall tendency for synthetic data to improve its precision, density and coverage in both CC12M and ImageNet-1k datasets, suggesting that more detailed captions, grounded on the real images, help generate data that lies within the support of the reference dataset. This also aligns with our synthetic experimental results showing that generalizing to general (lower complexity) prompts results in mode concentration and larger distributional gap (Figure~\ref{fig:synthetic_general_cat}). \looseness-1

Figures~\ref{fig:marginal_dinov2_SDexpansion_cc12m} and~\ref{fig:marginal_dinov2_SDexpansion_in1k} also show that advanced guidance methods and prompt expansion lead to overall better FDD than vanilla guidance. 
When it comes to coverage, prompt expansion also exhibits high performance compared to the vanilla guidance in most cases, although its benefits are less pronounced for higher complexity prompts.
The improvements in coverage come, in both cases, at the expense of precision and density. This is perhaps expected as 1) prompt expansion may include details via pre-trained language models that are not present in the real images, therefore deviating the generation process from the reference data\footnote{This can be understood as a mis-alignment between the likelihood estimation from the pre-trained language model and from the real image dataset.}; and 2) advanced guidance approaches may push the generation process outside of the real data manifold, resulting in reduced precision \wrt the vanilla guidance. \xf{The drop in precision and density shows the trade-off of creativity and fidelity to the real data distribution, emphasizing a cautious usage of T2I models for different downstream applications. We give a discussion in Appendix~\ref{app:trade-off-creativity-fidelity}.}\looseness-1

Interestingly, LDMv1.5 model has better overall performance (lower FDD) compared to LDMv3.5L model across different guidance methods, prompt expansion, and prompt lengths. However, as shown in section~\ref{sec:reference_free}, LDMv3.5L performs better than LDMv1.5 on reference-free quality and consistency metrics, only falling short in diversity, as shown in Figures~\ref{fig:diversity_SDexpansion_cc12m}, \ref{fig:quality_SDexpansion_cc12m}, and \ref{fig:consistency_SDexpansion_cc12m}. This suggests that the diversity is a key characteristic of real-world image distributions. Thus, synthetic data from models that fail to capture the diverse nature of the real world may lead to lower general performance in real-world applications even though the models advance the state-of-the-art generation quality in reference-free settings. Similar findings may be observed for the ImageNet-1k dataset, presented in Figures~\ref{fig:diversity_SDexpansion_in1k}, \ref{fig:quality_SDexpansion_in1k}, \ref{fig:consistency_SDexpansion_in1k},  and~\ref{fig:marginal_dinov2_SDexpansion_in1k}.\looseness-1

\subsection{Combining prompt expansion and advanced guidance methods}
\label{para:combining}
\begin{figure}[t]
    \centering
    \begin{subfigure}[ht]{0.20\textwidth}
        \includegraphics[width=0.85\linewidth]{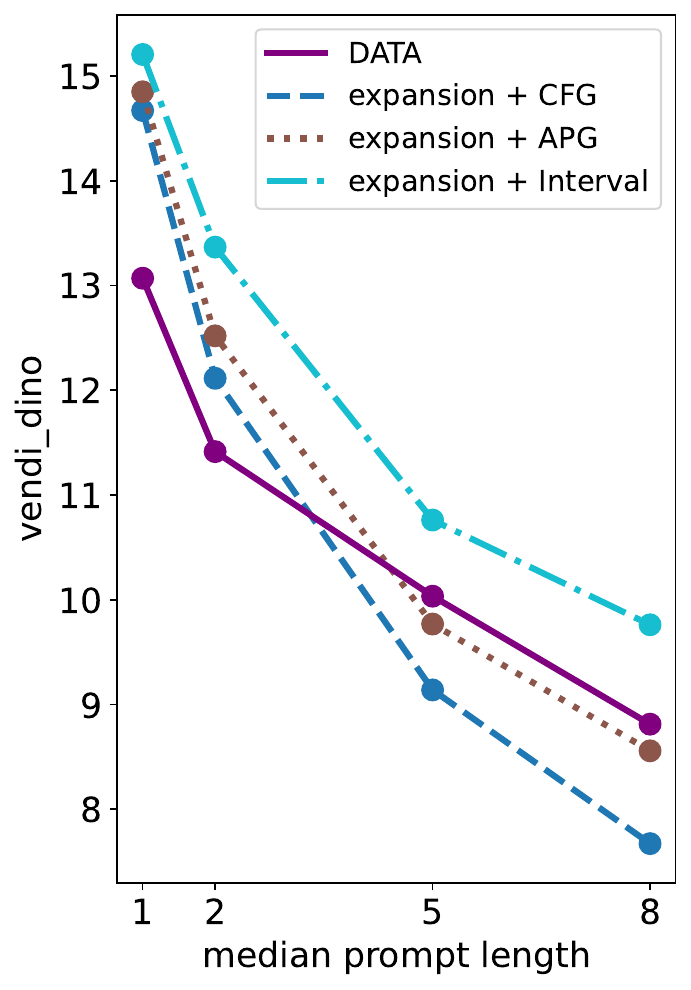}
        \caption{Diversity}
        \label{fig:vendi_dinov2_expansionguide_cc12m}
    \end{subfigure}
    \begin{subfigure}[ht]{0.20\textwidth}
        \includegraphics[width=0.85\textwidth]{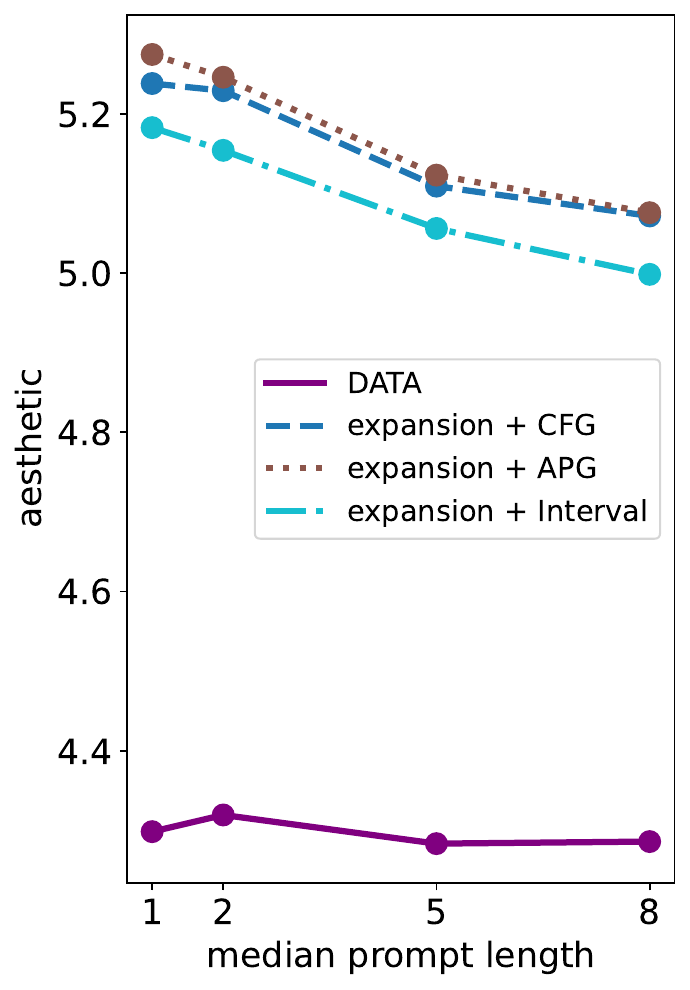}
        \caption{Quality}
        \label{fig:aesthetic_expansionguide_cc12m}
    \end{subfigure}
    \begin{subfigure}[ht]{0.20\textwidth}
        \includegraphics[width=0.85\textwidth]{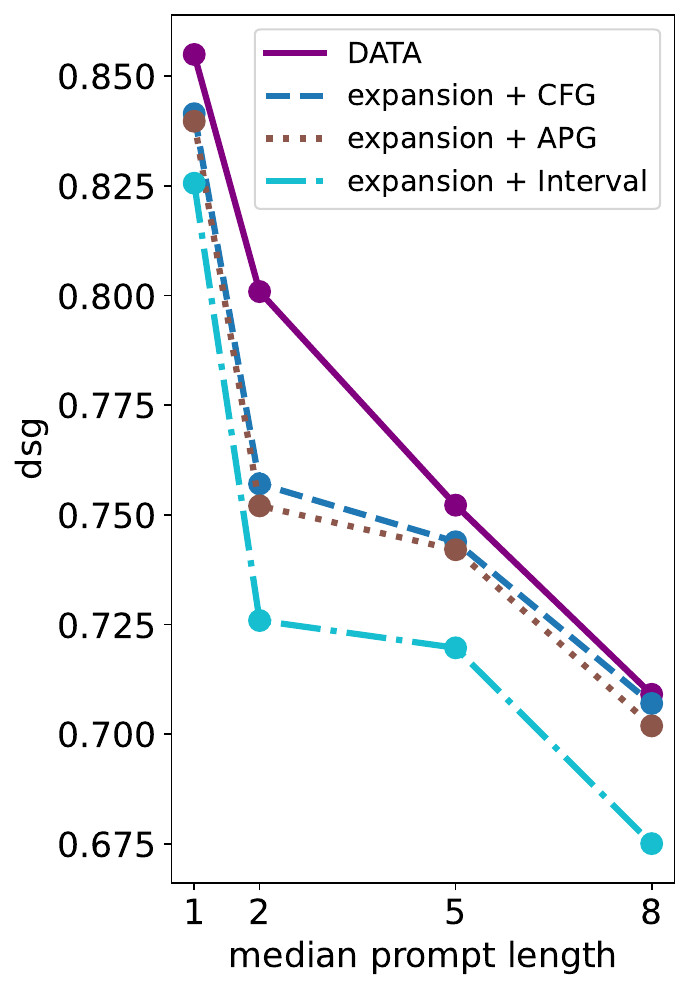}
        \caption{Consistency}
        \label{fig:dsg_expansionguide_cc12m}
    \end{subfigure}
    \begin{subfigure}[ht]{0.20\textwidth}
        \includegraphics[width=0.85\textwidth]{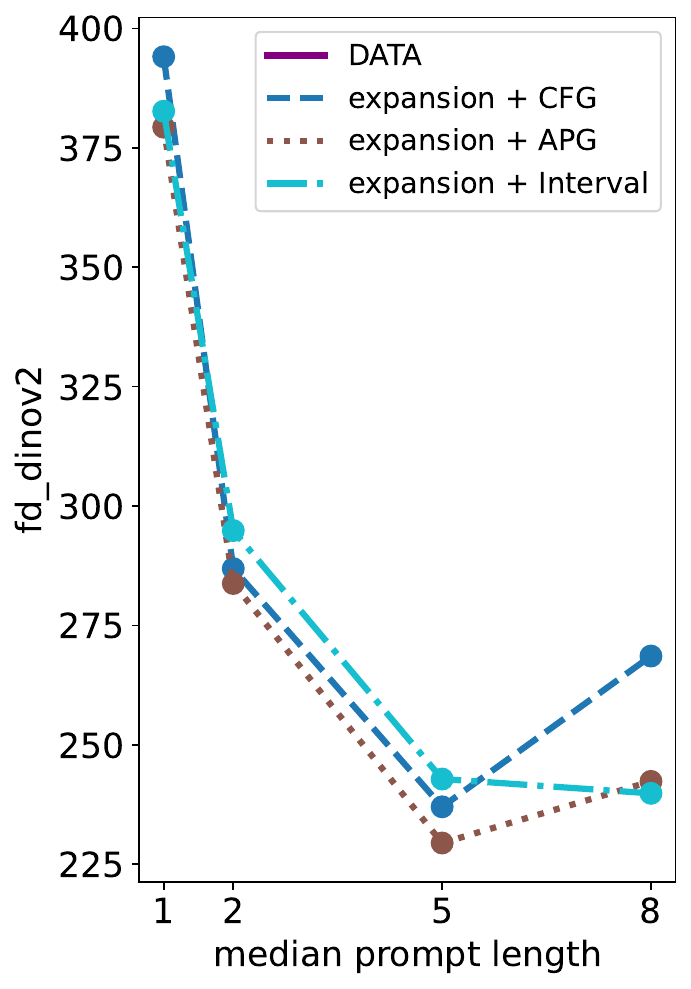}
        \caption{FDD}
        \label{fig:fdd_expansionguide_cc12m}
    \end{subfigure}
    \caption{\textbf{Effect of combining prompt expansion and guidance methods on the utility of synthetic data from LDMv3.5L using CC12M prompts.} This can further boost the diversity of synthetic images, with comparable quality, consistency, and FDD (expecially with APG).\looseness-1 }
    \label{fig:expansionguide_cc12m}
\end{figure}

Figure~\ref{fig:expansionguide_cc12m} shows the effect of combining prompt expansion and advanced guidance methods on the 3 utility axes of synthetic data, when using prompts from CC12M. We observe that the prompt expansion and advanced guidance methods can be combined to further boost the diversity of synthetic images. We note that in this case, Interval guidance appears to slightly suffer from aesthetic quality and shows considerably lower prompt consistency than both CFG and APG for prompts lengths higher than 1. Finally, when contrasting with real data, we observe that the diversity of the real data may be surpassed by that of synthetic data across prompt complexities, but this again comes at the cost of consistency. A closer look at advanced guidance methods and prompt expansion is provided in Appendix~\ref{app:closer_look}.\looseness-1

\section{Related Work}
\label{sec:related_work}
\textbf{Improving the synthetic images utility. \ \ }
Recent studies have yielded in-depth analyses on the utility of synthetic data from off-the-shelf T2I models~\citep{astolfi2024consistency,digin,lee2023holistic}, and have revealed that the impressive progress in image quality has come at the expense of generation diversity. 
Inference-time interventions have been introduced to improve over classifier free guidance (CFG) sampling~\citep{cfg}. These techniques involve either prompt rewriting~\citep{promptexpansion}, or alternative guidance signals. For example, APG~\citep{apg} proposes to adaptively adjust the guidance scale across different directions, CADS~\citep{cads} implements an annealing strategy that gradually uses less noisier guidance signal during the sampling process, and Interval-guidance~\citep{intervalguide} introduces a simpler approach by disabling the conditional guidance at the beginning of the sampling process. Auto-guidance~\citep{bvg} uses a bad-version of the model to guide the generations; however, in the context of open-source models, it is not trivial how to obtain bad-versions of the models that would provide useful guidance signals. Chamfer guidance~\citep{dallasen2025} leverages few examples of real images to guide the sampling process towards high-utility samples.\looseness-1

\textbf{Evaluation of T2I synthetic data. \ \ }
Numerous evaluation metrics have been proposed to assess the performance of T2I models, covering the quality, diversity, consistency utility axes. Many metrics need a reference set of real data to compare with. For example, Fréchet Inception Distance (FID)~\citep{fidmetric} has been widely used as an overall metric and measures the distance between the feature distributions of synthetic and real images. Precision and recall~\citep{prmetric}, followed by density and coverage~\citep{dcmetric} were proposed to evaluate the realism and diversity of synthetic images. To evaluate T2I models in the wild, researchers have introduced several reference-free metrics. When it comes to quality, aesthetics score~\citep{aesthetic,aesthetic_v2_5} has been built on top of CLIP~\citep{radford2021learning} to estimate how visually appealing synthetic images are. For diversity, Vendi score~\citep{vendiscore} was designed to evaluate the diversity of a set of images. Moreover, in ~\citet{w1kp}, a human-calibrated perceptual variability measure was introduced and then used to investigate the influence of linguistic features in the T2I generation process. This work focuses on the diversity axes exclusively, does not contrast results with real data, and does not consider any sampling intervention beyond vanilla guidance. For prompt-image consistency, CLIPscore~\citep{clipscore} has been used to measure the alignment between the generated image and the input text prompt. 
However, CLIPscore has been shown to be correlated with object numerosity~\citep{clipscorenumobject} and unable to handle detailed descriptions~\citep{longclip}. 
VQAscore~\citep{vqascore}, TIFA~\citep{tifa} and DSG~\citep{dsg} further leverage visual question answering (VQA) models for image-prompt consistency. Among them, TIFA and DSG decompose the captions into several granular questions and compute accuracy scores based on the answers to each question. Beyond these single-axis utility metrics, several analyses have been performed on T2I models from a multi-objective perspective~\citep{astolfi2024consistency,lee2023holistic,libevalgim}. Our study complements previous work by assessing the utility of synthetic data as a function on prompt complexity.\looseness-1

\section{Conclusion}
\label{sec:conclusion}
\textbf{Conclusion. } In this paper, we \xf{proposed a new evaluation framework} and presented an in-depth analysis of the utility of synthetic data as a function of prompt complexity, including synthetic and large-scale real data evaluations. Our synthetic experiments show that generalizing to more general conditions is harder than the other way round. Our study suggests that prompt complexity is a crucial axis to consider when sampling from T2I models, and requires more investigation when generating from very general prompts. \xf{Our findings include 1) the trend of utilities are non-linear, showing an asymmetry of prompt length generalization, 2) diversity does not collapse but plateaus as prompt length increases, suggesting an inherent “lower bound of diversity” in T2I models, 3) optimizing for reference-free metrics harms distributional fidelity, 4) combining guidance methods with prompt expansion achieve the most interesting trade-offs in utilities.} In addition, diversity is a key feature of real-world image distributions, which is still not properly captured by synthetic images from the state-of-the-art T2I models when no explicit prompt expansion is conducted during inference. 

\textbf{Limitations. } For datasets with long captions, our framework needs to use large and diverse enough image datasets to build paired image sets, which is not easy to get. Since DCI dataset has only 7805 image-caption pairs, we did not compare the utilities of synthetic images with real data.

\section*{Reproducibility statement}
We detailed our experimental setups, including datasets, hyperparameters, training and inference settings, and computing resources, in both main texts (Section~\ref{sec:synthetic} and \ref{sec:exp_setup}) and Appendices (Appendix~\ref{app:synthetic_exp_setting} and \ref{app:exp_details}). We provided full derivations of our theoretical results in Appendix~\ref{app:synthetic_proof}. All the datasets used in our paper are open-source and can be downloaded following the instructions given by dataset providers. We detailed the licences of datasets and packages used in Appendix~\ref{app:licences}. We provided detailed descriptions of our evaluation framework in Section~\ref{sec:pipeline} and Appendix~\ref{app:framework}, and our anonymous codes in supplementary materials for reviewing.

\section*{Acknowledgment}
AC's work was supported by the Institute of Information \& Communications Technology Planning \& Evaluation (IITP) grant funded by the Korean Government (MSIT) (No. RS-2024-00457882, National AI Research Lab Project) and his own CIFAR Canadian AI Chair. ZX is supported by the AIM program at Meta and the EDI in Research Scholarship of Mila. The authors thank Oscar Mañas, Andrei Nicolicioiu, and Nicola Dall'Asen for insightful discussions regarding the evaluation pipeline; Avery HW Ryoo for helpful suggestions on prompting language models; and Nicolas Beltran-Velez and Felix Friedrich for constructive feedback during the rebuttal.

\bibliographystyle{assets/plainnat}
\bibliography{paper}

\begin{thebibliography}{54}
\providecommand{\natexlab}[1]{#1}
\providecommand{\url}[1]{\texttt{#1}}
\expandafter\ifx\csname urlstyle\endcsname\relax
  \providecommand{\doi}[1]{doi: #1}\else
  \providecommand{\doi}{doi: \begingroup \urlstyle{rm}\Url}\fi

\bibitem[Ahmadi and Agrawal(2024)]{clipscorenumobject}
Saba Ahmadi and Aishwarya Agrawal.
\newblock An examination of the robustness of reference-free image captioning evaluation metrics.
\newblock In \emph{Findings of the Association for Computational Linguistics: EACL 2024}, pages 196--208, 2024.

\bibitem[Askari-Hemmat et~al.(2025)Askari-Hemmat, Pezeshki, Dohmatob, Bordes, Astolfi, Hall, Verbeek, Drozdzal, and Romero-Soriano]{askari2025improving}
Reyhane Askari-Hemmat, Mohammad Pezeshki, Elvis Dohmatob, Florian Bordes, Pietro Astolfi, Melissa Hall, Jakob Verbeek, Michal Drozdzal, and Adriana Romero-Soriano.
\newblock Improving the scaling laws of synthetic data with deliberate practice.
\newblock \emph{arXiv preprint arXiv:2502.15588}, 2025.

\bibitem[Astolfi et~al.(2024)Astolfi, Careil, Hall, Ma{\~n}as, Muckley, Verbeek, Soriano, and Drozdzal]{astolfi2024consistency}
Pietro Astolfi, Marlene Careil, Melissa Hall, Oscar Ma{\~n}as, Matthew Muckley, Jakob Verbeek, Adriana~Romero Soriano, and Michal Drozdzal.
\newblock Consistency-diversity-realism pareto fronts of conditional image generative models.
\newblock \emph{arXiv preprint arXiv:2406.10429}, 2024.

\bibitem[Betker et~al.(2023)Betker, Goh, Jing, Brooks, Wang, Li, Ouyang, Zhuang, Lee, Guo, et~al.]{bettercaptiondalle3}
James Betker, Gabriel Goh, Li~Jing, Tim Brooks, Jianfeng Wang, Linjie Li, Long Ouyang, Juntang Zhuang, Joyce Lee, Yufei Guo, et~al.
\newblock Improving image generation with better captions.
\newblock \emph{Computer Science. https://cdn. openai. com/papers/dall-e-3. pdf}, 2\penalty0 (3):\penalty0 8, 2023.

\bibitem[Changpinyo et~al.(2021)Changpinyo, Sharma, Ding, and Soricut]{datasetcc12m}
Soravit Changpinyo, Piyush Sharma, Nan Ding, and Radu Soricut.
\newblock {Conceptual 12M}: Pushing web-scale image-text pre-training to recognize long-tail visual concepts.
\newblock In \emph{CVPR}, 2021.

\bibitem[Chen et~al.(2023)Chen, Yu, Ge, Yao, Xie, Wu, Wang, Kwok, Luo, Lu, et~al.]{pixart}
Junsong Chen, Jincheng Yu, Chongjian Ge, Lewei Yao, Enze Xie, Yue Wu, Zhongdao Wang, James Kwok, Ping Luo, Huchuan Lu, et~al.
\newblock Pixart-$\alpha$: Fast training of diffusion transformer for photorealistic text-to-image synthesis.
\newblock \emph{arXiv preprint arXiv:2310.00426}, 2023.

\bibitem[Cho et~al.(2024)Cho, Hu, Baldridge, Garg, Anderson, Krishna, Bansal, Pont-Tuset, and Wang]{dsg}
Jaemin Cho, Yushi Hu, Jason~Michael Baldridge, Roopal Garg, Peter Anderson, Ranjay Krishna, Mohit Bansal, Jordi Pont-Tuset, and Su~Wang.
\newblock Davidsonian scene graph: Improving reliability in fine-grained evaluation for text-to-image generation.
\newblock In \emph{The Twelfth International Conference on Learning Representations}, 2024.

\bibitem[Dall'Asen et~al.(2025)Dall'Asen, Zhang, Hemmat, Hall, Verbeek, Romero-Soriano, and Drozdzal]{dallasen2025}
Nicola Dall'Asen, Xiaofeng Zhang, Reyhane~Askari Hemmat, Melissa Hall, Jakob Verbeek, Adriana Romero-Soriano, and Michal Drozdzal.
\newblock Increasing the utility of synthetic images through chamfer guidance.
\newblock \emph{arXiv preprint arXiv:2508.10631}, 2025.

\bibitem[Datta et~al.(2024)Datta, Ku, Ramachandran, and Anderson]{promptexpansion}
Siddhartha Datta, Alexander Ku, Deepak Ramachandran, and Peter Anderson.
\newblock Prompt expansion for adaptive text-to-image generation.
\newblock In \emph{Proceedings of the 62nd Annual Meeting of the Association for Computational Linguistics (Volume 1: Long Papers)}, August 2024.

\bibitem[Deng et~al.(2009)Deng, Dong, Socher, Li, Li, and Fei-Fei]{imagenet}
Jia Deng, Wei Dong, Richard Socher, Li-Jia Li, Kai Li, and Li~Fei-Fei.
\newblock Imagenet: A large-scale hierarchical image database.
\newblock In \emph{2009 IEEE conference on computer vision and pattern recognition}, pages 248--255. Ieee, 2009.

\bibitem[DeVries et~al.(2019)DeVries, Romero, Pineda, Taylor, and Drozdzal]{devries2019evaluation}
Terrance DeVries, Adriana Romero, Luis Pineda, Graham~W Taylor, and Michal Drozdzal.
\newblock On the evaluation of conditional gans.
\newblock \emph{arXiv preprint arXiv:1907.08175}, 2019.

\bibitem[discus0434 and Goswami(2023)]{aesthetic_v2_5}
discus0434 and Sayan Goswami.
\newblock aesthetic-predictor-v2-5: Siglip-based aesthetic score predictor.
\newblock \url{https://github.com/discus0434/aesthetic-predictor-v2-5}, 2023.

\bibitem[Du et~al.(2023)Du, Durkan, Strudel, Tenenbaum, Dieleman, Fergus, Sohl-Dickstein, Doucet, and Grathwohl]{composition_reducereuserecycle}
Yilun Du, Conor Durkan, Robin Strudel, Joshua~B Tenenbaum, Sander Dieleman, Rob Fergus, Jascha Sohl-Dickstein, Arnaud Doucet, and Will~Sussman Grathwohl.
\newblock Reduce, reuse, recycle: Compositional generation with energy-based diffusion models and mcmc.
\newblock In \emph{International conference on machine learning}, pages 8489--8510. PMLR, 2023.

\bibitem[Esser et~al.(2024)Esser, Kulal, Blattmann, Entezari, M{\"u}ller, Saini, Levi, Lorenz, Sauer, Boesel, et~al.]{sd35model}
Patrick Esser, Sumith Kulal, Andreas Blattmann, Rahim Entezari, Jonas M{\"u}ller, Harry Saini, Yam Levi, Dominik Lorenz, Axel Sauer, Frederic Boesel, et~al.
\newblock Scaling rectified flow transformers for high-resolution image synthesis.
\newblock In \emph{Forty-first international conference on machine learning}, 2024.

\bibitem[Fan et~al.(2024)Fan, Chen, Krishnan, Katabi, Isola, and Tian]{fan2024scaling}
Lijie Fan, Kaifeng Chen, Dilip Krishnan, Dina Katabi, Phillip Isola, and Yonglong Tian.
\newblock Scaling laws of synthetic images for model training... for now.
\newblock In \emph{Proceedings of the IEEE/CVF Conference on Computer Vision and Pattern Recognition}, pages 7382--7392, 2024.

\bibitem[Fellbaum(1998)]{wordnet}
Christiane Fellbaum.
\newblock \emph{WordNet: An electronic lexical database}.
\newblock MIT press, 1998.

\bibitem[Friedman and Dieng(2023)]{vendiscore}
Dan Friedman and Adji~Bousso Dieng.
\newblock The vendi score: A diversity evaluation metric for machine learning.
\newblock \emph{Transactions on Machine Learning Research}, 2023.
\newblock ISSN 2835-8856.

\bibitem[Hall et~al.(2024{\natexlab{a}})Hall, Bell, Ross, Williams, Drozdzal, and Soriano]{hall2024towards}
Melissa Hall, Samuel~J Bell, Candace Ross, Adina Williams, Michal Drozdzal, and Adriana~Romero Soriano.
\newblock Towards geographic inclusion in the evaluation of text-to-image models.
\newblock In \emph{Proceedings of the 2024 ACM Conference on Fairness, Accountability, and Transparency}, pages 585--601, 2024{\natexlab{a}}.

\bibitem[Hall et~al.(2024{\natexlab{b}})Hall, Mañas, Askari, Ibrahim, Ross, Astolfi, Ifriqi, Havasi, Benchetrit, Ullrich, Braga, Charnalia, Ryan, Rabbat, Drozdzal, Verbeek, and Soriano]{libevalgim}
Melissa Hall, Oscar Mañas, Reyhane Askari, Mark Ibrahim, Candace Ross, Pietro Astolfi, Tariq~Berrada Ifriqi, Marton Havasi, Yohann Benchetrit, Karen Ullrich, Carolina Braga, Abhishek Charnalia, Maeve Ryan, Mike Rabbat, Michal Drozdzal, Jakob Verbeek, and Adriana~Romero Soriano.
\newblock Evalgim: A library for evaluating generative image models, 2024{\natexlab{b}}.
\newblock \url{https://arxiv.org/abs/2412.10604}.

\bibitem[Hall et~al.(2024{\natexlab{c}})Hall, Ross, Williams, Carion, Drozdzal, and Romero-Soriano]{digin}
Melissa Hall, Candace Ross, Adina Williams, Nicolas Carion, Michal Drozdzal, and Adriana Romero-Soriano.
\newblock Dig in: Evaluating disparities in image generations with indicators for geographic diversity.
\newblock \emph{Transactions on Machine Learning Research}, 2024{\natexlab{c}}.

\bibitem[Han et~al.(2025)Han, Liu, Jiang, Yan, Zhang, Yuan, Peng, and Liu]{infinityautoregressive}
Jian Han, Jinlai Liu, Yi~Jiang, Bin Yan, Yuqi Zhang, Zehuan Yuan, Bingyue Peng, and Xiaobing Liu.
\newblock Infinity: Scaling bitwise autoregressive modeling for high-resolution image synthesis.
\newblock In \emph{Proceedings of the Computer Vision and Pattern Recognition Conference}, pages 15733--15744, 2025.

\bibitem[Hemmat et~al.(2024)Hemmat, Pezeshki, Bordes, Drozdzal, and Romero-Soriano]{feedbackguidance}
Reyhane~Askari Hemmat, Mohammad Pezeshki, Florian Bordes, Michal Drozdzal, and Adriana Romero-Soriano.
\newblock Feedback-guided data synthesis for imbalanced classification.
\newblock \emph{Transactions on Machine Learning Research}, 2024.
\newblock ISSN 2835-8856.
\newblock \url{https://openreview.net/forum?id=IHJ5OohGwr}.

\bibitem[Hessel et~al.(2021)Hessel, Holtzman, Forbes, Le~Bras, and Choi]{clipscore}
Jack Hessel, Ari Holtzman, Maxwell Forbes, Ronan Le~Bras, and Yejin Choi.
\newblock {CLIPS}core: A reference-free evaluation metric for image captioning.
\newblock In \emph{Proceedings of the 2021 Conference on Empirical Methods in Natural Language Processing}, pages 7514--7528, Online and Punta Cana, Dominican Republic, November 2021. Association for Computational Linguistics.

\bibitem[Heusel et~al.(2017)Heusel, Ramsauer, Unterthiner, Nessler, and Hochreiter]{fidmetric}
Martin Heusel, Hubert Ramsauer, Thomas Unterthiner, Bernhard Nessler, and Sepp Hochreiter.
\newblock Gans trained by a two time-scale update rule converge to a local nash equilibrium.
\newblock \emph{Advances in neural information processing systems}, 30, 2017.

\bibitem[Ho and Salimans(2022)]{cfg}
Jonathan Ho and Tim Salimans.
\newblock Classifier-free diffusion guidance.
\newblock In \emph{NeurIPS 2021 Workshop on Deep Generative Models and Downstream Applications}, 2022.

\bibitem[Ho et~al.(2020)Ho, Jain, and Abbeel]{ddpm}
Jonathan Ho, Ajay Jain, and Pieter Abbeel.
\newblock Denoising diffusion probabilistic models.
\newblock \emph{Advances in neural information processing systems}, 33:\penalty0 6840--6851, 2020.

\bibitem[Hu et~al.(2024)Hu, Wang, Fang, Fu, Cheng, and Yu]{dpgbench}
Xiwei Hu, Rui Wang, Yixiao Fang, Bin Fu, Pei Cheng, and Gang Yu.
\newblock Ella: Equip diffusion models with llm for enhanced semantic alignment.
\newblock \emph{arXiv preprint arXiv:2403.05135}, 2024.

\bibitem[Hu et~al.(2023)Hu, Liu, Kasai, Wang, Ostendorf, Krishna, and Smith]{tifa}
Yushi Hu, Benlin Liu, Jungo Kasai, Yizhong Wang, Mari Ostendorf, Ranjay Krishna, and Noah~A Smith.
\newblock Tifa: Accurate and interpretable text-to-image faithfulness evaluation with question answering.
\newblock In \emph{Proceedings of the IEEE/CVF International Conference on Computer Vision}, pages 20406--20417, 2023.

\bibitem[Karras et~al.(2024)Karras, Aittala, Kynk{\"a}{\"a}nniemi, Lehtinen, Aila, and Laine]{bvg}
Tero Karras, Miika Aittala, Tuomas Kynk{\"a}{\"a}nniemi, Jaakko Lehtinen, Timo Aila, and Samuli Laine.
\newblock Guiding a diffusion model with a bad version of itself.
\newblock \emph{Advances in Neural Information Processing Systems}, 37:\penalty0 52996--53021, 2024.

\bibitem[Kynk{\"a}{\"a}nniemi et~al.(2019)Kynk{\"a}{\"a}nniemi, Karras, Laine, Lehtinen, and Aila]{prmetric}
Tuomas Kynk{\"a}{\"a}nniemi, Tero Karras, Samuli Laine, Jaakko Lehtinen, and Timo Aila.
\newblock Improved precision and recall metric for assessing generative models.
\newblock \emph{Advances in neural information processing systems}, 32, 2019.

\bibitem[Kynk{\"a}{\"a}nniemi et~al.(2024)Kynk{\"a}{\"a}nniemi, Aittala, Karras, Laine, Aila, and Lehtinen]{intervalguide}
Tuomas Kynk{\"a}{\"a}nniemi, Miika Aittala, Tero Karras, Samuli Laine, Timo Aila, and Jaakko Lehtinen.
\newblock Applying guidance in a limited interval improves sample and distribution quality in diffusion models.
\newblock In \emph{The Thirty-eighth Annual Conference on Neural Information Processing Systems}, 2024.

\bibitem[Labs(2024)]{flux2024}
Black~Forest Labs.
\newblock Flux.
\newblock \url{https://github.com/black-forest-labs/flux}, 2024.

\bibitem[Lee et~al.(2023)Lee, Yasunaga, Meng, Mai, Park, Gupta, Zhang, Narayanan, Teufel, Bellagente, et~al.]{lee2023holistic}
Tony Lee, Michihiro Yasunaga, Chenlin Meng, Yifan Mai, Joon~Sung Park, Agrim Gupta, Yunzhi Zhang, Deepak Narayanan, Hannah Teufel, Marco Bellagente, et~al.
\newblock Holistic evaluation of text-to-image models.
\newblock \emph{Advances in Neural Information Processing Systems}, 36:\penalty0 69981--70011, 2023.

\bibitem[Lin et~al.(2024)Lin, Pathak, Li, Li, Xia, Neubig, Zhang, and Ramanan]{vqascore}
Zhiqiu Lin, Deepak Pathak, Baiqi Li, Jiayao Li, Xide Xia, Graham Neubig, Pengchuan Zhang, and Deva Ramanan.
\newblock Evaluating text-to-visual generation with image-to-text generation.
\newblock \emph{arXiv preprint arXiv:2404.01291}, 2024.

\bibitem[Liu et~al.(2022)Liu, Li, Du, Torralba, and Tenenbaum]{compositional_diffusion}
Nan Liu, Shuang Li, Yilun Du, Antonio Torralba, and Joshua~B Tenenbaum.
\newblock Compositional visual generation with composable diffusion models.
\newblock In \emph{European conference on computer vision}, pages 423--439. Springer, 2022.

\bibitem[Liu et~al.(2023)Liu, Gong, and Liu]{rectifiedflow}
Xingchao Liu, Chengyue Gong, and Qiang Liu.
\newblock Flow straight and fast: Learning to generate and transfer data with rectified flow.
\newblock In \emph{The Eleventh International Conference on Learning Representations}, 2023.

\bibitem[Naeem et~al.(2020)Naeem, Oh, Uh, Choi, and Yoo]{dcmetric}
Muhammad~Ferjad Naeem, Seong~Joon Oh, Youngjung Uh, Yunjey Choi, and Jaejun Yoo.
\newblock Reliable fidelity and diversity metrics for generative models.
\newblock In \emph{International conference on machine learning}, pages 7176--7185. PMLR, 2020.

\bibitem[Oquab et~al.(2023)Oquab, Darcet, Moutakanni, Vo, Szafraniec, Khalidov, Fernandez, Haziza, Massa, El-Nouby, et~al.]{dinov2}
Maxime Oquab, Timoth{\'e}e Darcet, Th{\'e}o Moutakanni, Huy Vo, Marc Szafraniec, Vasil Khalidov, Pierre Fernandez, Daniel Haziza, Francisco Massa, Alaaeldin El-Nouby, et~al.
\newblock Dinov2: Learning robust visual features without supervision.
\newblock \emph{arXiv preprint arXiv:2304.07193}, 2023.

\bibitem[Podell et~al.(2024)Podell, English, Lacey, Blattmann, Dockhorn, M{\"u}ller, Penna, and Rombach]{sdxl}
Dustin Podell, Zion English, Kyle Lacey, Andreas Blattmann, Tim Dockhorn, Jonas M{\"u}ller, Joe Penna, and Robin Rombach.
\newblock {SDXL}: Improving latent diffusion models for high-resolution image synthesis.
\newblock In \emph{The Twelfth International Conference on Learning Representations}, 2024.

\bibitem[Radford et~al.(2021)Radford, Kim, Hallacy, Ramesh, Goh, Agarwal, Sastry, Askell, Mishkin, Clark, et~al.]{radford2021learning}
Alec Radford, Jong~Wook Kim, Chris Hallacy, Aditya Ramesh, Gabriel Goh, Sandhini Agarwal, Girish Sastry, Amanda Askell, Pamela Mishkin, Jack Clark, et~al.
\newblock Learning transferable visual models from natural language supervision.
\newblock In \emph{International conference on machine learning}, pages 8748--8763. PmLR, 2021.

\bibitem[Rombach et~al.(2022)Rombach, Blattmann, Lorenz, Esser, and Ommer]{ldmrombach}
Robin Rombach, Andreas Blattmann, Dominik Lorenz, Patrick Esser, and Bj{\"o}rn Ommer.
\newblock High-resolution image synthesis with latent diffusion models.
\newblock In \emph{Proceedings of the IEEE/CVF conference on computer vision and pattern recognition}, pages 10684--10695, 2022.

\bibitem[Ronneberger et~al.(2015)Ronneberger, Fischer, and Brox]{unet}
Olaf Ronneberger, Philipp Fischer, and Thomas Brox.
\newblock U-net: Convolutional networks for biomedical image segmentation.
\newblock In \emph{International Conference on Medical image computing and computer-assisted intervention}, pages 234--241. Springer, 2015.

\bibitem[Sadat et~al.(2024)Sadat, Buhmann, Bradley, Hilliges, and Weber]{cads}
Seyedmorteza Sadat, Jakob Buhmann, Derek Bradley, Otmar Hilliges, and Romann~M. Weber.
\newblock {CADS}: Unleashing the diversity of diffusion models through condition-annealed sampling.
\newblock In \emph{The Twelfth International Conference on Learning Representations}, 2024.
\newblock \url{https://openreview.net/forum?id=zMoNrajk2X}.

\bibitem[Sadat et~al.(2025)Sadat, Hilliges, and Weber]{apg}
Seyedmorteza Sadat, Otmar Hilliges, and Romann~M Weber.
\newblock Eliminating oversaturation and artifacts of high guidance scales in diffusion models.
\newblock In \emph{The Thirteenth International Conference on Learning Representations}, 2025.

\bibitem[Schuhmann(2022)]{aesthetic}
Christoph Schuhmann.
\newblock Improved-aesthetic-predictor: Clip+mlp aesthetic score predictor.
\newblock \url{https://github.com/christophschuhmann/improved-aesthetic-predictor}, 2022.

\bibitem[Stein et~al.(2023)Stein, Cresswell, Hosseinzadeh, Sui, Ross, Villecroze, Liu, Caterini, Taylor, and Loaiza-Ganem]{fddmetric}
George Stein, Jesse Cresswell, Rasa Hosseinzadeh, Yi~Sui, Brendan Ross, Valentin Villecroze, Zhaoyan Liu, Anthony~L Caterini, Eric Taylor, and Gabriel Loaiza-Ganem.
\newblock Exposing flaws of generative model evaluation metrics and their unfair treatment of diffusion models.
\newblock \emph{Advances in Neural Information Processing Systems}, 36:\penalty0 3732--3784, 2023.

\bibitem[Tang et~al.(2024)Tang, Zhang, Xu, Lu, Li, Stenetorp, Lin, and T{\"u}re]{w1kp}
Raphael Tang, Crystina Zhang, Lixinyu Xu, Yao Lu, Wenyan Li, Pontus Stenetorp, Jimmy Lin, and Ferhan T{\"u}re.
\newblock Words worth a thousand pictures: Measuring and understanding perceptual variability in text-to-image generation.
\newblock In \emph{Proceedings of the 2024 Conference on Empirical Methods in Natural Language Processing}, pages 5441--5454, 2024.

\bibitem[Team et~al.(2025)Team, Kamath, Ferret, Pathak, Vieillard, Merhej, Perrin, Matejovicova, Ram{\'e}, Rivi{\`e}re, et~al.]{gemma3}
Gemma Team, Aishwarya Kamath, Johan Ferret, Shreya Pathak, Nino Vieillard, Ramona Merhej, Sarah Perrin, Tatiana Matejovicova, Alexandre Ram{\'e}, Morgane Rivi{\`e}re, et~al.
\newblock Gemma 3 technical report.
\newblock \emph{arXiv preprint arXiv:2503.19786}, 2025.

\bibitem[Tian et~al.(2023)Tian, Fan, Isola, Chang, and Krishnan]{tian2023stablerep}
Yonglong Tian, Lijie Fan, Phillip Isola, Huiwen Chang, and Dilip Krishnan.
\newblock Stablerep: Synthetic images from text-to-image models make strong visual representation learners.
\newblock \emph{Advances in Neural Information Processing Systems}, 36:\penalty0 48382--48402, 2023.

\bibitem[Urbanek et~al.(2024)Urbanek, Bordes, Astolfi, Williamson, Sharma, and Romero-Soriano]{dcidataset}
Jack Urbanek, Florian Bordes, Pietro Astolfi, Mary Williamson, Vasu Sharma, and Adriana Romero-Soriano.
\newblock A picture is worth more than 77 text tokens: Evaluating clip-style models on dense captions.
\newblock In \emph{Proceedings of the IEEE/CVF Conference on Computer Vision and Pattern Recognition (CVPR)}, pages 26700--26709, June 2024.

\bibitem[Wu et~al.(2025)Wu, Li, Zhou, Lin, Gao, Yan, Yin, Bai, Xu, Chen, et~al.]{qwenimage}
Chenfei Wu, Jiahao Li, Jingren Zhou, Junyang Lin, Kaiyuan Gao, Kun Yan, Sheng-ming Yin, Shuai Bai, Xiao Xu, Yilei Chen, et~al.
\newblock Qwen-image technical report.
\newblock \emph{arXiv preprint arXiv:2508.02324}, 2025.

\bibitem[Yoon et~al.(2024)Yoon, Hu, Weissburg, Qin, and Jeong]{yoon2024model}
Youngseok Yoon, Dainong Hu, Iain Weissburg, Yao Qin, and Haewon Jeong.
\newblock Model collapse in the self-consuming chain of diffusion finetuning: A novel perspective from quantitative trait modeling.
\newblock \emph{arXiv preprint arXiv:2407.17493}, 2024.

\bibitem[Zhai et~al.(2023)Zhai, Mustafa, Kolesnikov, and Beyer]{siglip}
Xiaohua Zhai, Basil Mustafa, Alexander Kolesnikov, and Lucas Beyer.
\newblock Sigmoid loss for language image pre-training.
\newblock In \emph{Proceedings of the IEEE/CVF international conference on computer vision}, pages 11975--11986, 2023.

\bibitem[Zhang et~al.(2024)Zhang, Zhang, Dong, Zang, and Wang]{longclip}
Beichen Zhang, Pan Zhang, Xiaoyi Dong, Yuhang Zang, and Jiaqi Wang.
\newblock Long-clip: Unlocking the long-text capability of clip.
\newblock In \emph{European Conference on Computer Vision}, pages 310--325. Springer, 2024.

\end{thebibliography}

\clearpage
\newpage
\beginappendix

\appendix

\renewcommand \thepart{}
\renewcommand \partname{}

\startcontents[appendix]

\printcontents[appendix]{l}{1}

\clearpage

\section{Details on the synthetic experiments}
\label{app:synthetic}
\subsection{Experimental settings}
\label{app:synthetic_exp_setting}
We use a mixture of four Gaussians as our training distribution. Each Gaussian is in each quadrant respectively. The means for each Gaussian is [-3, 3], [-3, -3], [3, 3], and [3, -3]. The covariance for each Gaussian is [[0.7,0],[0,0.7]]. We hypothetically link each Gaussian to a conditional prompt, \ie \texttt{white cat} for the first quadrant, \texttt{white dog} for the second quadrant, \texttt{black dog} for the third quadrant and \texttt{black cat} for the fourth quadrant.

We train conditional U-Net-based diffusion model with DDPM schedule using two types of conditional prompts: one is fine-grained with \texttt{white cat}, \texttt{white dog}, \texttt{black cat}, \texttt{black dog}, each corresponds to a Gaussian in one quadrant; the other is general with \texttt{cat}, \texttt{dog}, \texttt{white}, and \texttt{black}, each corresponds to two Gaussians.

Following~\citet{ldmrombach,sd35model}, we add a pooled conditional information together with the timestep embedding, and also incorporate the token-wise conditional information into the U-Net with an Attention mechanism. To get the pooled conditional information, we use a linear layer, while to get the token-wise embedding, we initialize a vocabulary dictionary from scratch and train it together with the diffusion model. We train the model with batch size 512, Adam optimizer, learning rate 1e-4 with linear warmup and exponential decay ($\gamma=0.99$), training epochs 250 in total. All training and inference are done using single NVIDIA V100 32GB GPU.

During training, we adopt a 50\% probability of dropping the conditional information, enabling the classifier-free guidance (CFG)~\citep{cfg} during inference. We use the ancestral sampler in DDPM~\citep{ddpm} with 1000 steps for inference. We conduct the inference in a generalization setting, \ie if the model is trained with general prompts, the inference is done with fine-grained prompts, and vice versa. We use forward Kullback-Leibler divergence and Fr\'{e}chet Distance to evaluate the generated samples. We sample 10,000 data points for evaluation.

\subsection{Derivations}
\label{app:synthetic_proof}
\textbf{From general conditions to fine-grained conditions. \ \ } Given a conditional diffusion model with parameters $\theta$ trained on general conditions, timestep $t \in \{1,2,...,T\}$, independent general conditions $c_{\text{g}}^i, i \in \mathbb{N}$, we have access to their score functions $s_\theta(x_t|c_{\text{g}}^i) = \nabla_{x_t} \log p_{\theta}(x_t|c_{\text{g}}^i) $. Supposing that a fine-grained condition $c_{\text{f}}$ comprises of several general conditions $c_{\text{g}}^i, i\in \{1, 2, ..., K\}$, we have
\begin{align}
    p_{\theta}(x_t| c_{\text{f}}) &= p_{\theta}(x_t| \cap_{i \in \{1,2,...,K\}} c_{\text{g}}^i) \\ 
    &= \frac{p_{\theta}(x_t, c_{\text{g}}^1, c_{\text{g}}^2, ..., c_{\text{g}}^K)}{p_{\theta}(c_{\text{g}}^1, c_{\text{g}}^2, ..., c_{\text{g}}^K)} \\
    &= \frac{p_{\theta}(x_t) \prod_{i \in \{1,2,...,K\}} p_{\theta}(c_{\text{g}}^i|x_t)}{\prod_{i \in \{1,2,...,K\}} p_{\theta}(c_{\text{g}}^i)} \\
    &= \frac{p_{\theta}(x_t) \prod_{i \in \{1,2,...,K\}} \frac{p_{\theta}(x_t|c_{\text{g}}^i)p_{\theta}(c_{\text{g}}^i)}{p_{\theta}(x_t)}}{\prod_{i \in \{1,2,...,K\}} p_{\theta}(c_{\text{g}}^i)} \\
    &= p_{\theta}(x_t) \prod_{i \in \{1,2,...,K\}} \frac{p_{\theta}(x_t|c_{\text{g}}^i)}{p_{\theta}(x_t)}
\end{align}
\begin{align}
    s_{\theta}(x_t|c_{\text{f}}) &= \nabla_{x_t} \log p_{\theta}(x_t| c_{\text{f}}) \\
    &= \nabla_{x_t} \log p_{\theta}(x_t) + \nabla_{x_t} \sum_{i\in \{1,2,...,K\}} \left(\log p_{\theta}(x_t|c_{\text{g}}^i) - \log p_{\theta}(x_t) \right) \\
    &= s_{\theta} (x_t) + \sum_{i \in \{1,2,...,K\}} (s_{\theta}(x_t|c_{\text{g}}^i) - s_{\theta}(x_t))
\end{align}

\textbf{From fine-grained conditions to general conditions. \ \ } Given a conditional diffusion model with parameters $\theta$ trained on fine-grained conditions, timestep $t \in \{1,2,...,T\}$, independent fine-grained conditions $c_{\text{f}}^i, i \in \mathbb{N}$, we have access to their score functions $s_{\theta}(x_t|c_{\text{f}}^i) = \nabla_{x_t} \log p_{\theta}(x_t|c_{\text{f}}^i)$. Supposing that a general condition $c_{\text{g}}$ can be decomposed into different fine-grained conditions $c_{\text{f}}^i, i \in \{1,2,...,M\}$, we have
\begin{align}
    p_{\theta}(x_t|c_{\text{g}}) = \sum_{i \in \{1,2,...,M\}}p_{\theta}(x_t|c_{\text{f}}^i)
\end{align}
\begin{align}
    s_{\theta}(x_t|c_{\text{g}}) &= \nabla_{x_t} \log \sum_{i \in \{1,2,...,M\}}p_{\theta}(x_t|c_{\text{f}}^i) \\
    &= \frac{1}{\sum_{j \in \{1,2,...,M\}}p_{\theta}(x_t|c_{\text{f}}^j)} \nabla_{x_t} \left(\sum_{i \in \{1,2,...,M\}}p_{\theta}(x_t|c_{\text{f}}^i) \right) \\
    &= \frac{1}{\sum_{j \in \{1,2,...,M\}}p_{\theta}(x_t|c_{\text{f}}^j)} \sum_{i\in \{1,2,...,M\}} \nabla_{x_t} p_{\theta}(x_t|c_{\text{f}}^i) \\
    &= \frac{1}{\sum_{j \in \{1,2,...,M\}}p_{\theta}(x_t|c_{\text{f}}^j)} \sum_{i\in\{1,2,...,M\}} \left(p_{\theta}(x_t|c_{\text{f}}^i) \nabla_{x_t} \log p_{\theta}(x_t|c_{\text{f}}^i)\right) \\
    &= \sum_{i\in\{1,2,...,M\}} \left(\frac{p_{\theta}(x_t|c_{\text{f}}^i)}{{\sum_{j \in \{1,2,...,M\}}p_{\theta}(x_t|c_{\text{f}}^j)}} s_{\theta}(x_t|c_{\text{f}}^i)\right)
\end{align}

\subsection{Relationship with the real T2I settings}
\label{app:synthetic_relation}
\xf{We use the synthetic experiment as a motivation to emphasize the importance of studying the axis of prompt complexity in the diffusion / flow model setting. The synthetic experiments clearly show an asymmetry of difficulty w.r.t. conditioning complexity when the probabilistic assumptions hold.}

\xf{The conditional independence assumption in synthetic setting derivation is a simplification. Real-world text encoders (\eg CLIP/T5) produce entangled representations where concepts (\eg “yellow” and “banana”) can be highly correlated and content dependent. However, we believe this distinction does not invalidate our conclusion regarding the asymmetry of difficulty between generalizing to general ("OR") vs. specific ("AND") prompts.}

\subsubsection{On how this affects the “OR is harder than AND” statement}

\xf{\textbf{AND Operator} (Generalizing to longer prompts):}

\xf{In our toy example (Ideal Case): The “black” and “dog” concepts are independent. Their guidance vectors are orthogonal. Our Equation (2) is a probabilistically sound formula, and the naive sum works perfectly, as shown by the KL/FD scores (KL divergence of 0.93 and Frechet distance of 1.32).}

\xf{The violation of the assumption doesn't make this operator "harder" in the sense of being impractical, but degrades it. It changes Equation (2) from a probabilistic law into a practical but imperfect heuristic, probably leading to some failure modes. For example, the guidance vectors for “white” and “dog” are non-orthogonal if they are correlated as captured by the CLIP / T5 model. Our Equation (2), by naively adding these vectors, leads to an over-magnification of this shared, correlated signal, potentially leading to some artifacts (low diversity, over-saturation, etc.) in the generation.}

\xf{\textbf{OR Operator} (Generalizing to shorter prompts):}

\xf{This is already hard in theory as we explain in Eq. 1 in our manuscript, where the score function for "OR" operator is intractable in diffusion models. The fact that tokens in real-world prompts are not perfectly “mutually exclusive” adds another layer of intractability, making this “OR” generalization even more difficult.}

\xf{Given the theoretical analysis and not over-interpret our empirical results, some of our empirical results also verifies this asymmetry. For aesthetic quality with CC12M and DCI settings (Fig. 3, (a) and (c)), we observe a sharper decrease towards shorter prompts compared to longer prompts, which empirically supports our statement.}

\subsubsection{On explaining CC12M vs. ImageNet behaviors}

\xf{Our theoretical analysis also helps explain some empirical divergences in results of CC12M and ImageNet. CC12M relies on composition of tokens (similar to the “AND” and “OR” operator logic), where we see similar trends as in the synthetic experiments. ImageNet experiments rely on specificity (using semantically richer tokens rather than more tokens) and the ImageNet hierarchy does not strictly follow the combinatorial logic of these mathematical derivations.}

\section{Details on the benchmarking framework}
\label{app:framework}
\subsection{Captioning step}
\label{app:captioning}
We use Gemma3~\citep{gemma3} model (\texttt{gemma-3-12b-it} version) to conduct the captioning of images from datasets with general captions (\eg CC12M~\citep{datasetcc12m}). The system prompt is: \texttt{You strictly follow the user's instructions.} The user prompts used for captioning to different complexities are as follows:
\begin{enumerate}
    \item \texttt{Mention the main subjects in the image with only 1 word. Do not use specific identifiers, such as names. Format the caption as such: <<CAPTION>>.}
    \item \texttt{Write a short caption with 3 or less words. Use only one adjective. Do not use specific identifiers, such as names. Format the caption as such: <<CAPTION>>.}
    \item \texttt{Write a caption with 6 or less words, describing at most two main foreground subjects. Do not use specific identifiers, such as names. Format the caption as such: <<CAPTION>>.}
    \item \texttt{Write a caption with 8 or less words, describing at most two main foreground subjects, as well as the background. Do not use specific identifiers, such as names. Format the caption as such: <<CAPTION>>.}
\end{enumerate}

\subsection{Pairing step}
\label{app:paring}
For datasets with general captions, we use the SigLip model~\citep{siglip} to embed both images and texts. Our implementation uses the OpenClip~\footnote{\url{https://github.com/mlfoundations/open_clip}} library and the SigLip version \texttt{ViT-SO400M-14-SigLIP-384}. %

\subsection{Alignment step}
\label{app:align}
We present the alignment step in Algorithm~\ref{algo:alignment}.

\begin{algorithm}
    \caption{Alignment step}
    \begin{algorithmic}[1]
        \State \textbf{Input:} $\bar{\mathcal{I}}^{ik}, k \in \{1,2, ..., K\}, i \in \mathcal{N}_k^{\rm{p}}$; $N_{\rm{gen}}$.
        \State breaksignal $\gets \texttt{False}$
        \While{\textit{not} breaksignal}
            \State CommonImagesBefore $\leftarrow$ $\bigcap_{k \in \{1,2,...,K\}} \bigcup_{i \in \mathcal{N}_k^{\rm{p}}} \bar{\mathcal{I}}^{ik}$
            \For{$j \in \{1,2,\dots,K\}$}
                \For{$l \in \mathcal{N}^{\rm{p}}_j$}
                    \State $\bar{\mathcal{I}}^{lj}$ $\leftarrow$ $\bar{\mathcal{I}}^{lj} \cap \rm{CommonImagesBefore}$
                    \If{$| \bar{\mathcal{I}}^{lj} |< N_{\rm{gen}}$}
                        \State {\textbf{del}} $l$ from $\mathcal{N}_j^{\rm{p}}$
                    \EndIf
                \EndFor
            \EndFor
            \State CommonImagesAfter $\leftarrow$ $\bigcap_{k \in \{1,2,...,K\}} \bigcup_{i \in \mathcal{N}_k^{\rm{p}}} \bar{\mathcal{I}}^{ik}$
            \State breaksignal $\leftarrow$ \rm{len}(CommonImagesBefore) == \rm{len}(CommonImagesAfter)
        \EndWhile
        \State \Return $\bar{\mathcal{I}}^{ik}, k \in \{1,2, ..., K\}, i \in \mathcal{N}_k^{\rm{p}}$
    \end{algorithmic}
    \label{algo:alignment}
\end{algorithm}

\subsection{Evaluation metrics}
To compute evaluation metrics, we use the following libraries:
\begin{enumerate}
    \item Reference-based metrics: \texttt{dgm-eval}~\citep{fddmetric}
    \item Vendi score: \texttt{Vendi-Score}~\citep{vendiscore}
    \item Aesthetic score: \texttt{aesthetic-predictor-v2.5}~\citep{aesthetic_v2_5}
    \item DSG score: \texttt{Eval-GIM}~\citep{libevalgim}
\end{enumerate}

\section{Experimental details}
\label{app:exp_details}
\subsection{Dataset details}
\label{app:dataset_details}
\textit{CC12M.} We investigate the effect of increasing prompt detail and length on the utility axes of synthetic data. Following the procedure described in Section~\ref{sec:pipeline}, we generate captions for randomly sampled one million images at four different complexity levels, each corresponding to a distinct complexity level. Lower complexity prompts are shorter and contain fewer details about the image. We follow the procedure in Section~\ref{sec:pipeline} and sample 5,000 captions per complexity for generation. As reference, many widely used evaluation prompt sets contain approximately 1,000 prompts~\citep{dsg,dpgbench}.
We produce 20 images per prompt, resulting in 100,000 generated images for each prompt complexity.
The statistics for different caption complexities are presented in Table~\ref{tab:stats_prompt_complexity}. We use \texttt{c\#} to represent complexity level of the prompts. We report the caption lengths statistics across complexities, and the number of real images retrieved using the 5,000 selected captions per complexity.
\begin{table}[ht]
\caption{Statistics on word lengths of different prompt complexities}
\label{tab:stats_prompt_complexity}
\centering
\begin{tabular}{@{}lllll@{}}
\toprule
               & \texttt{c1}    & \texttt{c2}  & \texttt{c3}  & \texttt{c4} %
\\ \midrule
Avg \# word    & 1.0007         & 2.1539       & 4.9094       & 7.5410       %
\\
Std \# word    & 0.0501         & 0.3621       & 0.7032       & 0.8849       %
\\
Median \# word & 1.0000         & 2.0000       & 5.0000       & 8.0000       %
\\ 
\# real images covered & 61,334  & 57,925       & 55,075      &46,066    \\  \bottomrule
\end{tabular}
\end{table}

\textit{ImageNet-1k.} We examine the effect of concept specificity on the utility axes of synthetic data. We follow the procedure in Section~\ref{sec:pipeline} and extract hyponym relations to vary the specificity of ImageNet-1k class labels. For example, the class \texttt{siberian husky} (highest specificity) has the parent class \texttt{sled dog} and grandparent class \texttt{working dog} (lowest specificity). Since these class labels possess intrinsic semantic complexity, we use class-label specificity as a proxy for complexity. To generate images, we use the prompt ``Image of a \texttt{<CLASS>}''. We utilize all 1,000 ImageNet-1k classes and generate 50 images per class, resulting in 50,000 images, matching the size of the ImageNet-1k evaluation set.

\textit{DCI.} We further explore prompt complexity using \emph{very long} prompts. DCI's detailed captions consist of multiple sentences, densely describing various aspects of each image. We construct prompts as described in follow the procedure in Section~\ref{sec:pipeline}, \ie progressively concatenating sentences to each caption. We cap the maximum prompt length at 100 words. We sample 1,000 detailed captions and generate 5,398 prompts of varying complexity. For each prompt, we generate 50 images. 

\subsection{Guidance methods' hyperparameters}
\label{app:guidance_hyperparams}
For inference, we use Euler discrete sampling for all guidance methods and sample 28 steps. The timestep scale for all models is from 0 to 1. To find hyperparamers for our setting, we start from the original papers' setting (when the information is available for the models used in our paper) and then empirically test the generation based on a randomly selected set of 10 prompts. Since many methods are tested on class-conditional models, the hyperparameters are not always directly applicable to our setting. Empirically, we find that we should keep more conditional information to achieve enough image-prompt consistency compared to class-conditional models.

\textit{APG~\citep{apg}.} This guidance method has three hyperparameters: effect of the parallel component $\eta$, effect of the rescaling threshold $r$, and effect of the momentum strength $\beta$. We follow the original paper and set $\eta = 0 $ for all settings. Other hyperparameters are presented in Table~\ref{tab:apg_hyperparams}.

\begin{table}[ht]
    \centering
    \caption{Hyperparameters for APG guidance method.}
    \begin{tabular}{c|cccc} \toprule
        params & \textbf{LDMv1.5} & \textbf{LDM-XL} & \textbf{LDMv3.5M} & \textbf{LDMv3.5L} \\ \midrule
        $r$ & 7.5 & 15 & 10 & 10 \\
        $\beta$ & -0.75 & -0.5 & -0.5 & -0.5 \\ \bottomrule
    \end{tabular}
    \label{tab:apg_hyperparams}
\end{table}

\textit{Interval~\citep{intervalguide}.} This guidance method has two hyperparameters: $\tau_{\rm{lo}}$ and $\tau_{\rm{hi}}$. The conditional information is only applied between $\tau_{\rm{lo}}$ and $\tau_{\rm{hi}}$. The hyperparameters used in our analyses are presented in Table~\ref{tab:interval_hyperparams}.
\begin{table}[ht]
    \centering
    \caption{Hyperparameters for interval guidance method.}
    \begin{tabular}{c|cccc} \toprule
        params & \textbf{LDMv1.5} & \textbf{LDM-XL} & \textbf{LDMv3.5M} & \textbf{LDMv3.5L} \\ \midrule
        $\tau_{lo}$ & 0.08 & 0.08 & 0.3 & 0.3 \\
        $\tau_{hi}$ & 0.81 & 0.81 & 0.95 & 0.95 \\ \bottomrule
    \end{tabular}
    \label{tab:interval_hyperparams}
\end{table}

\textit{CADS~\citep{cads}.} This guidance method has two main hyperparameters to control how the conditional information is annealed: $\tau_1$ and $\tau_2$, where the guidance scale is set to 0 for time steps between $\tau_2$ and 1.0, then linearly increases between time steps $\tau_1$ and $\tau_2$ from 0.0 to 1.0, and is kept to 1.0 between time steps 0.0 and $\tau_1$. In addition, CADS also has hyperparameters for noise scale $s$ and for mixing factor $\phi$. We keep $\phi = 1$ for all settings following the original paper. We present the hyperparameters' choice of CADS for different models in Table~\ref{tab:cads_hyperparams}. Some of the $\tau_2$ values are larger than 1.0, which means that we always keep some conditional information during sampling. This can help achieve better image-prompt consistency.
\begin{table}[ht]
    \centering
    \caption{Hyperparameters for CADS guidance method.}
    \begin{tabular}{c|cccc} \toprule
        params & \textbf{LDMv1.5} & \textbf{LDM-XL} & \textbf{LDMv3.5M} & \textbf{LDMv3.5L} \\ \midrule
        $\tau_1$ & 0.8 & 0.6 & 0.85 & 0.85 \\
        $\tau_2$ & 1.3 & 1.0 & 1.25 & 1.25 \\
        $s$ & 0.1 & 0.3 & 0.3 & 0.3 \\ \bottomrule
    \end{tabular}
    \label{tab:cads_hyperparams}
\end{table}

\subsection{Prompt expansion}
\label{app:prompt_expansion}
We use the Gemma3 model~\citep{gemma3} to expand the prompts. We use the \texttt{gemma-3-12b-it} version. The prompt used for expansion is as follows:
\begin{itemize}
    \item system prompt: \texttt{The user will give a sentence that is a caption for an image. Based on the sentence, you will generate 20 different new sentences with around 30 words, that don't violate the original meaning, but with more possible details in that image. You will return the sentences in a list format.}
    \item user prompt: \texttt{The sentence given is: ``CAPTION''.}
\end{itemize}

\subsection{Compute resources}
We use our internal cluster to conduct all the experiments.
The captioning, pairing and alignment steps are conducted on NVIDIA V100 GPUs with 32GB memory. The image generation is conducted on NVIDIA H100 GPUs with 80GB memory. All the models are run in 16-bit precision and can fit into one GPU. The generation time for 100,000 images is around 175 GPU hours for LDMv3.5L~\citep{sd35model}, 80 GPU hours for LDMv3.5M~\citep{sd35model}, 40 GPU hours for LDM-XL~\citep{sdxl} and 4 GPU hours for LDMv1.5.

\subsection{Licenses for existing assets}
\label{app:licences}
We list licenses for all the models, datasets and libraries used in our paper in Table~\ref{tab:licences}.

\subsection{New assets}
Our code is under License CC BY-NC 4.0.

\section{Results for LDM-XL and LDMv3.5M}
\label{app:xl35m}
In this section, we present results for LDM-XL~\citep{sdxl} and LDMv3.5M~\citep{sd35model} models on CC12M dataset~\citep{datasetcc12m} and ImageNet-1k dataset~\citep{imagenet}, complementing the results in the main paper.

\begin{table}[t]
    \centering
    \caption{Licences for all the models, datasets and libraries used in our paper.}
    \small
    \begin{tabular}{c|c|c} \toprule
        \textbf{Asset} & \textbf{Type} & \textbf{Licence} \\ \midrule
        CC12M~\citep{datasetcc12m} & Dataset & Freely use for any purpose~\tablefootnote{\url{https://github.com/google-research-datasets/conceptual-12m?tab=License-1-ov-file\#readme}}  \\ 
        Imagenet-1k~\citep{imagenet} & Dataset & Custom non-commercial license \\ 
        DCI~\citep{dcidataset} & Dataset & CC BY-NC 4.0 and SA-1B dataset Licnese~\tablefootnote{\url{https://ai.meta.com/datasets/segment-anything-downloads/}} \\
        LDMv3.5L~\citep{sd35model} & Model & Stability Community License \\ 
        LDMv3.5M~\citep{sd35model} & Model & Stability Community License \\ 
        LDM-XL~\citep{sdxl} & Model & CreativeML Open RAIL++-M License \\ 
        LDMv1.5~\citep{ldmrombach} & Model & CreativeML Open RAIL-M License \\ 
        Gemma3~\citep{gemma3} & Model & Gemma License \\ 
        SigLip~\citep{siglip} & Model & Apache-2.0 \\ 
        Openclip~\tablefootnote{\url{https://github.com/mlfoundations/open_clip}} & Library & Openclip License~\tablefootnote{\url{https://github.com/mlfoundations/open_clip?tab=License-1-ov-file\#readme}} \\ 
        Vendi-Score~\citep{vendiscore} & Library & MIT License \\
        DGM-eval~\citep{fddmetric} & Library & MIT License \\
        Eval-GIM~\citep{libevalgim} & Library & CC BY-NC 4.0 \\
        Aesthetic-predictor-v2.5~\citep{aesthetic_v2_5} & Library & AGPL-3.0 \\ \bottomrule
    \end{tabular}
    \label{tab:licences}
\end{table}

\subsection{Utility axes measured by reference-free metrics}
\begin{figure}[t]
    \centering
    \begin{subfigure}[ht]{0.22\textwidth}
        \includegraphics[width=\linewidth]{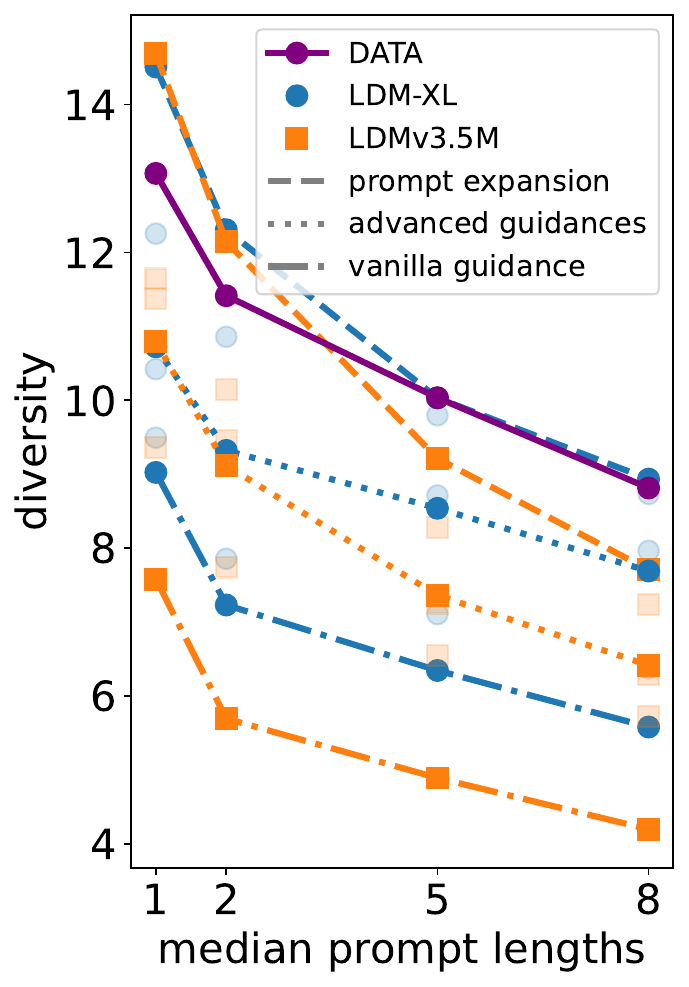}
        \caption{Diversitiy}
        \label{fig:diversity_SDexpansion_cc12m_xl35m}
    \end{subfigure}
    \begin{subfigure}[ht]{0.22\textwidth}
        \includegraphics[width=\textwidth]{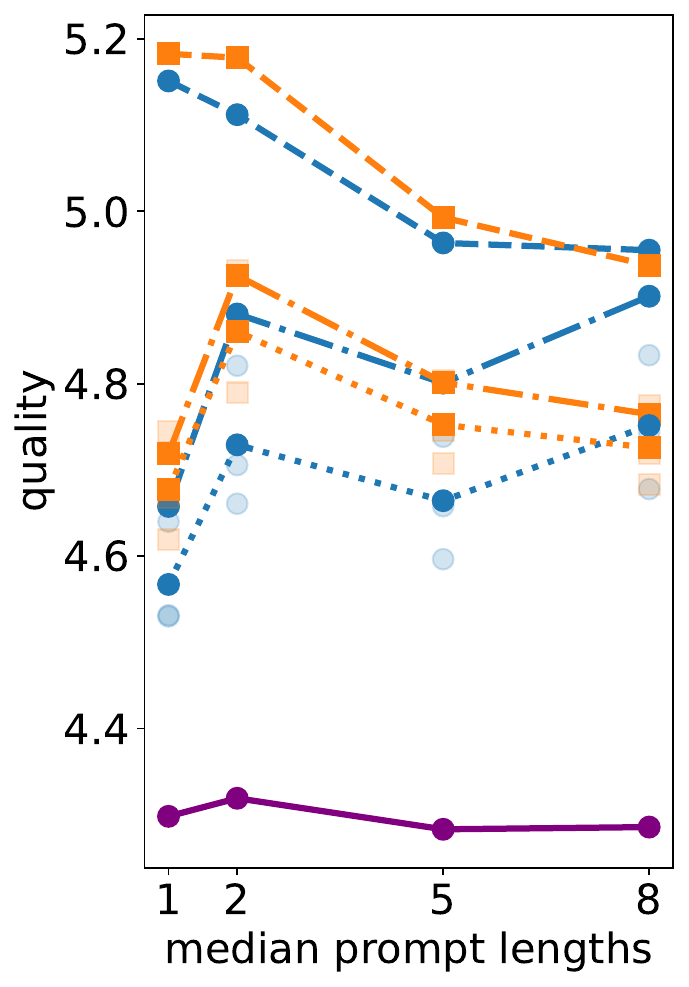}
        \caption{Quality}
        \label{fig:quality_SDexpansion_cc12m_xl35m}
    \end{subfigure}
    \begin{subfigure}[ht]{0.22\textwidth}
        \includegraphics[width=\textwidth]{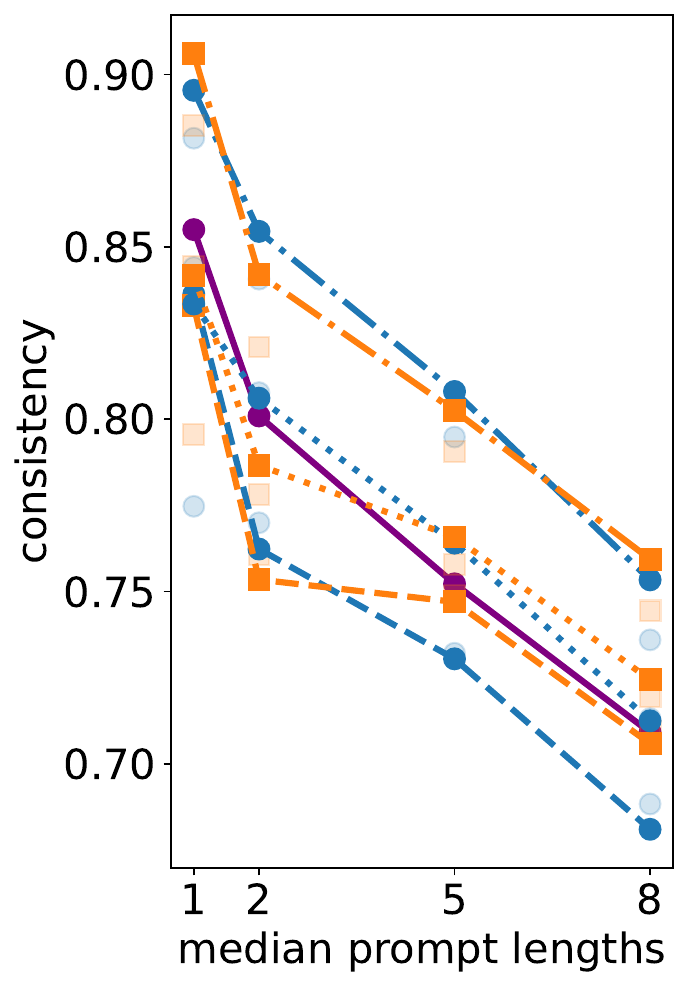}
        \caption{Consistency}
        \label{fig:consistency_SDexpansion_cc12m_xl35m}
    \end{subfigure}
    \caption{\textbf{Reference-free utility metrics of synthetic data using CC12M prompts.} Diversity (Vendi), quality (aesthetics) and consistency (DSG) metrics of LDM-XL and LDMv3.5M generations when using: 1) vanilla guidance (CFG), 2) prompt expansion, and 3) advanced guidance methods, for which transparent markers correspond to different methods and the solid marker is the average over methods. 
    Both advanced guidance methods and prompt expansion lead to improved diversity over the vanilla guidance. Prompt expansion from shorter captions can surpass the real data diversity. Prompt expansion leads to quality improvements, whereas advanced guidance methods slightly reduces quality \wrt vanilla guidance.} 
    \label{fig:utilities_SDexpansion_cc12m_xl35m}
\end{figure}

\begin{figure}[ht]
    \centering
    \begin{subfigure}[ht]{0.22\textwidth}
        \includegraphics[width=\linewidth]{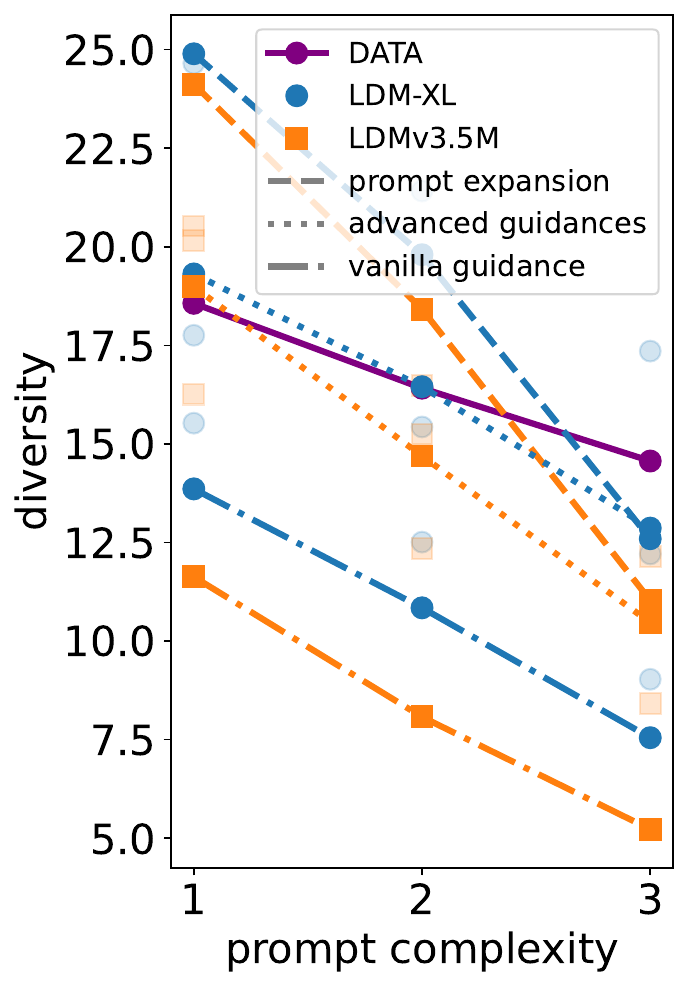}
        \caption{Diversity}
        \label{fig:diversity_SDexpansion_in1k_xl35m}
    \end{subfigure}
    \begin{subfigure}[ht]{0.22\textwidth}
        \includegraphics[width=\textwidth]{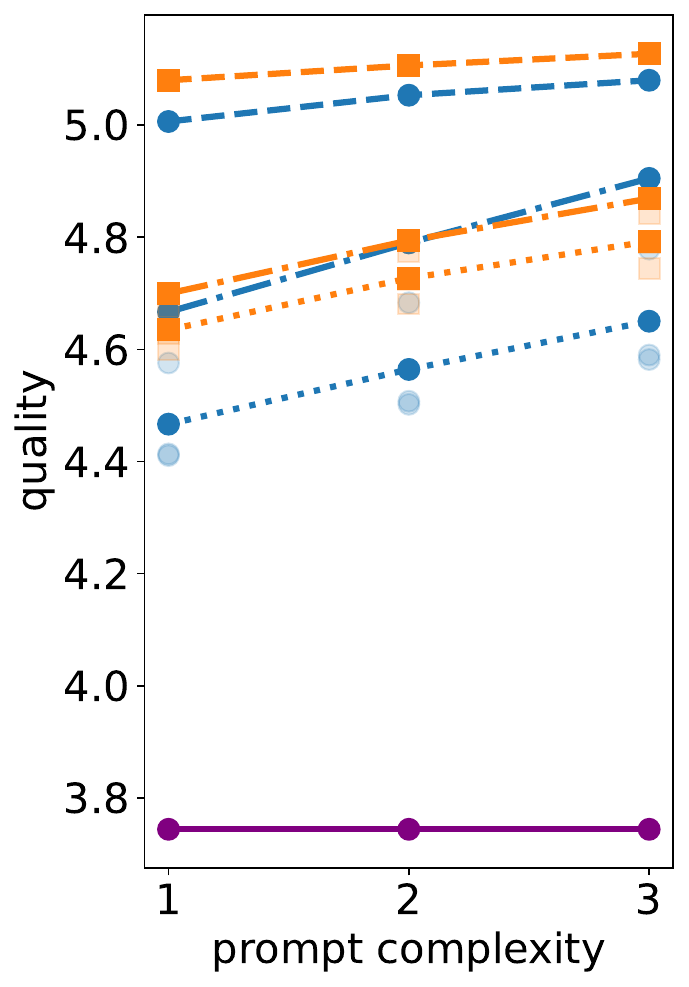}
        \caption{Quality}
        \label{fig:quality_SDexpansion_in1k_xl35m}
    \end{subfigure}
    \begin{subfigure}[ht]{0.22\textwidth}
        \includegraphics[width=\textwidth]{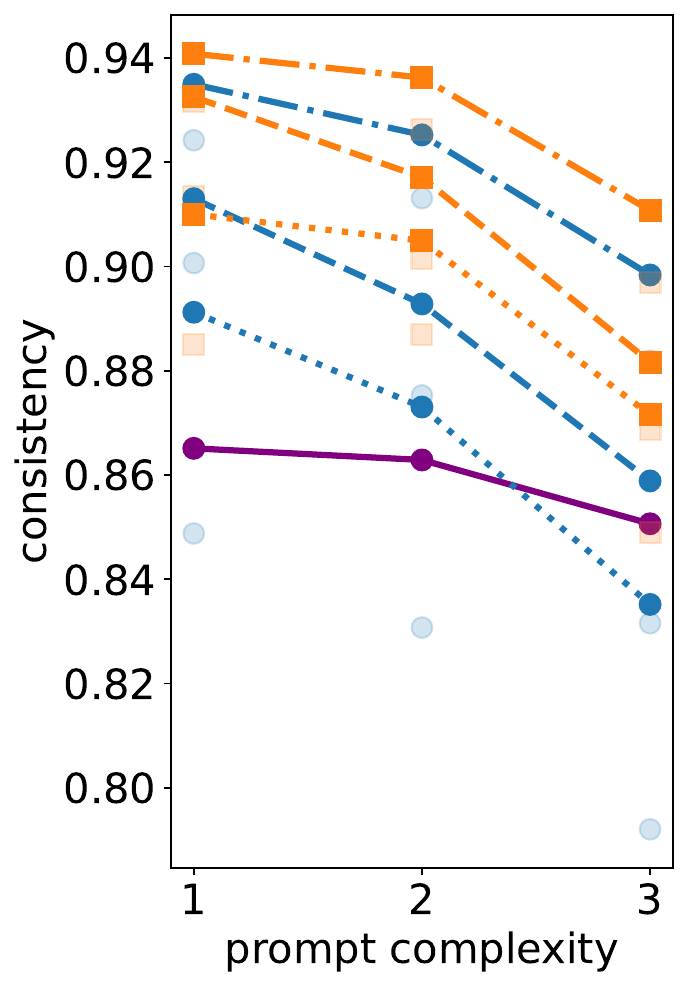}
        \caption{Consistency}
        \label{fig:consistency_SDexpansion_in1k_xl35m}
    \end{subfigure}
    \caption{\textbf{Reference-free utility metrics of synthetic data using ImageNet-1k class labels.}  Diversity (Vendi), quality (aesthetics) and consistency (DSG) metrics for LDM-XL and LDMv3.5M generations with: 1) vanilla guidance (CFG), 2) prompt expansion, and 3) advanced guidance methods, for which transparent markers correspond to different methods and the solid marker is the average over methods. Prompt complexity shows the specificity of the class label used for generation. Both advanced guidance methods and prompt expansion lead to higher diversity, but advanced guidance methods lead to slightly lower quality. Prompt expansion and advanced guidances with shorter captions can yield higher diversity compared to the real data. Both prompt expansion and advanced guidances have a negative impact on consistency.}
    \label{fig:utilities_SDexpansion_in1k_xl35m}
\end{figure}

We measure the utility of synthetic data in terms of quality (aesthetics), diversity (Vendi) and prompt consistency (DSG) with reference-free metrics. We present results using the prompts from CC12M in Figure~\ref{fig:utilities_SDexpansion_cc12m_xl35m} and ImageNet-1k in Figure~\ref{fig:utilities_SDexpansion_in1k_xl35m}.
The results show the same trend as presented in Section~\ref{sec:reference_free}.

\subsection{Utility axes measured by reference-based metrics}
\begin{figure}[t]
    \centering
    \begin{subfigure}[ht]{0.22\textwidth}
        \includegraphics[width=\textwidth]{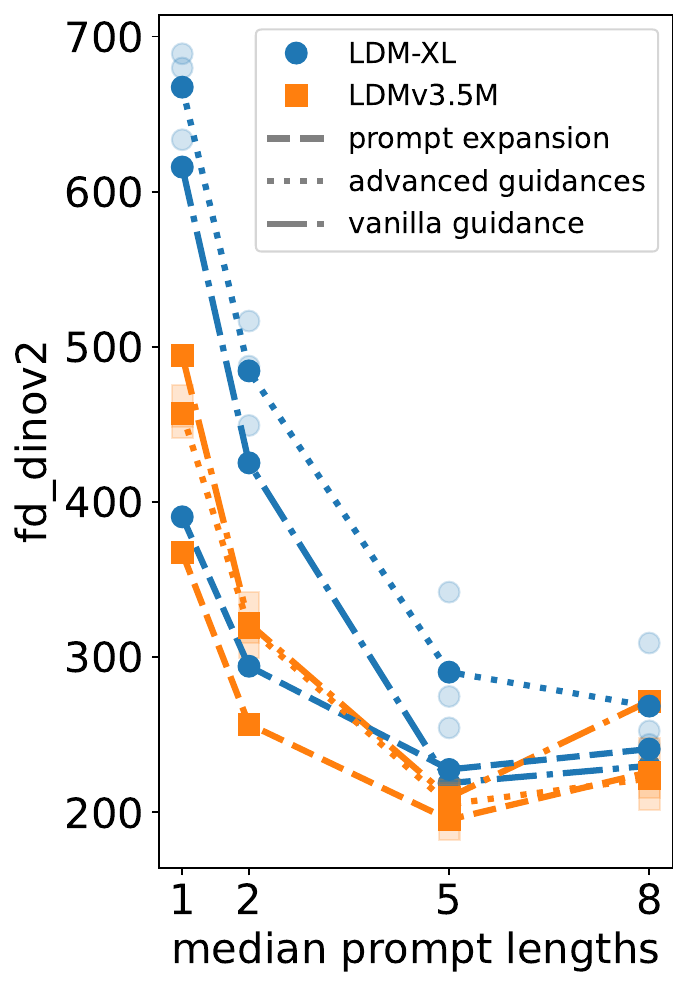}
        \caption{FDD}
        \label{fig:fd_dinov2_SDexpansion_cc12m_xl35m}
    \end{subfigure}
    \begin{subfigure}[ht]{0.22\textwidth}
        \includegraphics[width=\textwidth]{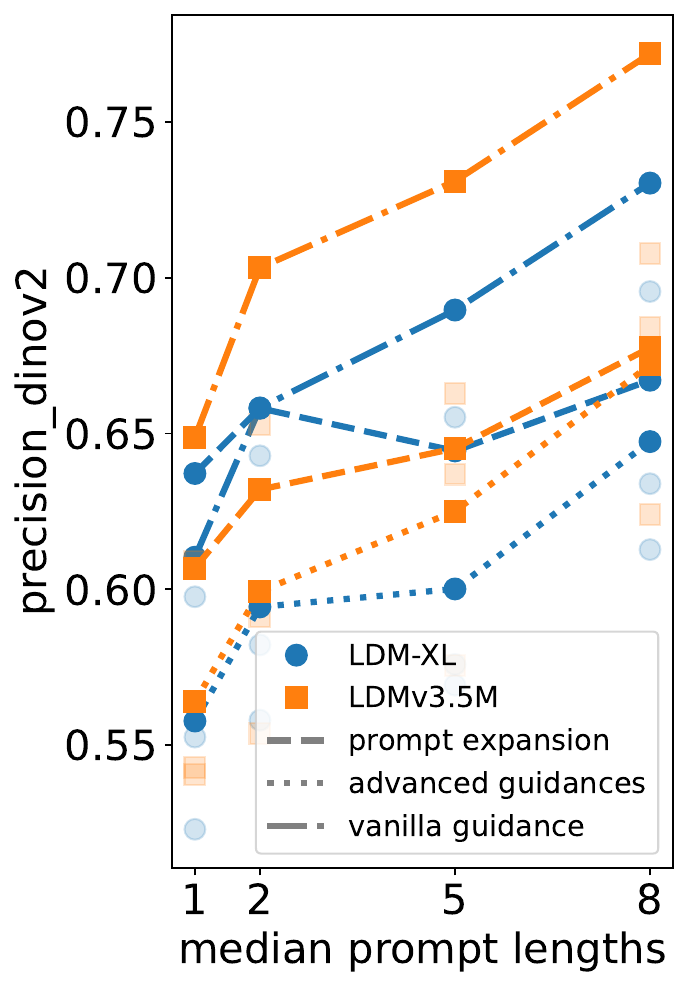}
        \caption{Precision}
        \label{fig:precision_dinov2_SDexpansion_cc12m_xl35m}
    \end{subfigure}
    \begin{subfigure}[ht]{0.22\textwidth}
        \includegraphics[width=\linewidth]{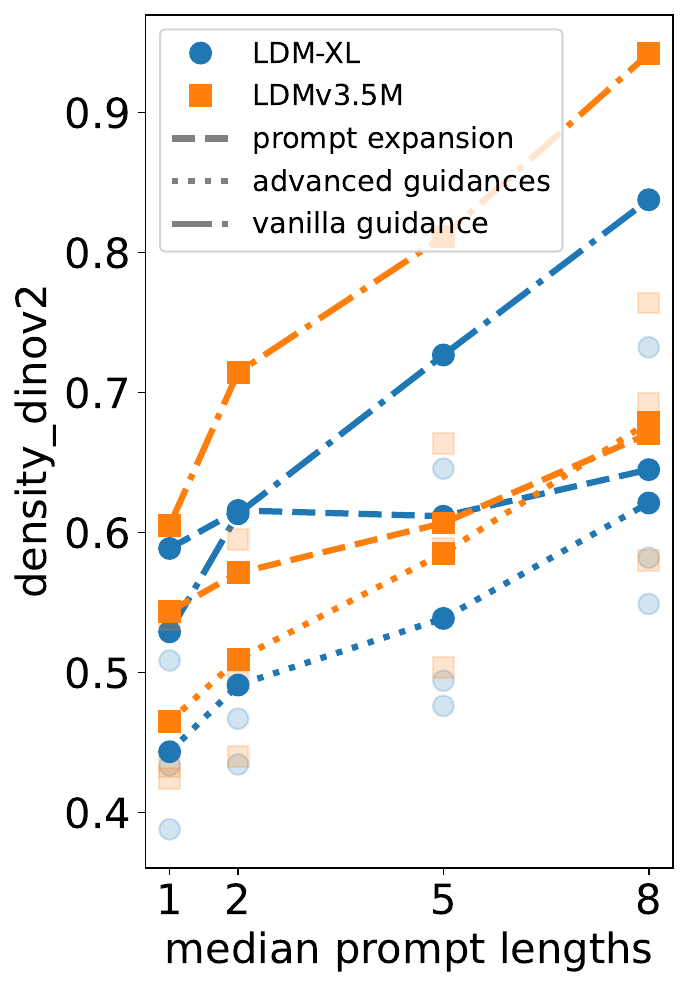}
        \caption{Density}
        \label{fig:density_dinov2_SDexpansion_cc12m_xl35m}
    \end{subfigure}
    \begin{subfigure}[ht]{0.22\textwidth}
        \includegraphics[width=\textwidth]{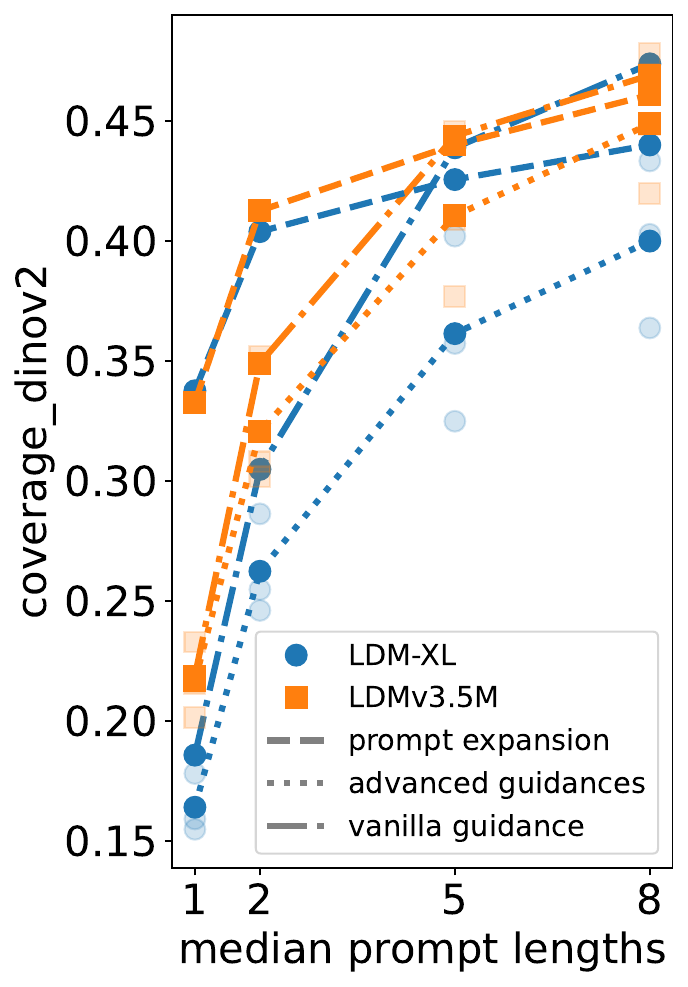}
        \caption{Coverage}
        \label{fig:coverage_dinov2_SDexpansion_cc12m_xl35m}
    \end{subfigure}
    \caption{\textbf{Reference-based utility metrics of synthetic data using CC12M prompts.} Fr\'{e}chet distance (FDD), precision, density and coverage in the DINOv2 feature space for LDM-XL and LDMv3.5M generations with: (1) vanilla guidance (CFG), (2) prompt expansion, and (3) advanced guidance methods, for which transparent markers correspond to different methods and the solid marker is the average over methods. Prompt expansion lead to better FDD for both models. Both prompt expansion and advanced guidances sacrifice precision and density.}
\label{fig:marginal_dinov2_SDexpansion_cc12m_xl35m}
\end{figure}

\begin{figure}[t]
    \centering
    \begin{subfigure}[ht]{0.22\textwidth}
        \includegraphics[width=\textwidth]{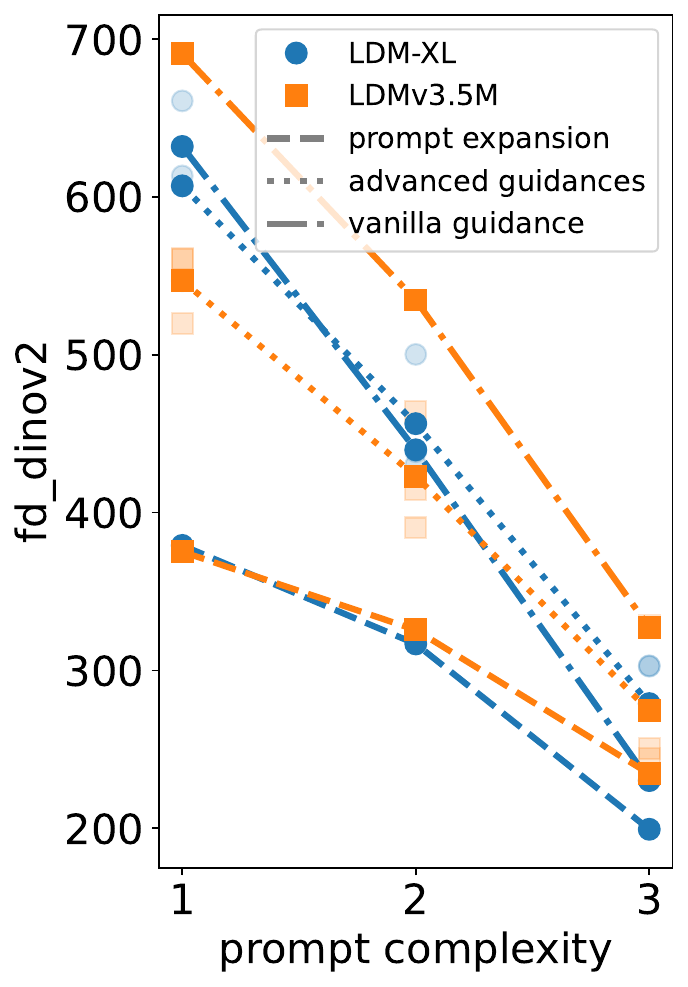}
        \caption{FDD}
        \label{fig:fd_dinov2_SDexpansion_in1k_xl35m}
    \end{subfigure}
    \begin{subfigure}[ht]{0.22\textwidth}
        \includegraphics[width=\textwidth]{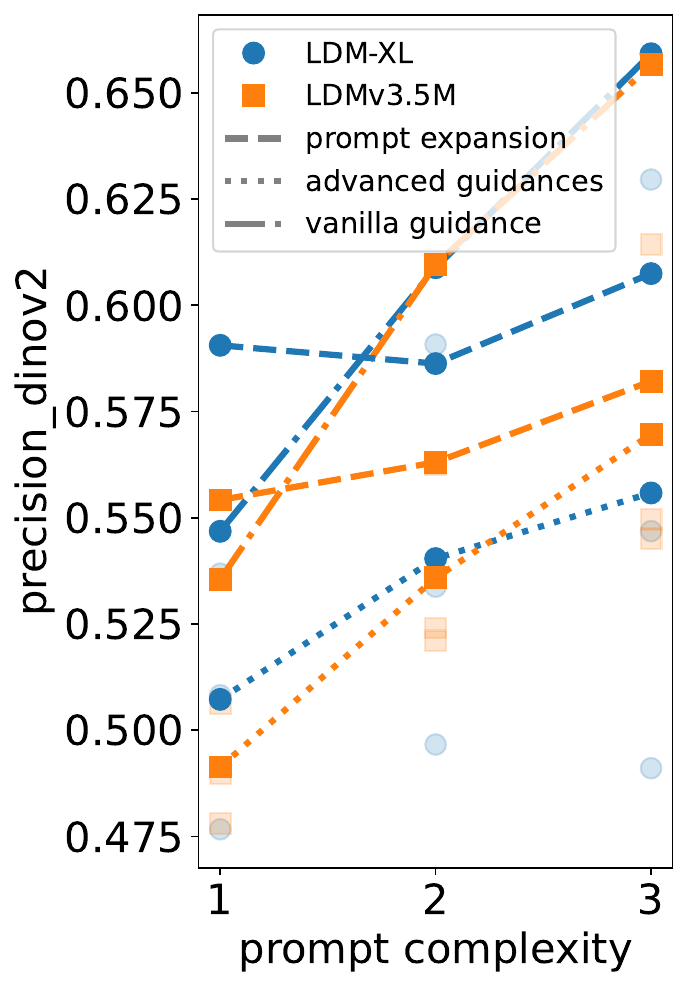}
        \caption{Precision}
        \label{fig:precision_dinov2_SDexpansion_in1k_xl35m}
    \end{subfigure}
    \begin{subfigure}[ht]{0.22\textwidth}
        \includegraphics[width=\linewidth]{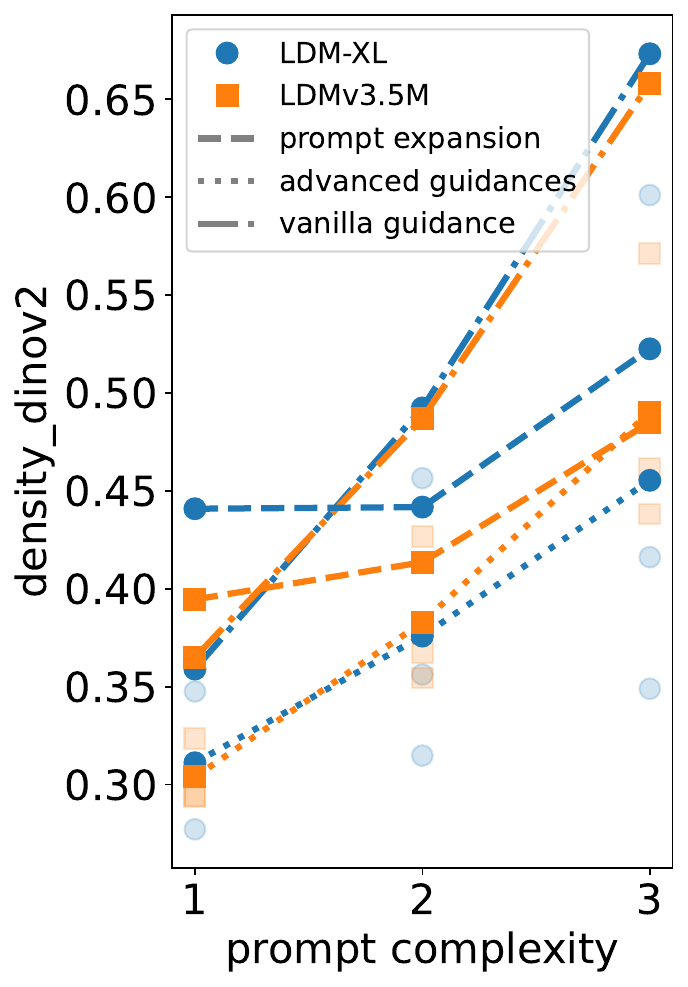}
        \caption{Density}
        \label{fig:density_dinov2_SDexpansion_in1k_xl35m}
    \end{subfigure}
    \begin{subfigure}[ht]{0.22\textwidth}
        \includegraphics[width=\textwidth]{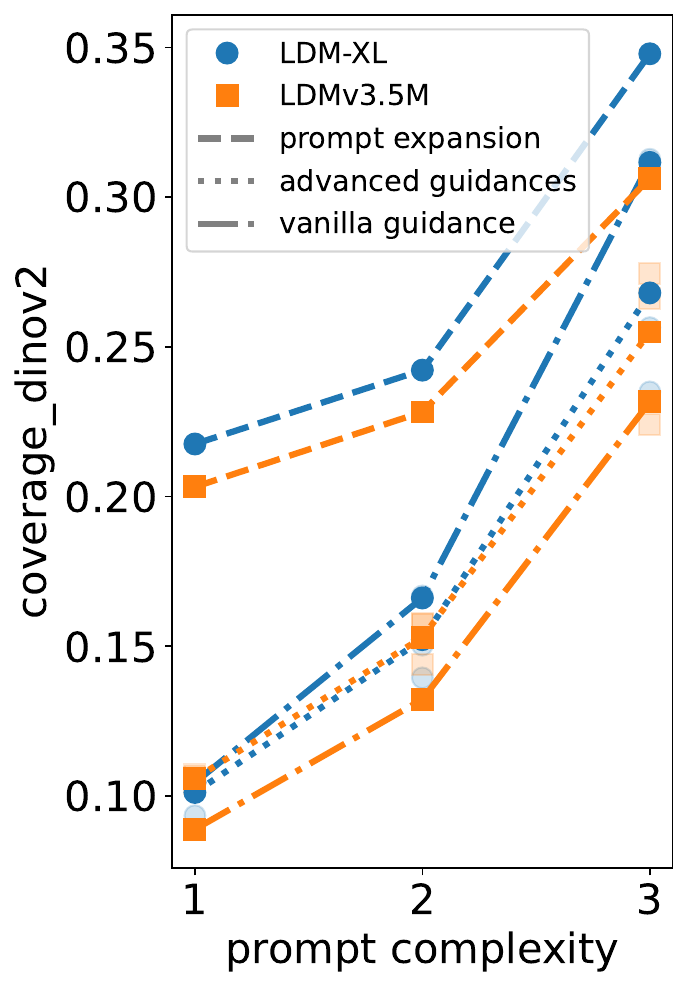}
        \caption{Coverage}
        \label{fig:coverage_dinov2_SDexpansion_in1k_xl35m}
    \end{subfigure}
    \caption{\textbf{Reference-based utility metrics of synthetic data using ImageNet-1k class labels.} FDD, precision, density, and coverage metrics in the DINOv2 feature space for LDM-XL and LDMv3.5M with: 1) vanilla guidance (CFG), 2) prompt expansion, and 3) advanced guidance methods, for which transparent markers correspond to different methods and the solid marker is the average over methods. Both advanced guidance methods and prompt expansion lead to better FDD, and improve or match the coverage of vanilla guidance, while sacrificing precision and density in most cases. For LDMv3.5M, we see a more prominent effect of advanced guidance methods and prompt expansion than for LDM-XL. Advanced guidance methods do not change the trend of the metrics \wrt prompt complexity, while the prompt expansion shapes the trend of the metrics differently.}
\label{fig:marginal_dinov2_SDexpansion_in1k_xl35m}
\end{figure}

Reference-free metrics are suitable for evaluating the utility of synthetic data in the wild. In our analysis, the prompts are extracted from existing dataset, containing both images and texts, and so enabling the use of reference-based metrics for evaluation. Unless otherwise specified, we use the DINOv2~\citep{dinov2} feature space to compute these metrics, since it has been found to correlate well with human judgement~\citep{hall2024towards,fddmetric}.
The CC12M and ImageNet-1k results are presented in Figures~\ref{fig:marginal_dinov2_SDexpansion_cc12m_xl35m} and~\ref{fig:marginal_dinov2_SDexpansion_in1k_xl35m}, respectively. 

The results are similar to those for LDMv1.5 and LDMv3.5L presented in Section~\ref{para:reference_based}. It is worth noting that we employed more aggressive hyperparameters of advanced guidance methods for LDM-XL model as shown in Tables~\ref{tab:apg_hyperparams}, \ref{tab:interval_hyperparams} and \ref{tab:cads_hyperparams}. This leads to larger differences for reference-based metrics since the generated images can be out side of the support of the reference dataset.

\begin{figure}[t]
    \centering
    \begin{subfigure}[ht]{0.28\textwidth}
        \includegraphics[width=\linewidth]{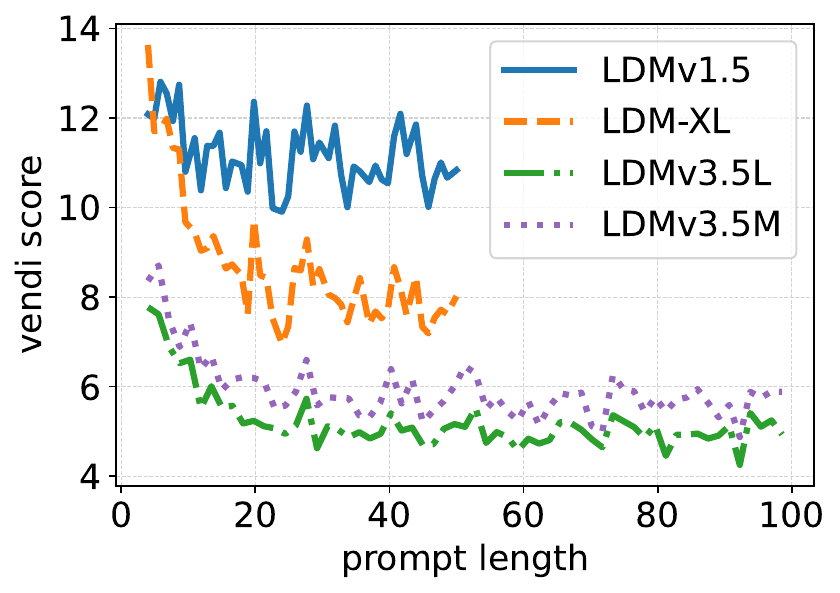}
        \caption{Diversity}
        \label{fig:dcisd_vendi}
    \end{subfigure}
    \begin{subfigure}[ht]{0.28\textwidth}
        \includegraphics[width=\linewidth]{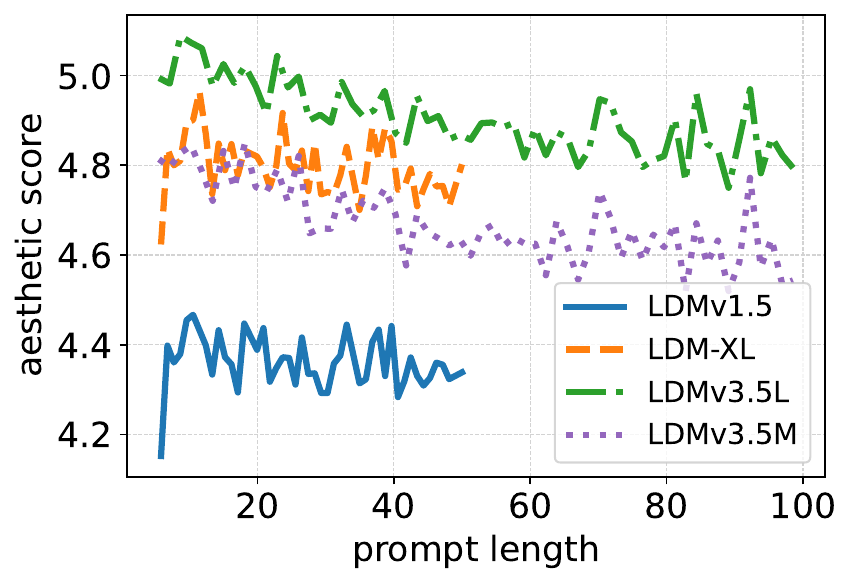}
        \caption{Quality}
        \label{fig:dcisd_aesthetic}
    \end{subfigure}
    \begin{subfigure}[ht]{0.28\textwidth}
        \includegraphics[width=\textwidth]{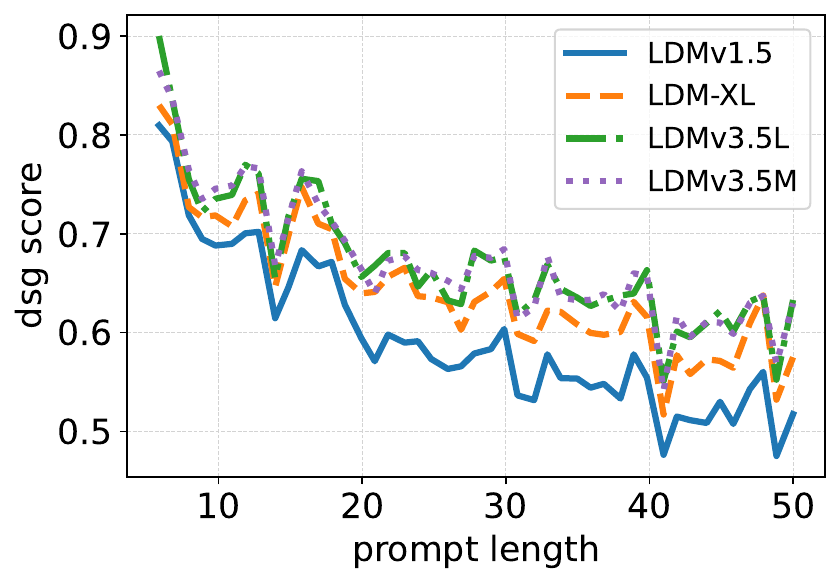}
        \caption{Consistency}
        \label{fig:dcisd_dsg}
    \end{subfigure}
    \caption{\textbf{Reference-free utility metrics of synthetic data using DCI prompts of increasing complexity.} For all models, diversity decreases at first and then plateaus, while consistency score continuously decreases with the increase of the prompt length. The quality of LDMv3.5 decreases as prompt complexity increases.}
    \label{fig:DCI_SD}
\end{figure}

\subsection{DCI results}
\label{app:dci}
DCI dataset opens the possibility to evaluate the utility of synthetic data for notably longer prompts. We perform reference-free evaluations on synthetic data generated from DCI prompts by binning the prompts into different length chunks and computing the average diversity, aesthetic and consistency scores for each bin. We present the results in Figure~\ref{fig:DCI_SD}~\footnote{LDMv1.5 and LDM-XL models stop at prompt length of $50$, given the $77$ token limit of their text encoder.}.
We observe that the diversity decreases and then plateaus as we increase the prompt length. For most LDM models, the plateau starts at a prompt length of $\sim30$ words. However, the plateau region of LDMv1.5 seems to occur earlier than for its successor models. However, the consistency tends to decrease as we increase the prompt length. As the prompt becomes increasing detailed, the consistency metric keeps assessing whether all the information is captured, emphasizing the limitations of current T2I models to follow very long and detailed prompts. Interestingly, for the most recent LDMv3.5 models, image aesthetics also seems to slightly decrease as a function of prompt length.

\section{Human evaluation}
\label{app:human_eval}
Grounding our automatic metrics with human evaluation is crucial for validating our findings. We conducted human evaluations, specifically designed to assess whether our automatic diversity metric (Vendi score) \xf{and consistency metric (DSG score) aligns with human perception. }

\subsection{Vendi score}
We employed a two-alternative forced-choice (2AFC) task. In each trial, annotators were shown two image mosaics (16 images each), generated from prompts of two different complexity levels. They were asked: ``Which set of images is more diverse?'' Images were generated from CC12M grounded prompts of different complexities using the LDMv3.5L model. For each trial, we randomly sampled two prompts, each from a distinct complexity level to generate the two mosaics for comparison. The study involved 9 annotators, who collectively provided 900 pairwise comparison annotations, ensuring comprehensive coverage of all complexity level pairs. We calculated the win rate for lower-complexity prompts against higher-complexity prompts. A win rate greater than 0.50 indicates that, on average, humans perceive images from lower-complexity prompts as more diverse. The results are summarized in Table~\ref{tab:human_eval}.

Our analysis yields two key findings that validate our approach:
\begin{enumerate}
    \item \textbf{Human perception aligns with our main findings.} In all pairwise comparisons, the win rate for the lower-complexity prompts was all above 0.50 (e.g., 0.620 for complexity 1 vs. 3; 0.794 for 0 vs. 3). This proves that humans consistently perceive the sets of images generated from simpler prompts as more diverse, mirroring the trend identified by our automatic metrics.
    \item \textbf{Vendi score strongly correlates with human preference.} There is a clear positive correlation between the magnitude of the Vendi score difference (values in parentheses) and the human-annotated win rate. For instance, the largest Vendi score difference (3.00, between levels 0 and 3) corresponds to the most decisive human judgment (0.794 win rate), while the smallest difference (0.605, between levels 2 and 3) corresponds to a judgment barely above chance (0.531 win rate). This provides strong evidence that the Vendi score used in our work is a meaningful and valid proxy for the degree of human-perceived diversity.
\end{enumerate}

\begin{table}[h]
    \centering
    \caption{Human evaluation results for diversity. Values represent the win rate of images generated from prompts of the row complexity level being judged as more diverse than those from the column level. Values in parentheses show the corresponding Vendi score difference between the two levels.}
    \begin{tabular}{l|lll}
    \toprule
        Complexity level & 2 & 3 & 4 \\ \midrule
        1 & 0.589 (1.63) & 0.645 (2.40) & 0.794 (3.00) \\
        2 & - & 0.563 (0.77) & 0.620 (1.37) \\
        3 & - & - & 0.531 (0.61) \\ \bottomrule
    \end{tabular}
    \label{tab:human_eval}
\end{table}

\subsection{DSG score}
\xf{DSG uses a two-stage process: }
\xf{(1) Decomposition: An LLM (Gemma-3-27B-Instruct, the largest model in the series released in 2025) decomposes the single complex prompt (e.g., "A white dog is playing on the grassy field.") into a set of simple, atomic questions (e.g., "Is there a dog?", "Is the dog white?", "Is the field grassy?", …). }
\xf{(2) VQA: The VQA model (default as in DSG score paper) is then only required to answer these simple, unit-level yes/no questions, a task it can perform more reliably compared to answering a single long complex question.}

\xf{To further validate our automatic evaluated results with DSG score, we conduct a human evaluation for both stages.}

\xf{\paragraph{Stage 1: prompt decomposition.}}

\xf{We give the guideline on how the prompt is decomposed into atomic questions along with an example of decomposition presented to human annotators. Given a prompt, we present the prompt along with the decomposed questions generated by LLM to human evaluators, and ask them to check whether each question is a valid question according to decomposition rules. We ask 4 human annotators, and each of them annotates 25 prompts, yielding 100 sets of answers in total. Among all the decomposed questions, the percentage of correctly decomposed questions is 95.54\%, which effectively demonstrates the validity of the decomposed questions using LLM automatic generation.}

\xf{\paragraph{Stage 2: VQA model reliability.}}

\xf{Given the generated images from the original long prompt, we ask the human annotators to answer each decomposed question generated by LLM from the original long prompt. We ask 10 annotators and each evaluates 50 generated images using DCI prompts and gather in total 500 evaluation results. The human evaluation is small compared to the large-scale automatic evaluation, but it shows the same decreasing trend and a strong correlation between human evaluation results and automatic VLM results. Table~\ref{tab:dsg_stage2_human} shows the human evaluation results for Stage 2.}

\begin{table}[!tbp]
\caption{DSG metrics computed using human evaluation and automatic VLM}
\label{tab:dsg_stage2_human}
\centering
\begin{tabular}{@{}lllllllll@{}}
\toprule
\# words   & 10   & 13   & 18   & 23   & 28   & 35   & 41   & 50   \\ \midrule
\# samples & 58   & 54   & 66   & 61   & 65   & 66   & 65   & 75   \\
Human DSG  & 0.92 & 0.90 & 0.85 & 0.87 & 0.83 & 0.82 & 0.81 & 0.81 \\
VQA DSG    & 0.95 & 0.89 & 0.81 & 0.86 & 0.85 & 0.81 & 0.79 & 0.82 \\ \midrule
           & \multicolumn{8}{c}{Pearson corr 0.89; R2 0.68}        \\ \bottomrule
\end{tabular}
\end{table}

\section{Additional models}
\label{app:additional_models}
\begin{figure}[t]
    \centering
    \begin{subfigure}[ht]{0.20\textwidth}
        \includegraphics[width=\linewidth]{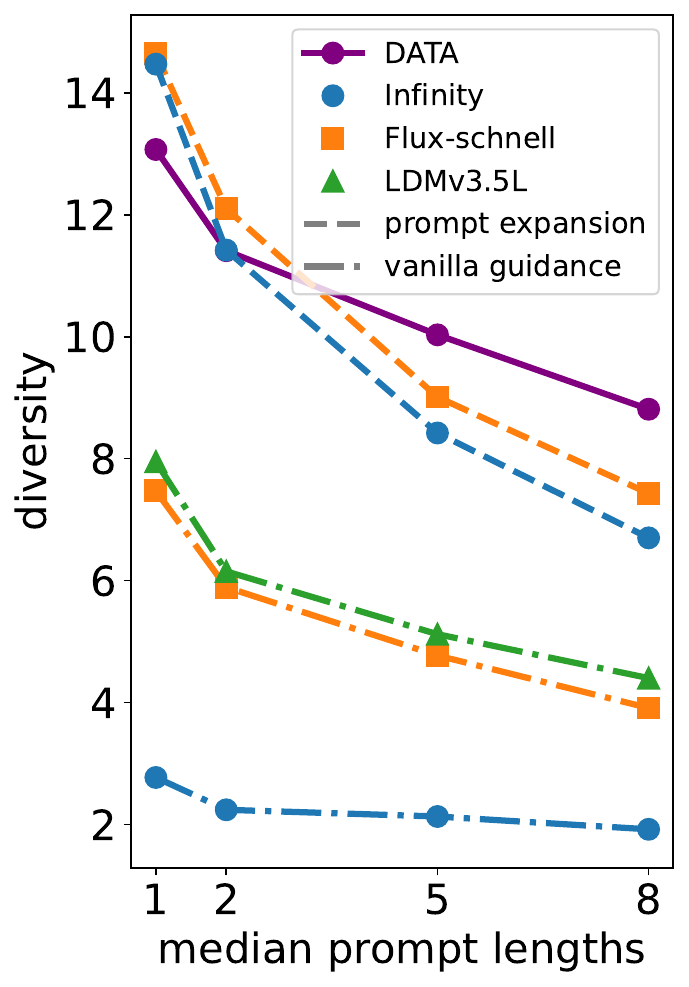}
        \caption{Diversitiy}
        \label{fig:diversity_InfinityFluxexpansion_cc12m}
    \end{subfigure}
    \begin{subfigure}[ht]{0.20\textwidth}
        \includegraphics[width=\textwidth]{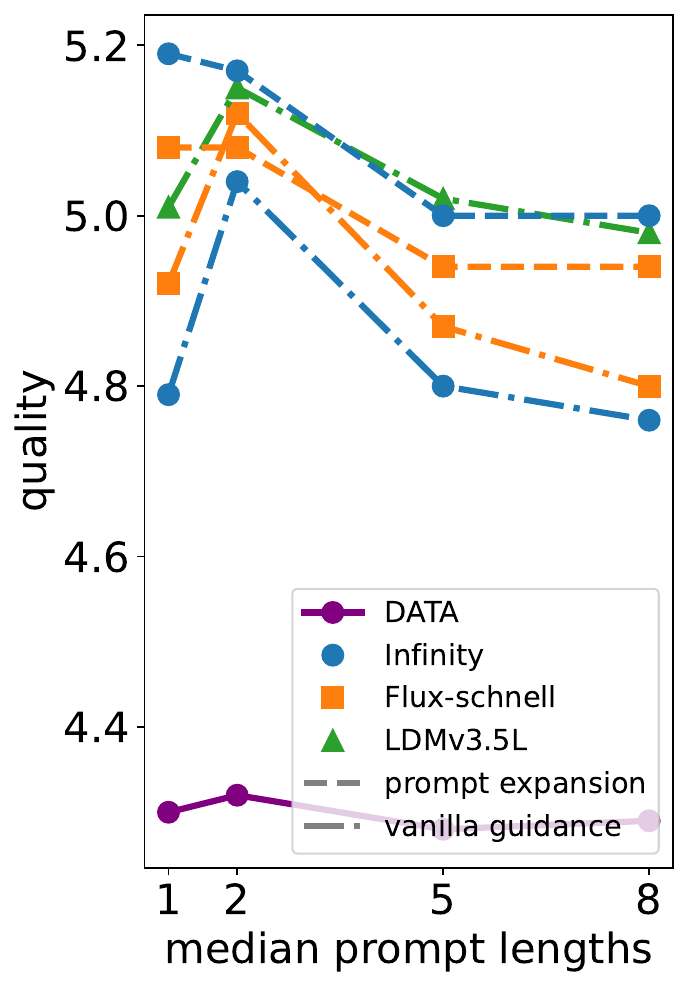}
        \caption{Quality}
        \label{fig:quality_InfinityFluxexpansion_cc12m}
    \end{subfigure}
    \begin{subfigure}[ht]{0.20\textwidth}
        \includegraphics[width=\textwidth]{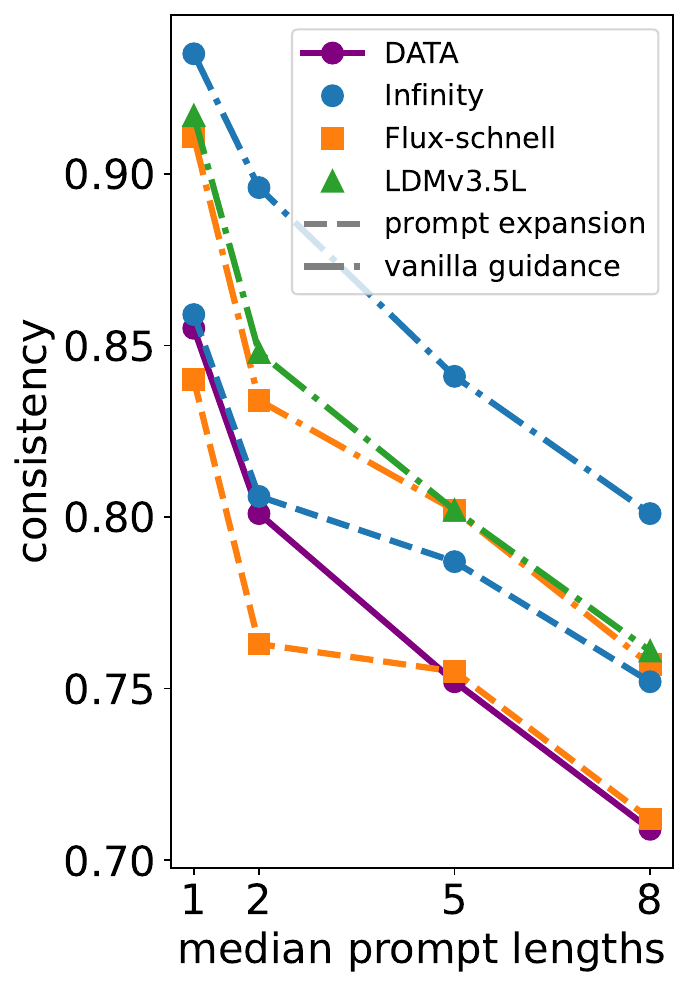}
        \caption{Consistency}
        \label{fig:consistency_InfinityFluxexpansion_cc12m}
    \end{subfigure}
    \caption{\textbf{Reference-free utility metrics of synthetic data using CC12M prompts.} Diversity (Vendi), quality (aesthetics) and consistency (DSG) metrics of Infinity and Flux-schnell generations when using: 1) vanilla guidance (CFG) and 2) prompt expansion. 
    Prompt expansion leads to improved diversity over the vanilla guidance. Prompt expansion from shorter captions can surpass the real data diversity. Prompt expansion leads to quality improvements. Infinity model with vanilla guidance has very low diversity in generations, while prompt expansion can effectively recover the diversity.} 
    \label{fig:utilities_InfinityFluxexpansion_cc12m}
\end{figure}

Our selection of models cover latent diffusion models and flow models with a chronological perspective. We can compare different guidance methods over all these models selected. However, there are some other models that may also raise interests for the research community, \eg guidance distilled models that uses no classifier-free guidance, and visual autoregreessive models. We present the utility of the synthetic data generated from FLUX-schnell~\citep{flux2024} (a distilled diffusion model) and Infinity~\citep{infinityautoregressive} (a visual autoregressive model) using CC12M prompts in Figure~\ref{fig:utilities_InfinityFluxexpansion_cc12m}. We include vanilla guidance and prompt expansion for both models.

The Infinity model has the lowest diversity compared to the LDM model series. The LDMv3.5L model has a diversity metric of 4.4 as the lowest, while Infinity can only effectively generate 2 different images on average across prompt lengths.
Prompt expansion boosts the diversity metric for all the complexities. Similar to the LDM model series, the diversity of the synthetic images generated using level-1 prompts surpass that of the real data. Interestingly, the decrease of the diversity along the prompt complexity axis is more pronounced than the LDMv1.5 and LDMv3.5L models as shown in Figure~\ref{fig:diversity_SDexpansion_cc12m}. This indicates that the prompt may put more constraints on a visual autoregressive model, \eg Infinity, when exploring the image space compared to diffusion or flow models.
Nevertheless, the Infinity model shows good quality that surpasses the real data and the LDMv1.5 model. It is on par with or a bit lower in terms of aesthetic quality compared to the LDMv3.5L model.
Consistency-wise, the Infinity achieves the best compared to all the LDM model series. This may also correlate with the low diversity of the Infinity model as the model is more faithful to the prompt and thus limit the exploration in the image space.

As for the FLUX-schnell model, it performs similarly to the LDMv3.5L model on three utility axes, with a small but noticeable drop in utility, especially on the axes of diversity and consistency. The gap tends to become larger when the prompt complexity increases.

\section{Additional results}
\label{app:additional_results}

\subsection{Reference-based metrics using ImageNet-1k class labels}
\label{app:reference_based_in1k}
\begin{figure}[t]
    \centering
    \begin{subfigure}[ht]{0.20\textwidth}
        \includegraphics[width=\textwidth]{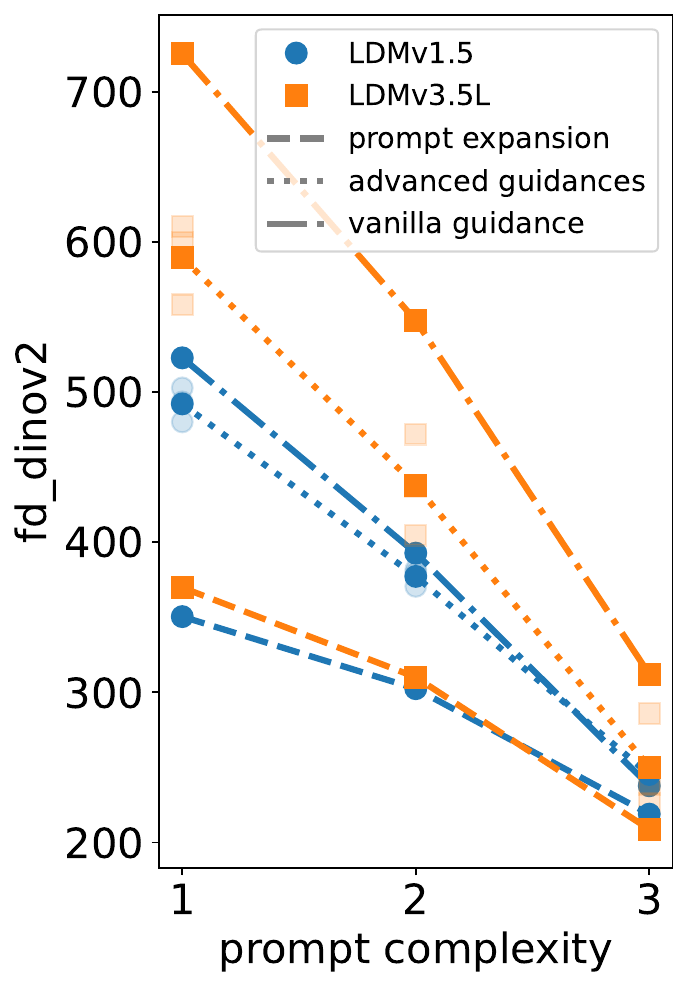}
        \caption{FDD}
        \label{fig:fd_dinov2_SDexpansion_in1k}
    \end{subfigure}
    \begin{subfigure}[ht]{0.20\textwidth}
        \includegraphics[width=\textwidth]{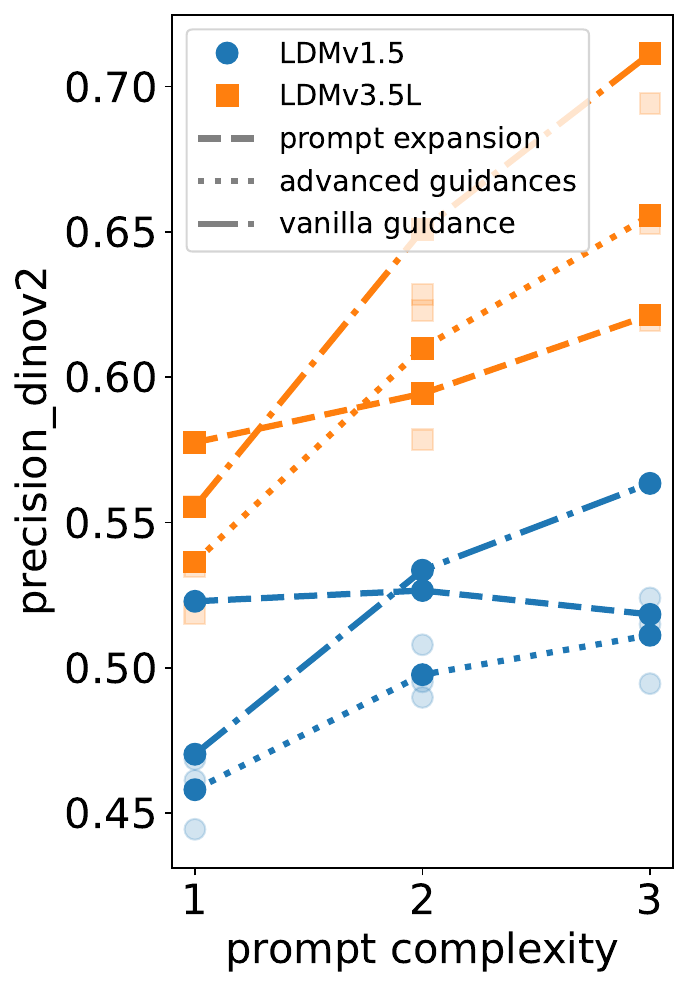}
        \caption{Precision}
        \label{fig:precision_dinov2_SDexpansion_in1k}
    \end{subfigure}
    \begin{subfigure}[ht]{0.20\textwidth}
        \includegraphics[width=\linewidth]{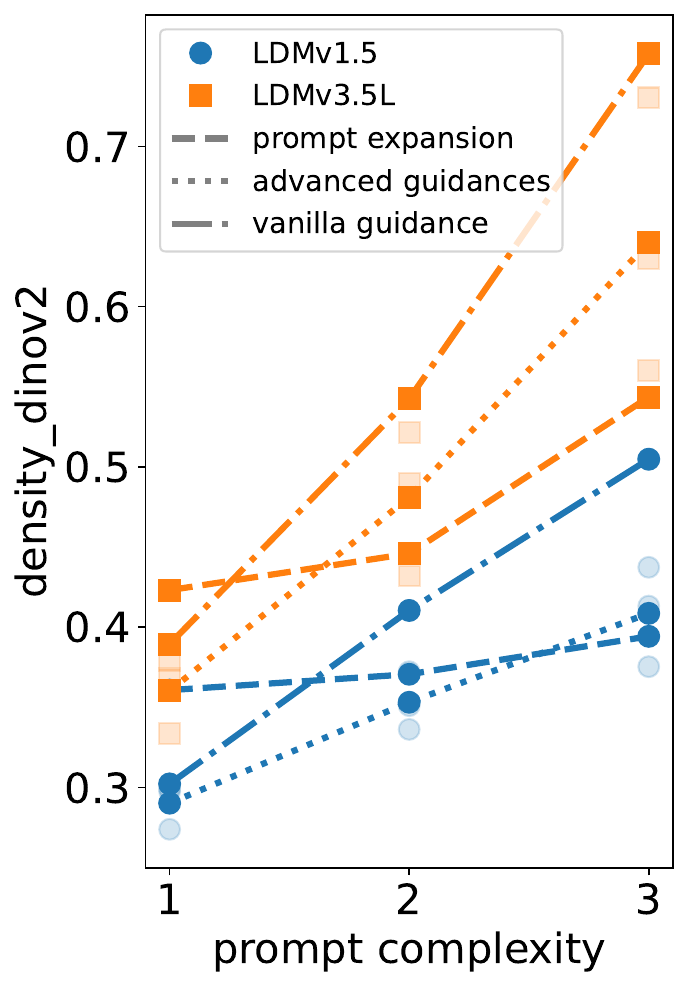}
        \caption{Density}
        \label{fig:density_dinov2_SDexpansion_in1k}
    \end{subfigure}
    \begin{subfigure}[ht]{0.20\textwidth}
        \includegraphics[width=\textwidth]{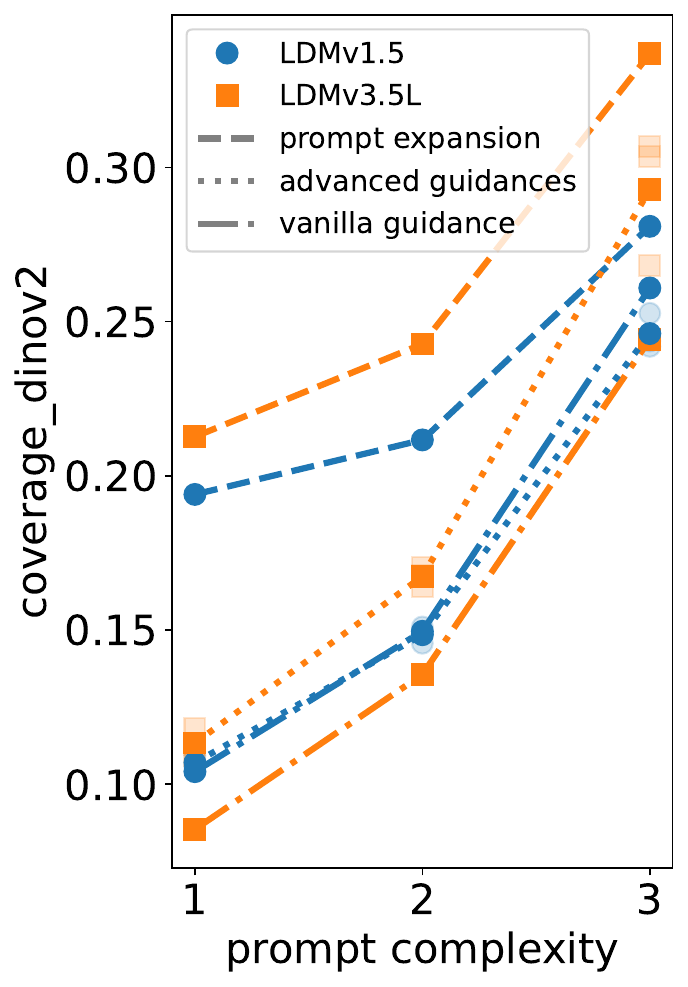}
        \caption{Coverage}
        \label{fig:coverage_dinov2_SDexpansion_in1k}
    \end{subfigure}
    \caption{\textbf{Reference-based utility metrics of synthetic data using ImageNet-1k class labels.} FDD, precision, density, and coverage metrics in the DINOv2 feature space for LDMv1.5 and LDMv3.5L with: 1) vanilla guidance (CFG), 2) prompt expansion, and 3) advanced guidance methods, for which transparent markers correspond to different methods and the solid marker is the average over methods. Both advanced guidance methods and prompt expansion lead to better FDD, and improve or match the coverage of vanilla guidance, while sacrificing precision and density in most cases. For LDMv3.5L, we see a more prominent effect of advanced guidance methods and prompt expansion than for LDMv1.5. Advanced guidance methods do not change the trend of the metrics \wrt prompt complexity, while the prompt expansion shapes the trend of the metrics differently.}
\label{fig:marginal_dinov2_SDexpansion_in1k}
\end{figure}

Figure~\ref{fig:marginal_dinov2_SDexpansion_in1k} shows the reference-based metrics using ImageNet-1k class labels. The findings are similar with the CC12M dataset, as presented in section~\ref{para:reference_based}.

\subsection{Trade-off of creativity and fidelity when using inference-time interventions}
\label{app:trade-off-creativity-fidelity}
\xf{Figs.~\ref{fig:marginal_dinov2_SDexpansion_cc12m} and \ref{fig:marginal_dinov2_SDexpansion_in1k} show that using prompt expansion and advanced guidance will result in a drop in precision and density, revealing an under-explored but important trade-off between creativity and fidelity. We suggest cautious usage of T2I models when applying to different downstream applications. We provide a discussion over this trade-off in the following.}

\xf{\textbf{Beneficial Scenarios:}}

\begin{enumerate}
    \item \xf{Creative ideation and co-creation. Beyond simple augmentation, prompt expansion acts as an exploratory partner for artists or designers. It uses the LLM's “imagination” to brainstorm novel, high-aesthetic compositions that diverge from the dataset's common tropes.}
    \item \xf{Combating prompt-level mode collapse. For simple prompts (e.g., a dog), T2I models can default to a typical representation (e.g., a golden retriever). Prompt expansion explicitly forces intra-class diversity by generating a set of specific instances, preventing the model from settling on a single, average output.}
    \item \xf{Training for OOD Robustness: “hallucinations” create data that is plausible but statistically rare in the reference set, making it a useful source for training downstream models that are robust to out-of-distribution contexts.}
\end{enumerate}

\xf{\textbf{Detrimental Scenarios:}}
\begin{enumerate}
    \item \xf{Bias amplification and stereotype entrenchment: This is a critical risk we will add. The LLM's prior, used for expansion, is not neutral. It can inject societal biases (e.g., related to gender or race for prompts like "a doctor" or "a CEO"), amplifying stereotypes by hard-coding them into the prompt before the T2I model ever runs.}
    \item \xf{Loss of user intent and control. When a user desires generality (e.g., “a simple icon of a bird”), prompt expansion’s forced specificity is a failure to respect the user's intent. It overrides their desire for an “average” or “iconic” representation.}
\end{enumerate}

\subsection{Statistical significance of the trends}
\label{app:95ci}
\xf{To ensure the statistical significnace of the observed trends, we compute the mean and 95\% confidence interval of the reference-free metrics across complexities using 5,000 randomly sampled CC12M prompts. The results are shown in Table~\ref{tab:95ci_diversity}, \ref{tab:95ci_quality}, and \ref{tab:95ci_consistency} for diversity, quality, and consistency respectively. The confidence intervals ($95\% \text{ CI}$) are narrow, indicating high stability in the measurement. Notably, the intervals between consecutive complexity levels are non-overlapping, confirming that the main trends at each complexity are statistically significant.}

\begin{table}[!htbp]
\centering
\caption{95\% Confidence Interval for Diversity using CC12M prompts}
\label{tab:95ci_diversity}
\small
\begin{tabular}{@{}lllll@{}}
\toprule
complexity             & 1            & 2            & 3            & 4           \\ \midrule
\textbf{LDMv1.5} + CFG          & 9.228±0.094  & 7.649±0.079  & 6.957±0.067  & 6.483±0.059 \\
+ APG                   & 10.163±0.093 & 8.701±0.081  & 7.921±0.070  & 7.496±0.063 \\
+ CADS                  & 11.112±0.095 & 9.557±0.081  & 8.496±0.072  & 7.902±0.064 \\
+ Interval              & 11.934±0.089 & 10.378±0.082 & 9.314±0.070  & 8.839±0.066 \\
+ Prompt expansion, CFG & 15.074±0.089 & 12.977±0.097 & 10.882±0.092 & 9.725±0.081 \\
\textbf{LDMv3.5L} + CFG   & 5.597±0.053  & 5.493±0.044  & 4.294±0.032  & 3.966±0.028 \\
+ APG                   & 8.978±0.090  & 7.331±0.079  & 6.089±0.065  & 5.302±0.054 \\
+ CADS                  & 9.864±0.108  & 7.690±0.093  & 5.954±0.067  & 5.214±0.056 \\
+ Interval              & 10.707±0.088 & 9.463±0.086  & 7.668±0.073  & 6.788±0.064 \\
+ Prompt expansion, CFG & 14.671±0.098 & 12.114±0.107 & 9.140±0.100  & 7.672±0.083 \\ \bottomrule
\end{tabular}
\end{table}

\begin{table}[!htbp]
\centering
\caption{95\% Confidence Interval for Quality using CC12M prompts}
\label{tab:95ci_quality}
\small
\begin{tabular}{@{}lllll@{}}
\toprule
complexity             & 1           & 2           & 3           & 4           \\ \midrule
\textbf{LDMv1.5} + CFG          & 4.085±0.011 & 4.297±0.013 & 4.335±0.012 & 4.357±0.012 \\
+ APG                   & 4.035±0.011 & 4.213±0.012 & 4.214±0.011 & 4.233±0.011 \\
+ CADS                  & 4.013±0.011 & 4.195±0.012 & 4.239±0.011 & 4.254±0.011 \\
+ Interval              & 4.017±0.009 & 4.163±0.012 & 4.204±0.011 & 4.213±0.011 \\
+ Prompt expansion, CFG & 4.645±0.009 & 4.620±0.009 & 4.472±0.010 & 4.415±0.010 \\
\textbf{LDMv3.5L} + CFG               & 5.011±0.011 & 5.155±0.012 & 5.016±0.013 & 4.981±0.013 \\
+ APG                   & 5.071±0.009 & 5.179±0.010 & 5.012±0.012 & 4.967±0.012 \\
+ CADS                  & 5.019±0.010 & 5.118±0.011 & 4.977±0.012 & 4.940±0.013 \\
+ Interval              & 5.042±0.008 & 5.101±0.009 & 4.987±0.009 & 4.924±0.011 \\
+ Prompt expansion, CFG & 5.238±0.008 & 5.229±0.009 & 5.110±0.010 & 5.072±0.010 \\ \bottomrule
\end{tabular}
\end{table}

\begin{table}[!htbp]
\centering
\caption{95\% Confidence Interval for Consistency using CC12M prompts}
\label{tab:95ci_consistency}
\small
\begin{tabular}{@{}lllll@{}}
\toprule
complexity             & 1             & 2             & 3             & 4             \\ \midrule
\textbf{LDMv1.5} + CFG          & 0.9026±0.0026 & 0.8153±0.0029 & 0.7630±0.0025 & 0.7175±0.0022 \\
+ APG                   & 0.8853±0.0028 & 0.8011±0.0030 & 0.7520±0.0026 & 0.7036±0.0023 \\
+ CADS                  & 0.8724±0.0029 & 0.7928±0.0030 & 0.7438±0.0026 & 0.6981±0.0023 \\
+ Interval              & 0.8573±0.0030 & 0.7788±0.0031 & 0.7313±0.0026 & 0.6822±0.0023 \\
+ Prompt expansion, CFG & 0.8072±0.0034 & 0.7130±0.0034 & 0.6902±0.0028 & 0.6502±0.0024 \\
\textbf{LDMv3.5L} + CFG               & 0.9235±0.0023 & 0.8404±0.0028 & 0.8001±0.0024 & 0.7622±0.0025 \\
+ APG                   & 0.9063±0.0025 & 0.8270±0.0029 & 0.7983±0.0024 & 0.7606±0.0021 \\
+ CADS                  & 0.8546±0.0031 & 0.7881±0.0031 & 0.7636±0.0025 & 0.7251±0.0022 \\
+ Interval              & 0.8677±0.0029 & 0.7901±0.0031 & 0.7715±0.0025 & 0.7378±0.0022 \\
+ Prompt expansion, CFG & 0.8473±0.0031 & 0.7494±0.0032 & 0.7432±0.0026 & 0.7083±0.0023 \\ \bottomrule
\end{tabular}
\end{table}

\subsection{Closer look at models and guidances}
\label{app:closer_look}
\begin{figure}[t]
    \centering
    \begin{subfigure}[ht]{0.2\textwidth}
        \includegraphics[width=\linewidth]{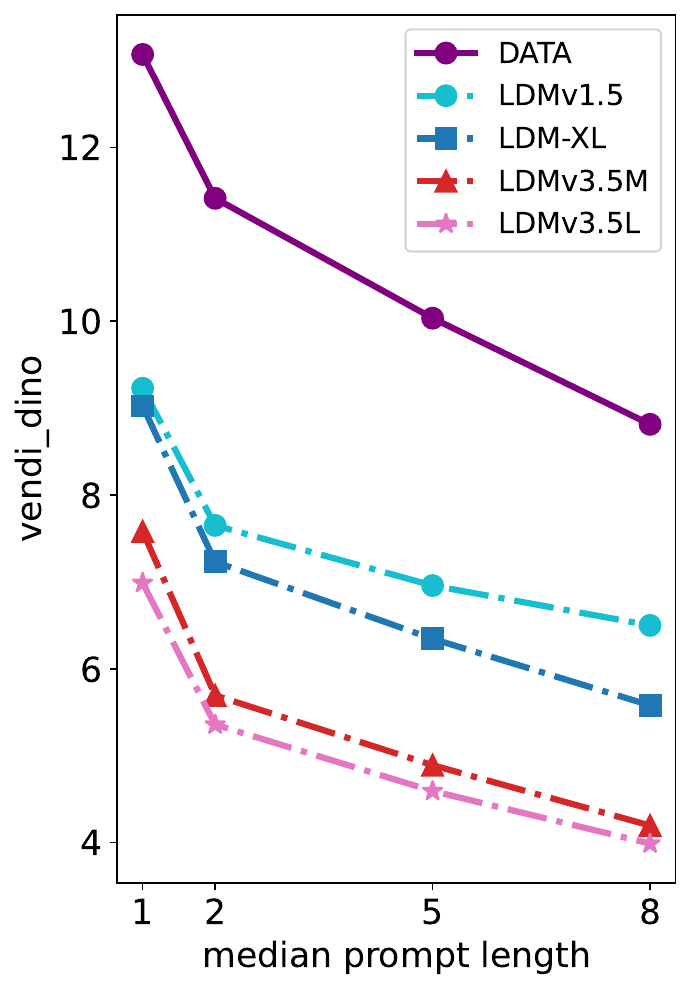}
        \caption{Diversity}
        \label{fig:vendi_dinov2_model_cc12m}
    \end{subfigure}
    \begin{subfigure}[ht]{0.2\textwidth}
        \includegraphics[width=\textwidth]{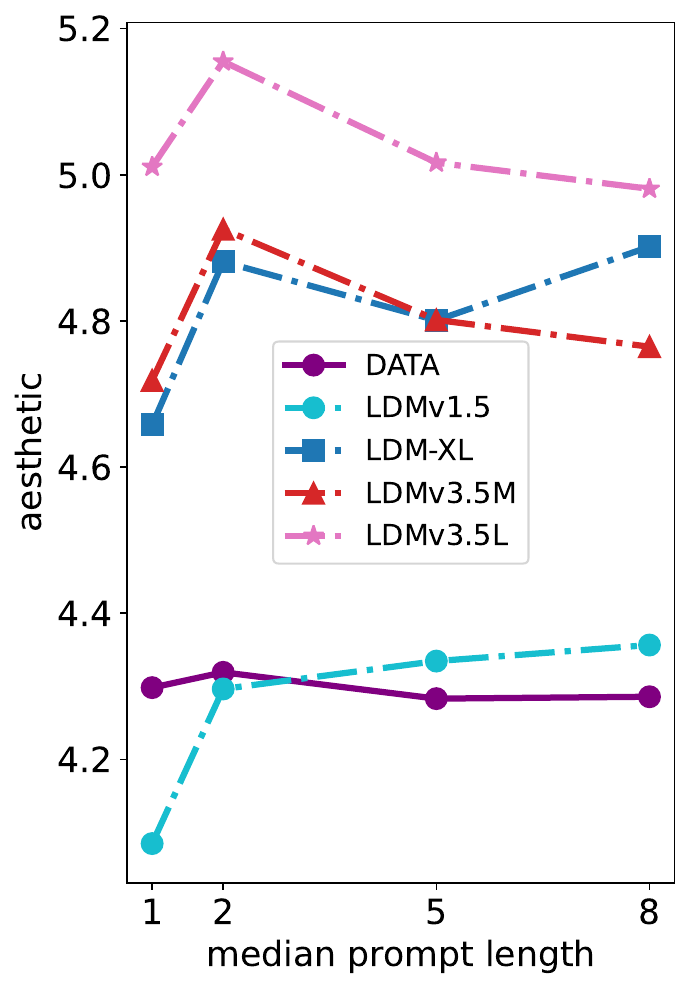}
        \caption{Quality}
        \label{fig:aesthetic_model_cc12m}
    \end{subfigure}
    \begin{subfigure}[ht]{0.2\textwidth}
        \includegraphics[width=\textwidth]{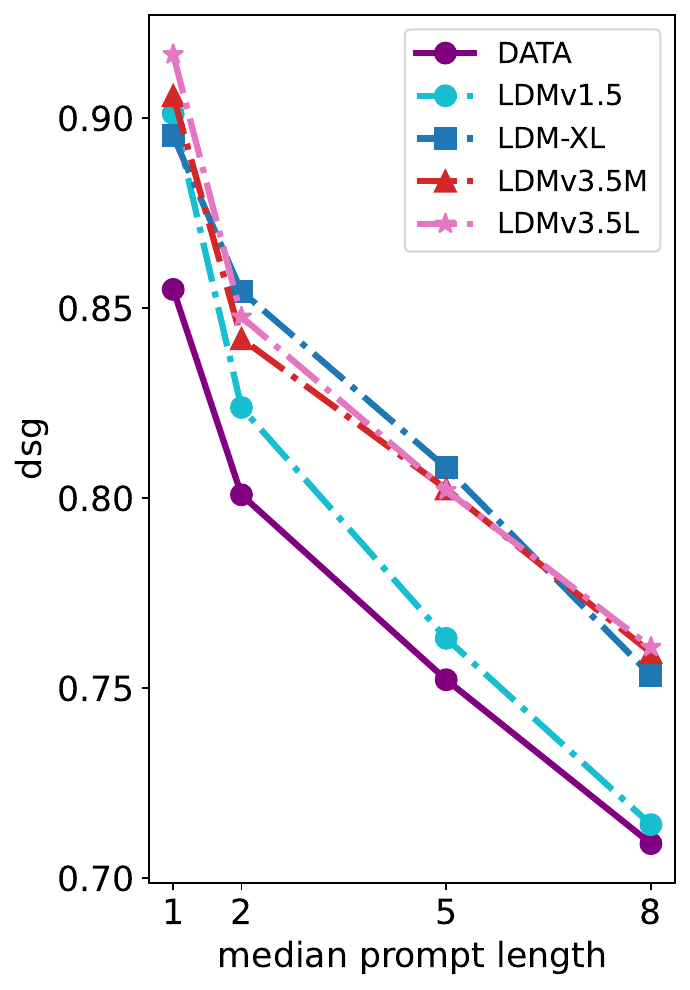}
        \caption{Consistency}
        \label{fig:dsg_model_cc12m}
    \end{subfigure}
    \caption{\textbf{Reference-free utility metrics of synthetic data from different LDM models using CC12M prompts.} We show Vendi score, aesthetic score, and DSG score using different LDM models with CFG. We see a decrease of diversity across model release time -- \ie more recent models tend to have lower diversity. Interestingly, LDMv3.5L model shows less diversity than LDMv3.5M model. When it comes to image quality (aesthetics score), LDMv3.5L exhibits the best performance, followed by LDM-XL and LDMv3.5M. LDMv1.5 shows the worst aesthetic score, which is however similar to that of the dataset. When it comes to image-prompt consistency, all models show relatively high DSG scores. However, LDMv1.5's consistency drops more sharply than the consistency of other models.}
    \label{fig:CFG_model_cc12m}
\end{figure}

\paragraph{A closer look at different models.}
Figure~\ref{fig:CFG_model_cc12m} presents reference-free metrics (Vendi, aesthetic, and DSG scores) comparing different LDM models on the CC12M dataset when using CFG guidance. We observe that Vendi score and aesthetics score follow opposite model ranking trends, with LDMv3.5L showing the best aesthetic quality and the lowest diversity performance. As for image-prompt consistency, we observe that LDMv1.5 model falls short from other models by a large margin.

\begin{figure}[t]
    \centering
    \begin{subfigure}[ht]{0.2\textwidth}
        \includegraphics[width=\textwidth]{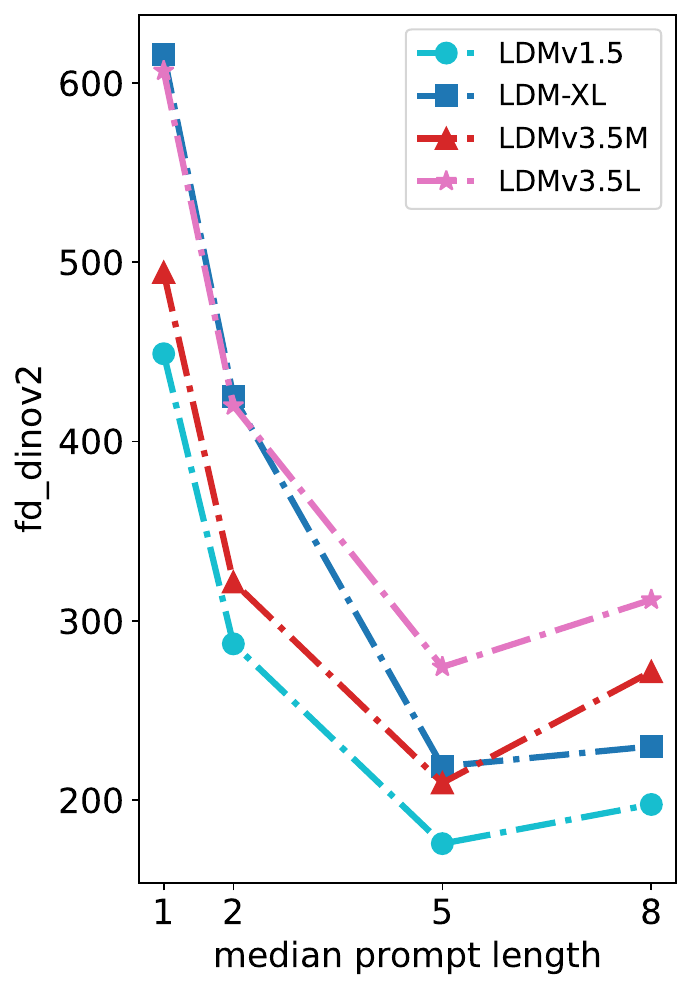}
        \caption{FDD}
        \label{fig:fd_dinov2_model_cc12m}
    \end{subfigure}
    \begin{subfigure}[ht]{0.2\textwidth}
        \includegraphics[width=\linewidth]{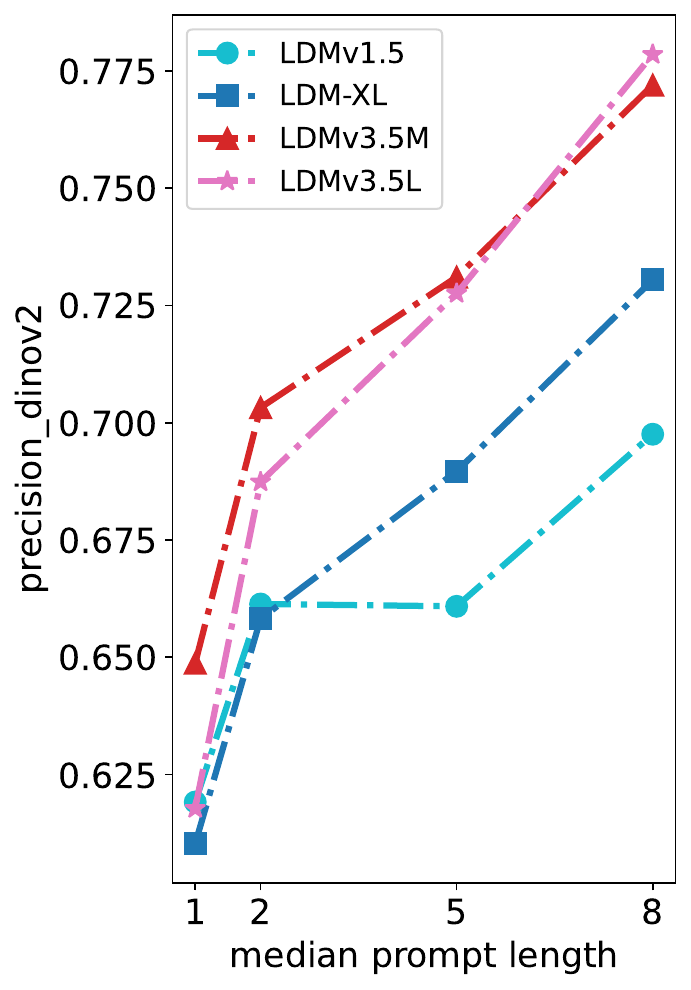}
        \caption{Precision}
        \label{fig:precision_dinov2_model_cc12m}
    \end{subfigure}
    \begin{subfigure}[ht]{0.2\textwidth}
        \includegraphics[width=\textwidth]{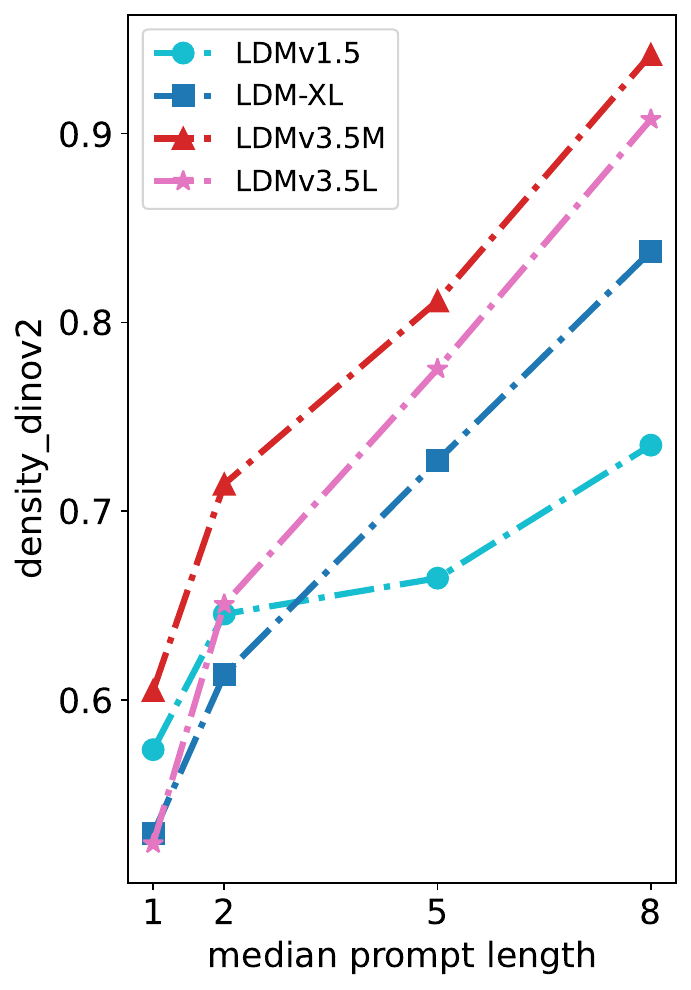}
        \caption{Density}
        \label{fig:density_dinov2_model_cc12m}
    \end{subfigure}
    \begin{subfigure}[ht]{0.2\textwidth}
        \includegraphics[width=\textwidth]{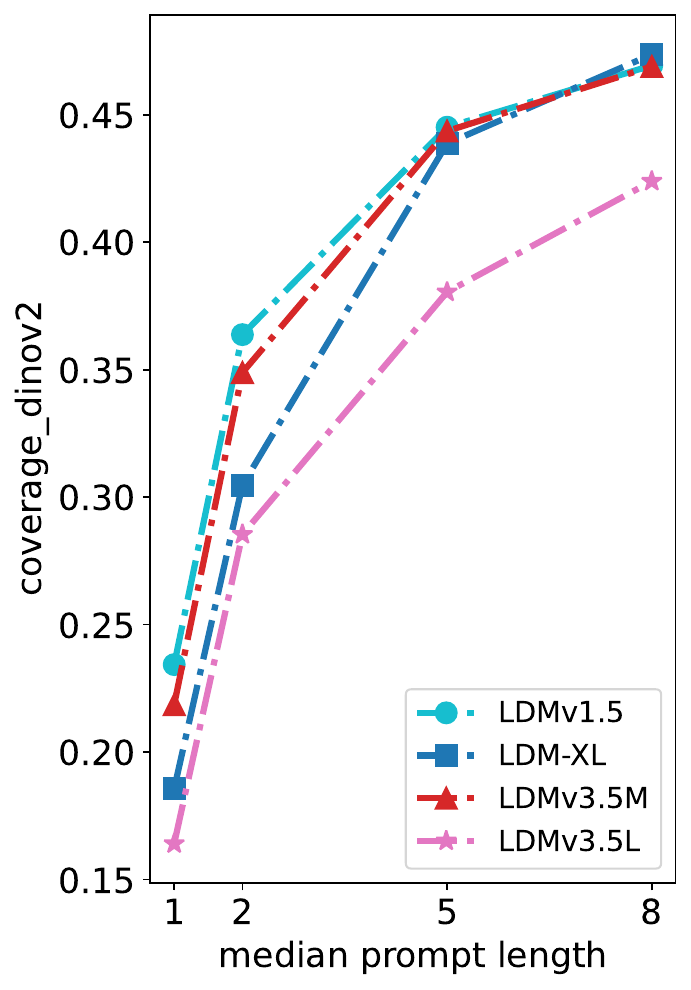}
        \caption{Coverage}
        \label{fig:coverage_dinov2_model_cc12m}
    \end{subfigure}
    \caption{\textbf{Reference-based utility metrics of synthetic data from different LDM models using CC12M prompts.} FDD, precision, density, and coverage metrics in the DINOv2 feature space for different LDM models when using CFG guidance scale 5. For FDD, LDMv1.5 shows the best performance. Interestingly, LDMv3.5M has better FDD than LDMv3.5L. As for precision and density, more recent models tend to have better performances. LDMv3.5M and LDMv1.5 models show similar coverage, with LDM-XL slightly falling behind, and LDMv3.5L presenting the lowest coverage.}
    \label{fig:CFG_model_cc12m_marginal}
\end{figure}

Figure~\ref{fig:CFG_model_cc12m_marginal} presents reference-based metrics comparing the same models. LDMv3.5M model works surprisingly well in terms of FDD compared to LDMv3.5L, considering that they both show similar reference-free diversity  and that LDMv3.5M has a lower aesthetic quality than LDMv3.5L. Looking into precision, density and coverage, we can better understand why this happens.
LDMv3.5M model shows similar precision and higher density than LDMv3.5L, while also showing higher coverage, leading to an overall better FDD. Our results on CC12M dataset show that LDMv3.5M may better represent the real data distribution than LDMv3.5L.

\begin{figure}[t]
    \centering
    \begin{subfigure}[ht]{0.2\textwidth}
        \includegraphics[width=\linewidth]{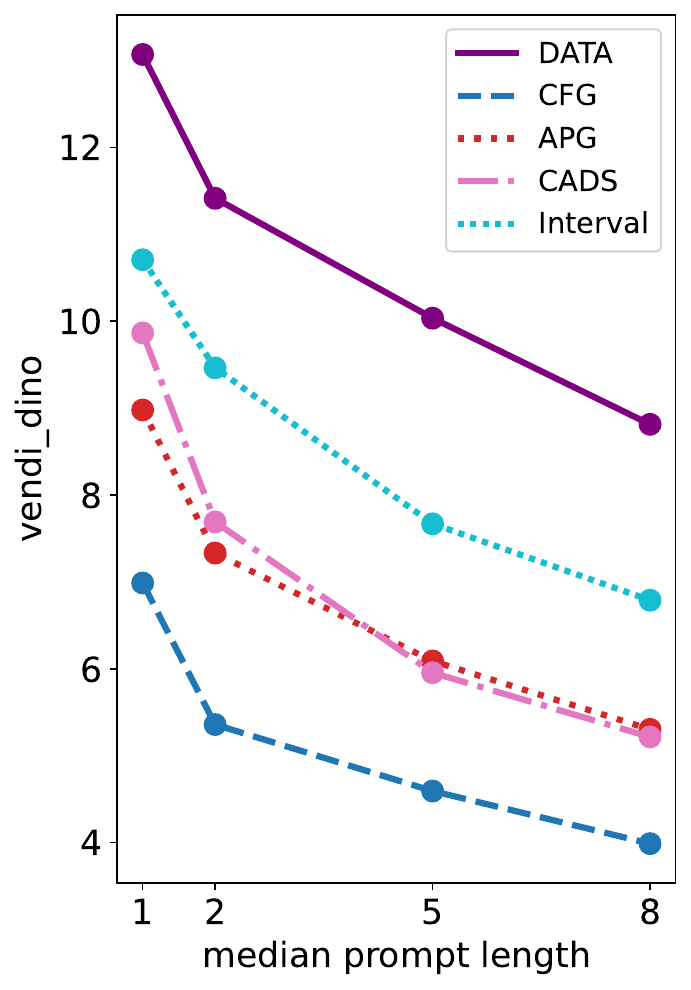}
        \caption{Diversity}
        \label{fig:vendi_dinov2_guidancemethods_cc12m}
    \end{subfigure}
    \begin{subfigure}[ht]{0.2\textwidth}
        \includegraphics[width=\textwidth]{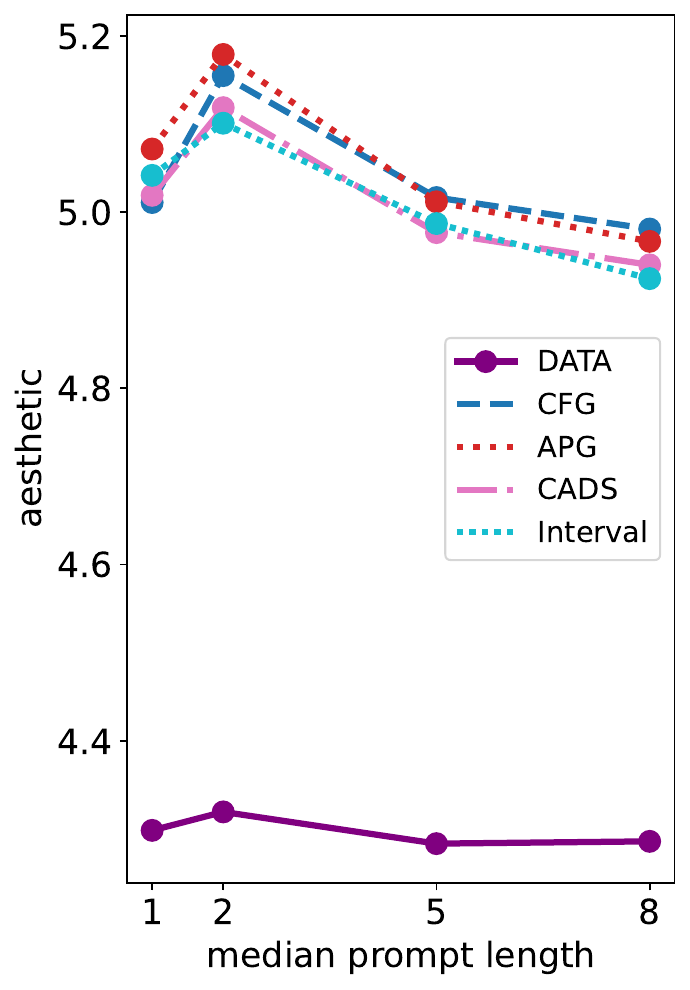}
        \caption{Quality}
        \label{fig:aesthetic_guidancemethods_cc12m}
    \end{subfigure}
    \begin{subfigure}[ht]{0.2\textwidth}
        \includegraphics[width=\textwidth]{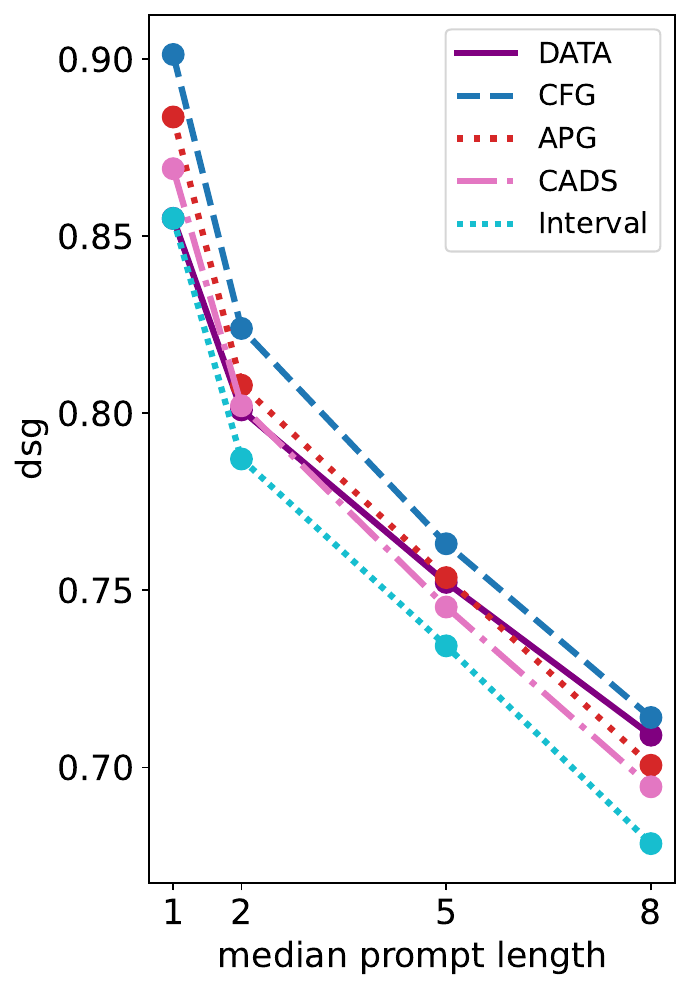}
        \caption{Consistency}
        \label{fig:dsg_guidancemethods_cc12m}
    \end{subfigure}
    \begin{subfigure}[ht]{0.2\textwidth}
        \includegraphics[width=\textwidth]{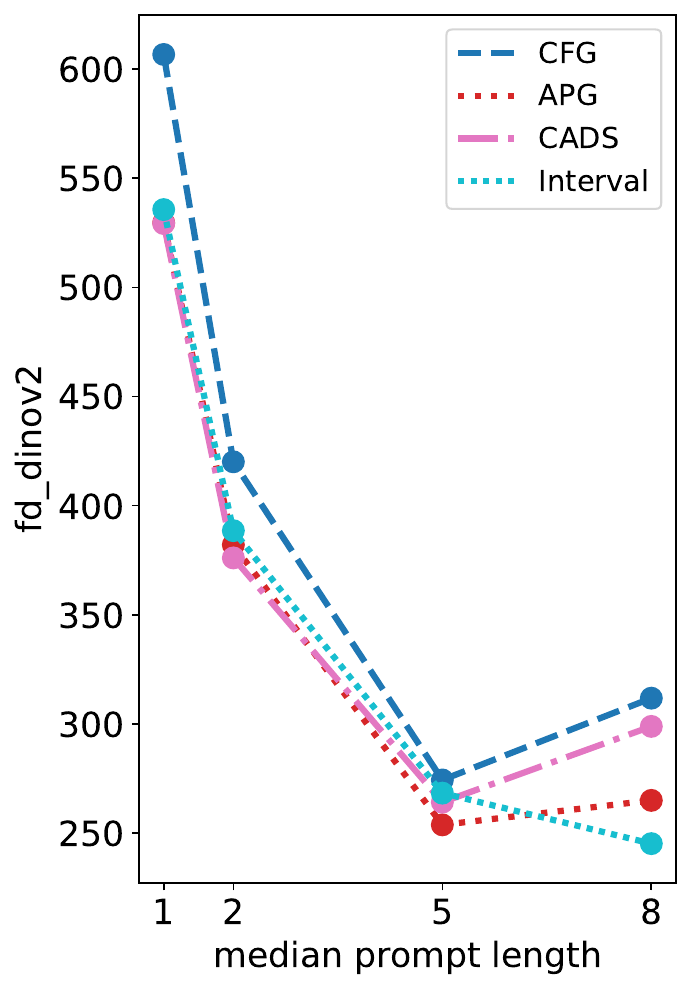}
        \caption{FDD}
        \label{fig:fd_dinov2_guidancemethods_cc12m}
    \end{subfigure}
    \caption{\textbf{Effect of guidance methods on the utility of synthetic data from LDMv3.5L using CC12M prompts.} 
    Diversity (Vendi), quality (aesthetics), consistency (DSG), and FD in DINOv2 feature space. All advanced guidance methods lead to higher diversity than the CFG baseline. APG has the smallest negative effect on consistency compared to CADS and Interval guidance. All advanced guidance methods lead to better FDD.}
    \label{fig:SD35L_guidancemethods_cc12m}
\end{figure}

\paragraph{A closer look at advanced guidance methods.} 
\label{para:guidancemethods_influence}
We now shift our focus to study the effect of different advanced guidance approaches on the utility axes of synthetic data. Figure~\ref{fig:SD35L_guidancemethods_cc12m} presents the results of reference-free metrics from LDMv3.5L model using CC12M prompts of increasing complexity. All advanced guidance methods lead to substantially higher diversity than the baseline CFG (previously referred to as vanilla guidance), with a stable aesthetic quality across different approaches. This is observed consistently across different prompt complexities. It is worth noting that APG leads to the lowest consistency drops \wrt CFG. APG primarily addresses the over-saturation problem in synthetic images by reducing the perpendicular direction guidance but still uses the conditional information at all inference steps, which might help the method achieve higher consistency than other the advanced guidance methods. However, CADS and Interval are more aggressive in removing the conditional information during the generation process, leading to overall lower consistencies across prompt complexities. As for reference-based metrics, all advanced guidance methods lead to better FDD. These results suggest that the advanced guidance methods may effectively improve the diversity of synthetic images while maintaining a stable quality and sufficiently good consistency, without sacrificing the overall reference-based metric (FDD). Finally, when contrasting with real data utility, we observe that: 1) None of the advanced guidance techniques bridges the gap between the diversity resulting from vanilla CFG guidance and the real data diversity for LDMv3.5L model. 2) All advanced guidance techniques have similar aesthetics, which is in all cases superior to the one of the real data. 3) Among advanced guidance methods, APG appears to sacrifice prompt-image consistency the least.

\begin{figure}[t]
    \centering
    \begin{subfigure}[ht]{0.2\textwidth}
        \includegraphics[width=\textwidth]{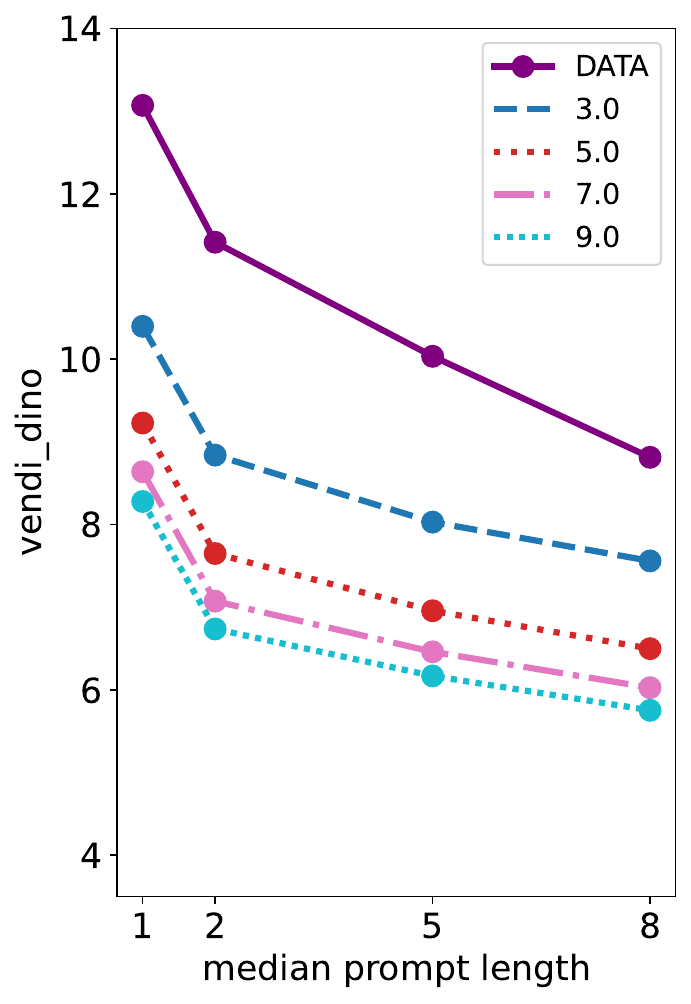}
        \caption{\centering \textcolor{white}{w} Diversity \\ \textcolor{white}{white} LDMv1.5}
        \label{fig:vendi_SD15_gscales_cc12m}
    \end{subfigure}
    \begin{subfigure}[ht]{0.2\textwidth}
        \includegraphics[width=\linewidth]{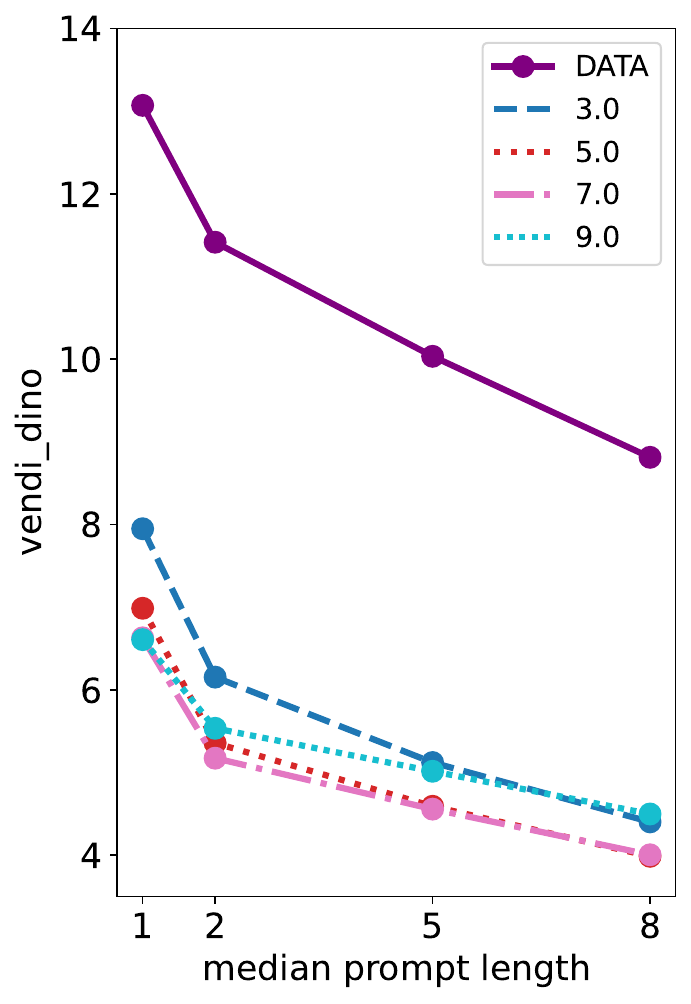}
        \caption{\centering \textcolor{white}{w} Diversity \\ \textcolor{white}{white} LDMv3.5L}
        \label{fig:vendi_SD35L_gscales_cc12m}
    \end{subfigure}
    \begin{subfigure}[ht]{0.2\textwidth}
        \includegraphics[width=\linewidth]{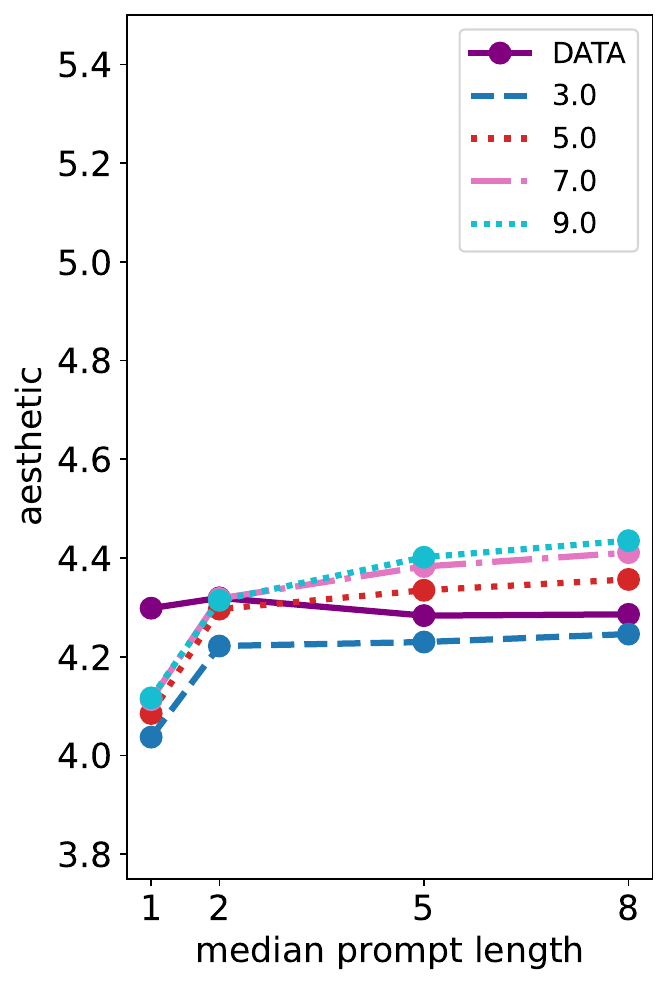}
        \caption{\centering \textcolor{white}{w} Quality \\ \textcolor{white}{white} LDMv1.5}
        \label{fig:aesthetic_SD15_gscales_cc12m}
    \end{subfigure}
    \begin{subfigure}[ht]{0.2\textwidth}
        \includegraphics[width=\textwidth]{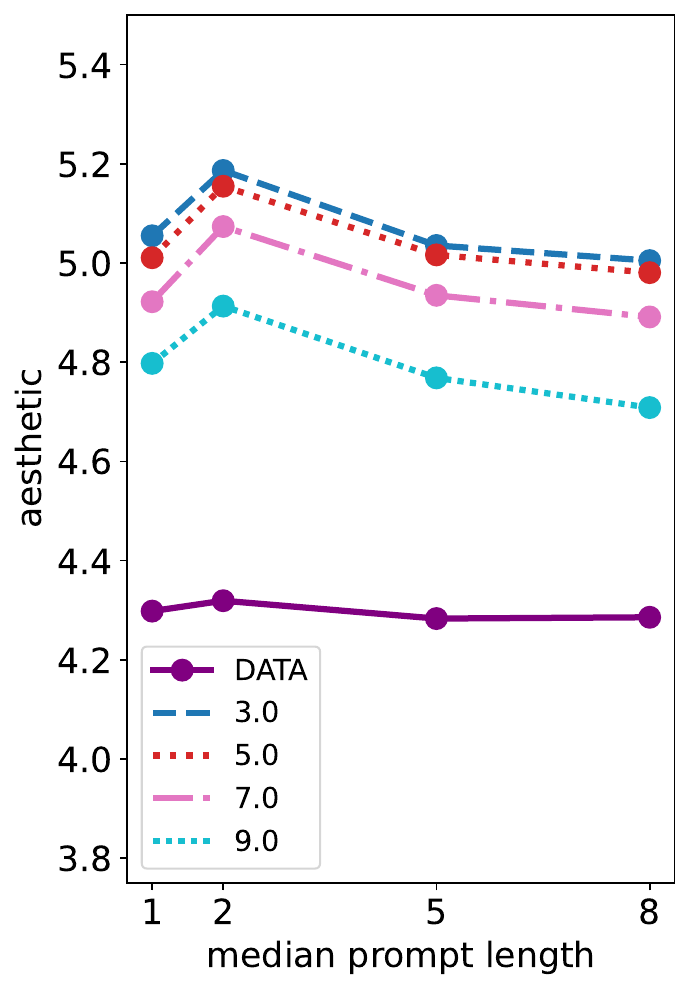}
        \caption{\centering \textcolor{white}{w} Quality \\ \textcolor{white}{white} LDMv3.5L}
        \label{fig:aesthetic_SD35L_gscales_cc12m}
    \end{subfigure}
    \caption{\textbf{Effect of guidance scale on diversity (Vendi) and quality (aesthetic) of synthetic images from CC12M prompts.} Metrics are computed in the DINOv2 feature space. For LDMv1.5, the diversity decreases as we increase the guidance scale, while for LDMv3.5L, the diversity first decreases and then increases again when using the highest guidance scale. LDMv1.5 shows better aesthetic quality when increasing the guidance scale while LDMv3.5L shows the opposite trend.}
    \label{fig:vendi_aesthetic_gscales__SD_cc12m}
\end{figure}

\begin{figure}[t]
    \centering
    \begin{subfigure}[ht]{0.2\textwidth}
        \includegraphics[width=\textwidth]{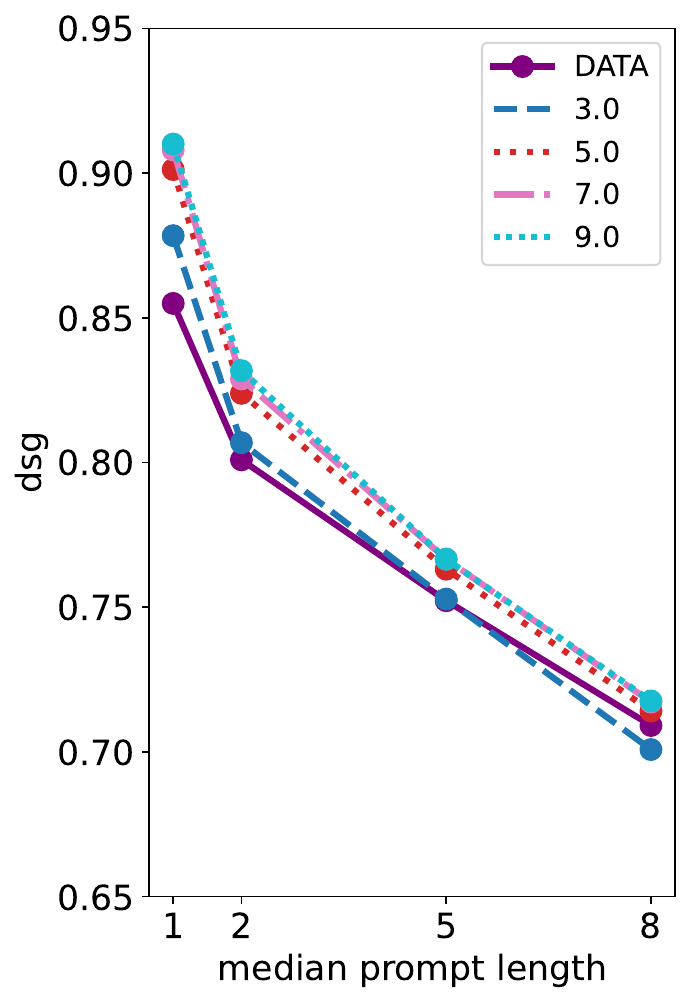}
        \caption{\centering \textcolor{white}{w} Consistency \\ \textcolor{white}{white} LDMv1.5}
        \label{fig:dsg_SD15_gscales_cc12m}
    \end{subfigure}
    \begin{subfigure}[ht]{0.2\textwidth}
        \includegraphics[width=\linewidth]{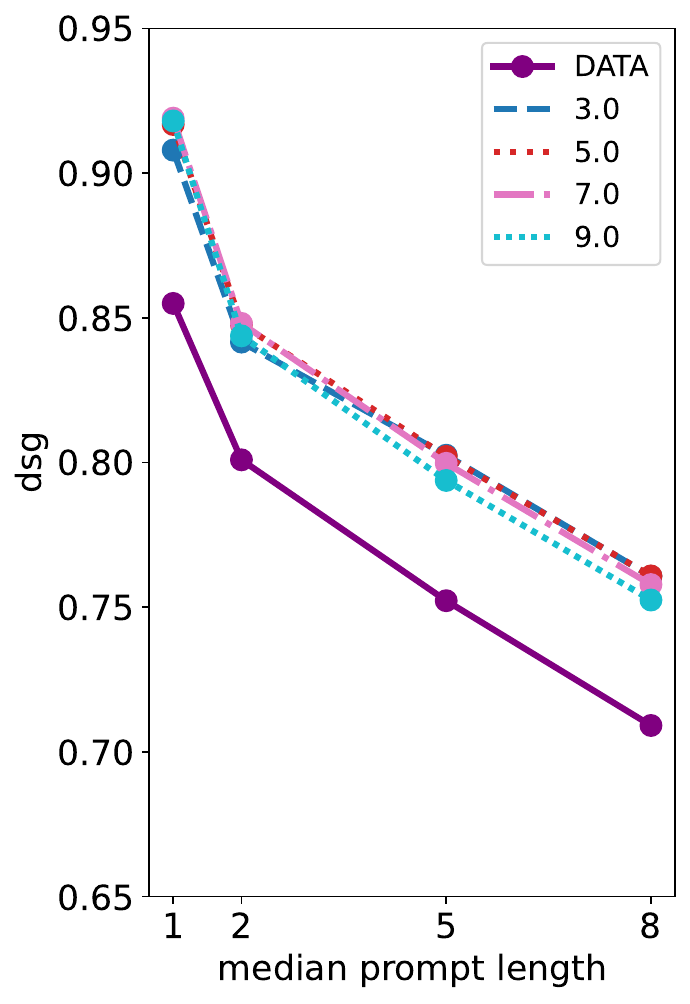}
        \caption{\centering \textcolor{white}{w} Consistency \\ \textcolor{white}{white} LDMv3.5L}
        \label{fig:dsg_SD35L_gscales_cc12m}
    \end{subfigure}
    \begin{subfigure}[ht]{0.2\textwidth}
        \includegraphics[width=\linewidth]{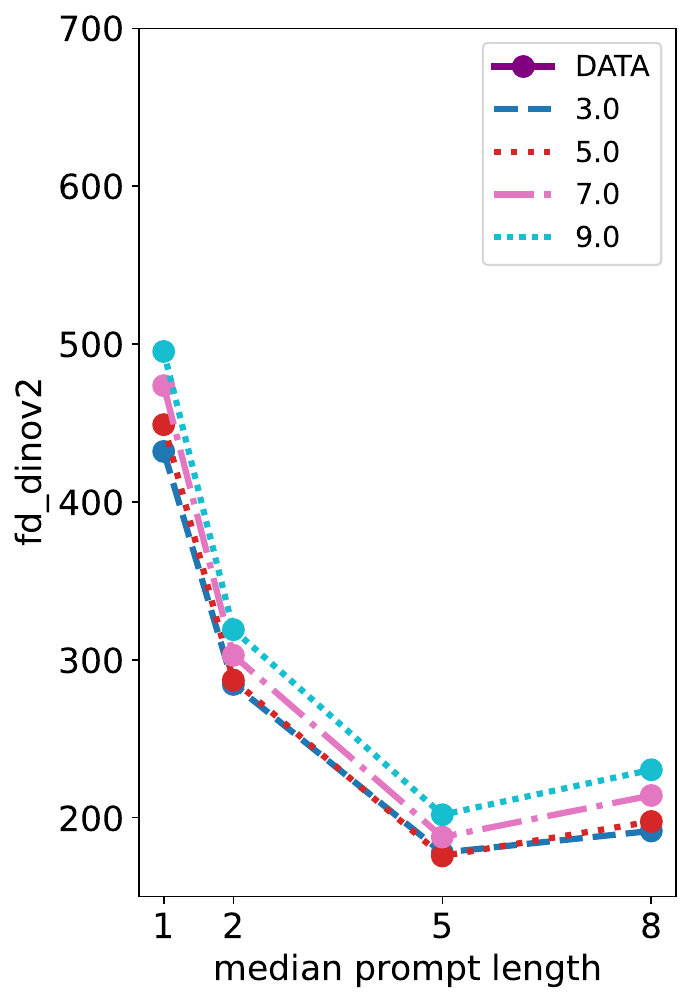}
        \caption{\centering \textcolor{white}{w} FDD \\ \textcolor{white}{white} LDMv1.5}
        \label{fig:fdd_SD15_gscales_cc12m}
    \end{subfigure}
    \begin{subfigure}[ht]{0.2\textwidth}
        \includegraphics[width=\textwidth]{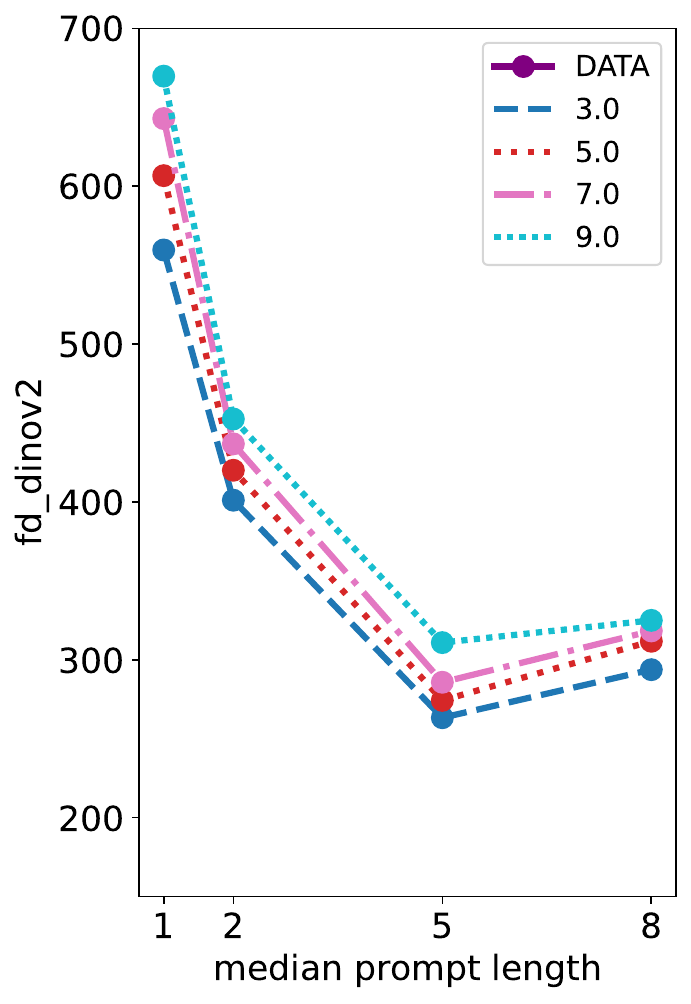}
    \caption{\centering \textcolor{white}{w} FDD \\ \textcolor{white}{white} LDMv3.5L}
        \label{fig:fdd_SD35L_gscales_cc12m}
    \end{subfigure}
    \caption{\textbf{Effect of guidance scale on image-prompt consistency (DSG) and FDD of synthetic images from CC12M prompts.} DSG score is not sensitive to guidance scale, especially for LDMv3.5L model. For LDMv1.5 model, the consistency slightly increases when employing larger guidance scales. As for FDD, both LDMv1.5 and LDMv3.5L models exhibit increase of FDD when generating images with larger guidance scale.}
    \label{fig:consistency_fdd_gscales__SD_cc12m}
\end{figure}

\paragraph{A closer look at guidance scales.}
In this paragraph, we study how guidance scale affects the utility of synthetic data. We present diversity (Vendi score) and aesthetic quality for LDMv1.5 and LDMv3.5L with guidance scales $3.0, 5.0 ,7.0$, and $9.0$ in Figure~\ref{fig:vendi_aesthetic_gscales__SD_cc12m}.
Perhaps unsurprisingly, we observe that increasing the guidance scale results in decreasing the diversity of synthetic images. However, the diversity captured by Vendi score increases for LDMv3.5L model when using very large guidance scale (\eg 9.0 in Figure~\ref{fig:vendi_SD35L_gscales_cc12m}). 
We hypothesize that this increase in Vendi score is due to the oversaturation in the synthetic images and the color range exceeding the color jittering range employed in the data augmentation process of DINOv2, making the DINOv2 features color-sensitive.
For different models and different metrics, the sensitivity to the guidance scale varies. For example, LDMv1.5 is more sensitive towards diversity while LDMv3.5L is more sensitive towards aesthetic quality as the guidance scale changes. Interestingly, LDMv1.5 has better aesthetic quality with higher guidance scales while LDMv3.5L shows the opposite trend.

We also depict consistency and FDD metrics using different guidance scales in Figure~\ref{fig:consistency_fdd_gscales__SD_cc12m}. Image-prompt consistency does not appear too sensitive to guidance scale, especially for LDMv3.5L. For LDMv1.5, the consistency slightly increases when employing larger guidance scales. As for FDD, both LDMv1.5 and LDMv3.5L exhibit increases of FDD when generating images with larger guidance scales.

\section{Additional qualitative samples}
\label{app:qualitative_samples}
\begin{figure}[!tbp]
    \centering
    \rotatebox[origin=c]{90}{\raisebox{0.5pt}{\textit{real data}}}
    \begin{subfigure}[h]{0.22\textwidth}
        \centering
        \includegraphics[width=\textwidth]{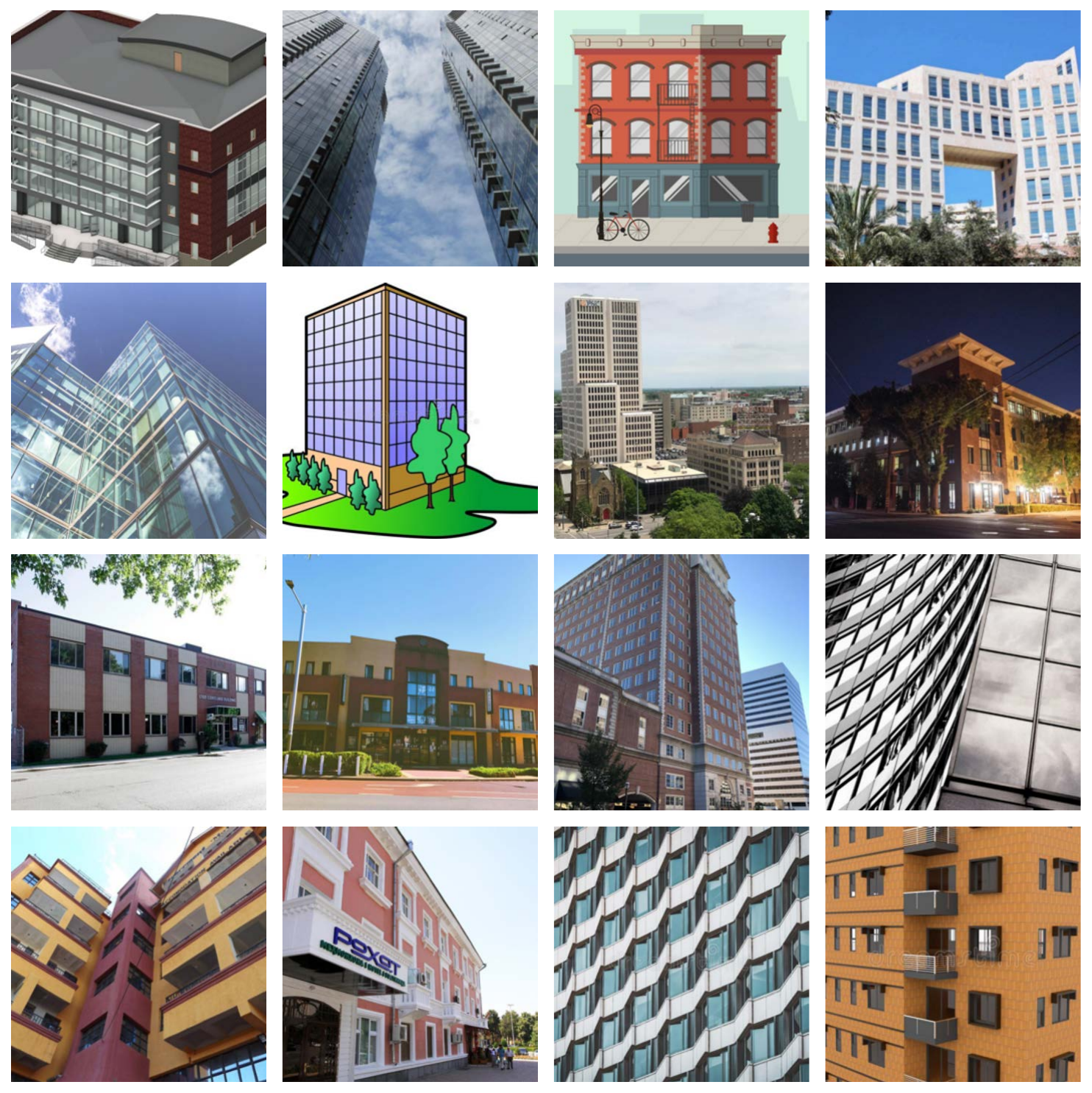}
        \captionfont{\text{15.87}}
        \label{fig:data_c1}
    \end{subfigure}
    \begin{subfigure}[h]{0.22\textwidth}
        \centering
        \includegraphics[width=\textwidth]{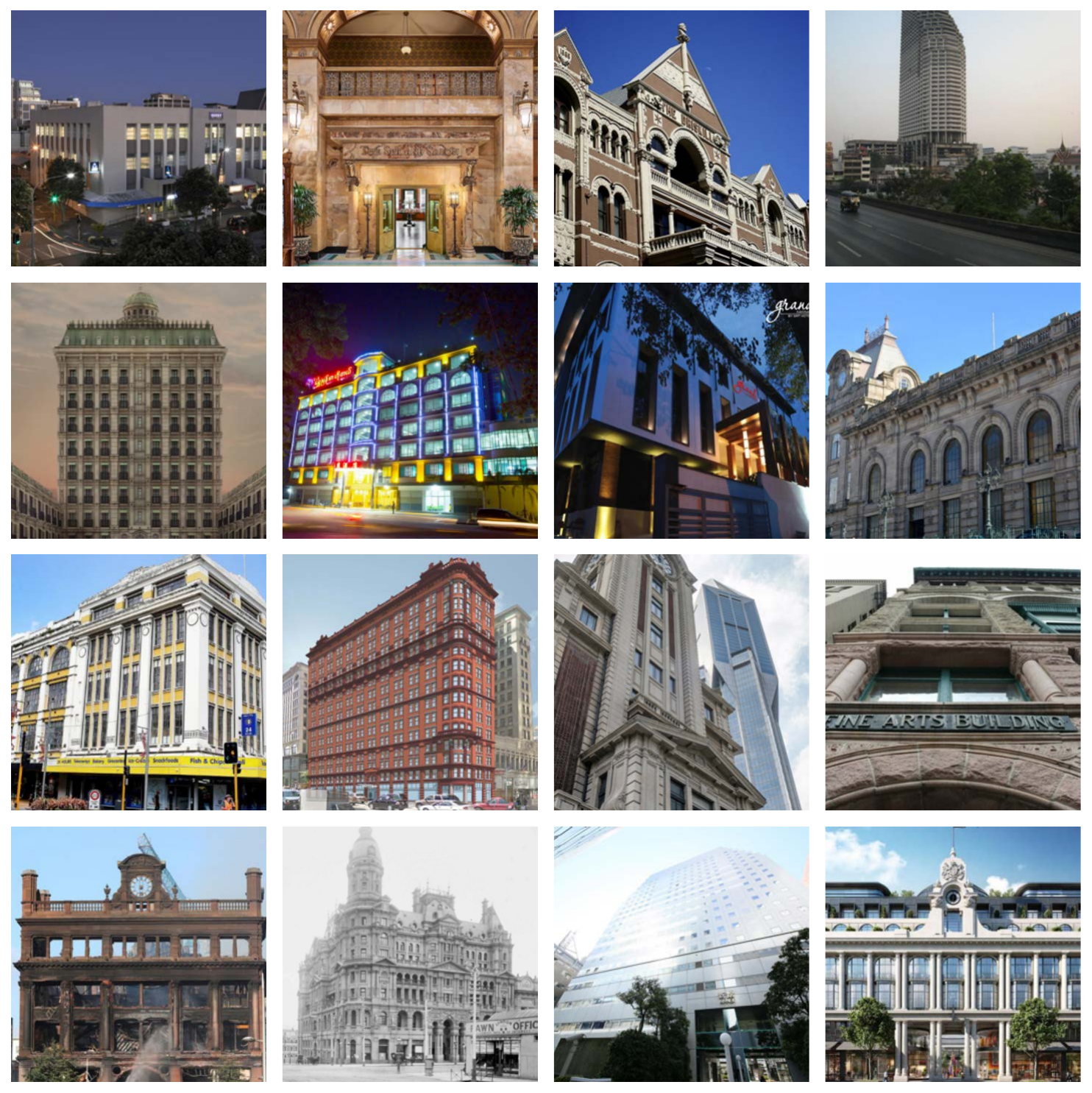}
        \captionfont{\text{15.49}}
        \label{fig:data_c2}
    \end{subfigure}
    \begin{subfigure}[h]{0.22\textwidth}
        \centering
        \includegraphics[width=\textwidth]{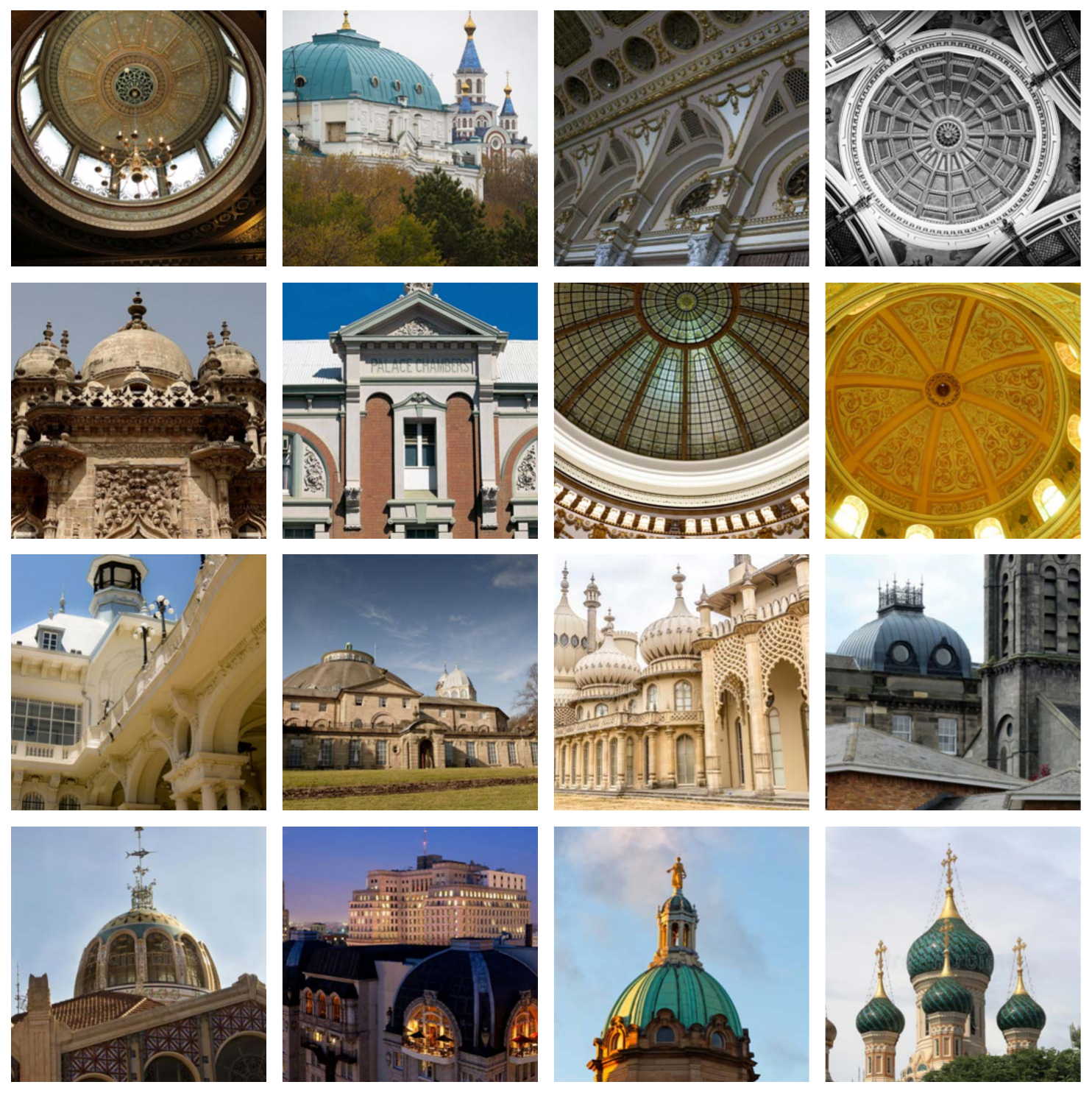}
        \captionfont{\text{14.04}}
        \label{fig:data_c3}
    \end{subfigure}
    \begin{subfigure}[h]{0.22\textwidth}
        \centering
        \includegraphics[width=\textwidth]{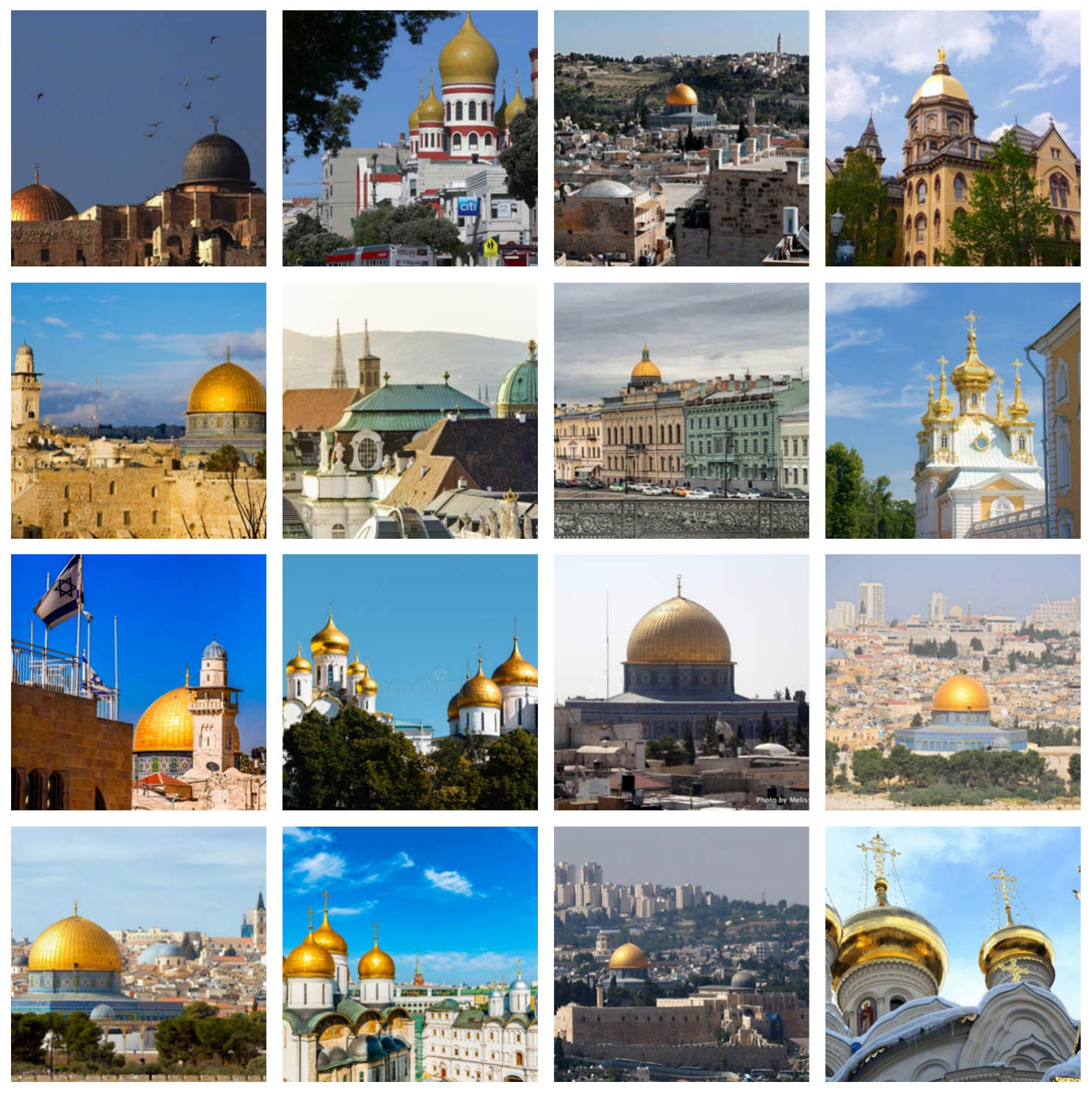}
        \captionfont{\text{8.47}}
        \label{fig:data_c4}
    \end{subfigure} \\ 
    \rotatebox[origin=c]{90}{\raisebox{0.5pt}{\textit{CFG}}}
    \begin{subfigure}[h]{0.22\textwidth}
        \centering
        \includegraphics[width=\textwidth]{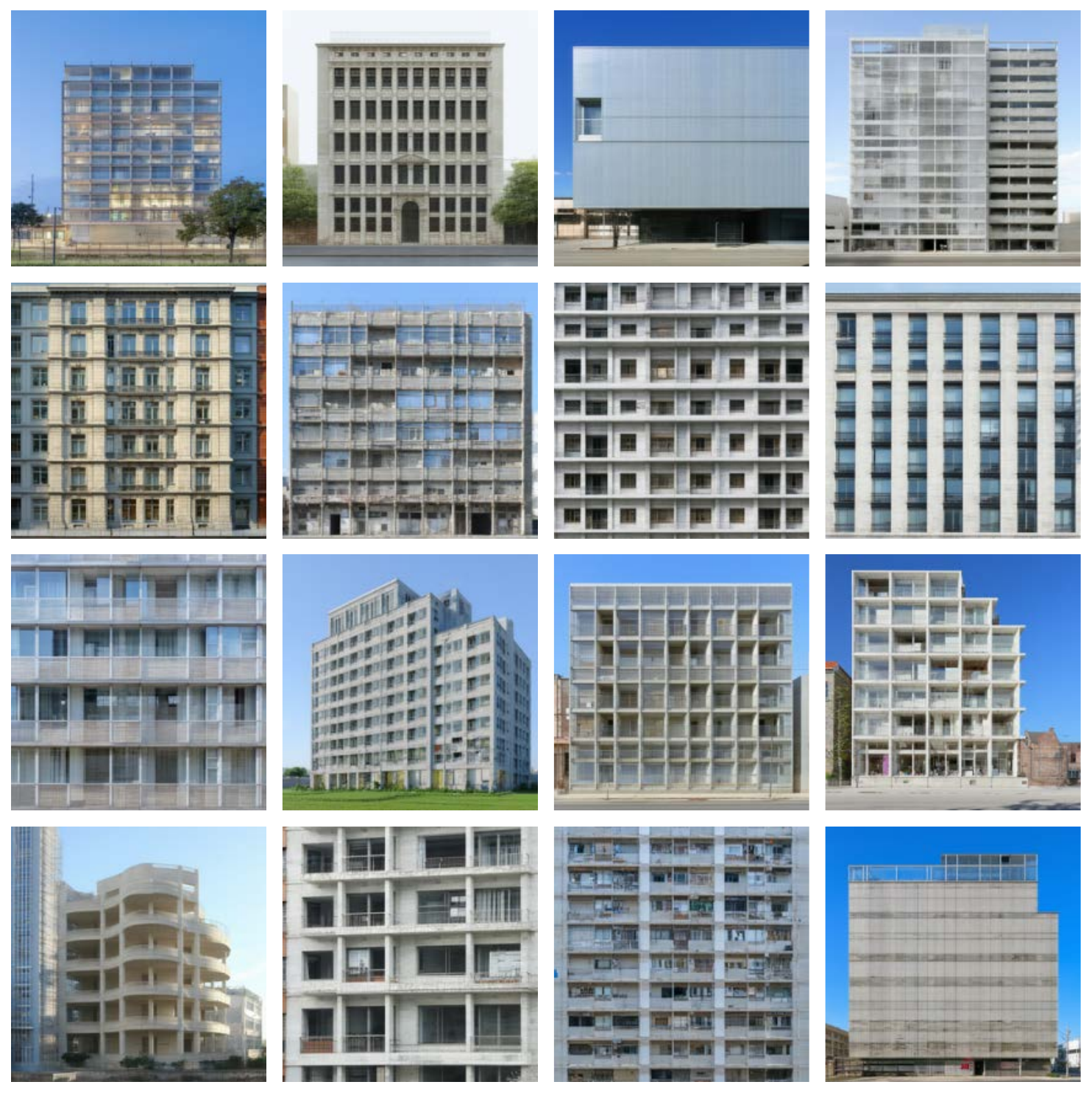}
        \captionfont{\text{8.72}}
        \label{fig:data_c0}
    \end{subfigure}
    \begin{subfigure}[h]{0.22\textwidth}
        \centering
        \includegraphics[width=\textwidth]{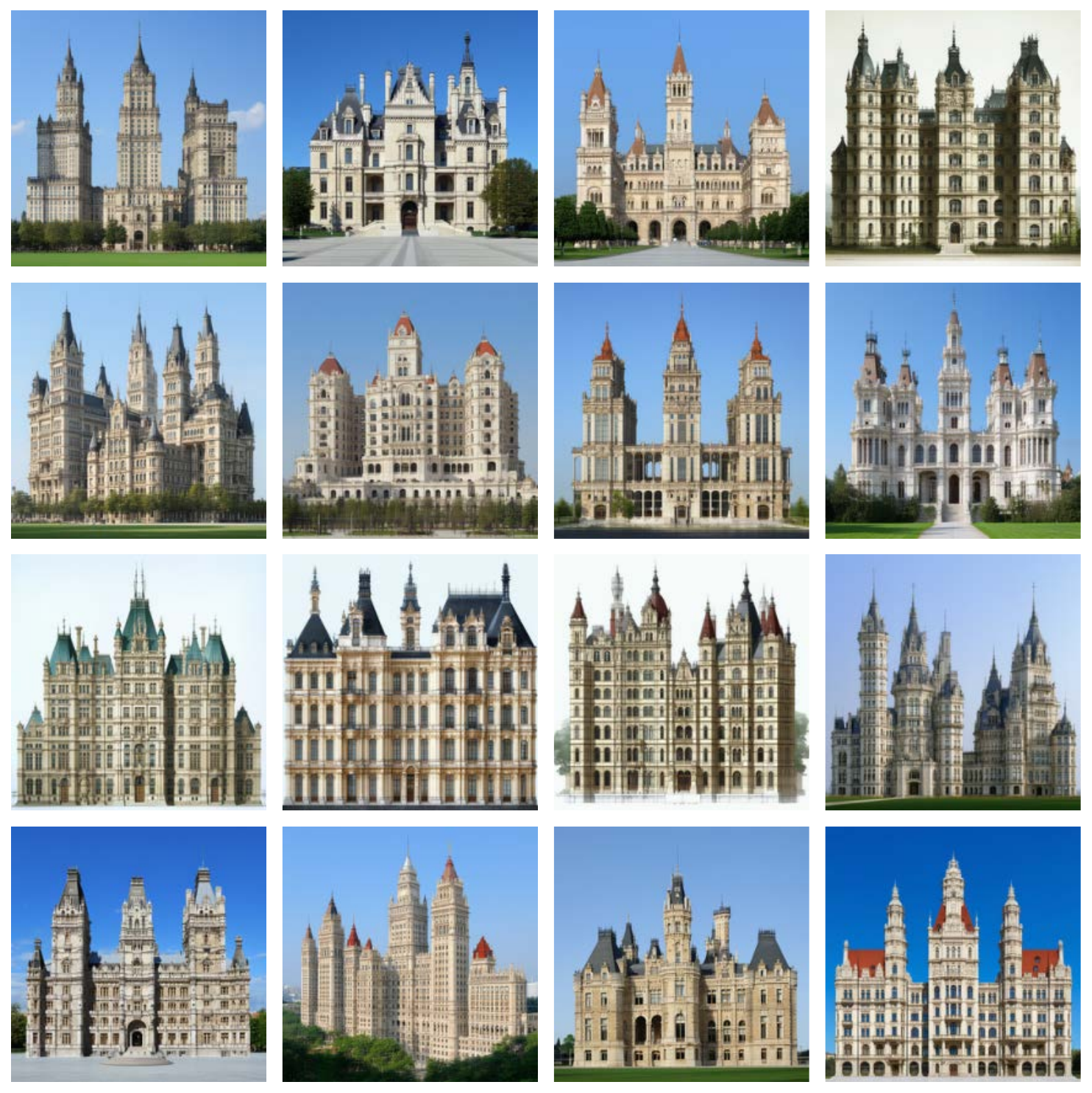}
        \captionfont{\text{6.07}}
        \label{fig:CFG_c2}
    \end{subfigure}
    \begin{subfigure}[h]{0.22\textwidth}
        \centering
        \includegraphics[width=\textwidth]{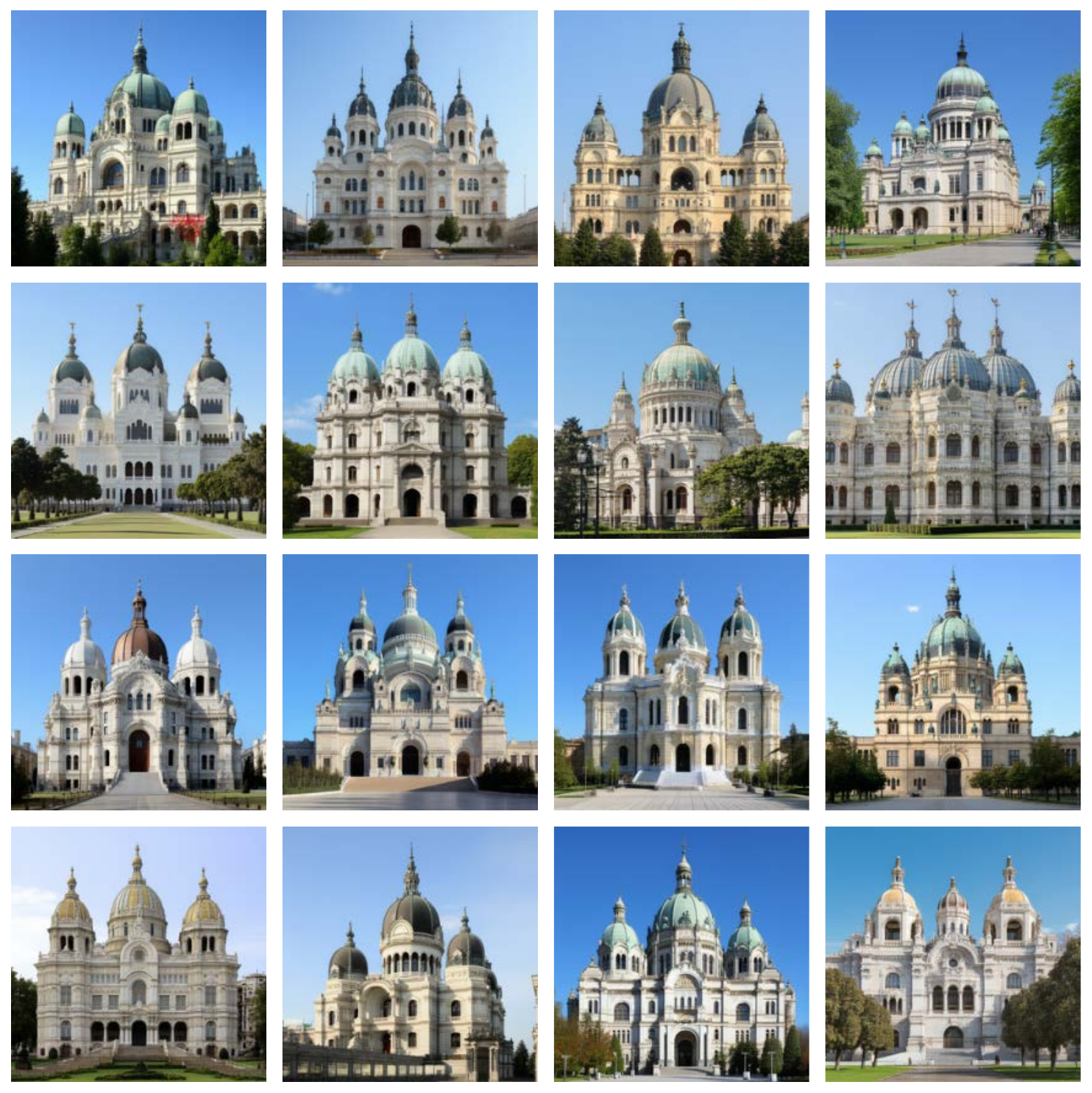}
        \captionfont{\text{4.90}}
        \label{fig:CFG_c3}
    \end{subfigure}
    \begin{subfigure}[h]{0.22\textwidth}
        \centering
        \includegraphics[width=\textwidth]{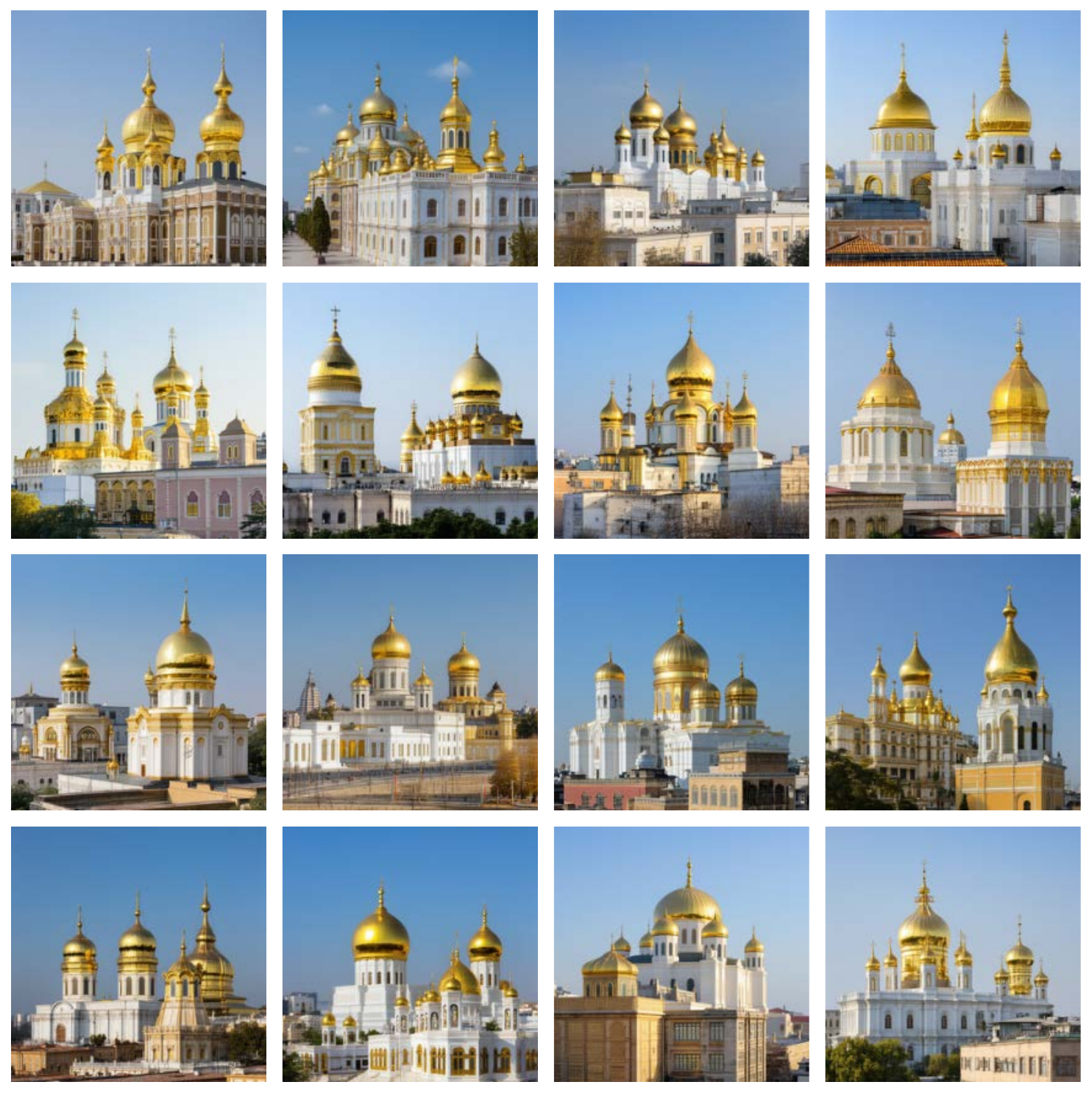}
        \captionfont{\text{4.37}}
        \label{fig:CFG_c4}
    \end{subfigure} \\ 
    \rotatebox[origin=c]{90}{\raisebox{0.5pt}{\textcolor{white}{ee} \textit{CADS guidance}}}
    \begin{subfigure}[h]{0.22\textwidth}
        \centering
        \includegraphics[width=\textwidth]{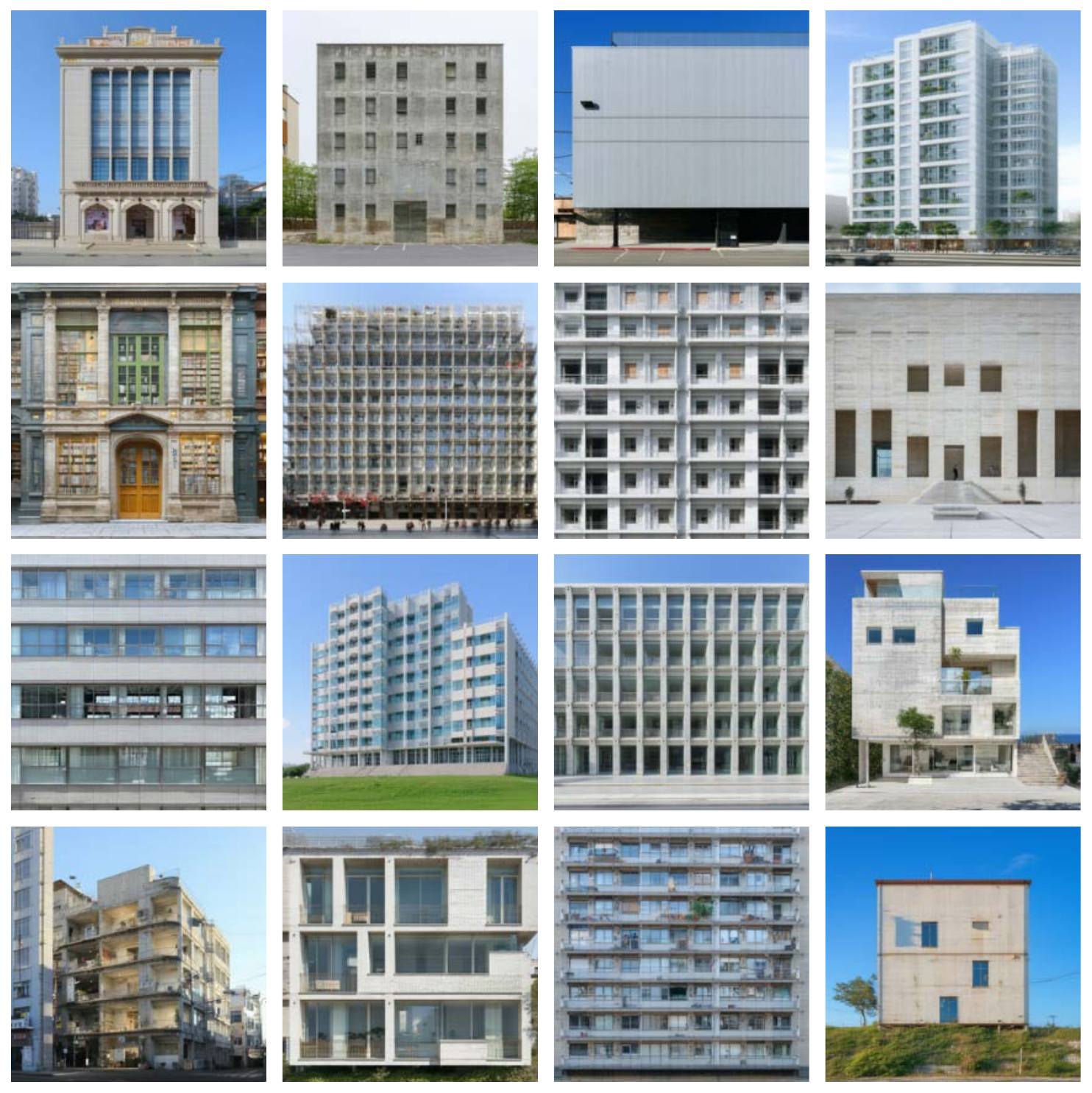}
        \captionfont{\text{11.87}}
        \label{fig:CADS_C1}
    \end{subfigure}
    \begin{subfigure}[h]{0.22\textwidth}
        \centering
        \includegraphics[width=\textwidth]{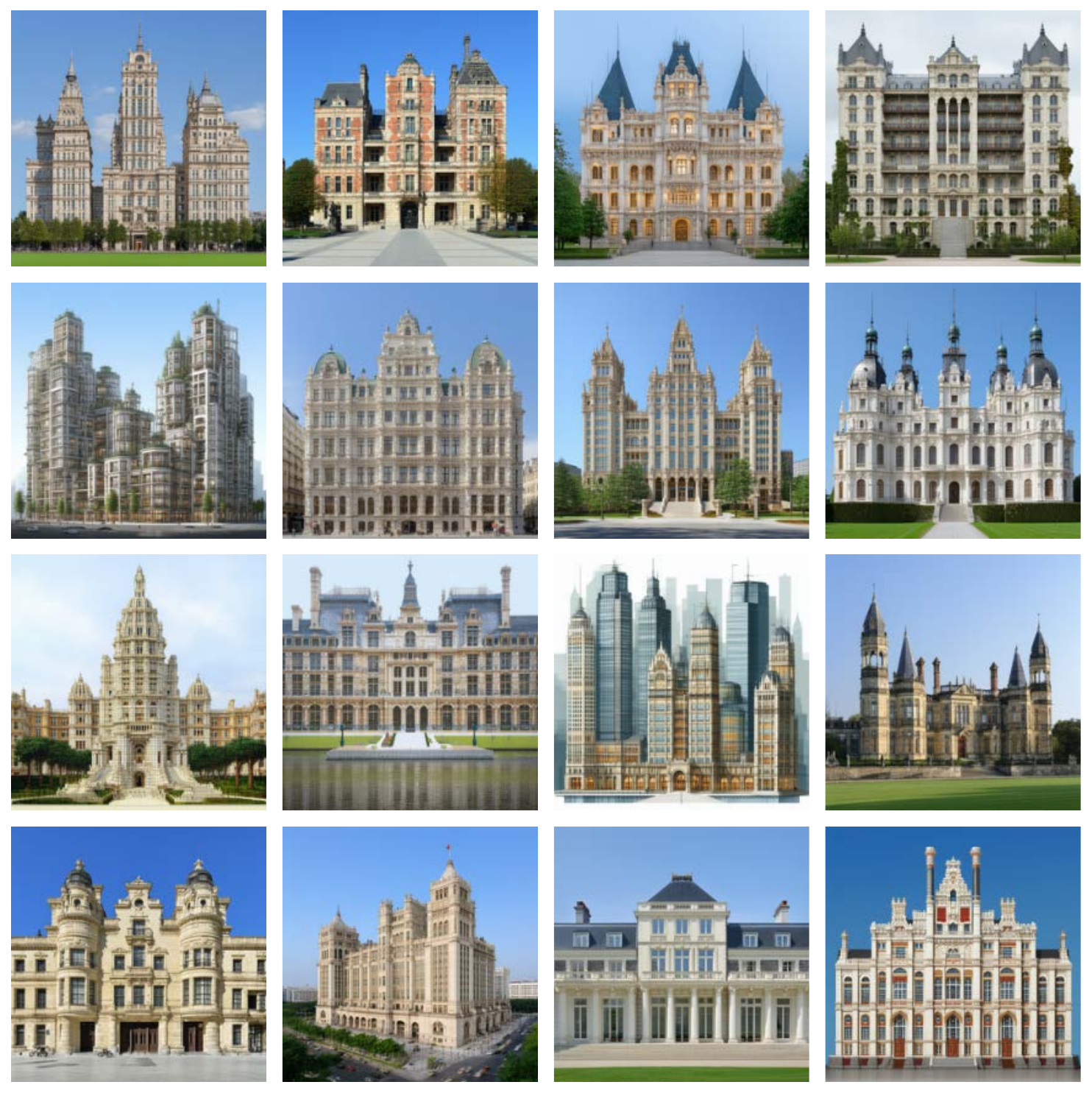}
        \captionfont{\text{10.43}}
        \label{fig:CADS_C2}
    \end{subfigure}
    \begin{subfigure}[h]{0.22\textwidth}
        \centering
        \includegraphics[width=\textwidth]{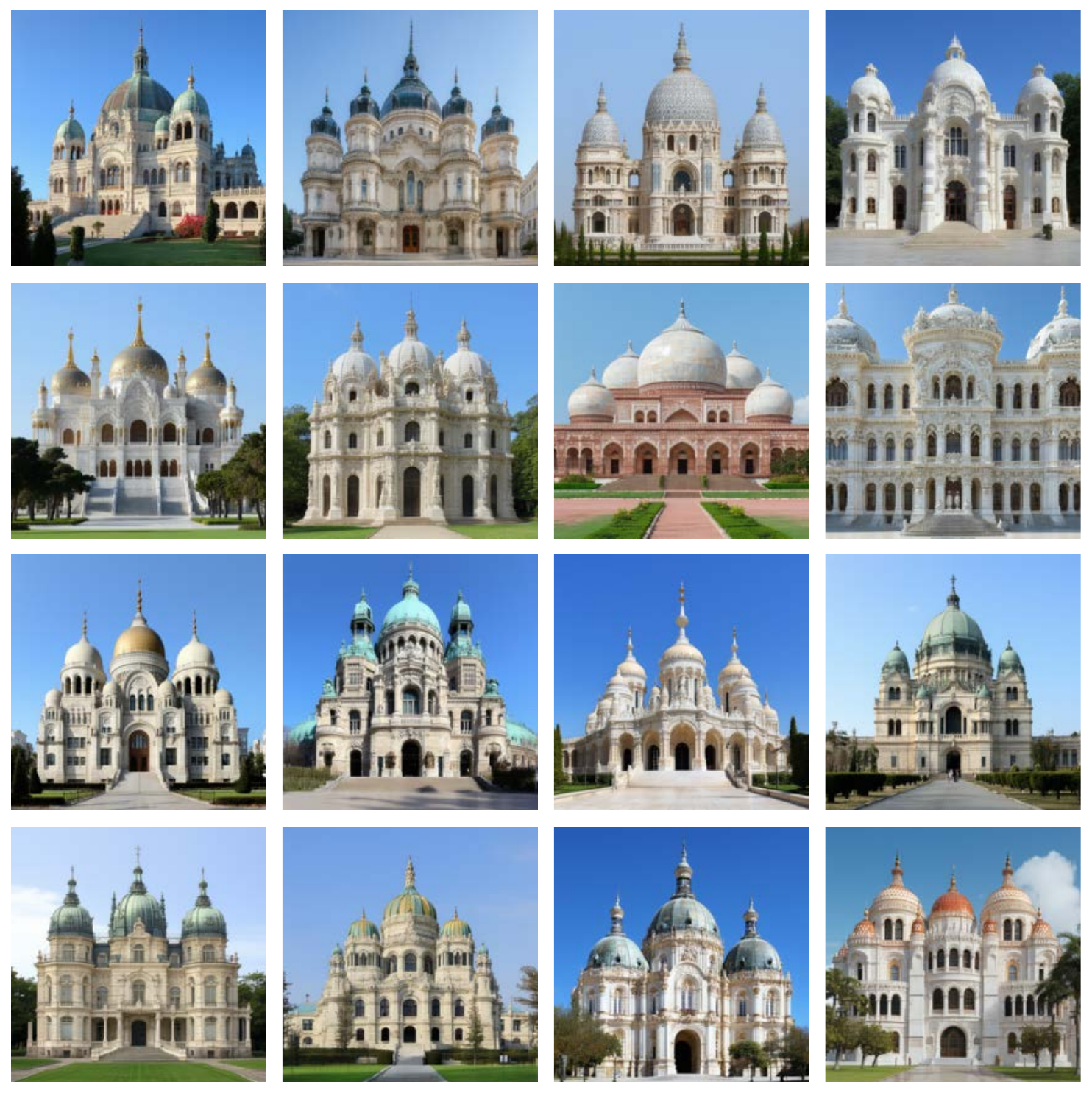}
        \captionfont{\text{7.03}}
        \label{fig:CADS_C3}
    \end{subfigure}
    \begin{subfigure}[h]{0.22\textwidth}
        \centering
        \includegraphics[width=\textwidth]{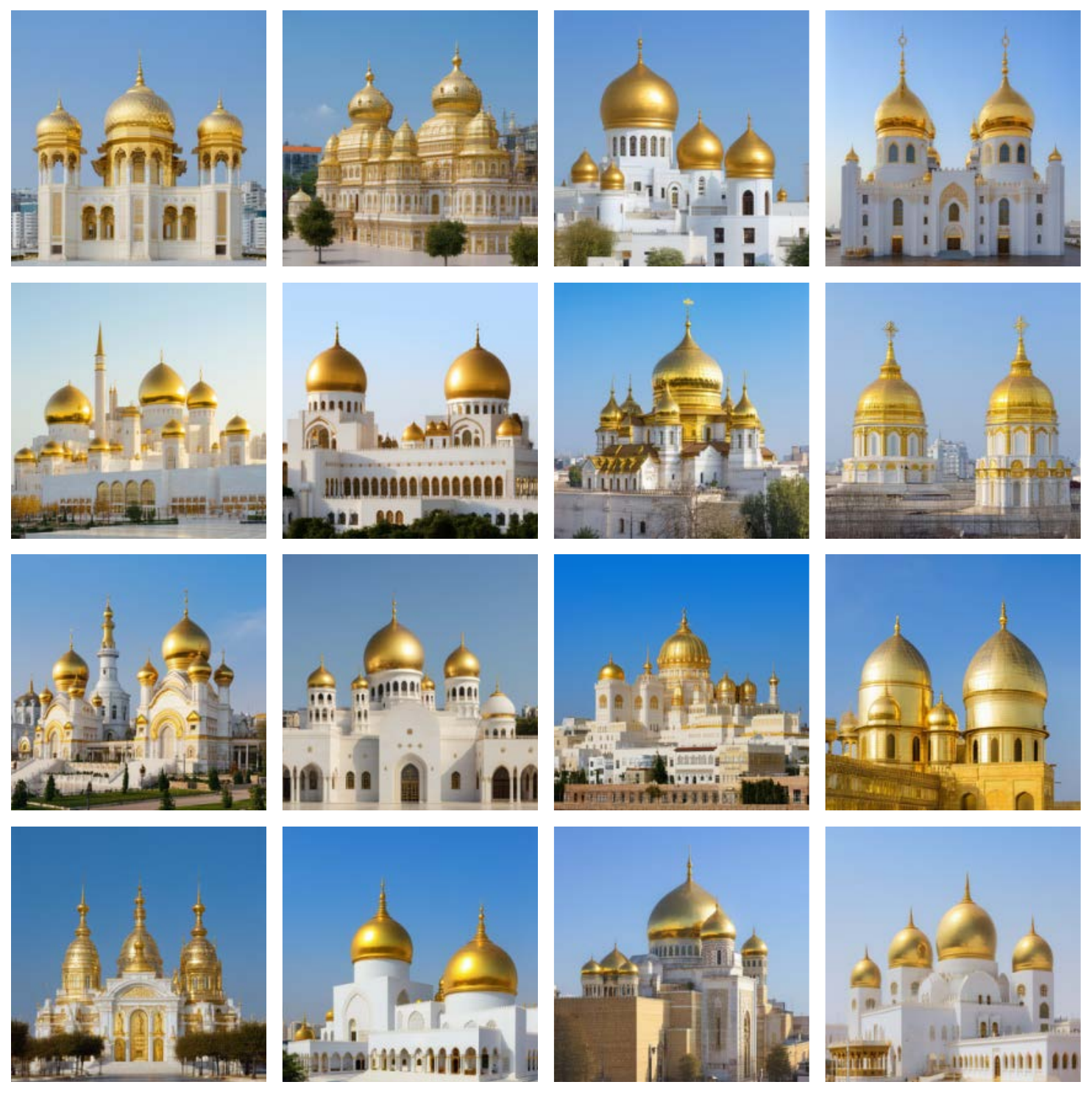}
        \captionfont{\text{5.51}}
        \label{fig:CADS_C4}
    \end{subfigure} \\
    \rotatebox[origin=c]{90}{\raisebox{0.5pt}{\textcolor{white}{whiteee} \textit{prompt expansion + CFG}}}
    \begin{subfigure}[h]{0.22\textwidth}
        \centering
        \includegraphics[width=\textwidth]{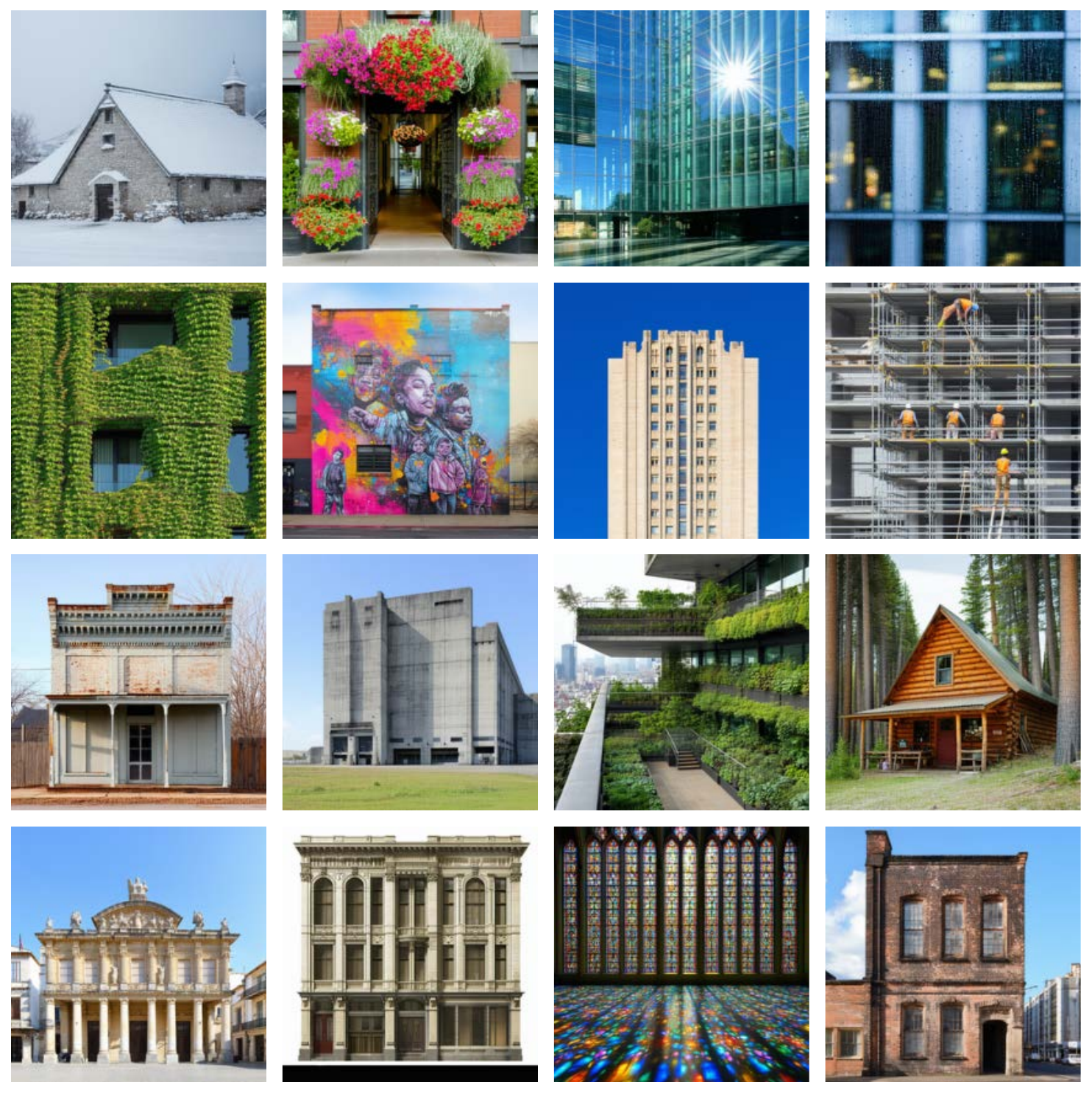}
        \captionfont{\text{17.12}}
        \caption{Building. \\ \textcolor{white}{spacing}}
        \label{fig:pmt_exp_c1}
    \end{subfigure}
    \begin{subfigure}[h]{0.22\textwidth}
        \centering
        \includegraphics[width=\textwidth]{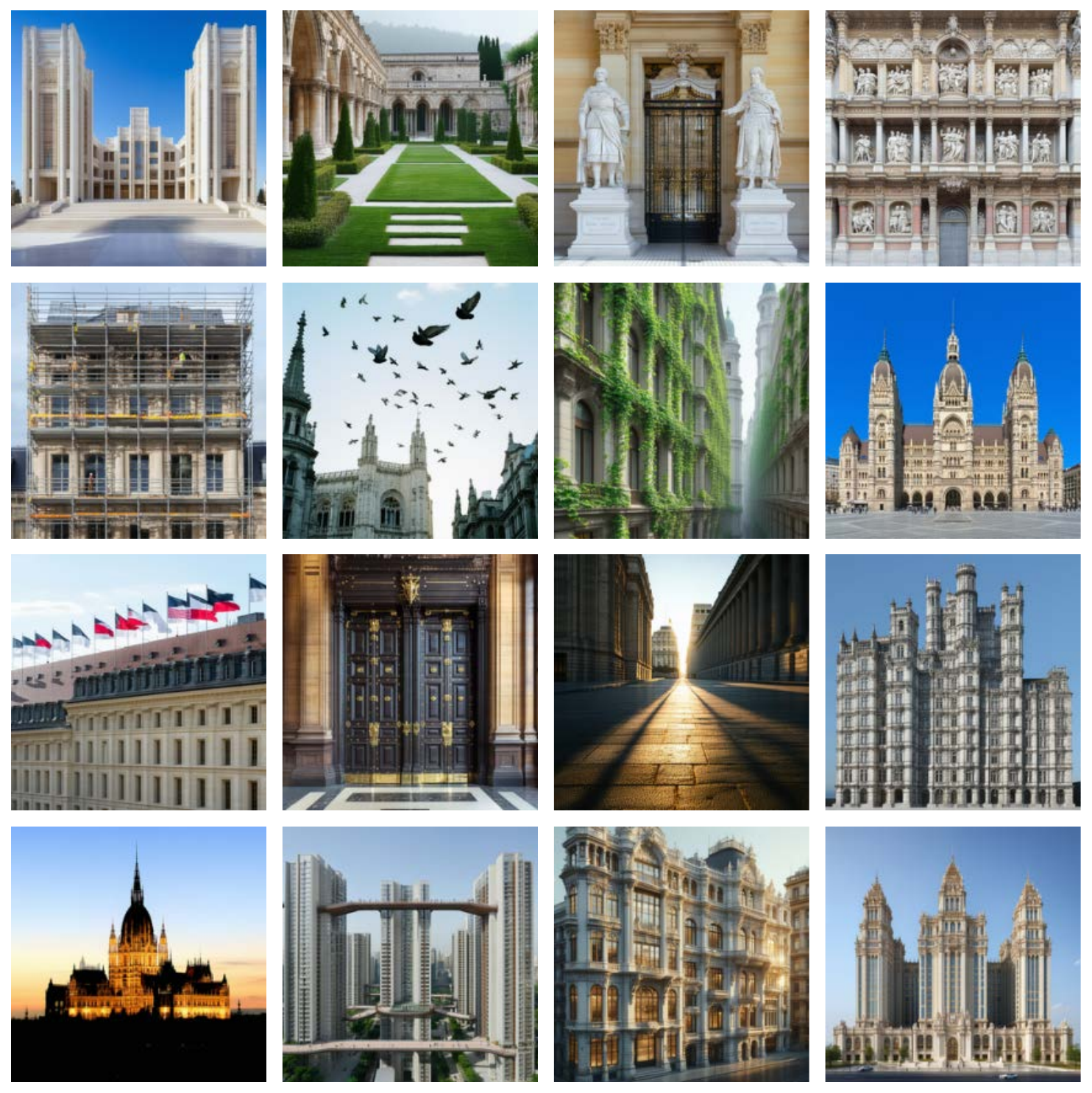}
        \captionfont{\text{15.68}}
        \caption{Grand building. \\ \textcolor{white}{spacing}}
        \label{fig:pmt_exp_c2}
    \end{subfigure}
    \begin{subfigure}[h]{0.22\textwidth}
        \centering
        \includegraphics[width=\textwidth]{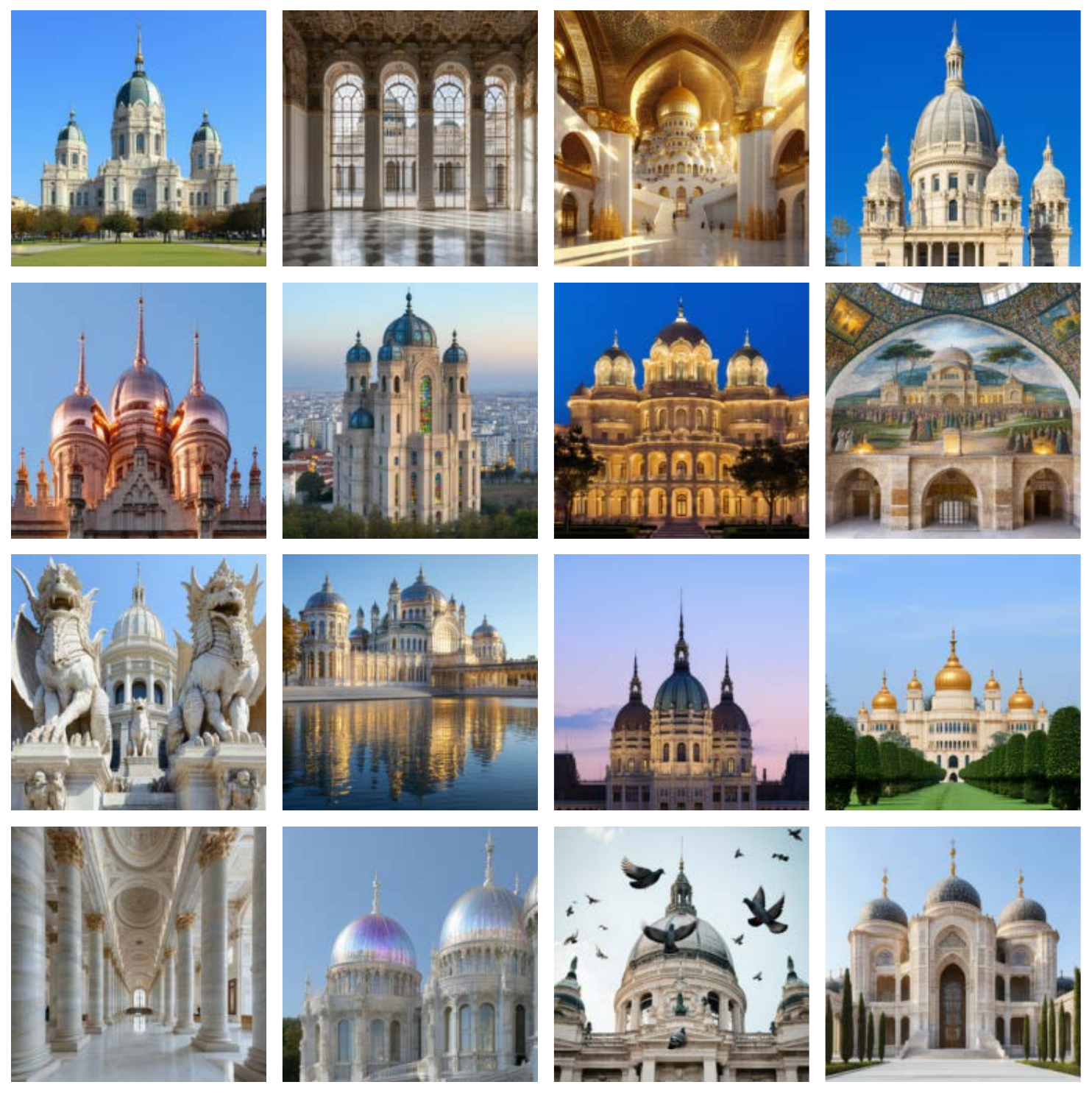}
        \captionfont{\text{11.79}}
        \caption{Grand building, ornate domes.}
        \label{fig:pmt_exp_c3}
    \end{subfigure}
    \begin{subfigure}[h]{0.22\textwidth}
        \centering
        \includegraphics[width=\textwidth]{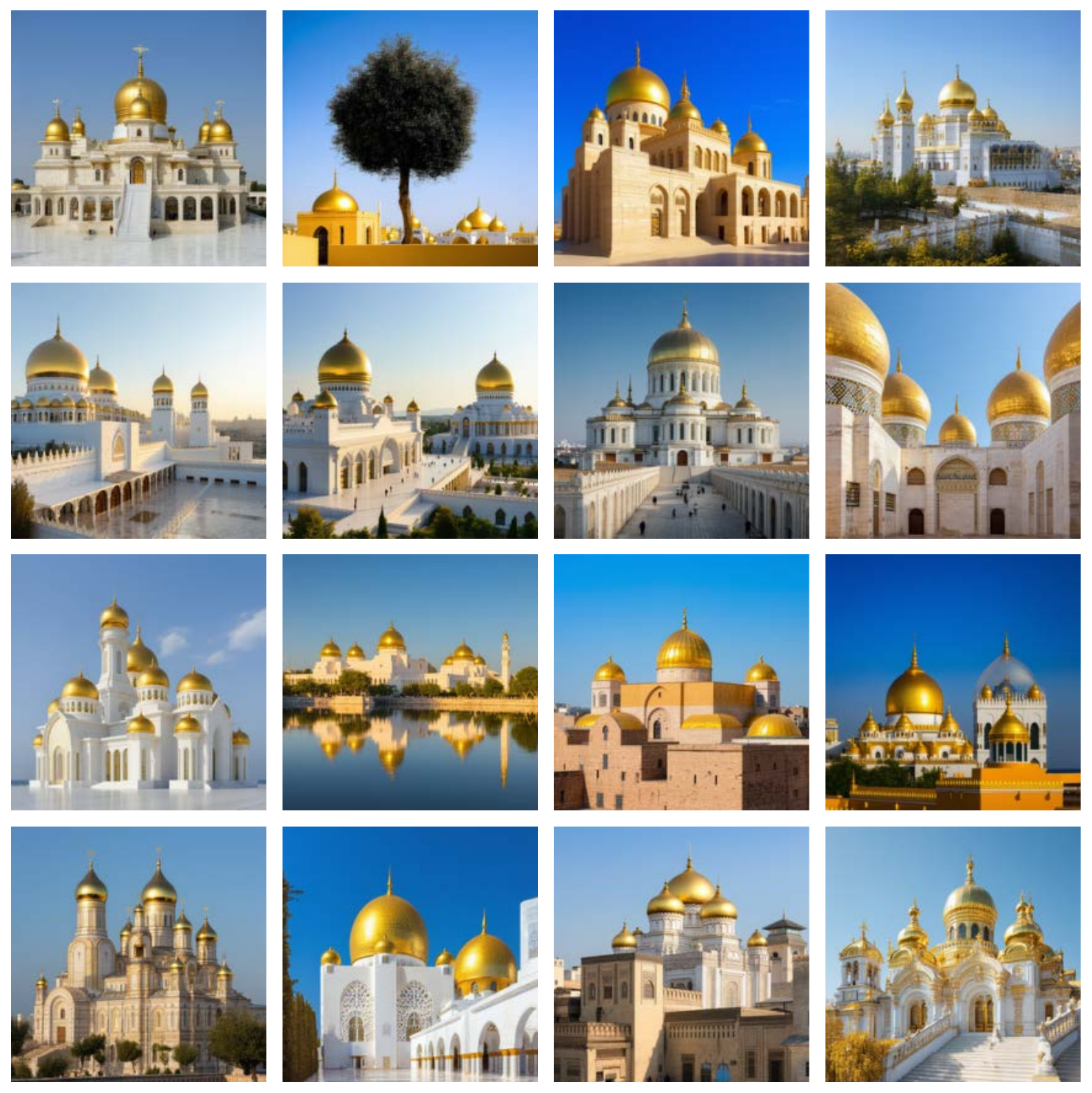}
        \captionfont{\text{5.73}}
        \caption{Golden domes and buildings against a clear sky.}
        \label{fig:pmt_exp_c4}
    \end{subfigure}
    \noindent %
    \begin{tikzpicture}
        \draw [->] (0.1\textwidth,0) -- (\textwidth,0);
    \end{tikzpicture}
    \raisebox{0.1pt}{\small\textit{prompt complexity increasing}}
    \caption{\textbf{Synthetic images across prompt complexity (row-wise) and sampling methods (column-wise).} The number below each mosaic corresponds to the diversity metric (Vendi score) of that set of images.
    Diversity decreases both visually and quantitatively as we increase prompt complexity. Advanced sampling methods (CADS and prompt expansion) improve the diversity of synthetic data. Image quality is not significantly affected by the prompt complexity, but prompt consistency decreases as we increase prompt complexity. CFG: classifier-free guidance.  \looseness-1 \looseness-1}
    \label{fig:header_visuals_pmt_complexity}
\end{figure}

In Figure~\ref{fig:header_visuals_pmt_complexity}, we show qualitative samples for LDMv3.5L model using vanilla guidance (CFG), CADS, and prompt expansion with prompts extracted from CC12M. Notably, we observe that as we increase prompt complexity, the generations appear closer to the real data and exhibit lower diversity, slightly lower prompt consistency, and similar image aesthetics. Moreover, inference-time interventions and their combinations (rows 3, 4) result in synthetic data with higher diversity than the CFG baseline (row 2).

\begin{figure}[t]
    \centering
    \rotatebox[origin=c]{90}{\raisebox{0.5pt}{\textit{real data}}}
    \begin{subfigure}[h]{0.23\textwidth}
        \centering
        \includegraphics[width=\textwidth]{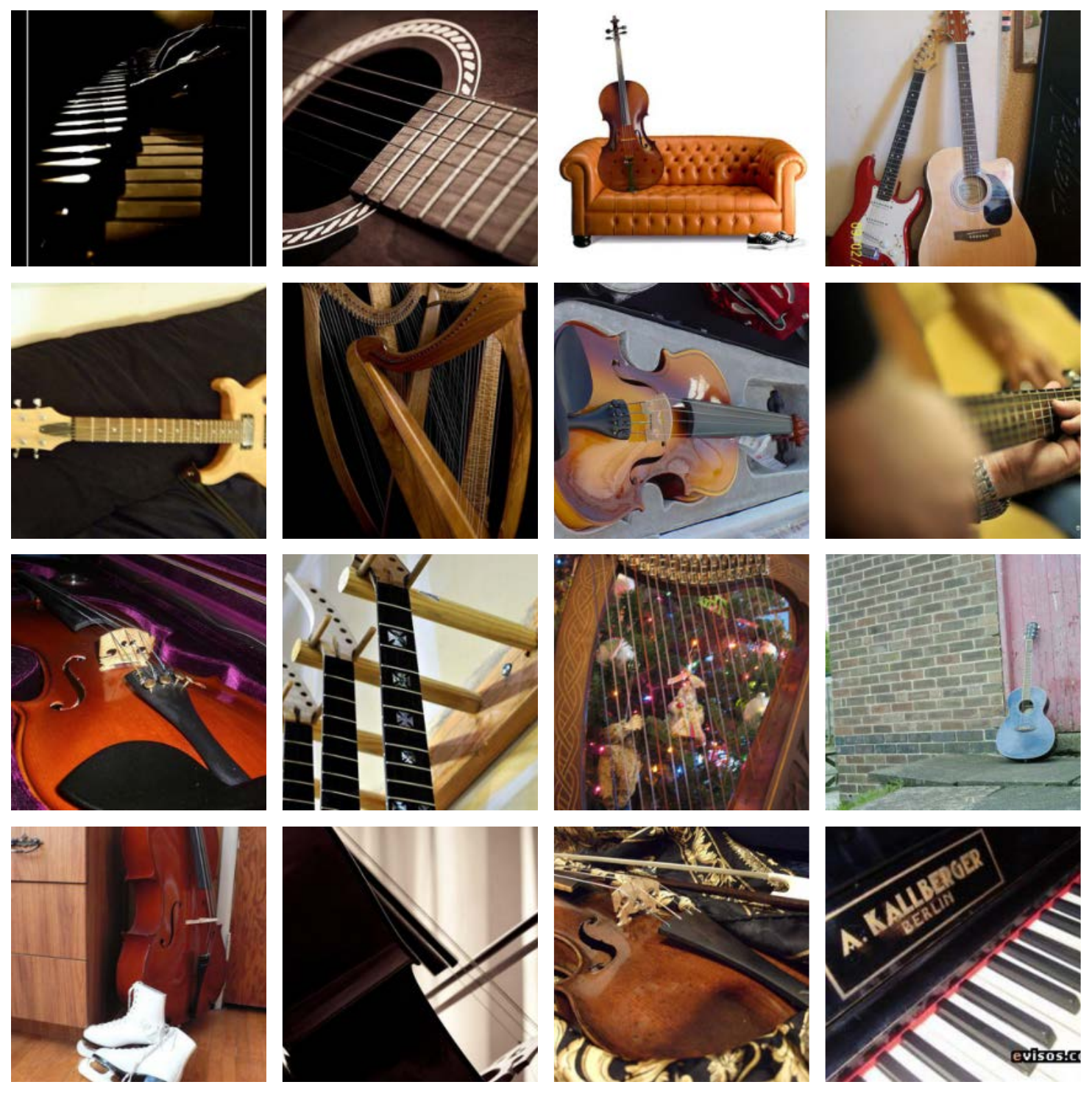}
        \captionfont{\text{19.32}}
        \label{fig:data_c1_in1k}
    \end{subfigure} \hspace{0.3em}
    \begin{subfigure}[h]{0.23\textwidth}
        \centering
        \includegraphics[width=\textwidth]{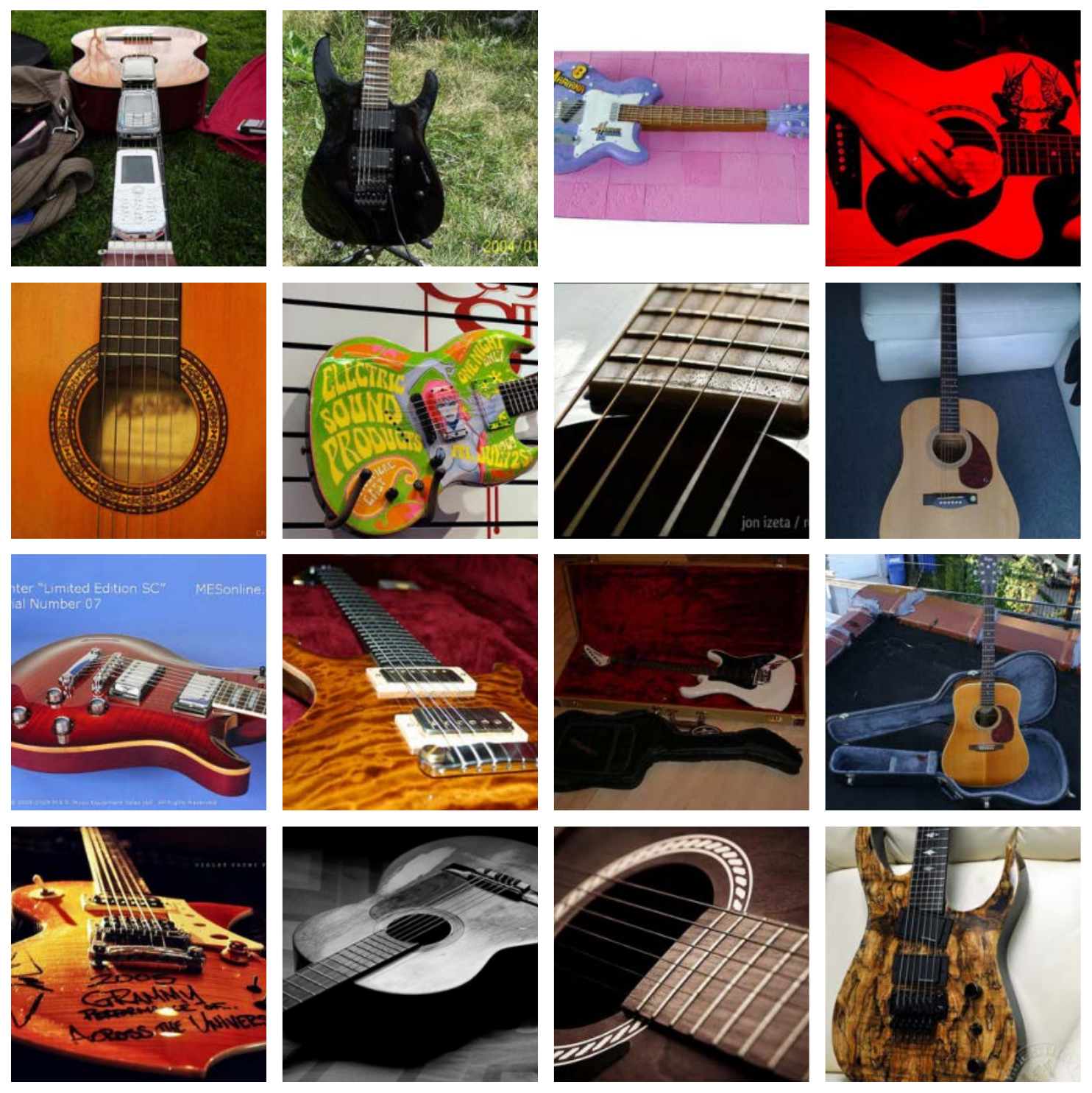}
        \captionfont{\text{14.55}}
        \label{fig:data_c2_in1k}
    \end{subfigure} \hspace{0.3em}
    \begin{subfigure}[h]{0.23\textwidth}
        \centering
        \includegraphics[width=\textwidth]{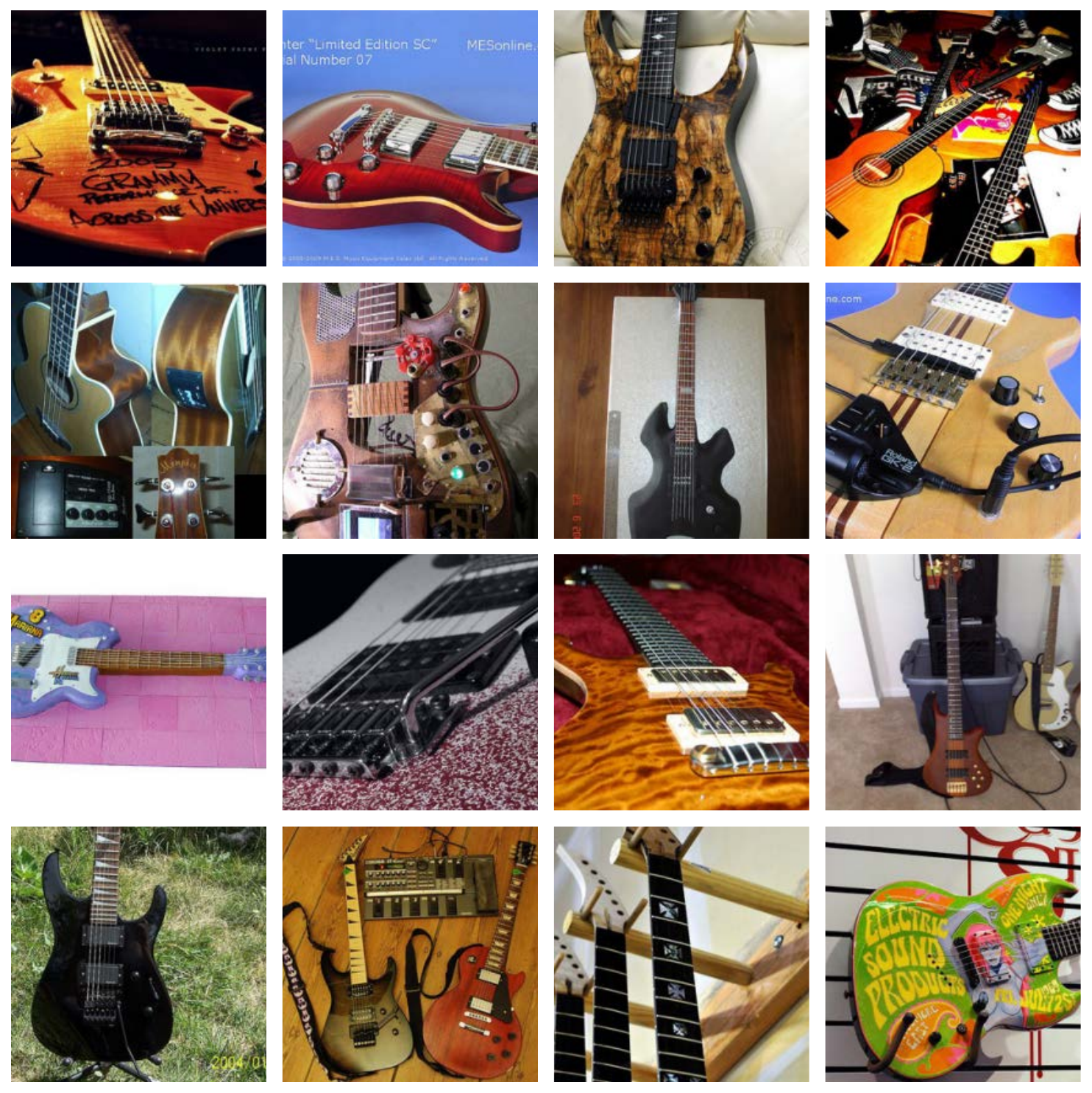}
        \captionfont{\text{13.21}}
        \label{fig:data_c3_in1k}
    \end{subfigure} \\
    \rotatebox[origin=c]{90}{\raisebox{0.5pt}{\textit{CFG}}}
    \begin{subfigure}[h]{0.23\textwidth}
        \centering
        \includegraphics[width=\textwidth]{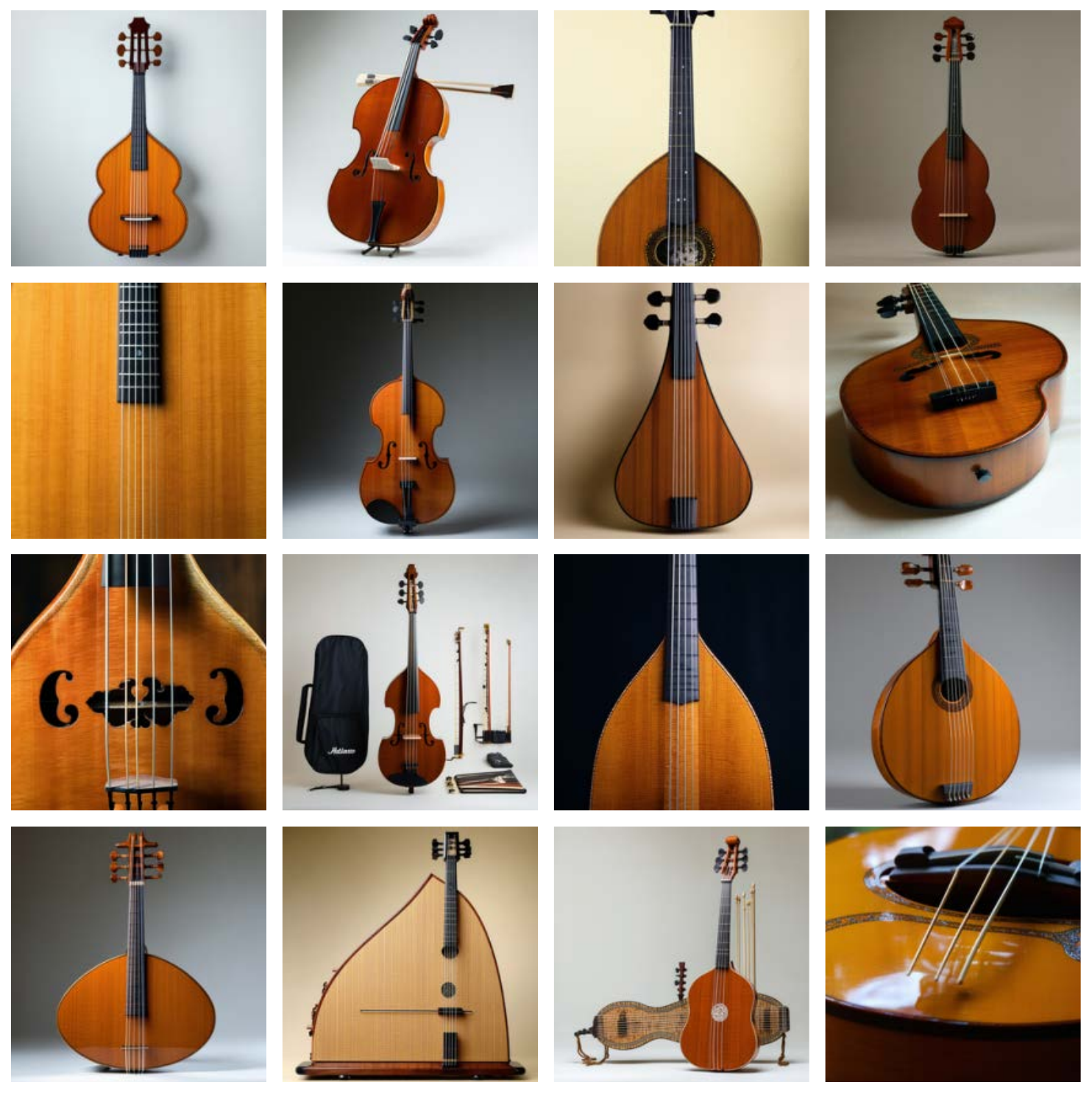}
        \captionfont{\text{5.78}}
        \label{fig:CFG_c1_in1k}
    \end{subfigure} \hspace{0.3em}
    \begin{subfigure}[h]{0.23\textwidth}
        \centering
        \includegraphics[width=\textwidth]{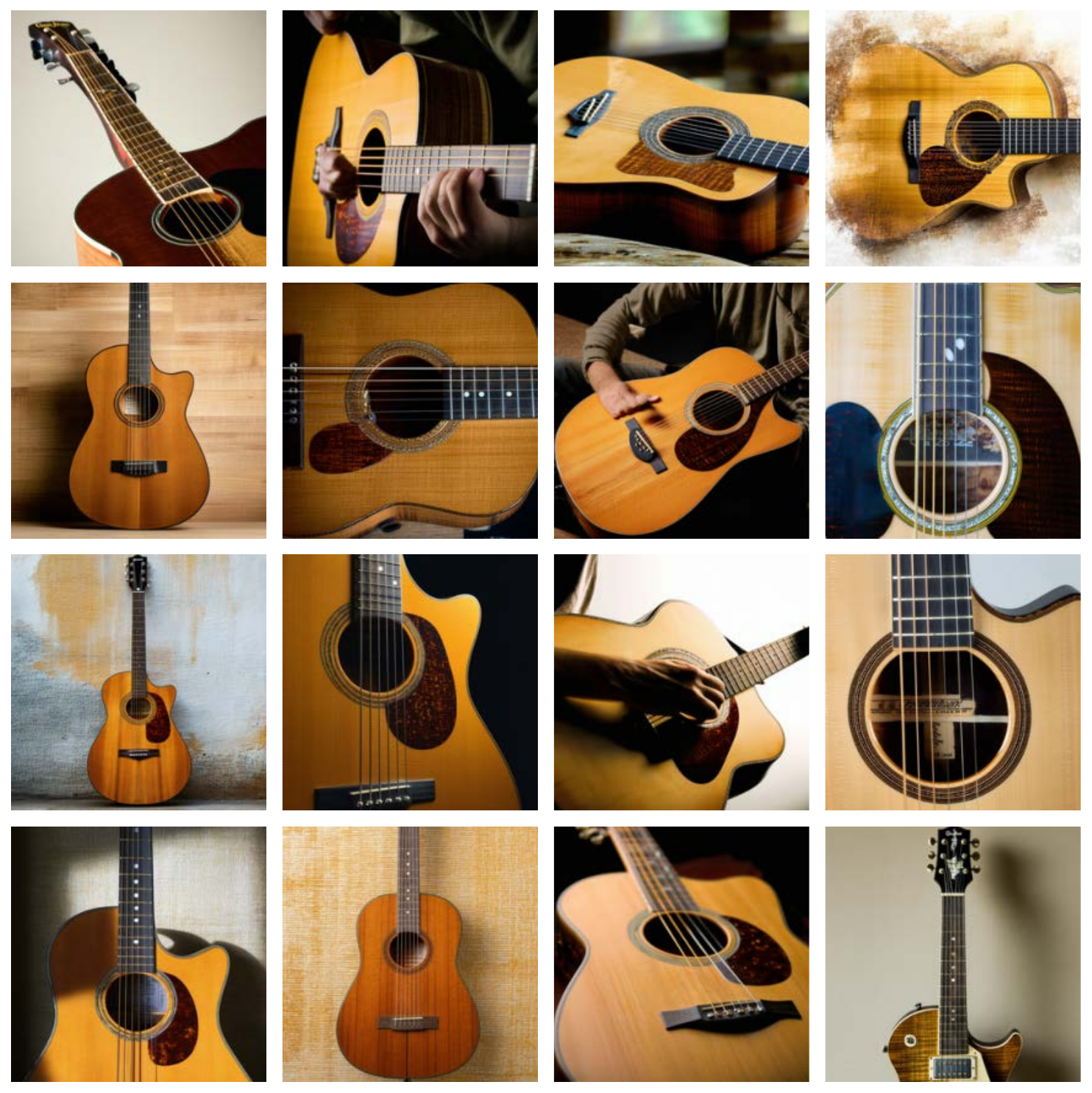}
        \captionfont{\text{5.68}}
        \label{fig:CFG_c2_in1k}
    \end{subfigure} \hspace{0.3em}
    \begin{subfigure}[h]{0.23\textwidth}
        \centering
        \includegraphics[width=\textwidth]{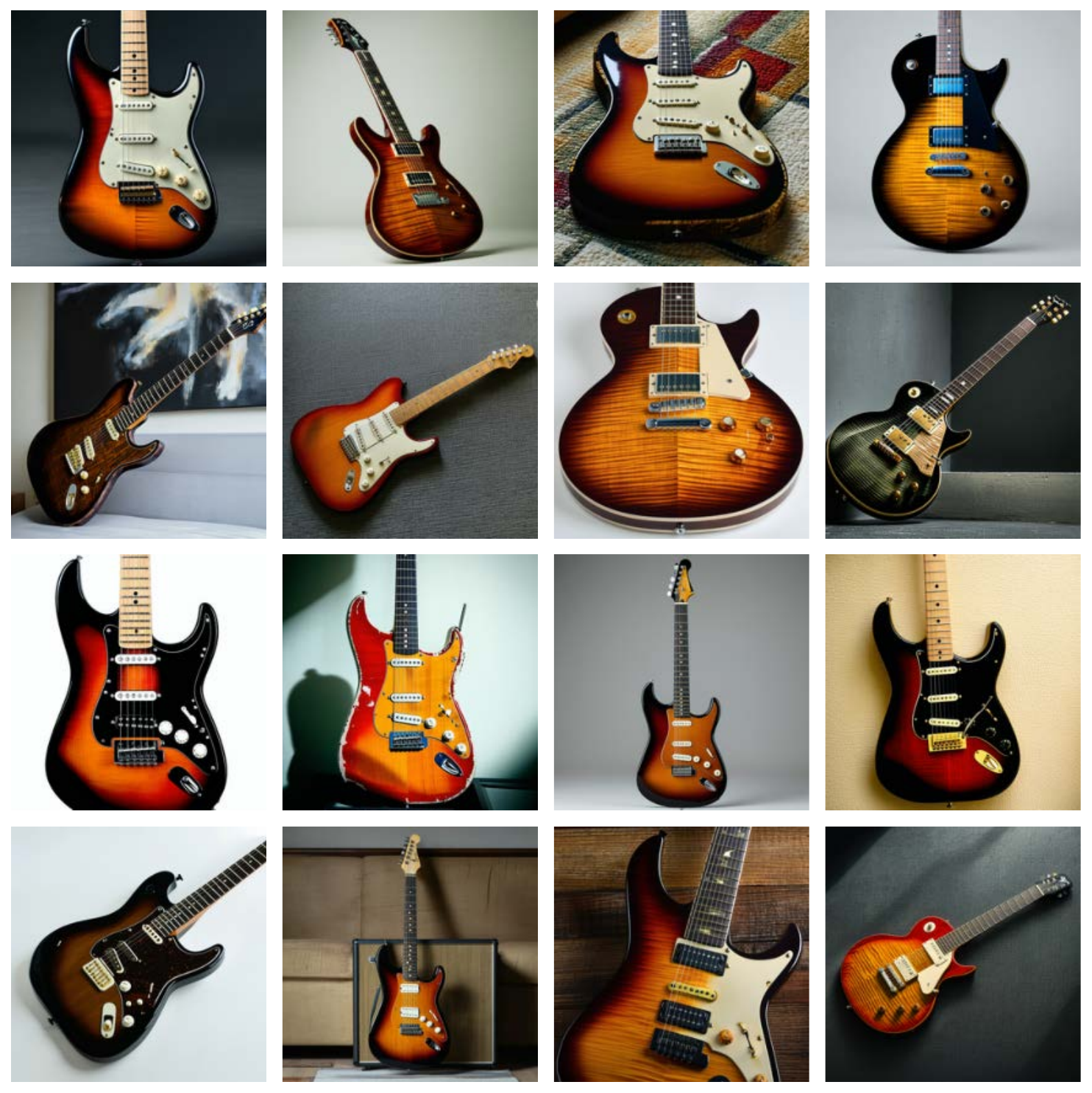}
        \captionfont{\text{4.19}}
        \label{fig:CFG_c3_in1k}
    \end{subfigure} \\
    \rotatebox[origin=c]{90}{\raisebox{0.5pt}{\textcolor{white}{ee} \textit{CADS guidance}}}
    \begin{subfigure}[h]{0.23\textwidth}
        \centering
        \includegraphics[width=\textwidth]{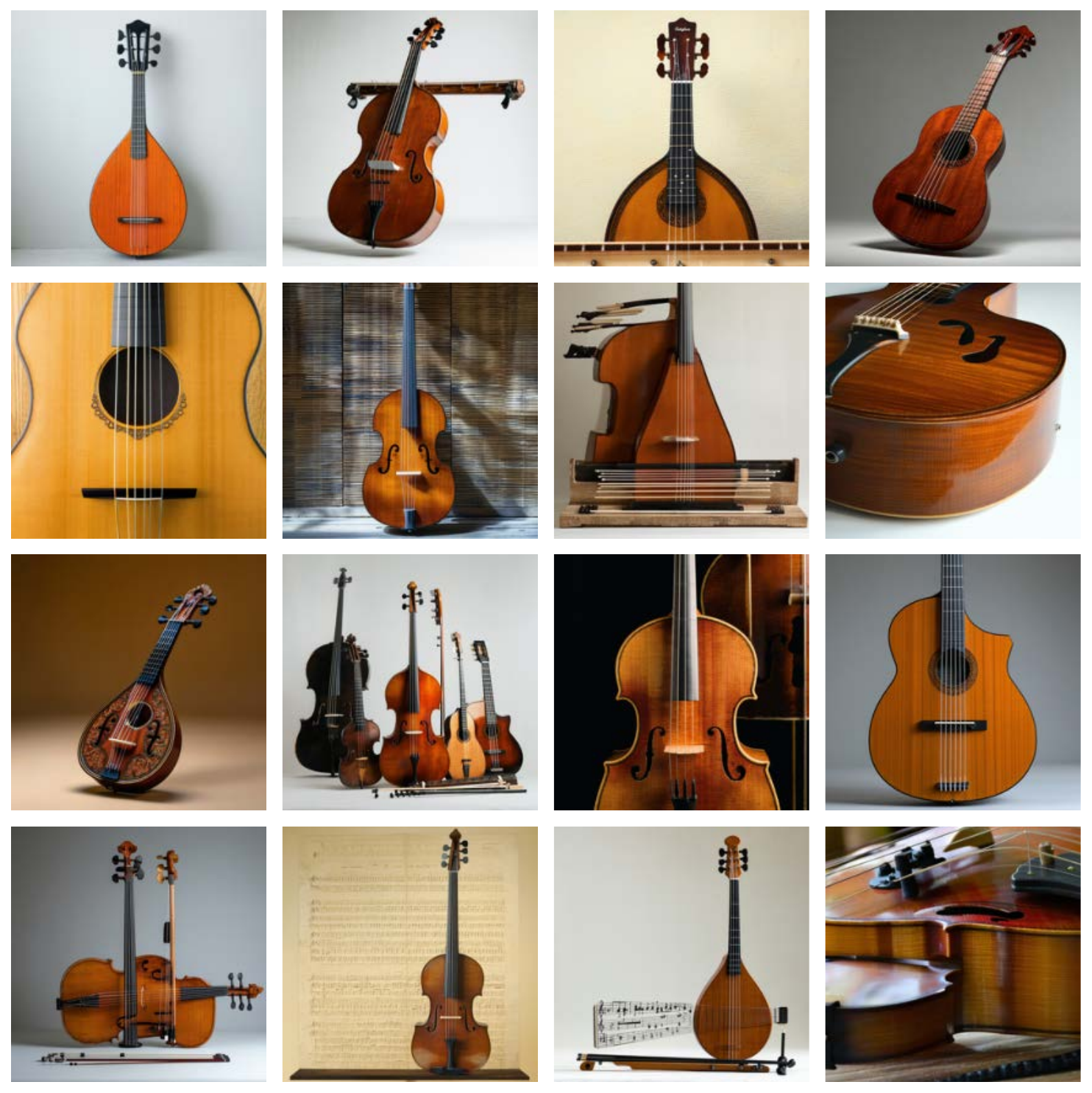}
        \captionfont{\text{7.04}}
        \label{fig:CADS_C1_in1k}
    \end{subfigure} \hspace{0.3em}
    \begin{subfigure}[h]{0.23\textwidth}
        \centering
        \includegraphics[width=\textwidth]{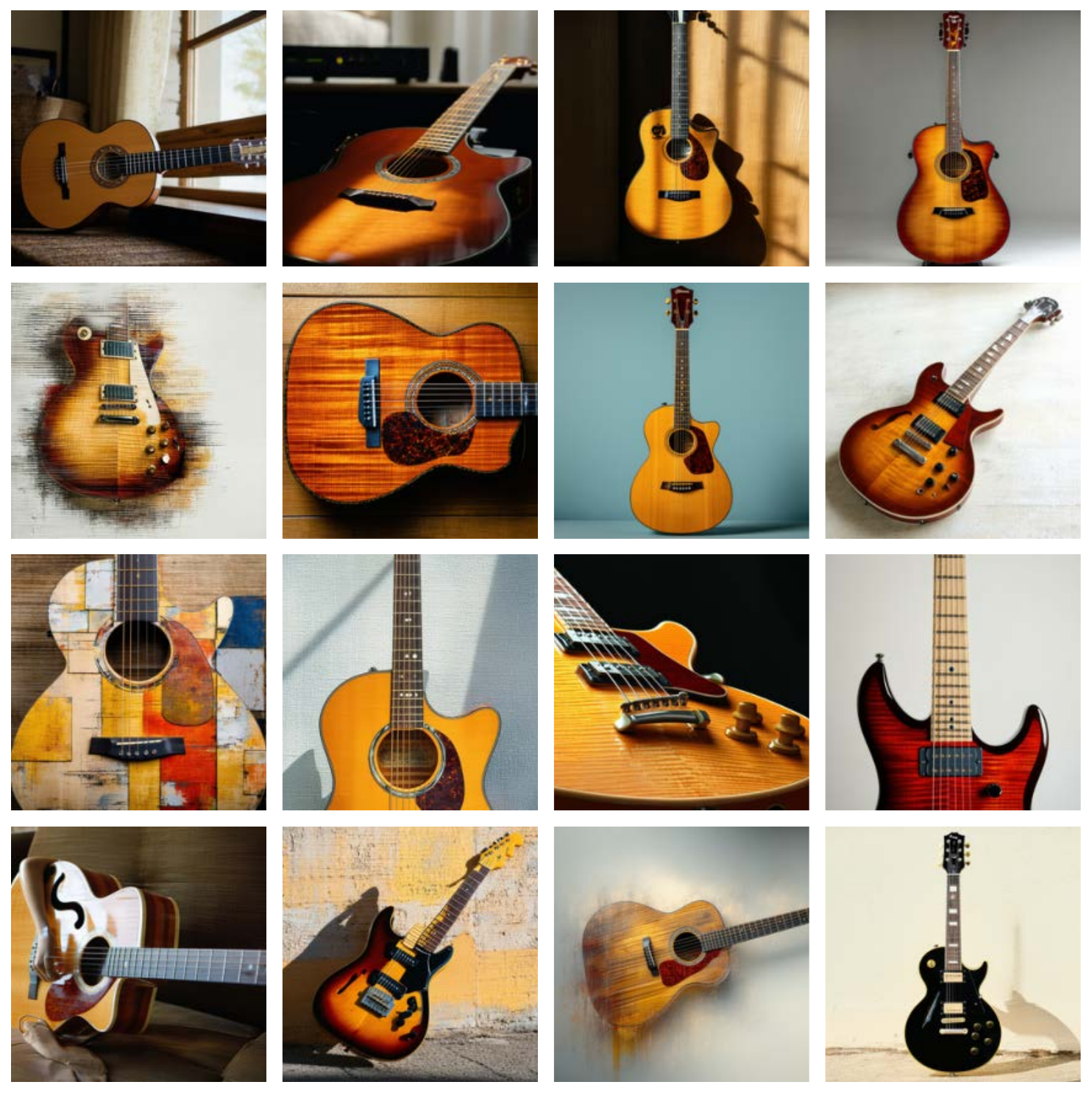}
        \captionfont{\text{6.70}}
        \label{fig:CADS_C2_in1k}
    \end{subfigure} \hspace{0.3em}
    \begin{subfigure}[h]{0.23\textwidth}
        \centering
        \includegraphics[width=\textwidth]{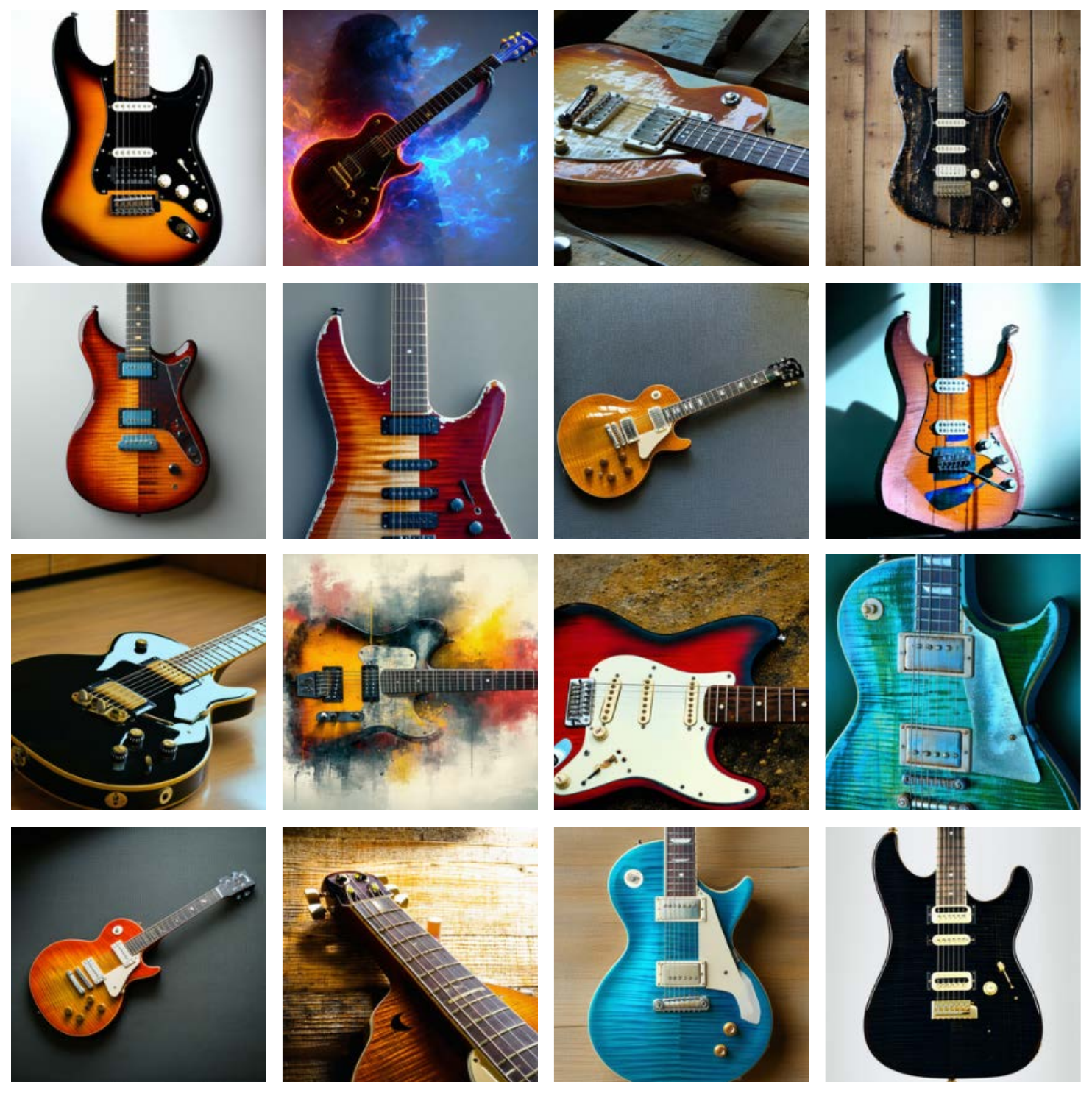}
        \captionfont{\text{6.30}}
        \label{fig:CADS_C3_in1k}
    \end{subfigure} \\
    \rotatebox[origin=c]{90}{\raisebox{0.5pt}{\textcolor{white}{whiteee} \textit{prompt expansion + CFG}}}
    \begin{subfigure}[h]{0.23\textwidth}
        \centering
        \includegraphics[width=\textwidth]{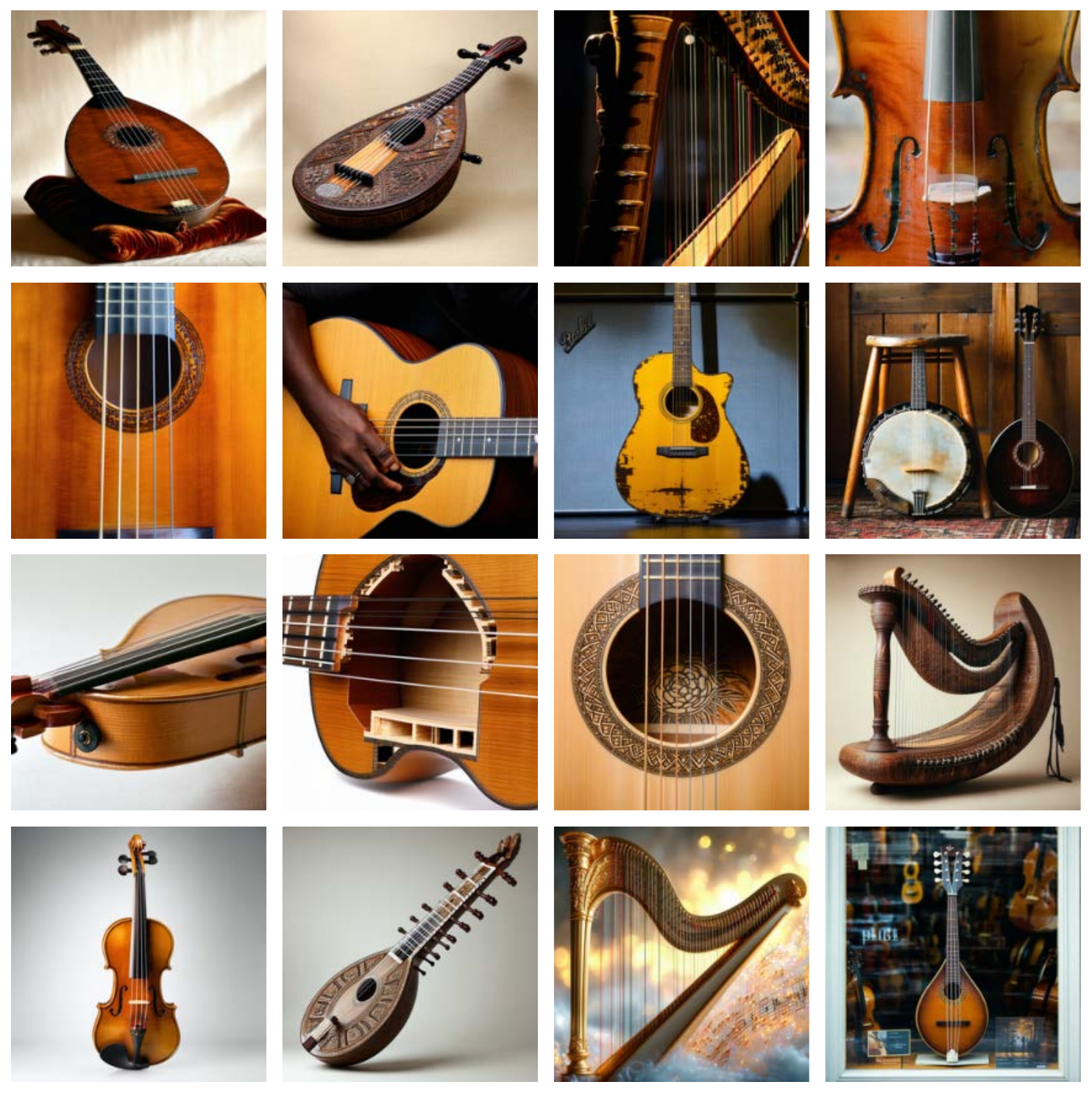}
        \captionfont{\text{12.65}}
        \caption{\centering \texttt{stringed instrument}}
        \label{fig:pmt_exp_c1_in1k}
    \end{subfigure} \hspace{0.3em}
    \begin{subfigure}[h]{0.23\textwidth}
        \centering
        \includegraphics[width=\textwidth]{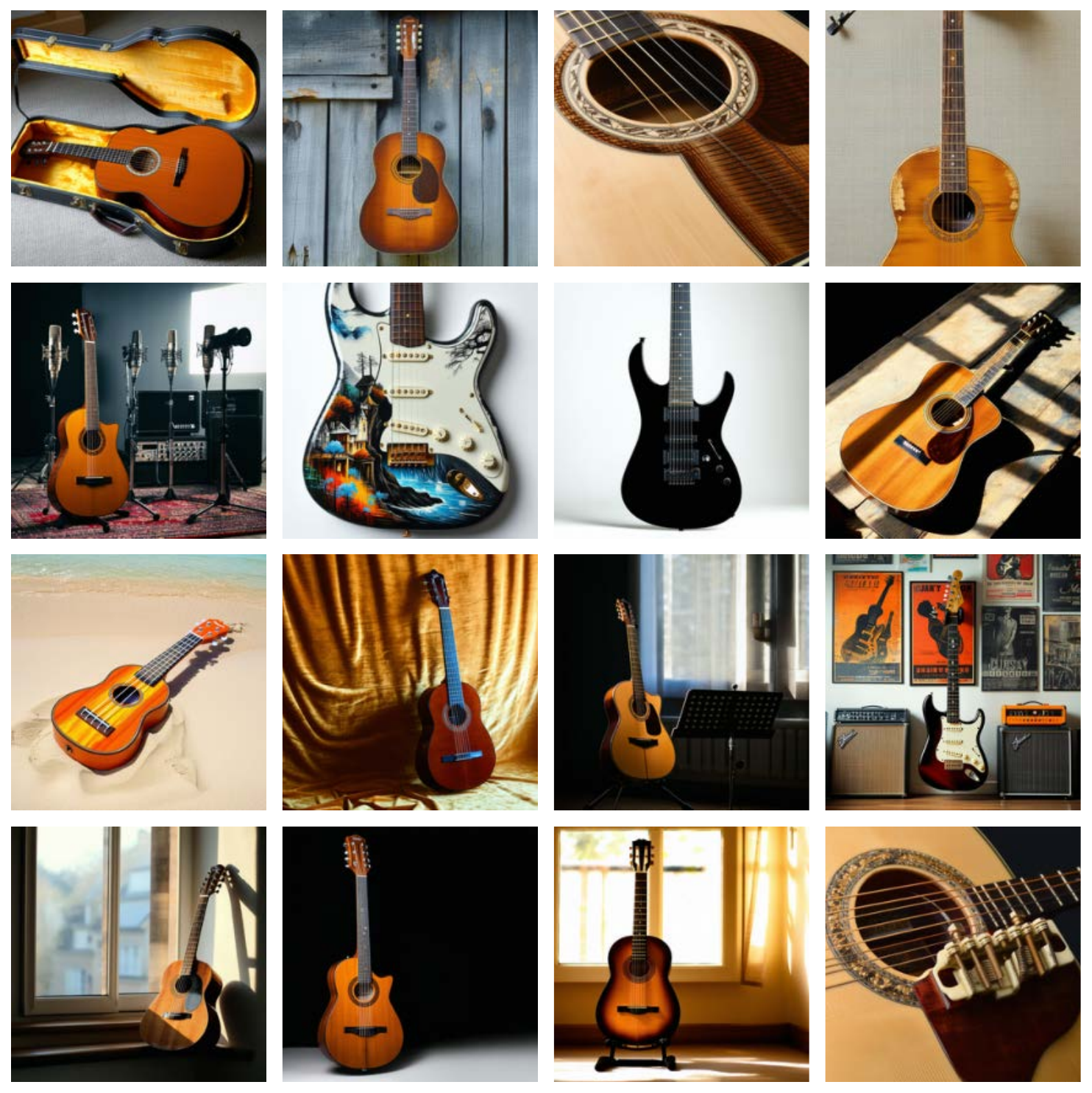}
        \captionfont{\text{10.55}}
       \caption{\centering \texttt{guitar}}
        \label{fig:pmt_exp_c2_in1k}
    \end{subfigure} \hspace{0.3em}
    \begin{subfigure}[h]{0.23\textwidth}
        \centering
        \includegraphics[width=\textwidth]{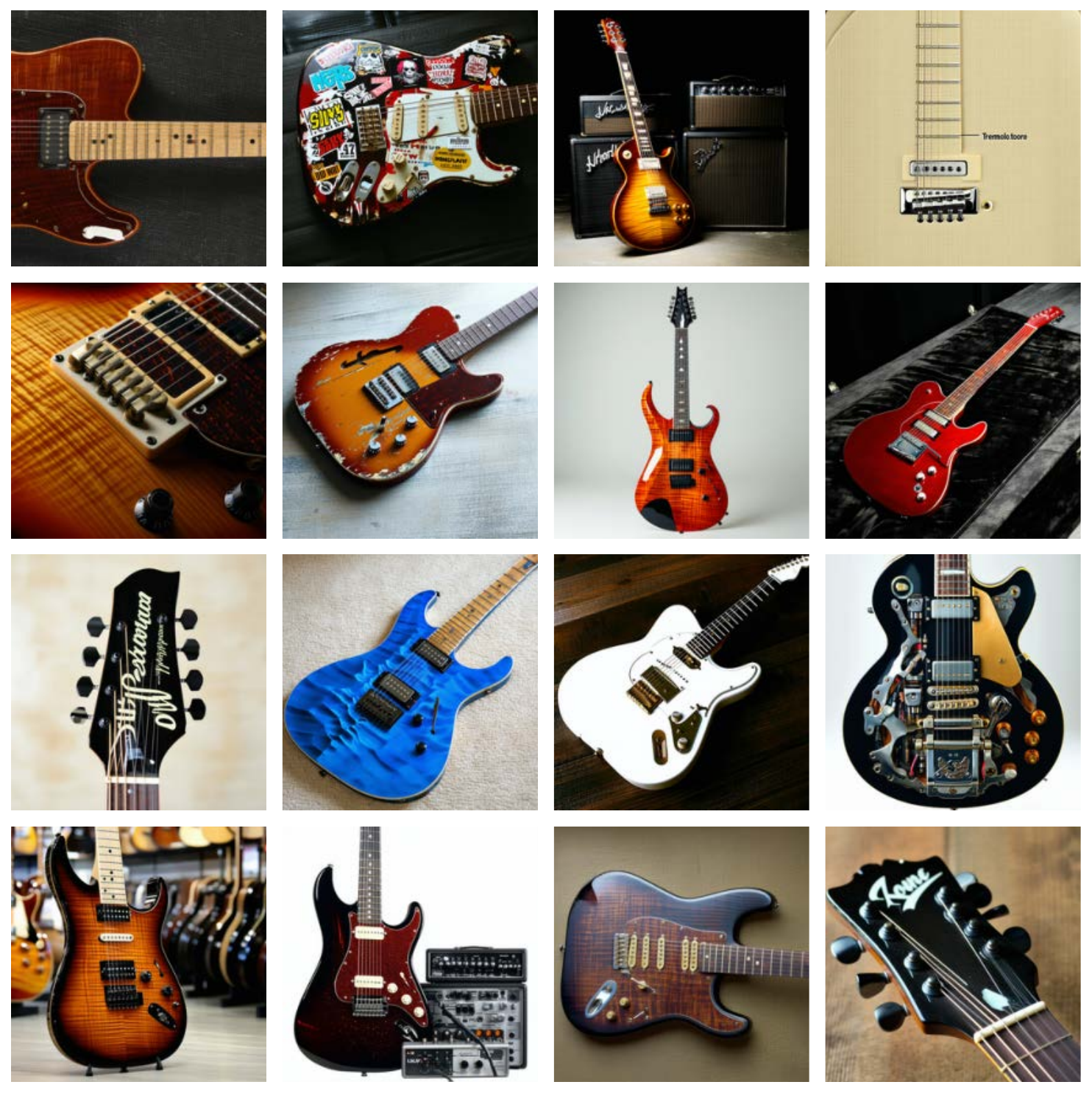}
        \captionfont{\text{7.92}}
        \caption{\texttt{electric guitar}}
        \label{fig:pmt_exp_c3_in1k}
    \end{subfigure}
    \noindent %
    \begin{tikzpicture}
        \draw [->] (0.1\textwidth,0) -- (\textwidth,0);
    \end{tikzpicture}
    \raisebox{0.3pt}{\textit{prompt complexity increasing}}
    \caption{\textbf{Synthetic images across prompt complexity (row-wise) and sampling methods (column-wise) using prompts from ImageNet-1k.} The number below each mosaic corresponds to the diversity metric (Vendi score) of that set of images.
    Diversity decreases both visually and quantitatively as we increase prompt complexity. Advanced sampling methods (CADS and prompt expansion) improve the diversity of synthetic data. Image quality is not significantly affected by the prompt complexity, but prompt consistency decreases as we increase prompt complexity. CFG: classifier-free guidance.}
    \label{fig:header_visuals_pmt_complexity_in1k}
\end{figure}

We show qualitative samples in Figure~\ref{fig:header_visuals_pmt_complexity_in1k} for LDMv3.5L model using vanilla guidance (CFG), CADS, and prompt expansion with class labels extracted from ImageNet-1k. According to WordNet~\citep{wordnet}, \texttt{``stringed instrument''} includes harp, violin, cello, guitar, piano, etc., \texttt{``guitar''} includes acoustic guitar and electric guitar.

\begin{figure}[ht]
    \centering
    \begin{subfigure}[ht]{0.25\textwidth}
        \includegraphics[width=\textwidth]{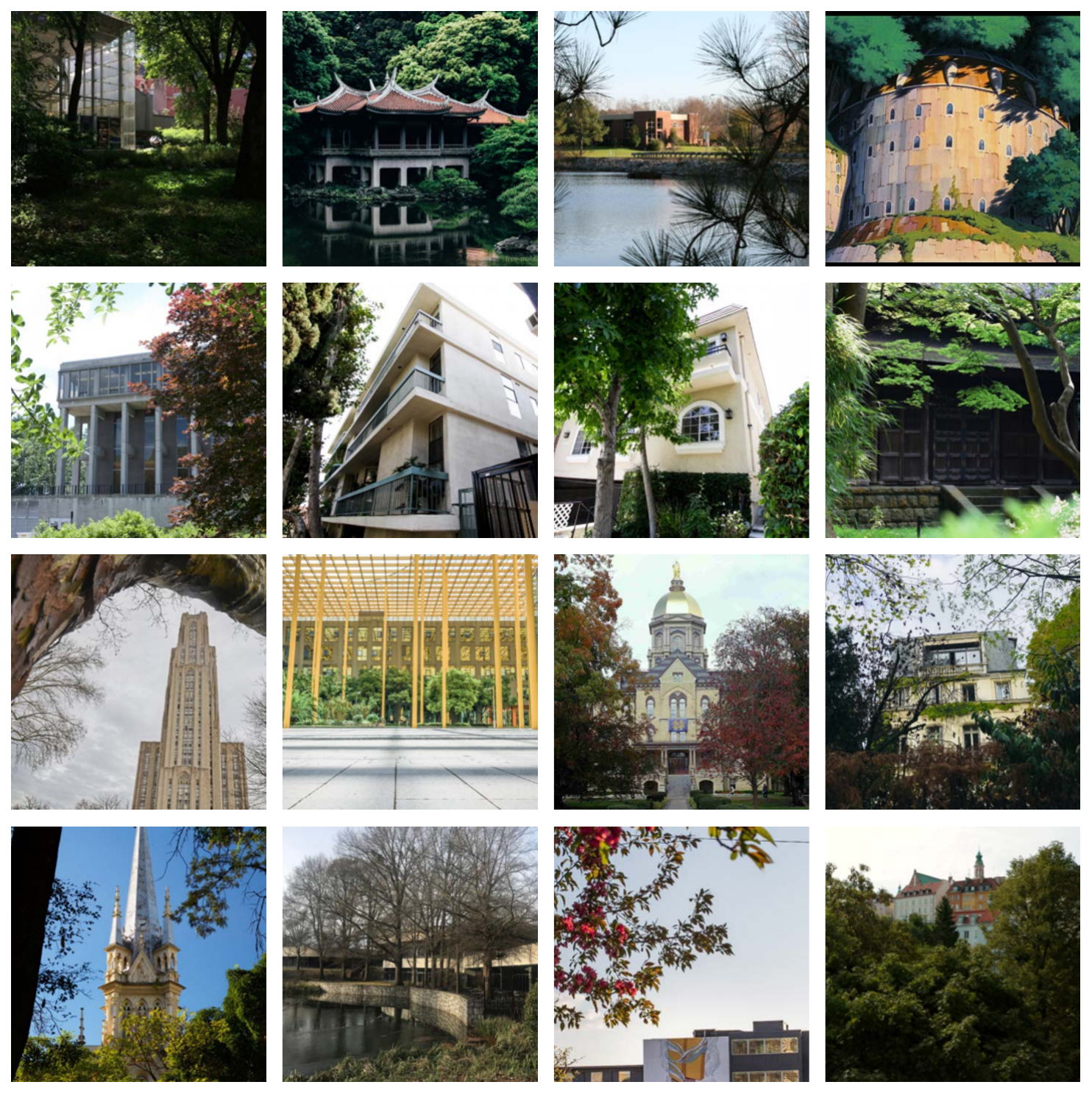}
        \caption{Dataset}
    \end{subfigure}
    \begin{subfigure}[ht]{0.25\textwidth}
        \includegraphics[width=\textwidth]{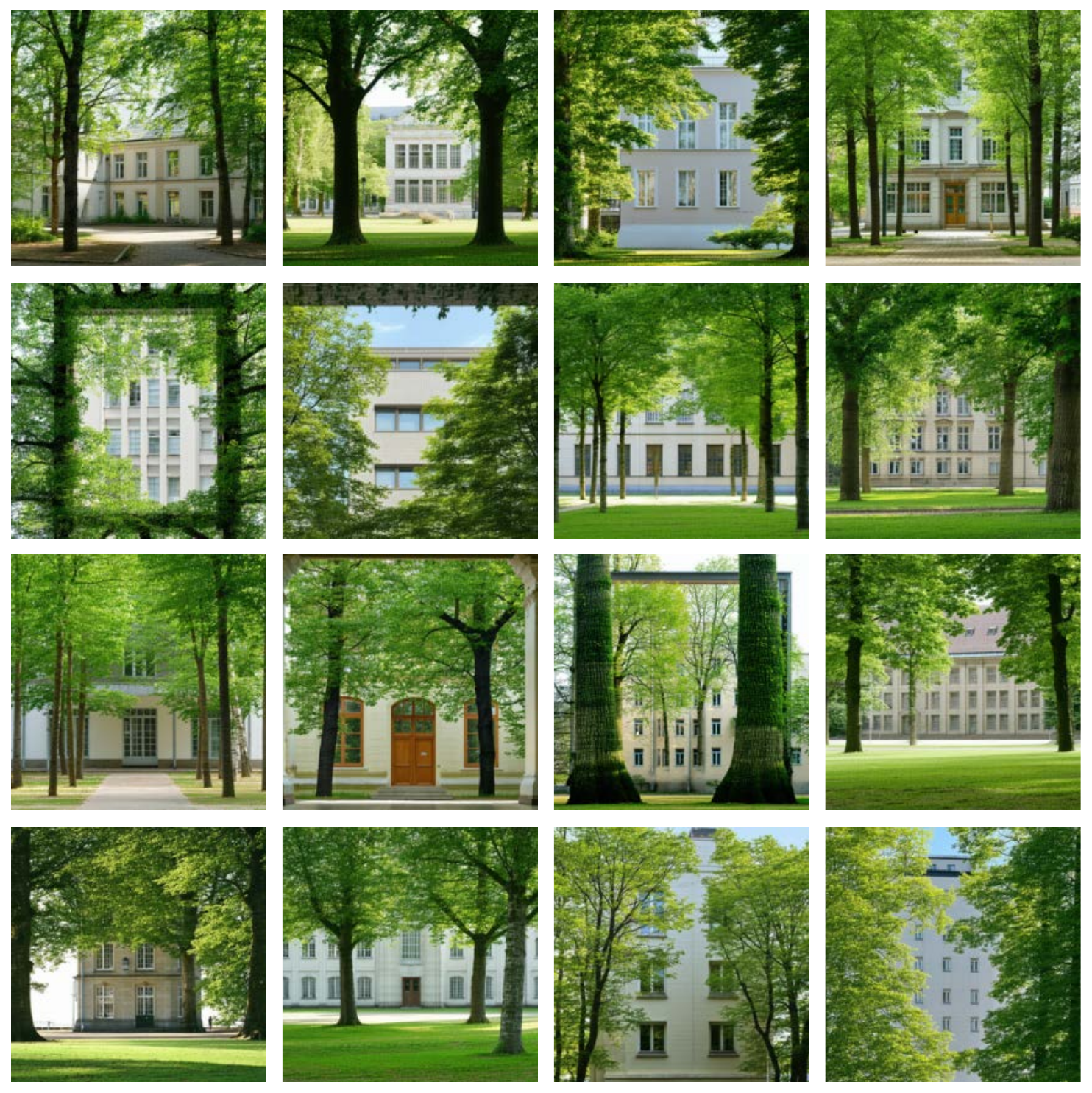}
        \caption{Vanilla guidance}
    \end{subfigure}
    \begin{subfigure}[ht]{0.25\textwidth}
        \includegraphics[width=\textwidth]{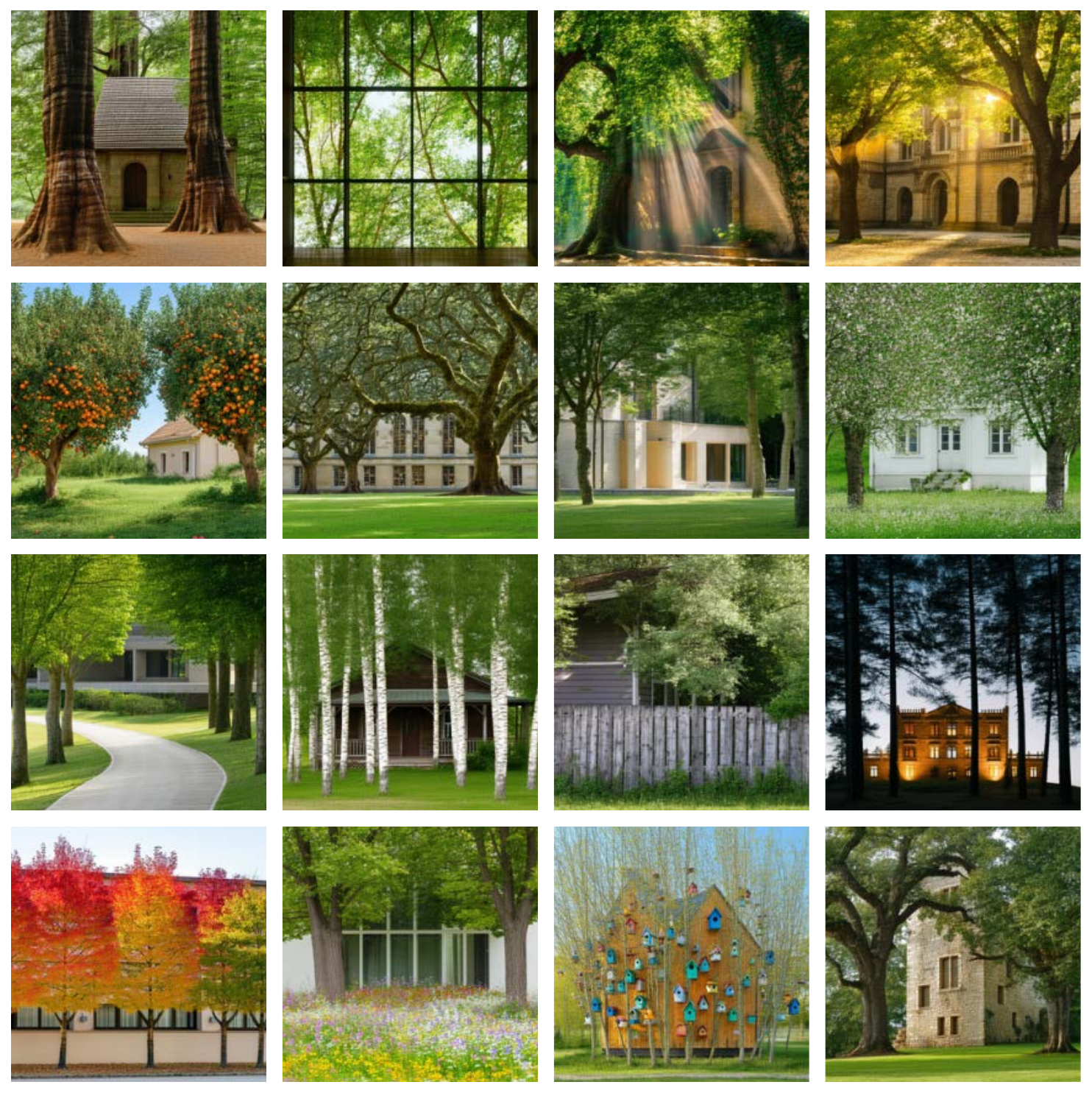}
        \caption{Prompt expansion}
    \end{subfigure} \\
    \begin{subfigure}[ht]{0.25\textwidth}
        \includegraphics[width=\textwidth]{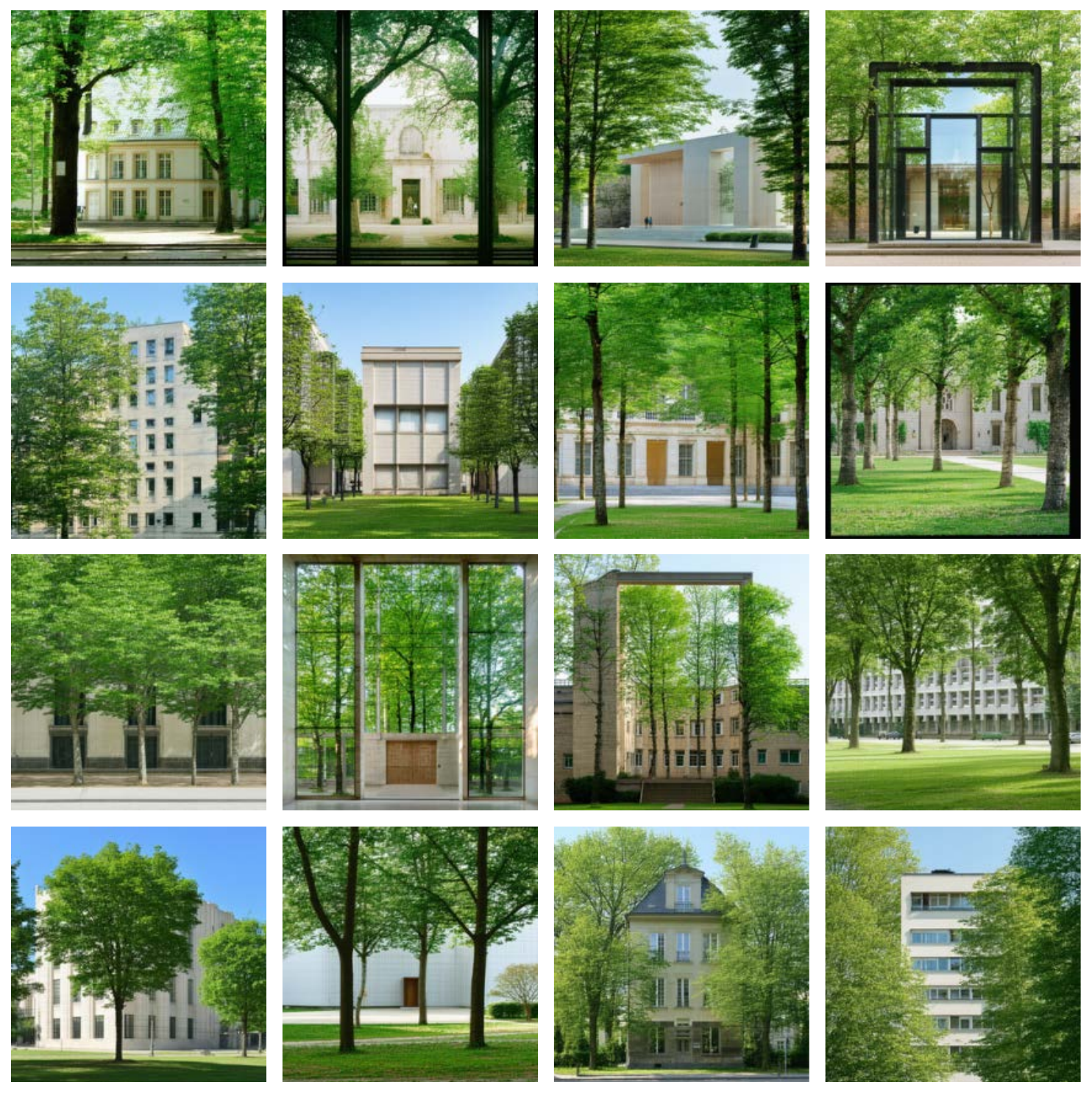}
        \caption{CADS guidance}
    \end{subfigure}
    \begin{subfigure}[ht]{0.25\textwidth}
        \includegraphics[width=\textwidth]{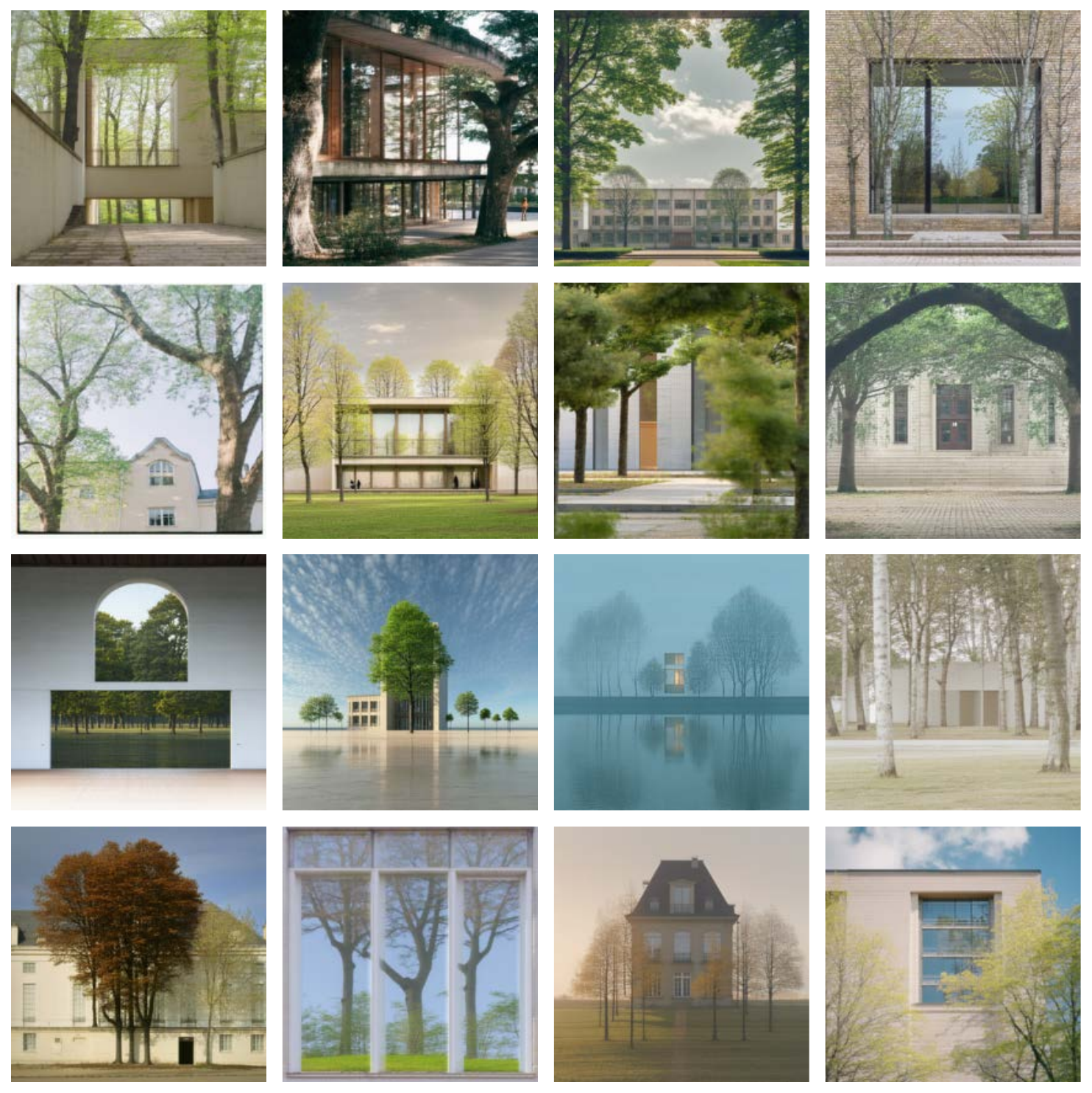}
        \caption{Interval guidance}
    \end{subfigure}
    \begin{subfigure}[ht]{0.25\textwidth}
        \includegraphics[width=\textwidth]{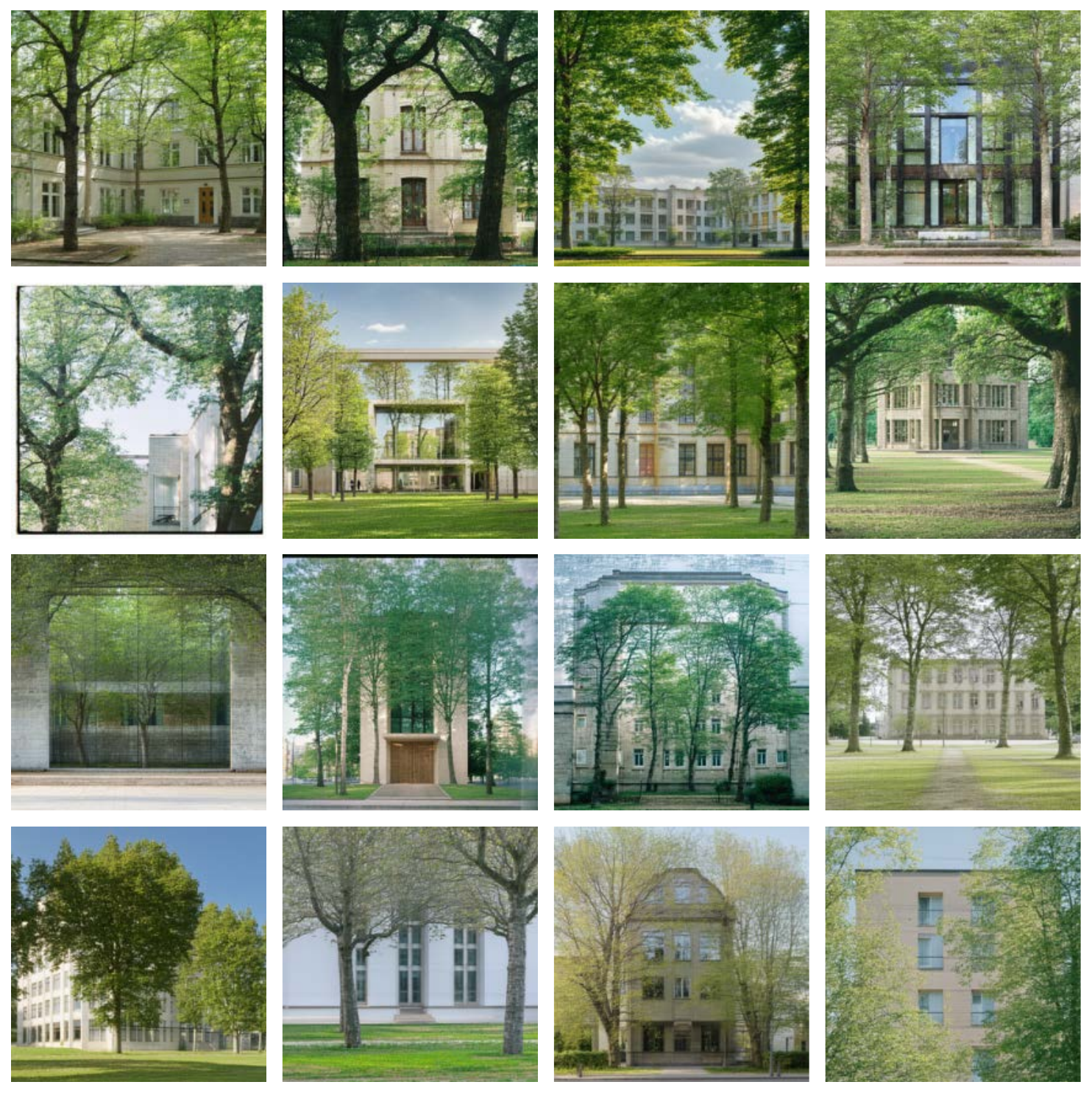}
        \caption{APG guidance}
    \end{subfigure}
    \caption{\textbf{Qualitative visuals for the prompt \texttt{``Trees frame a peaceful building''} using the LDMv3.5L model with different sampling settings.}}
    \label{fig:header_visuals_1}
\end{figure}

\begin{figure}[ht]
    \centering
    \begin{subfigure}[ht]{0.25\textwidth}
        \includegraphics[width=\textwidth]{Imgs/visual_non_human/qualitative_data_41_1.pdf}
        \caption{Dataset}
    \end{subfigure}
    \begin{subfigure}[ht]{0.25\textwidth}
        \includegraphics[width=\textwidth]{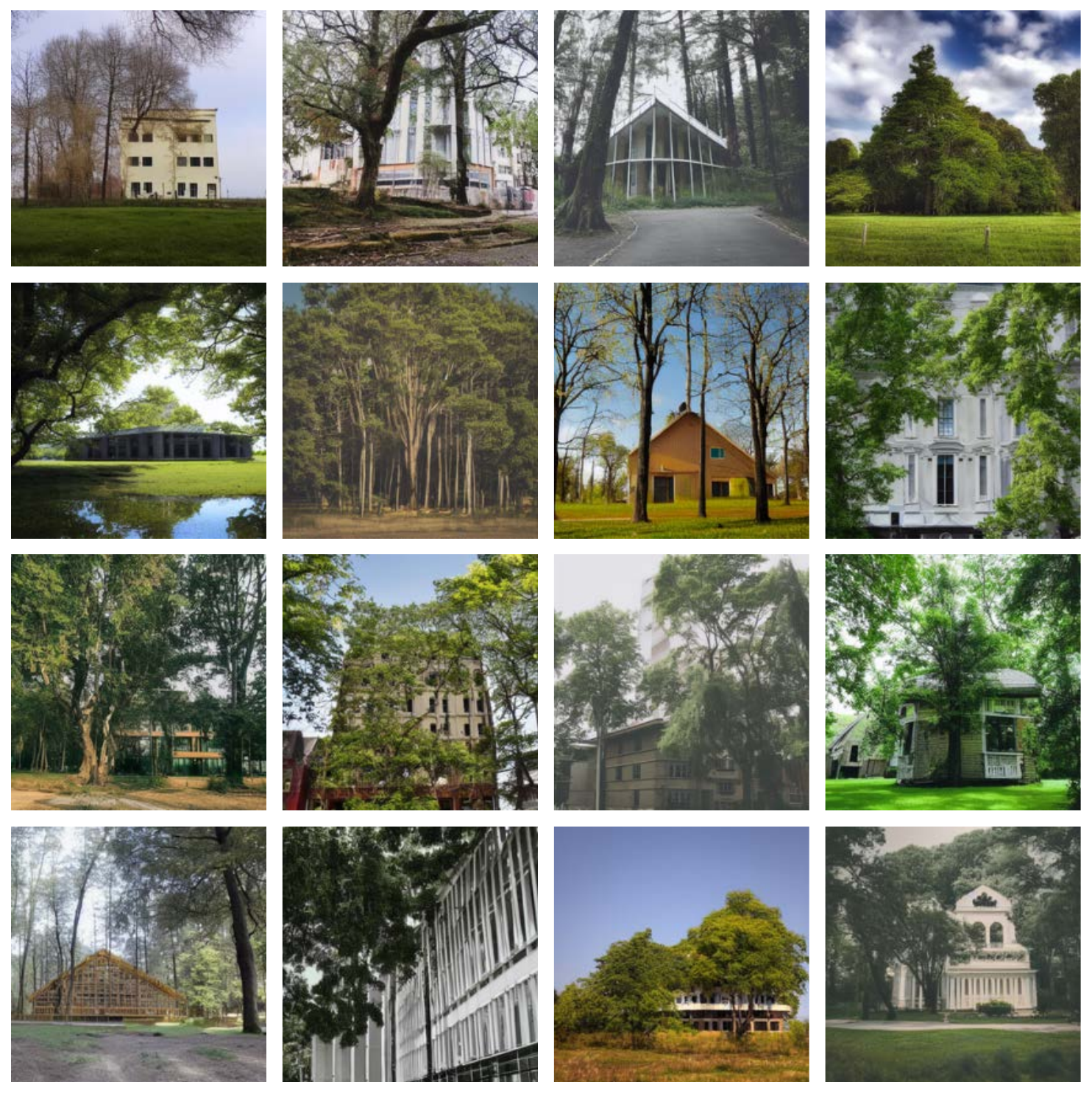}
        \caption{Vanilla guidance}
    \end{subfigure}
    \begin{subfigure}[ht]{0.25\textwidth}
        \includegraphics[width=\textwidth]{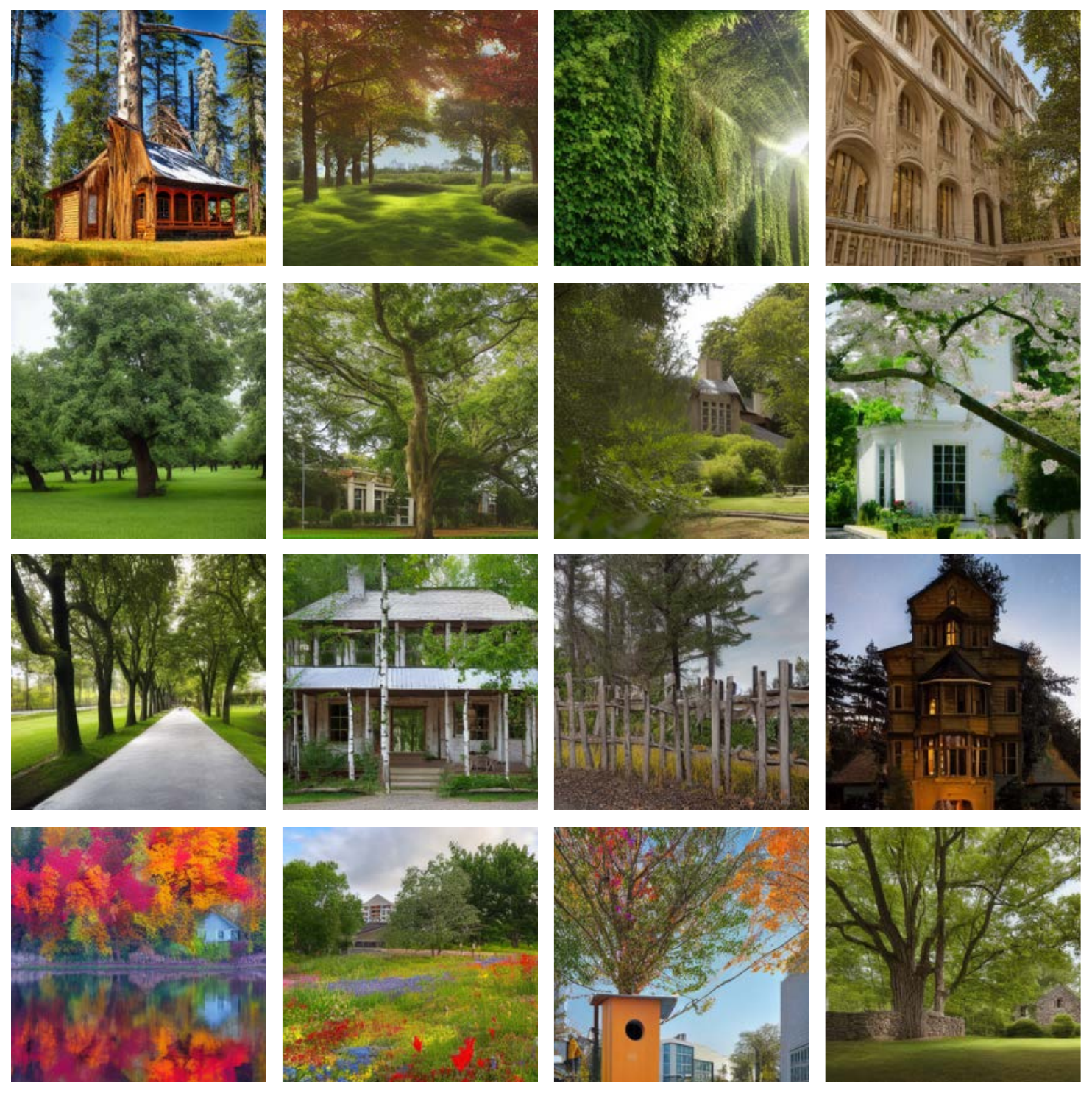}
        \caption{Prompt expansion}
    \end{subfigure} \\
    \begin{subfigure}[ht]{0.25\textwidth}
        \includegraphics[width=\textwidth]{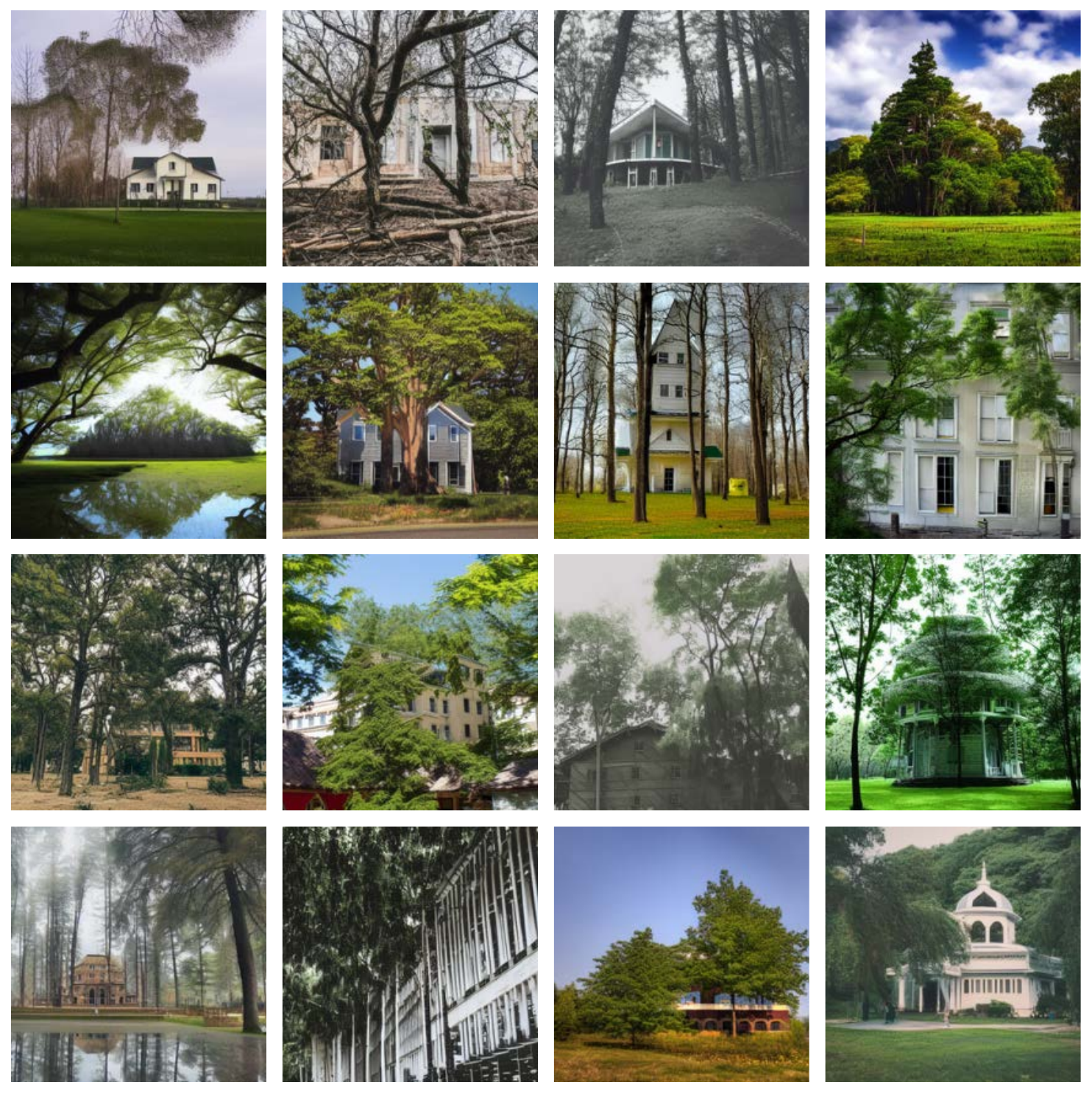}
        \caption{CADS guidance}
    \end{subfigure}
    \begin{subfigure}[ht]{0.25\textwidth}
        \includegraphics[width=\textwidth]{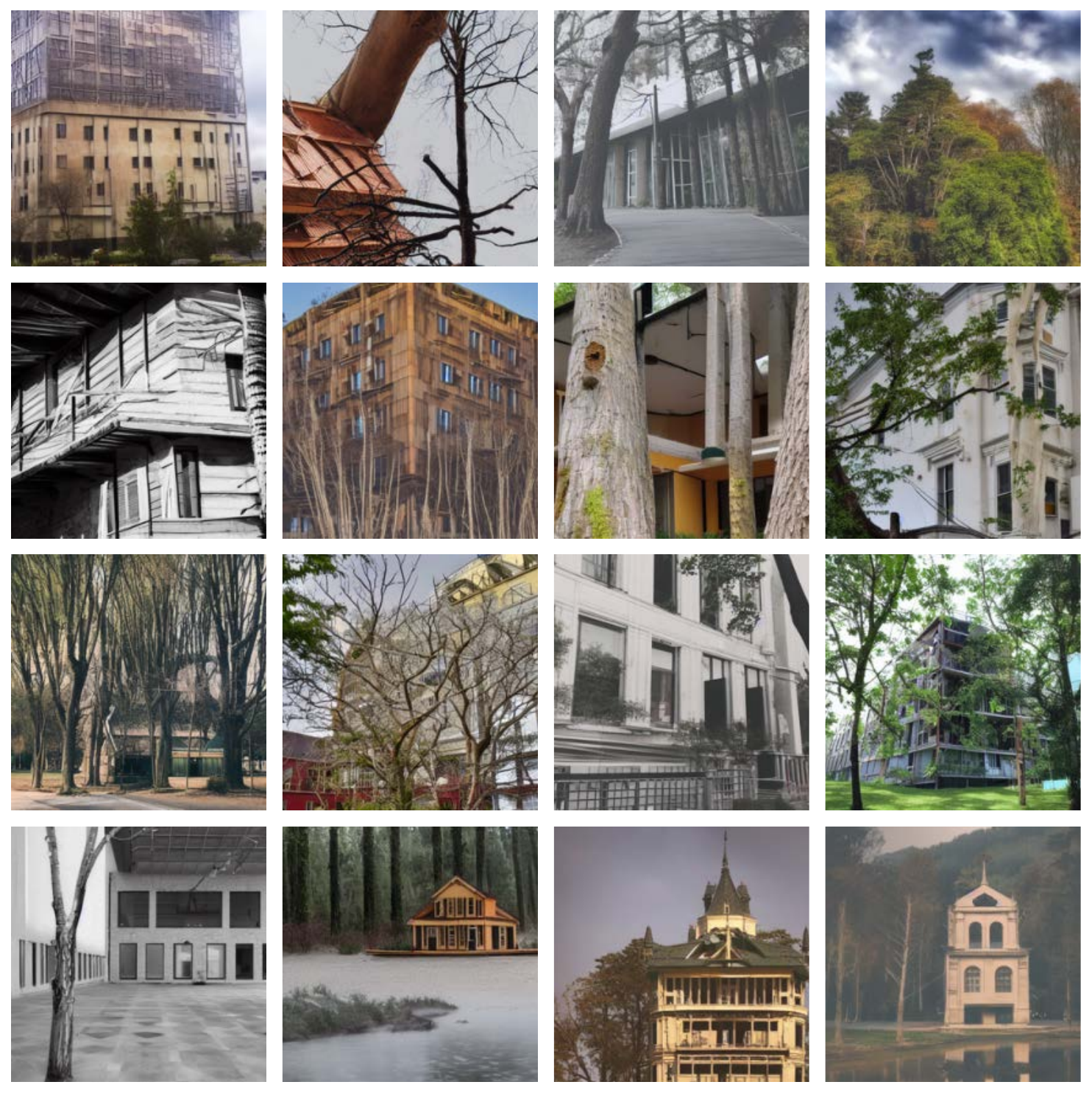}
        \caption{Interval guidance}
    \end{subfigure}
    \begin{subfigure}[ht]{0.25\textwidth}
        \includegraphics[width=\textwidth]{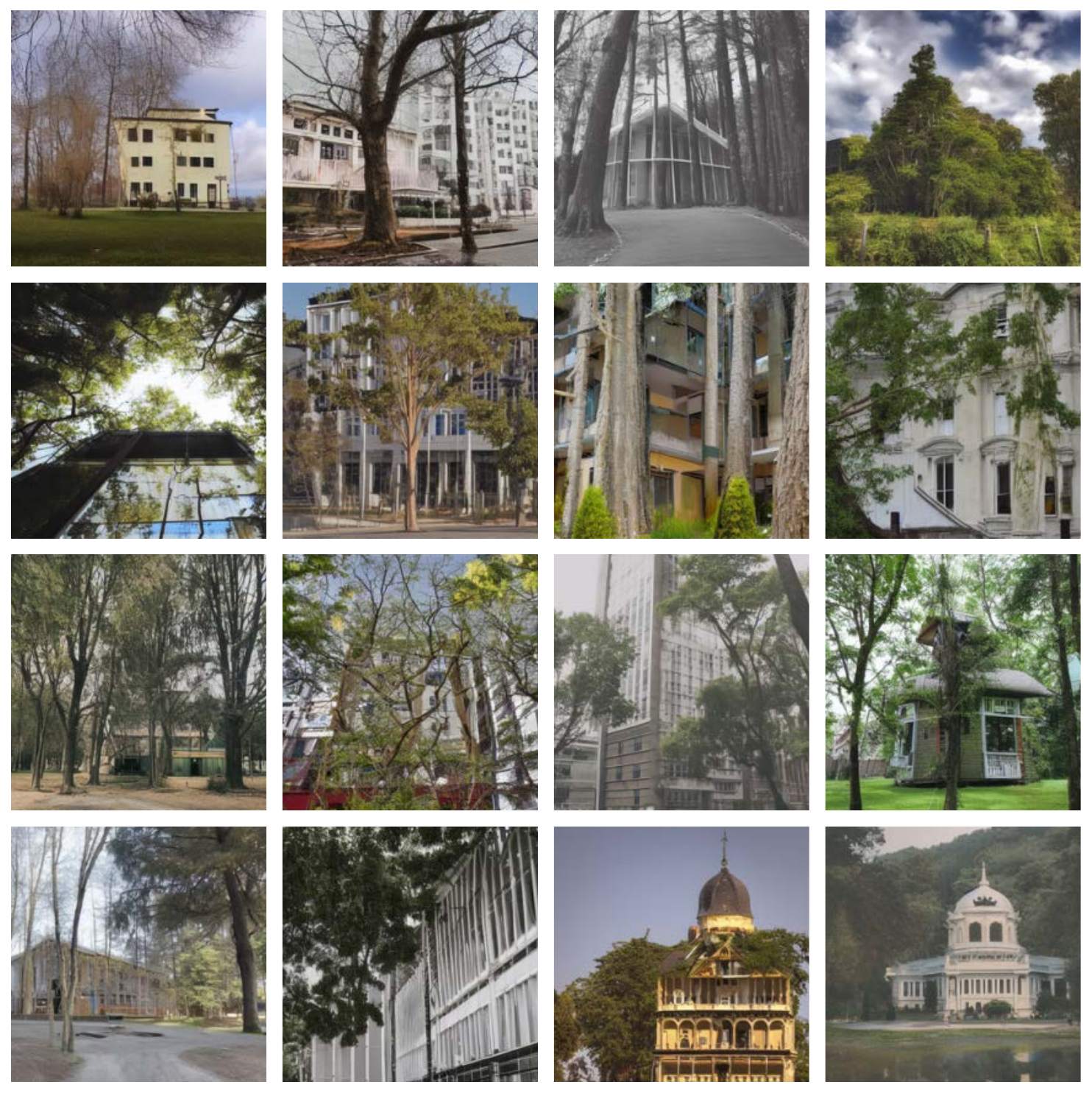}
        \caption{APG guidance}
    \end{subfigure}
    \caption{\textbf{Qualitative visuals for the prompt \texttt{``Trees frame a peaceful building''} using the LDMv1.5 model with different sampling settings.}}
    \label{fig:header_visuals_2}
\end{figure}

\begin{figure}[ht]
    \centering
    \begin{subfigure}[ht]{0.25\textwidth}
        \includegraphics[width=\textwidth]{Imgs/visual_non_human/qualitative_data_41_1.pdf}
        \caption{Dataset}
    \end{subfigure}
    \begin{subfigure}[ht]{0.25\textwidth}
        \includegraphics[width=\textwidth]{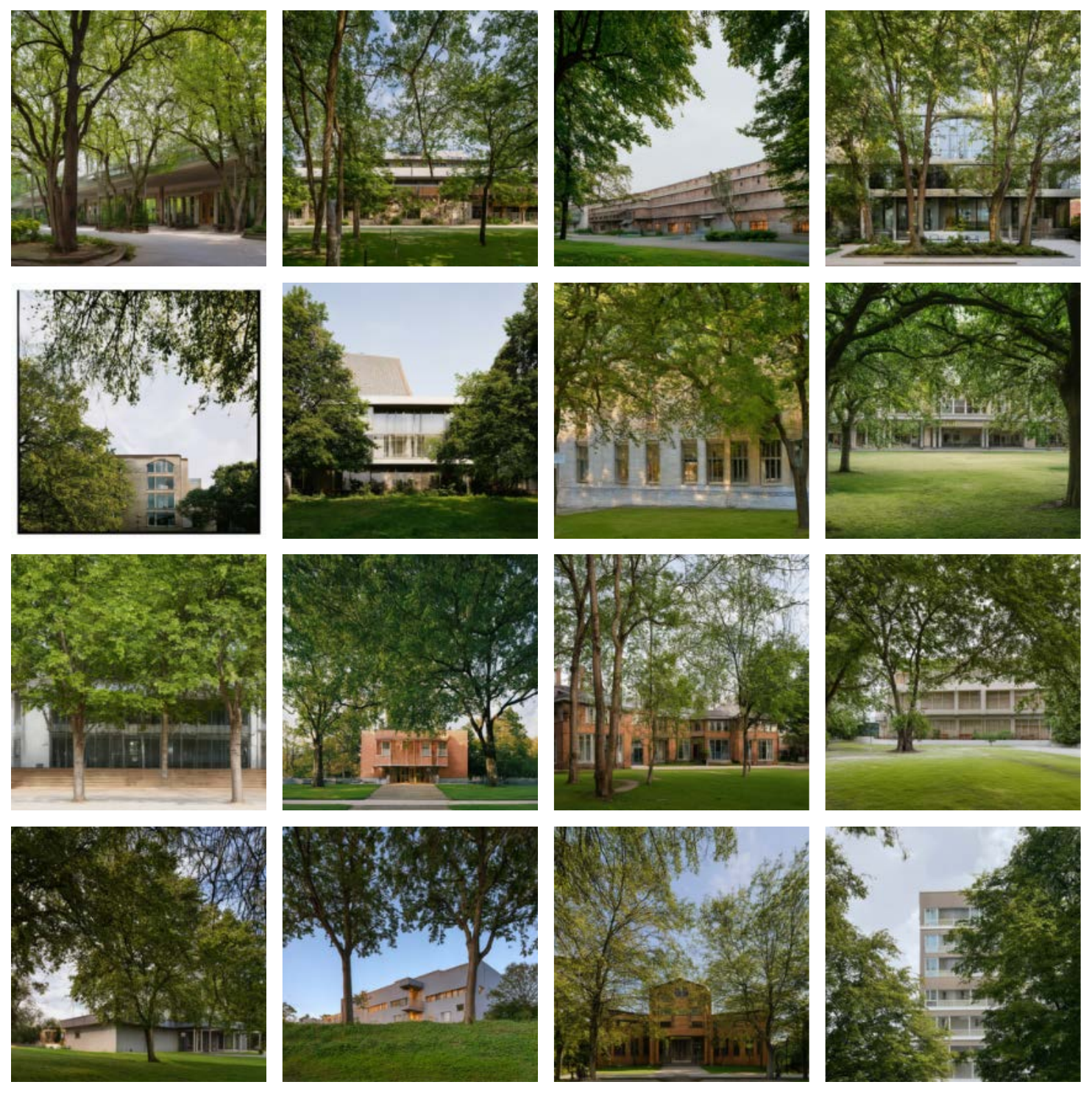}
        \caption{Vanilla guidance}
    \end{subfigure}
    \begin{subfigure}[ht]{0.25\textwidth}
        \includegraphics[width=\textwidth]{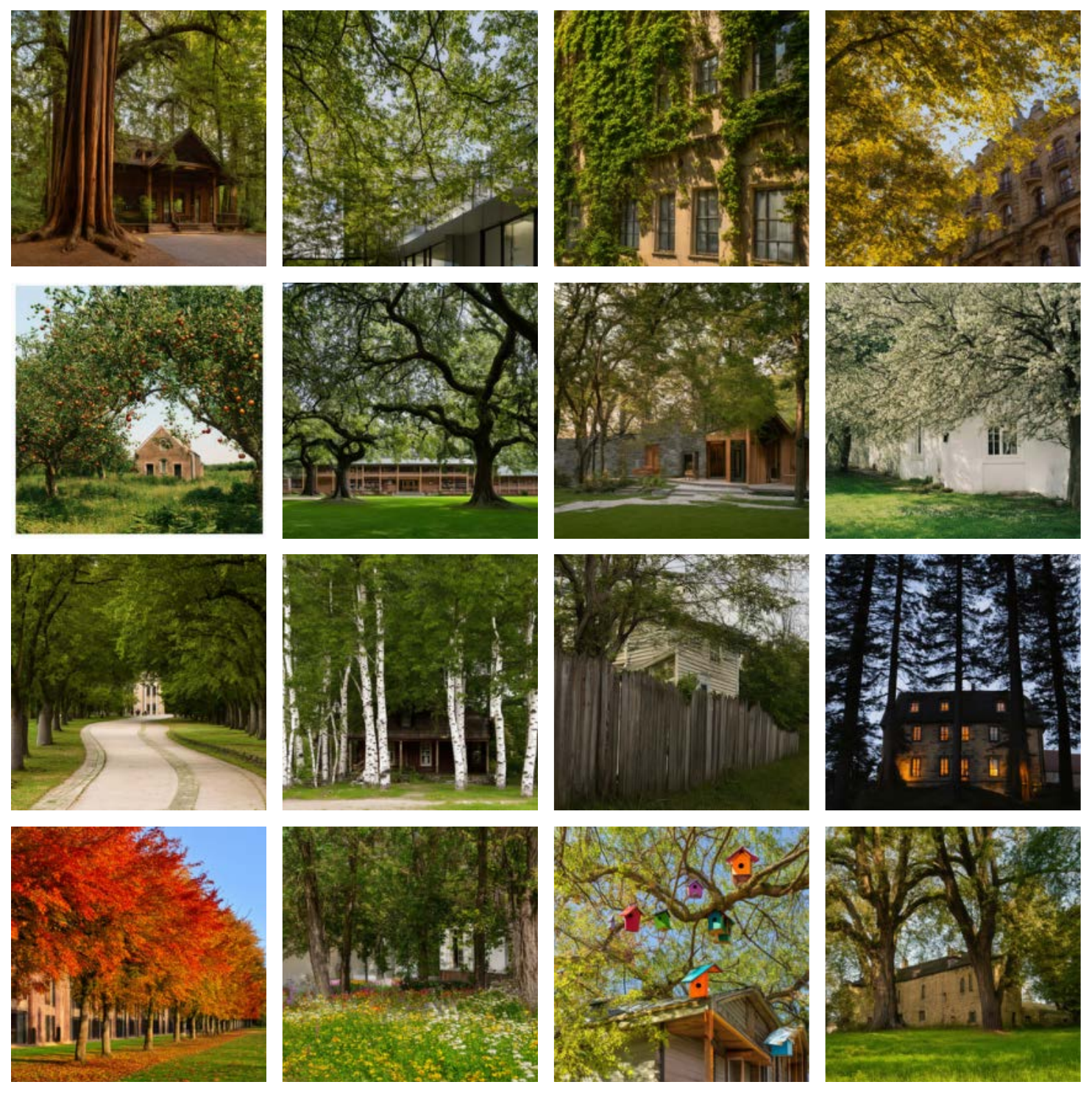}
        \caption{Prompt expansion}
    \end{subfigure} \\
    \begin{subfigure}[ht]{0.25\textwidth}
        \includegraphics[width=\textwidth]{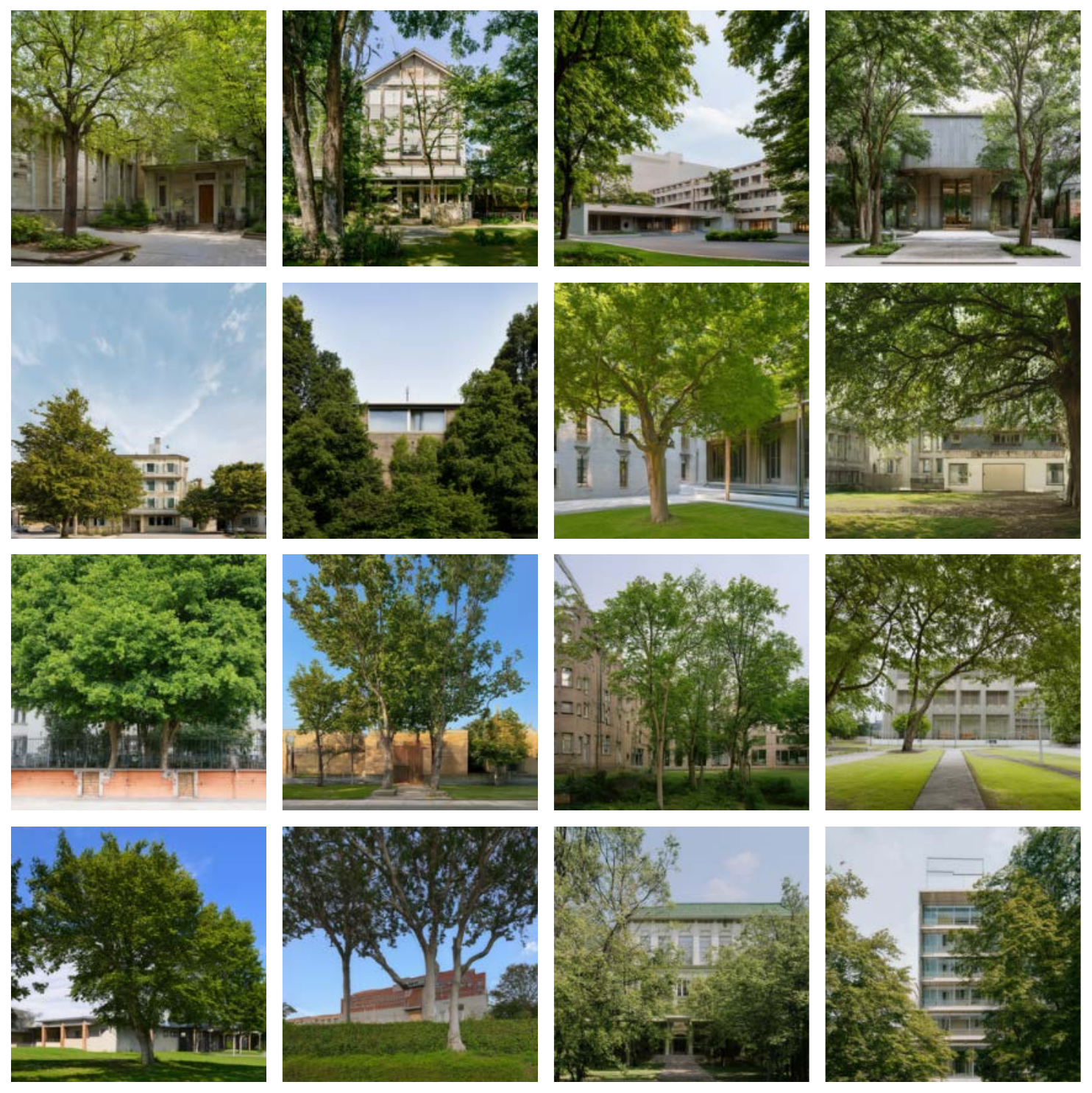}
        \caption{CADS guidance}
    \end{subfigure}
    \begin{subfigure}[ht]{0.25\textwidth}
        \includegraphics[width=\textwidth]{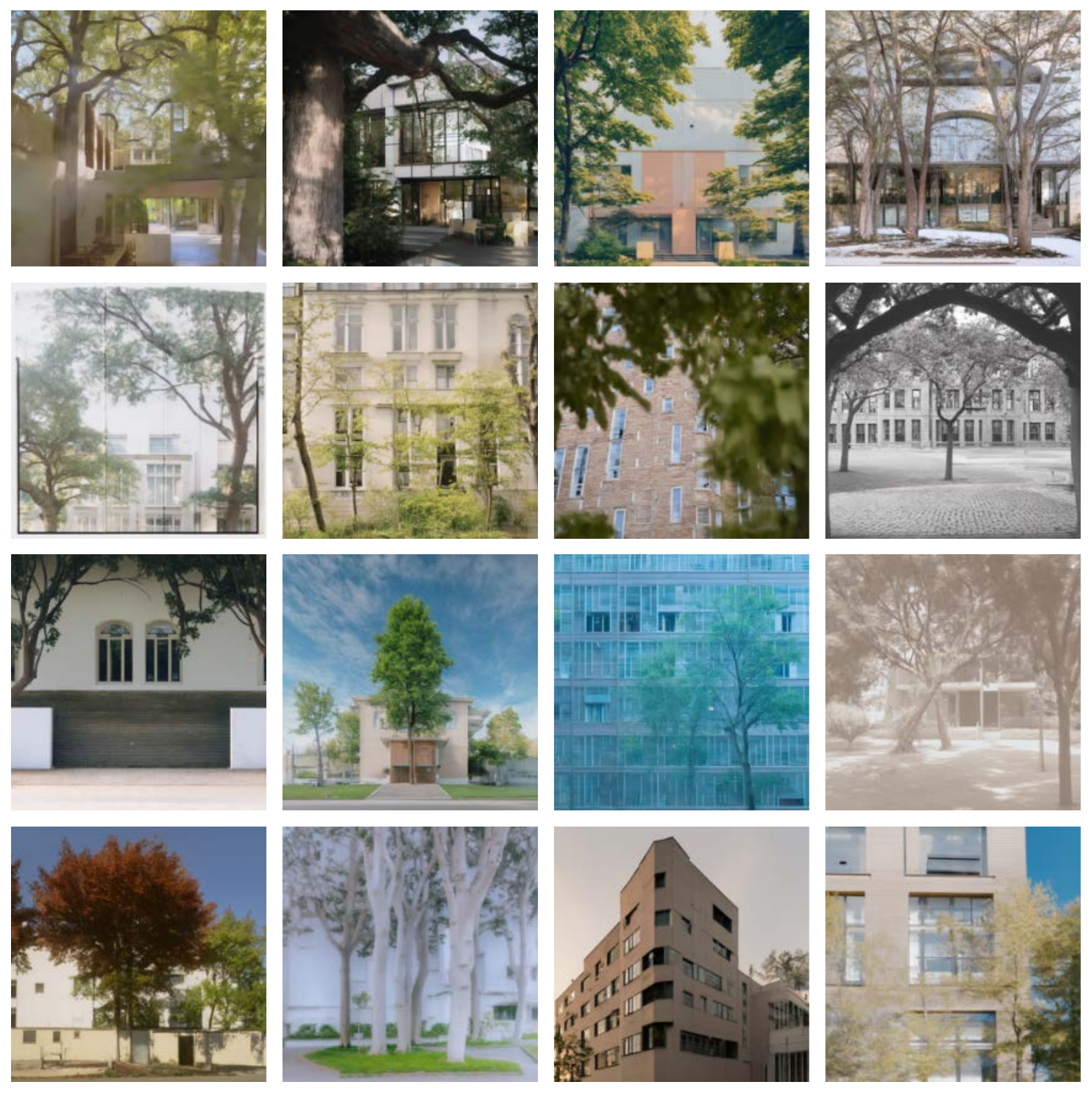}
        \caption{Interval guidance}
    \end{subfigure}
    \begin{subfigure}[ht]{0.25\textwidth}
        \includegraphics[width=\textwidth]{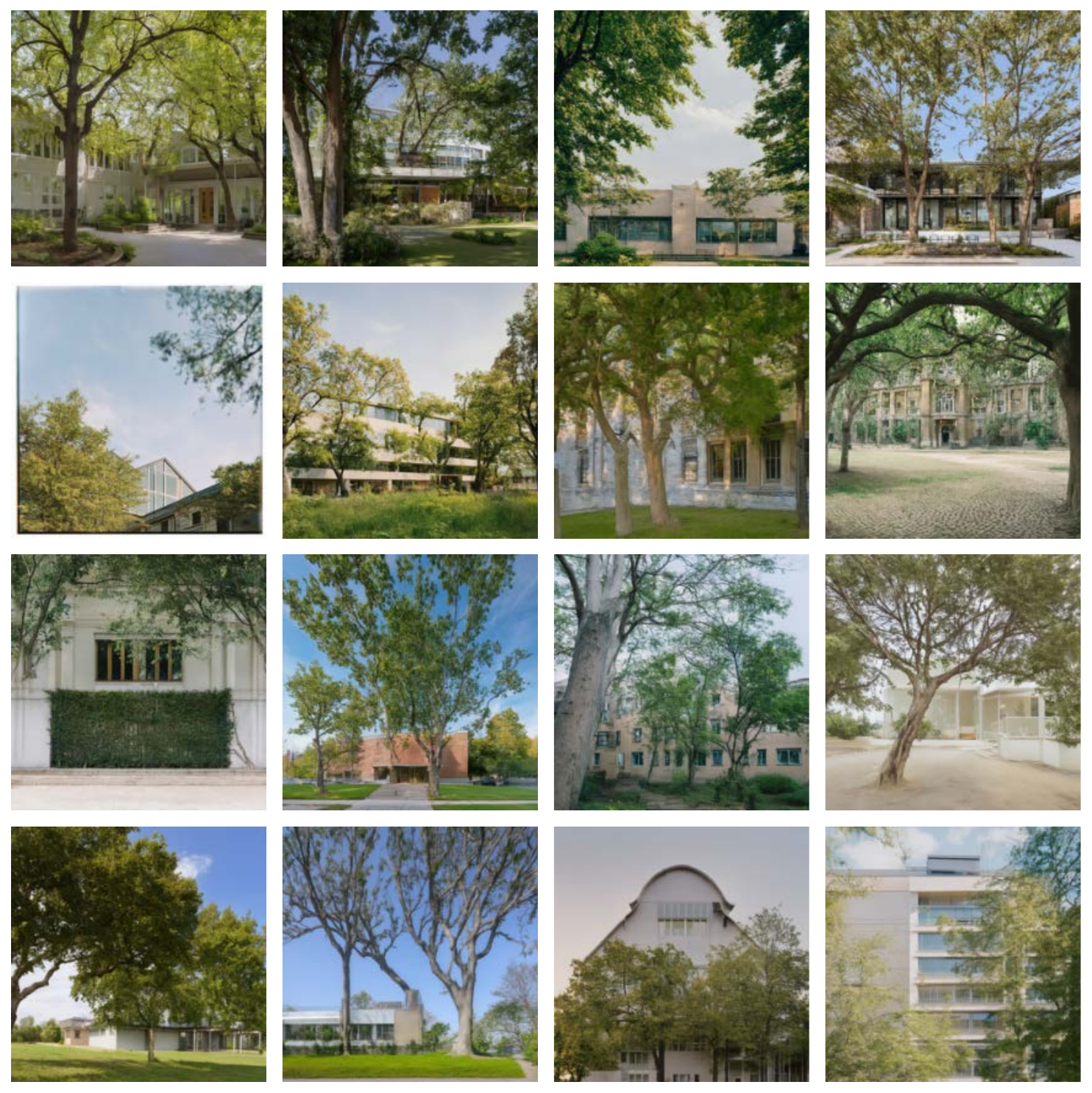}
        \caption{APG guidance}
    \end{subfigure}
    \caption{\textbf{Qualitative visuals for the prompt \texttt{``Trees frame a peaceful building''} using the LDMv3.5M model with different sampling settings.}}
    \label{fig:header_visuals_3}
\end{figure}

\begin{figure}[ht]
    \centering
    \begin{subfigure}[ht]{0.25\textwidth}
        \includegraphics[width=\textwidth]{Imgs/visual_non_human/qualitative_data_41_1.pdf}
        \caption{Dataset}
    \end{subfigure}
    \begin{subfigure}[ht]{0.25\textwidth}
        \includegraphics[width=\textwidth]{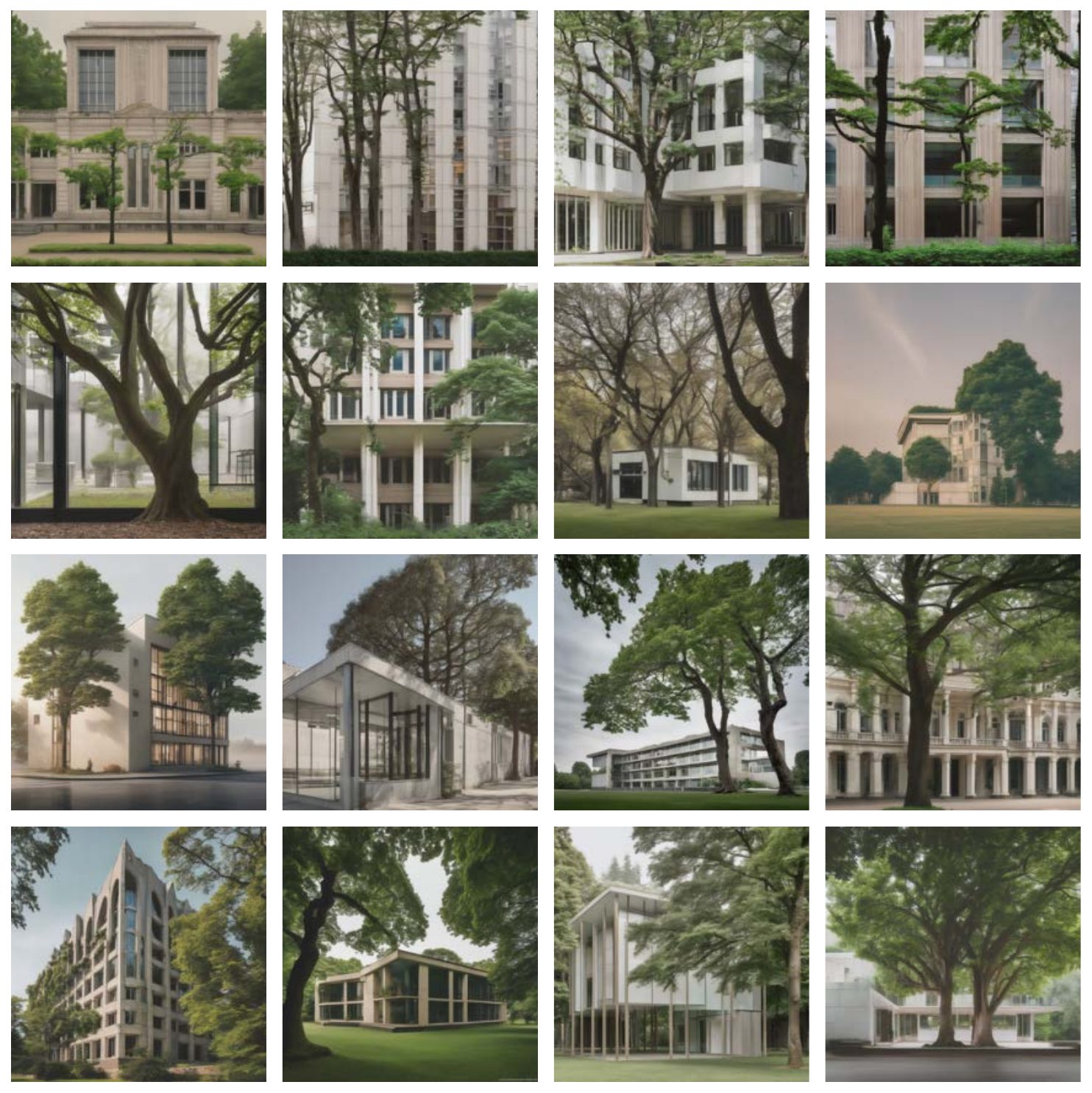}
        \caption{Vanilla guidance}
    \end{subfigure}
    \begin{subfigure}[ht]{0.25\textwidth}
        \includegraphics[width=\textwidth]{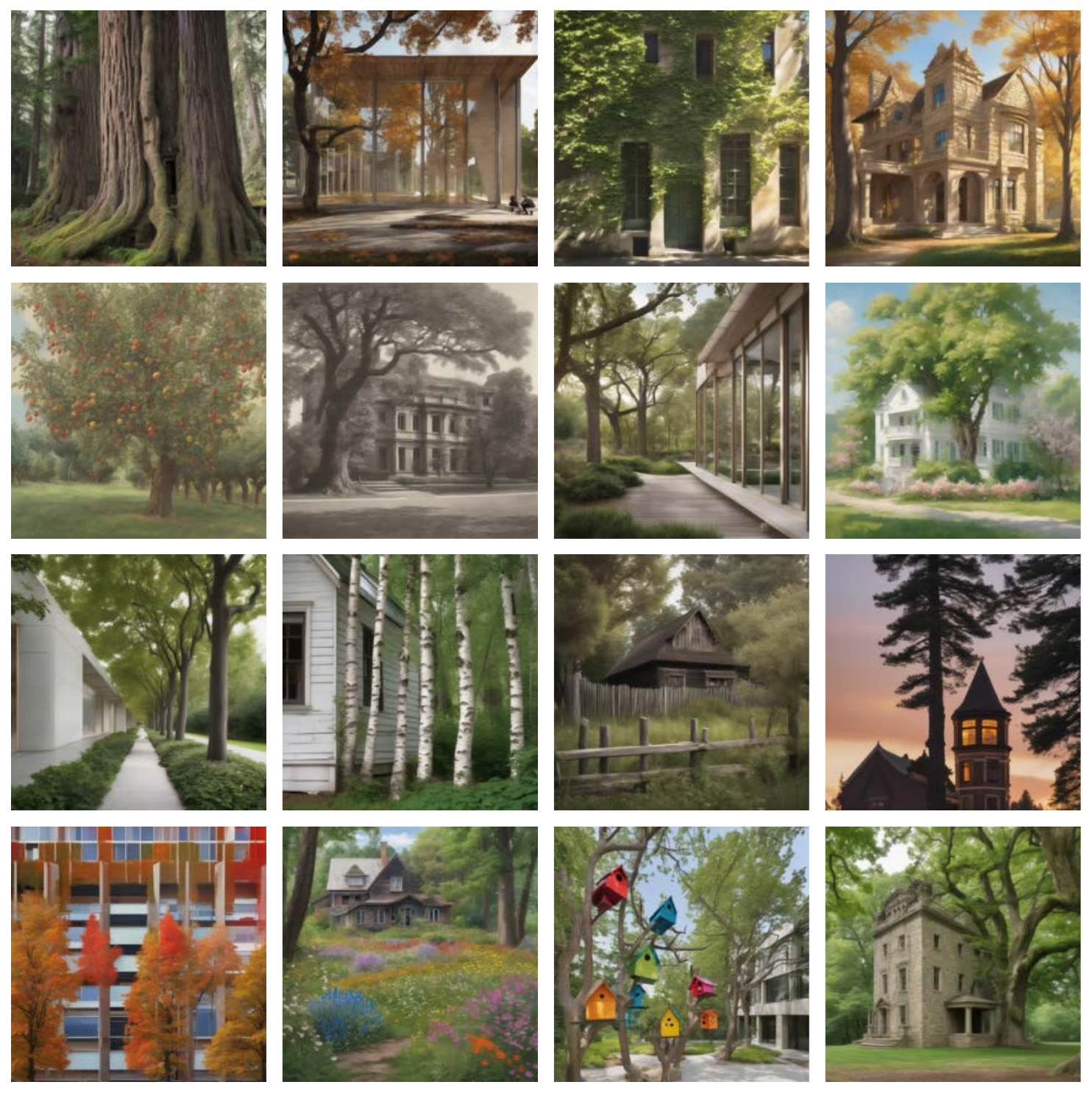}
        \caption{Prompt expansion}
    \end{subfigure} \\
    \begin{subfigure}[ht]{0.25\textwidth}
        \includegraphics[width=\textwidth]{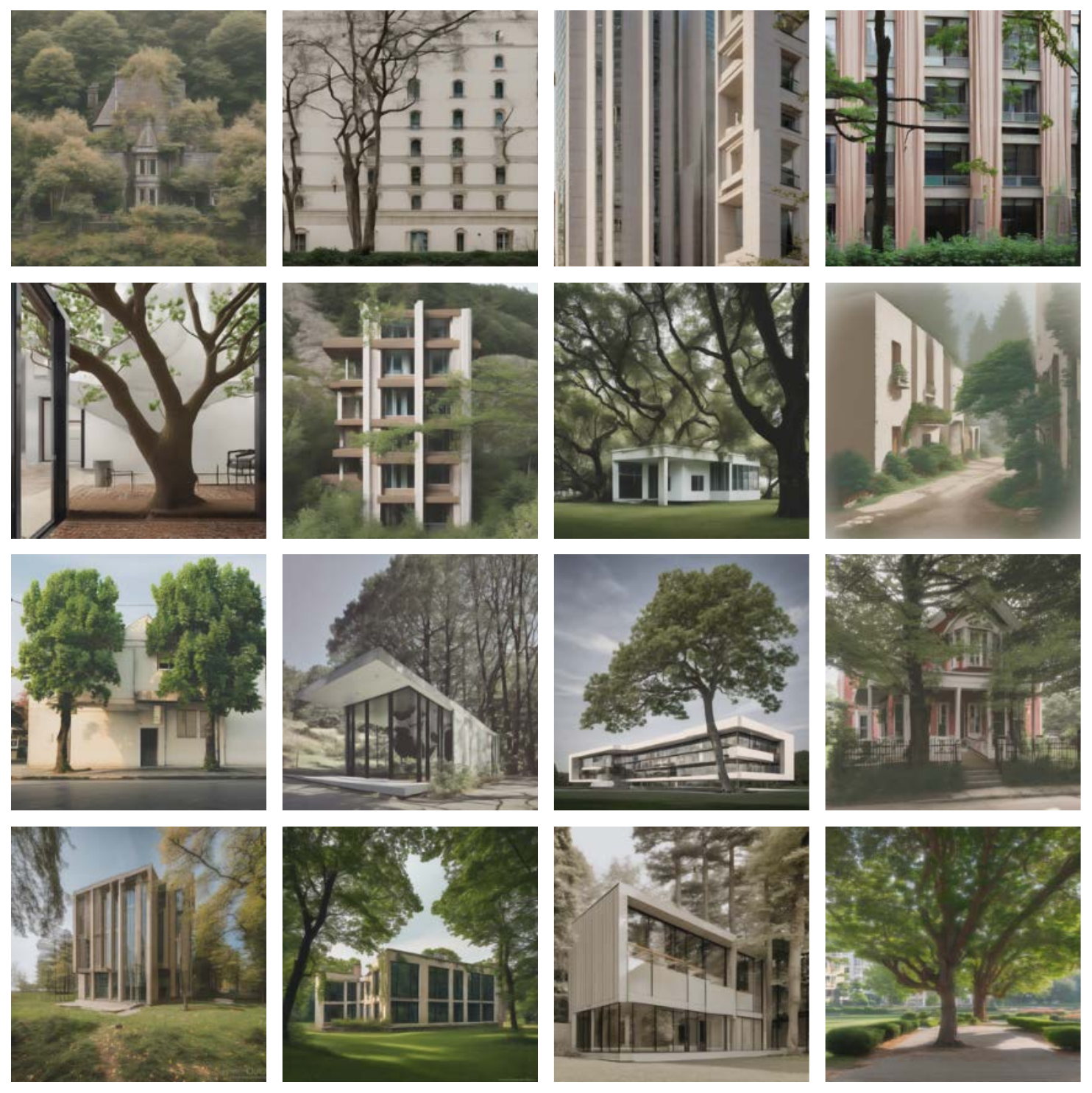}
        \caption{CADS guidance}
    \end{subfigure}
    \begin{subfigure}[ht]{0.25\textwidth}
        \includegraphics[width=\textwidth]{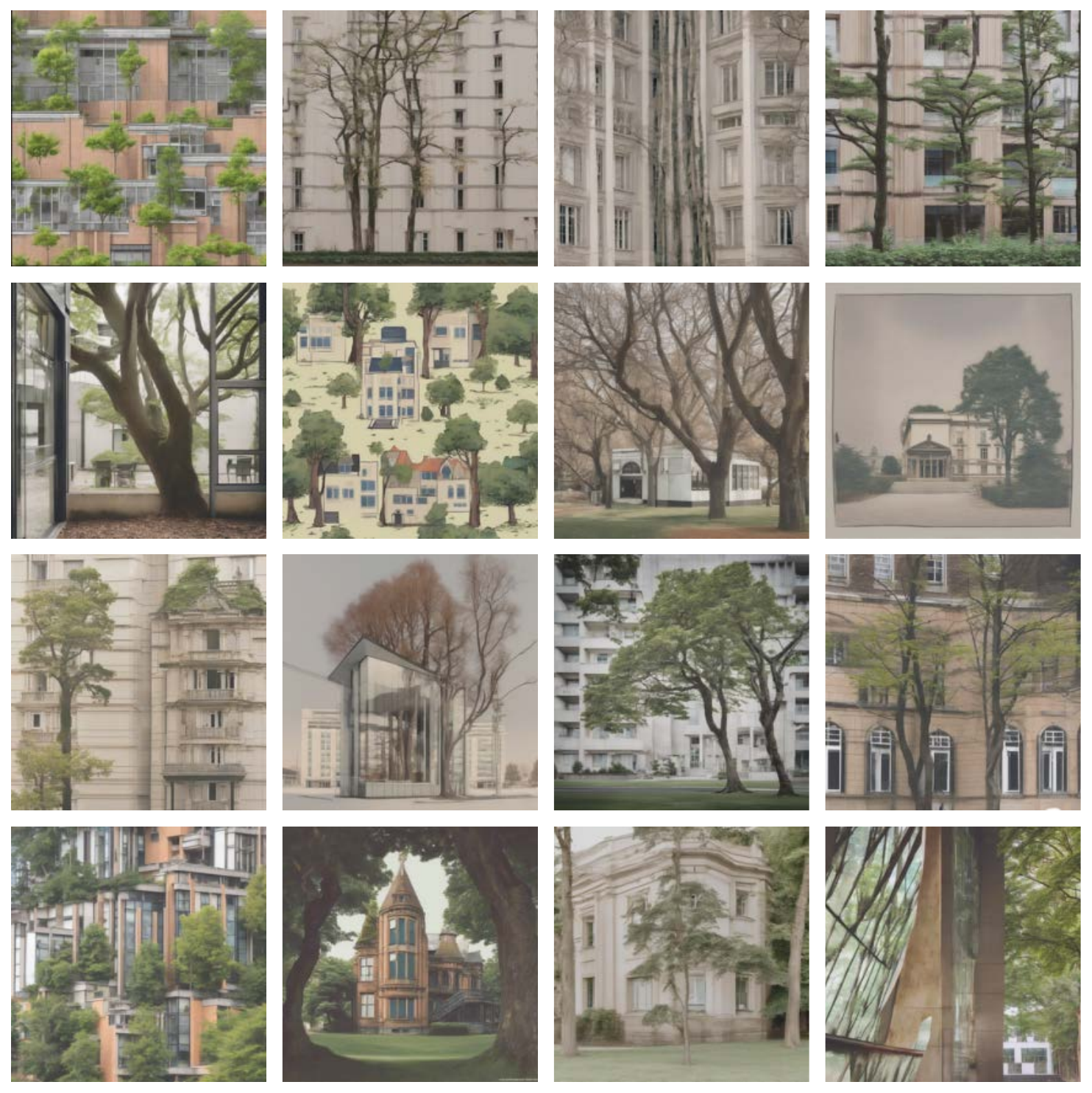}
        \caption{Interval guidance}
    \end{subfigure}
    \begin{subfigure}[ht]{0.25\textwidth}
        \includegraphics[width=\textwidth]{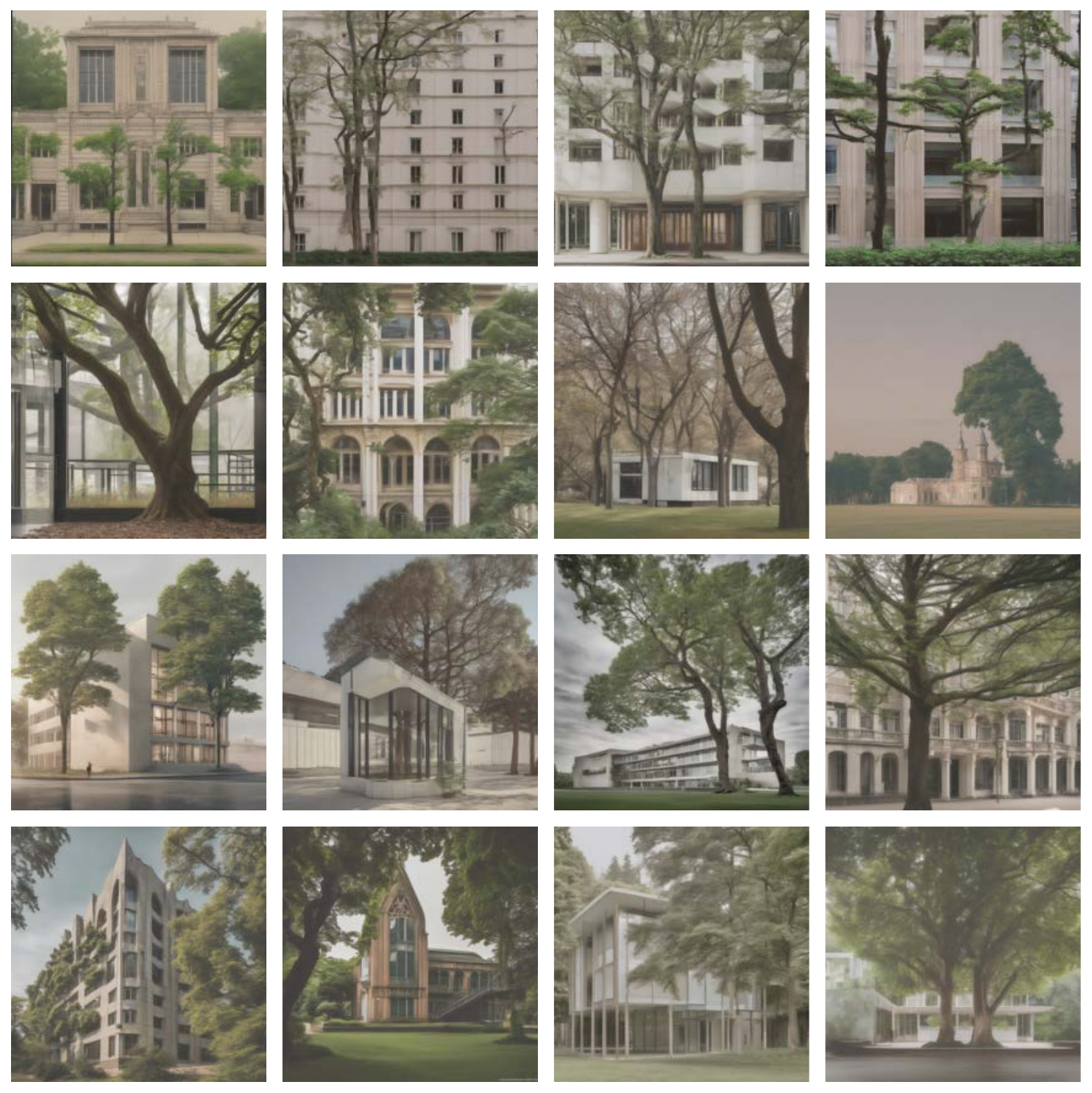}
        \caption{APG guidance}
    \end{subfigure}
    \caption{\textbf{Qualitative visuals for the prompt \texttt{``Trees frame a peaceful building''} using the LDM-XL model with different sampling settings.}}
    \label{fig:header_visuals_4}
\end{figure}

\begin{figure}[ht]
    \centering
    \begin{subfigure}[ht]{0.25\textwidth}
        \includegraphics[width=\textwidth]{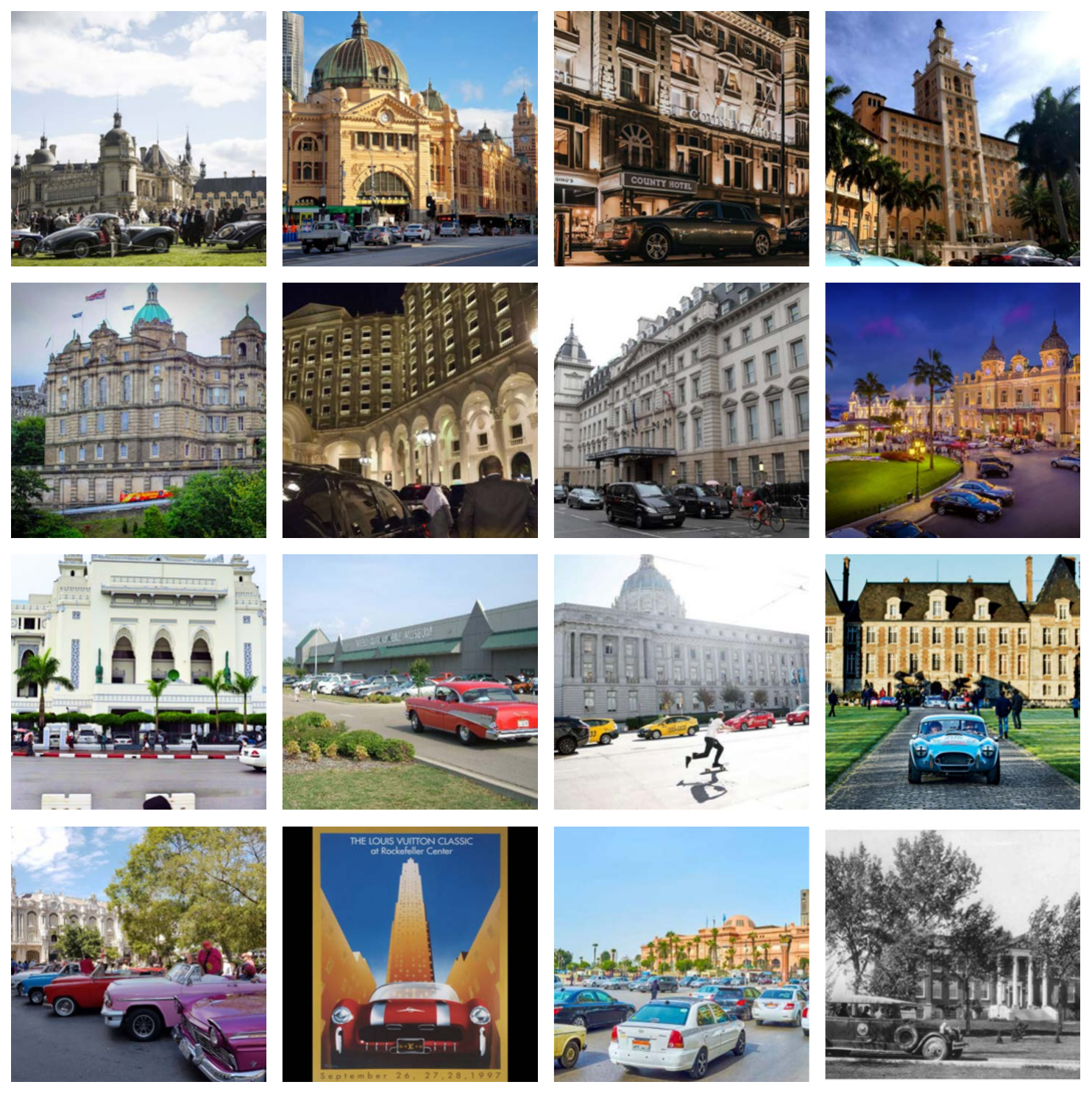}
        \caption{Dataset}
    \end{subfigure}
    \begin{subfigure}[ht]{0.25\textwidth}
        \includegraphics[width=\textwidth]{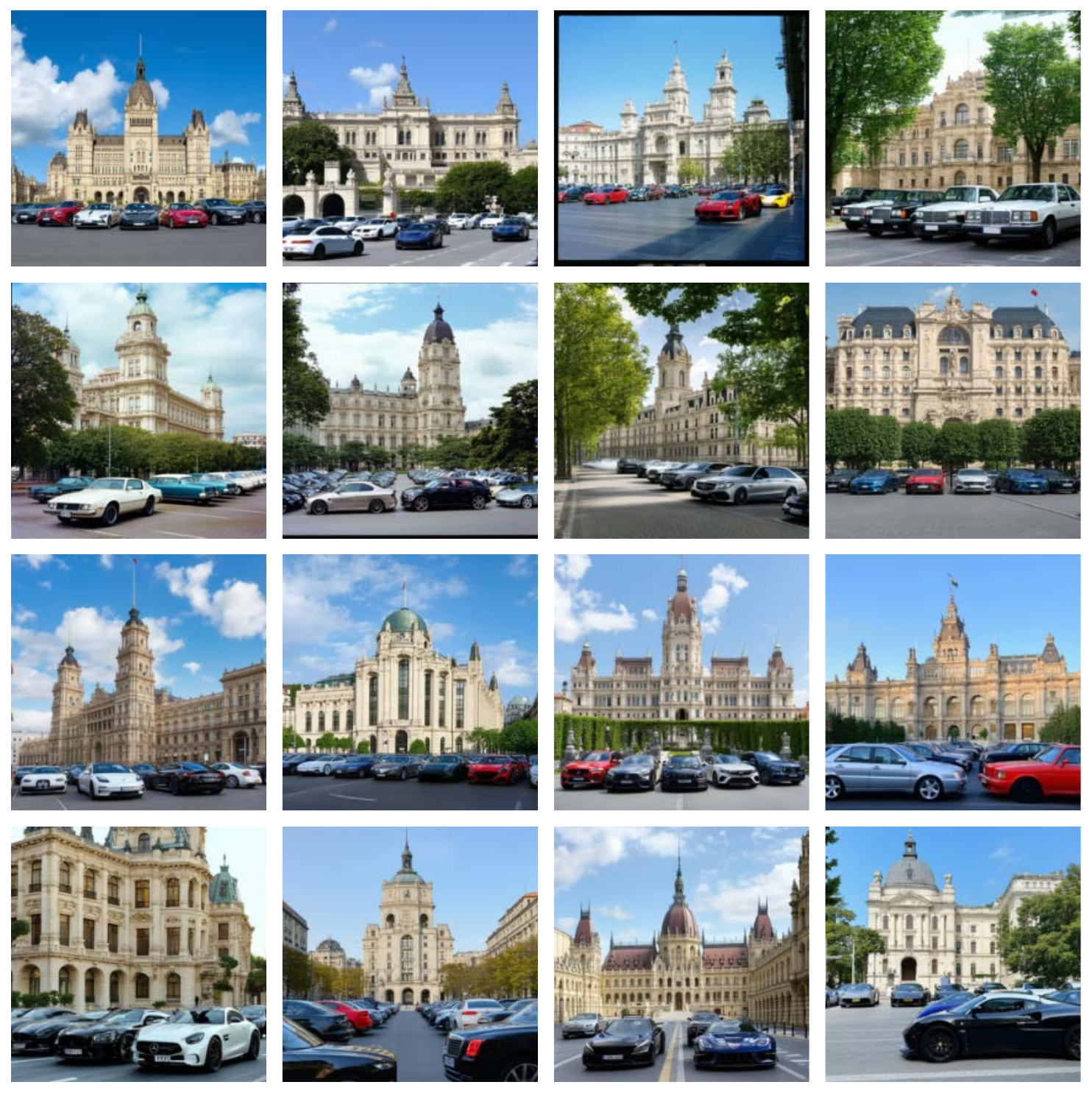}
        \caption{Vanilla guidance}
    \end{subfigure}
    \begin{subfigure}[ht]{0.25\textwidth}
        \includegraphics[width=\textwidth]{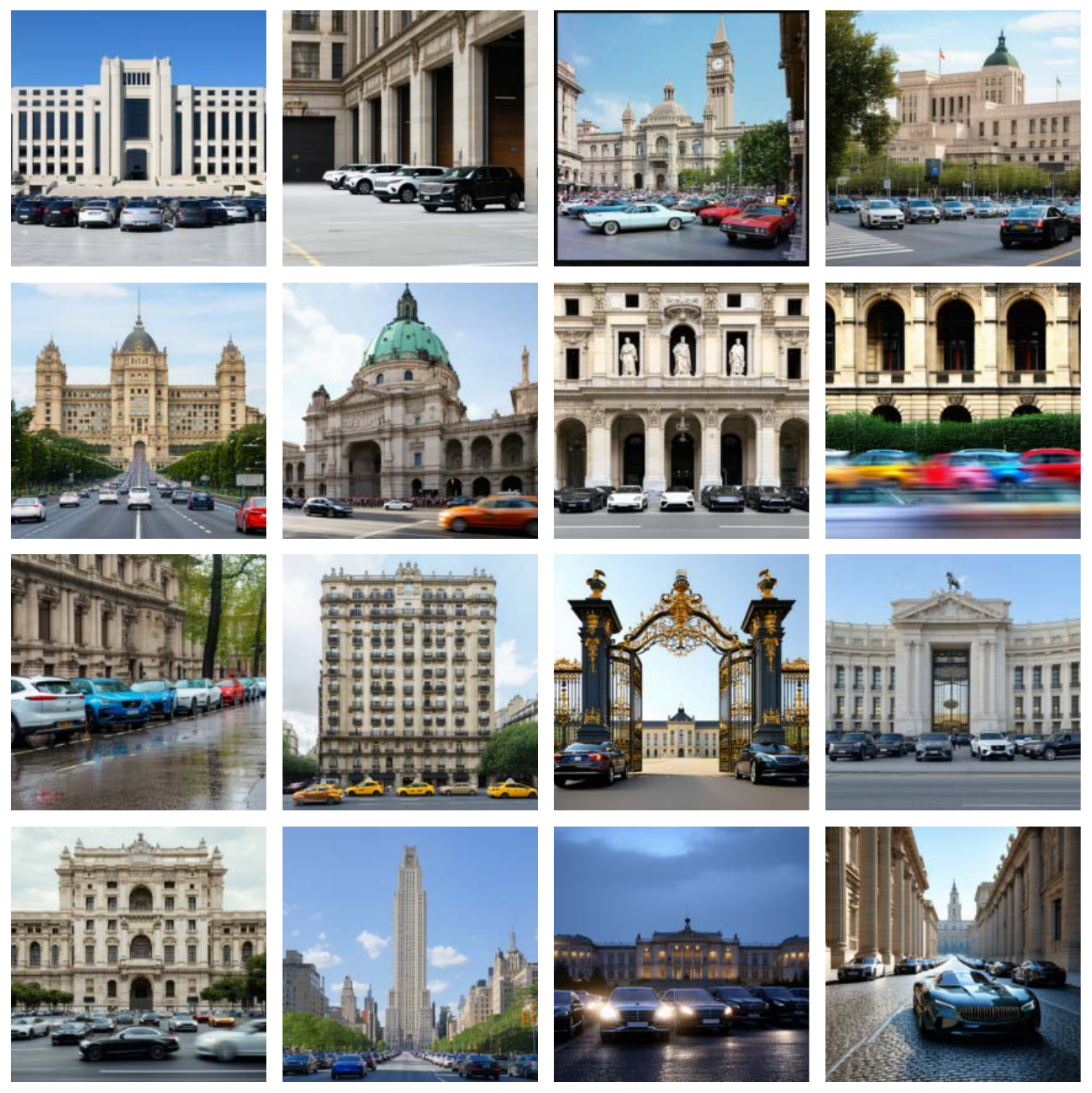}
        \caption{Prompt expansion}
    \end{subfigure} \\
    \begin{subfigure}[ht]{0.25\textwidth}
        \includegraphics[width=\textwidth]{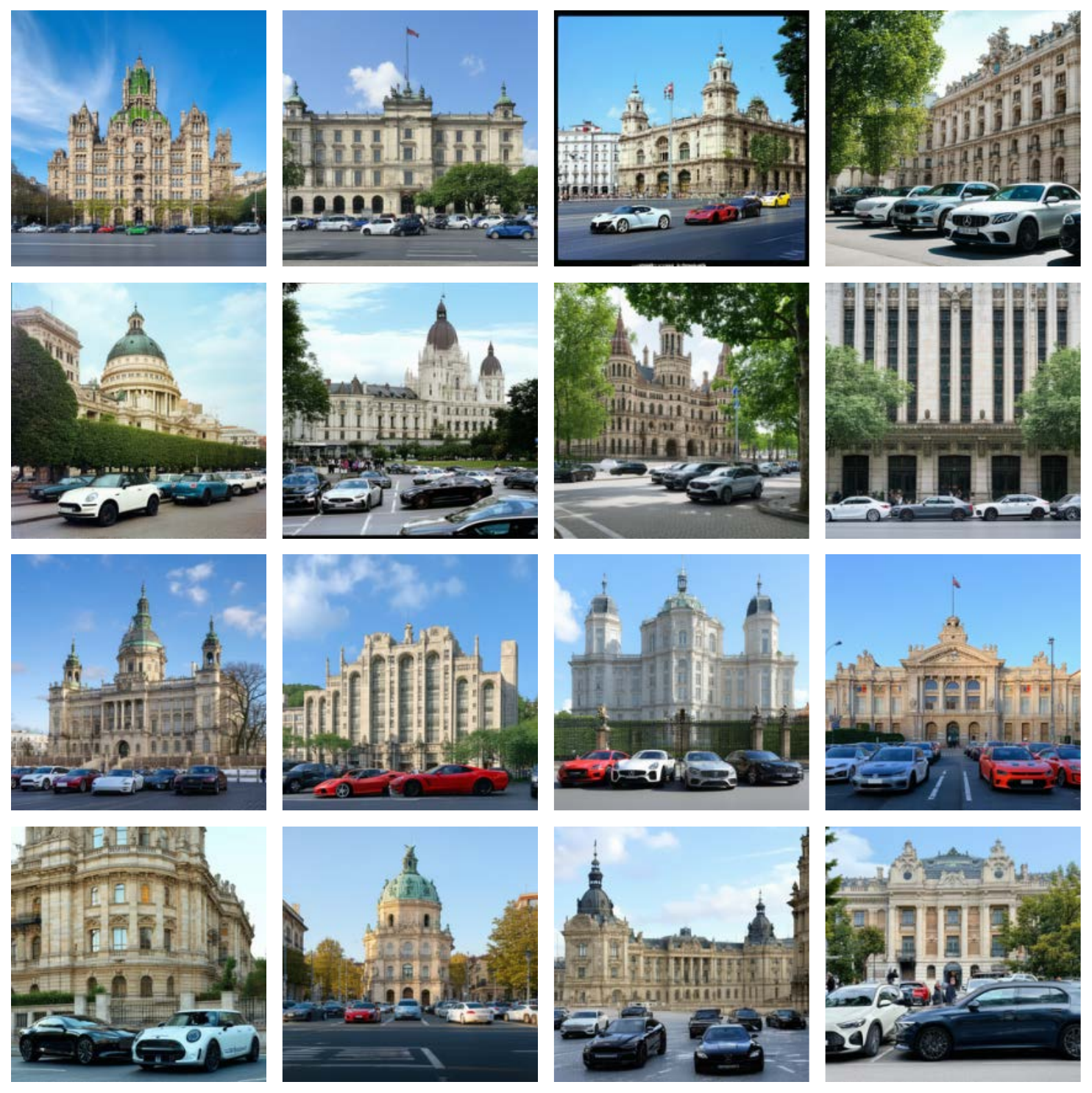}
        \caption{CADS guidance}
    \end{subfigure}
    \begin{subfigure}[ht]{0.25\textwidth}
        \includegraphics[width=\textwidth]{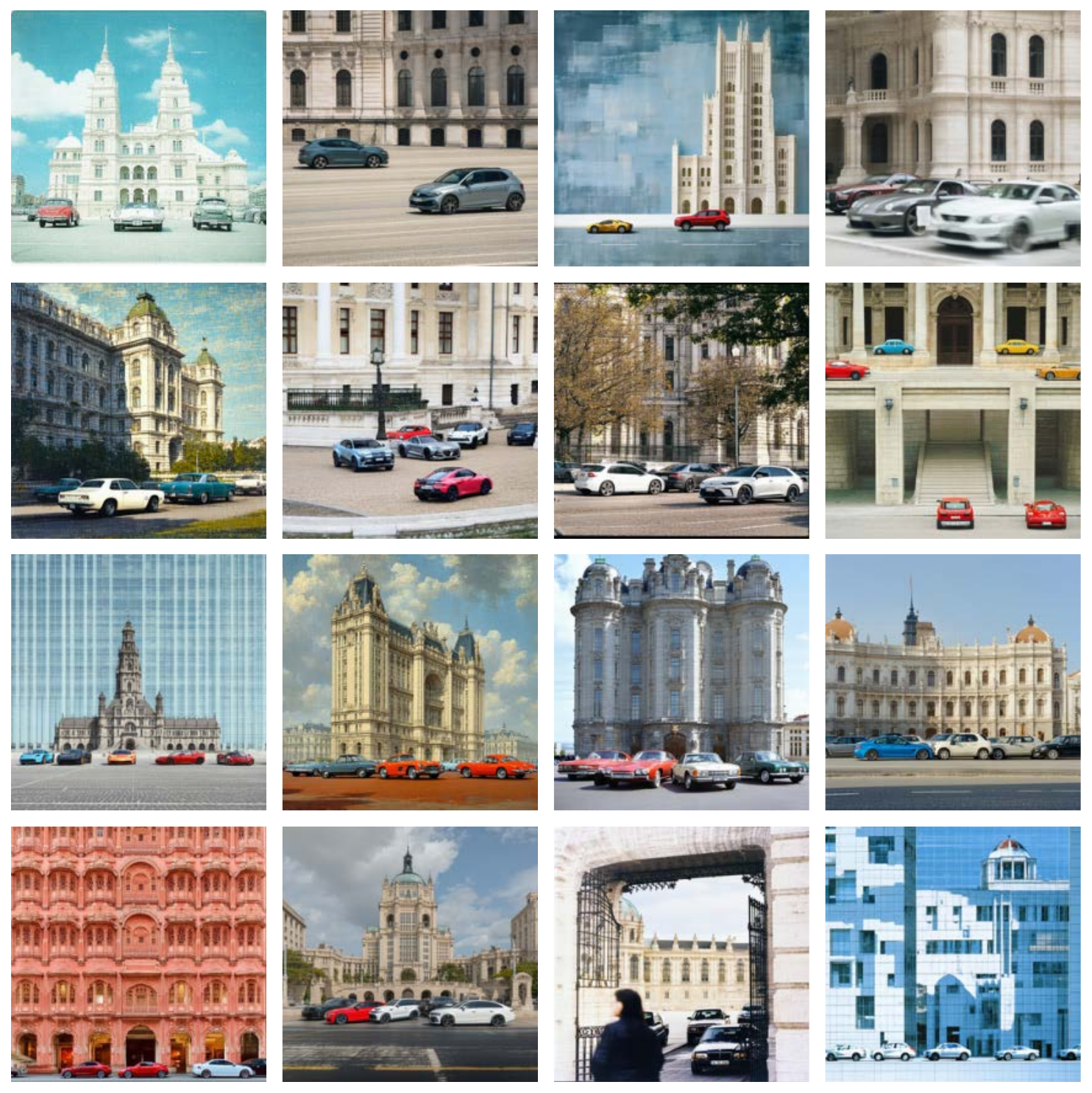}
        \caption{Interval guidance}
    \end{subfigure}
    \begin{subfigure}[ht]{0.25\textwidth}
        \includegraphics[width=\textwidth]{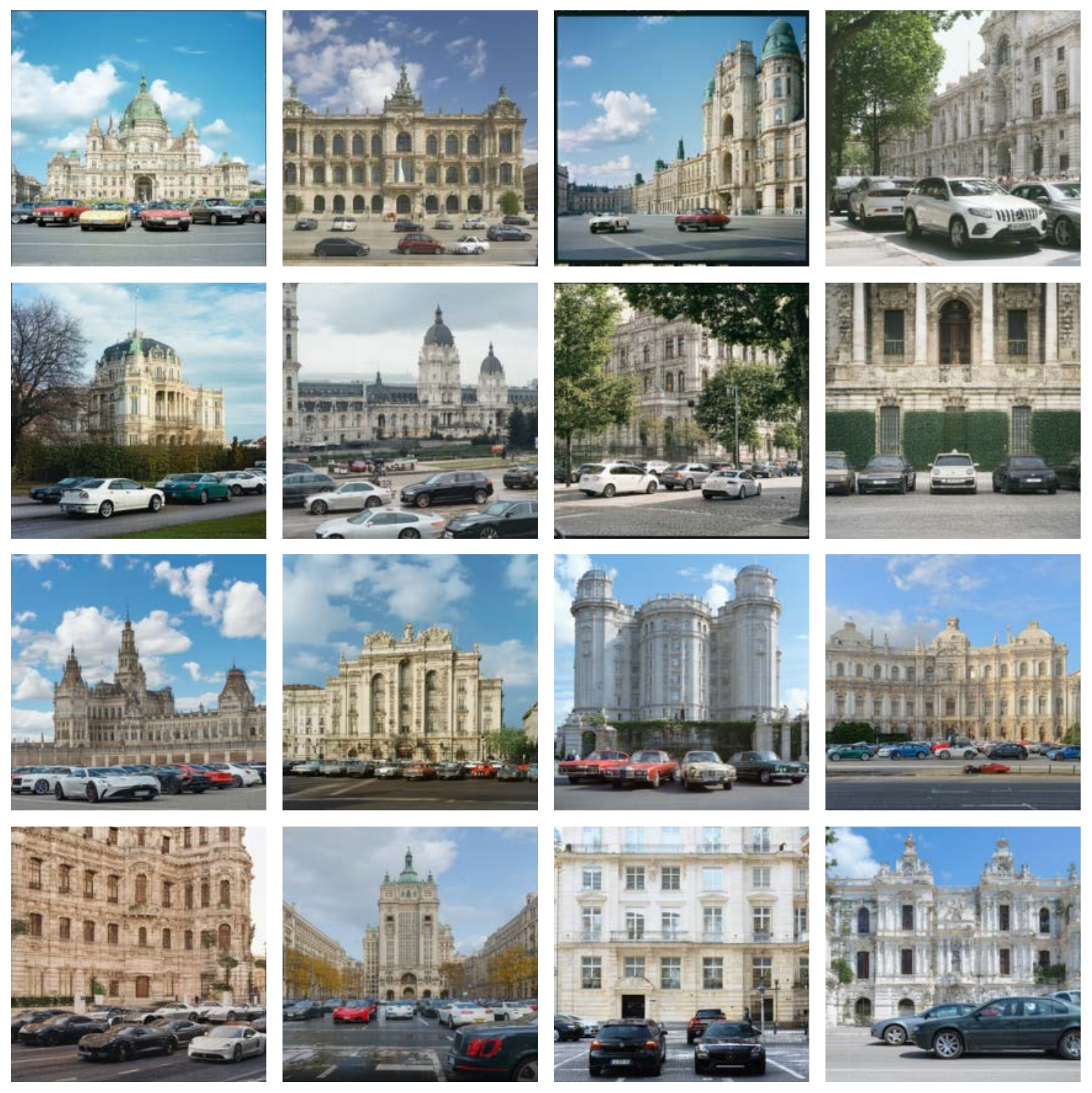}
        \caption{APG guidance}
    \end{subfigure}
    \caption{\textbf{Qualitative visuals for the prompt \texttt{``Cars and a grand building''} using the LDMv3.5L model with different sampling settings.}}
    \label{fig:header_visuals_5}
\end{figure}

\begin{figure}[ht]
    \centering
    \begin{subfigure}[ht]{0.25\textwidth}
        \includegraphics[width=\textwidth]{Imgs/visual_non_human/qualitative_data_7_2.pdf}
        \caption{Dataset}
    \end{subfigure}
    \begin{subfigure}[ht]{0.25\textwidth}
        \includegraphics[width=\textwidth]{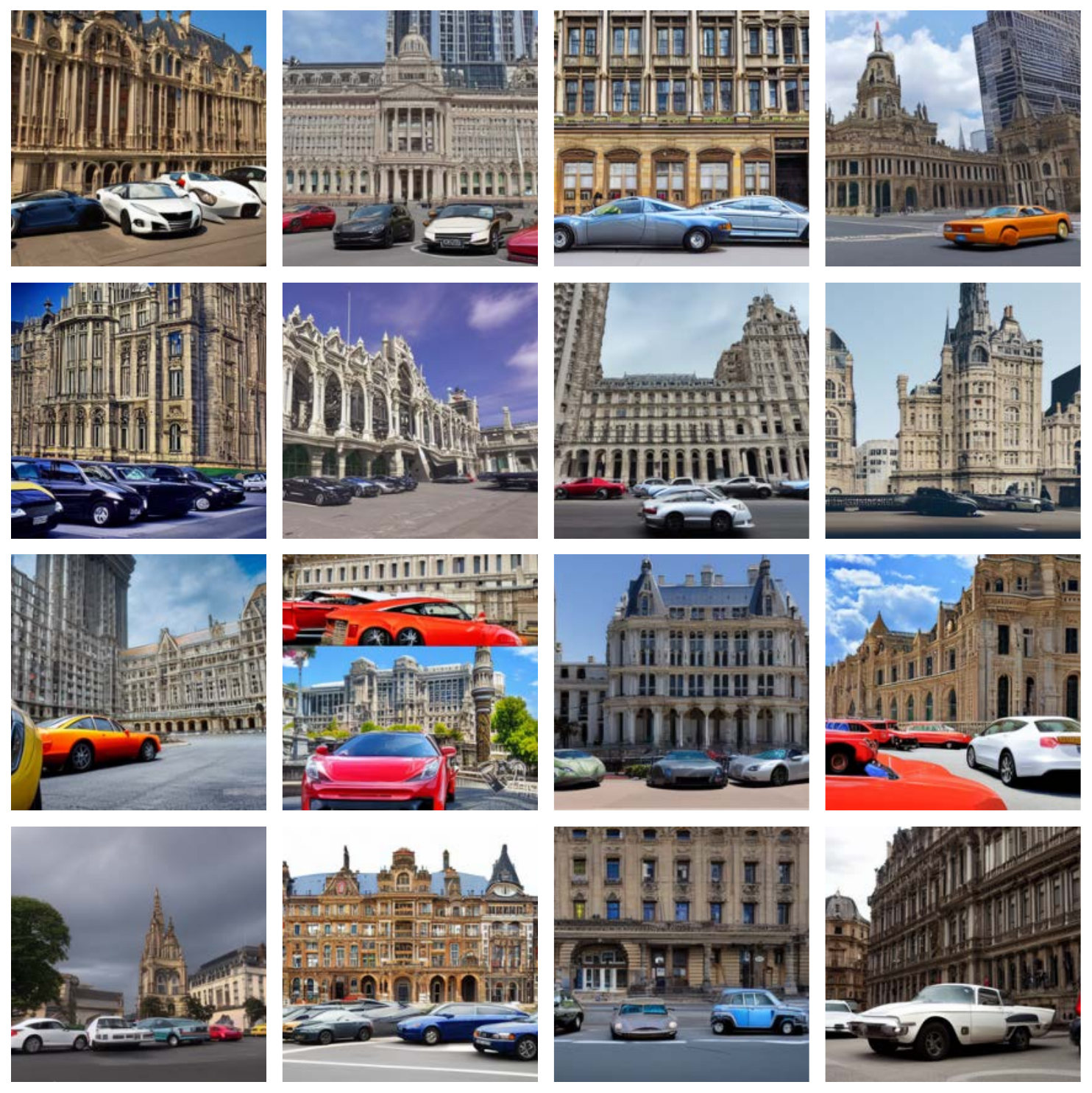}
        \caption{Vanilla guidance}
    \end{subfigure}
    \begin{subfigure}[ht]{0.25\textwidth}
        \includegraphics[width=\textwidth]{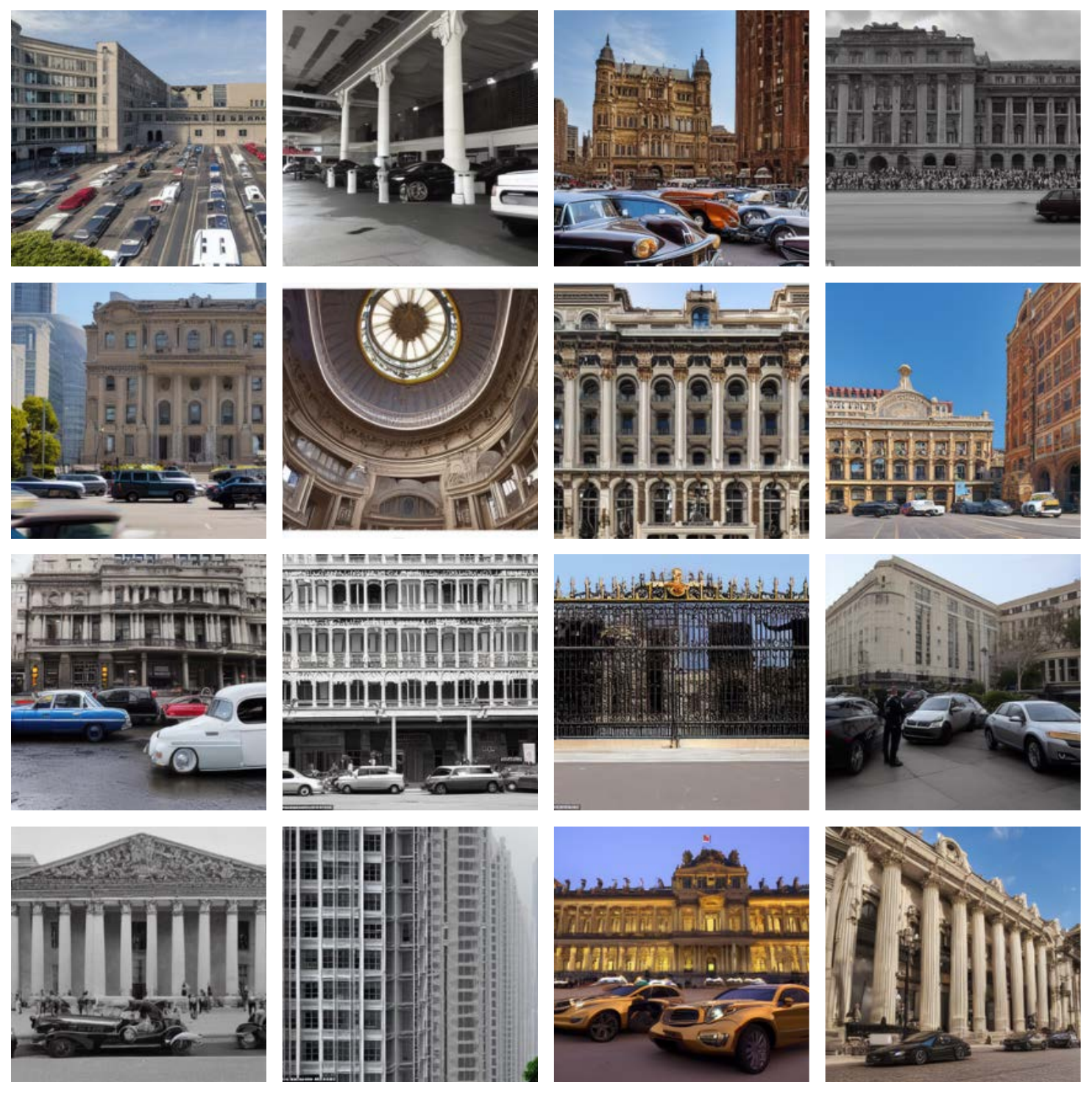}
        \caption{Prompt expansion}
    \end{subfigure} \\
    \begin{subfigure}[ht]{0.25\textwidth}
        \includegraphics[width=\textwidth]{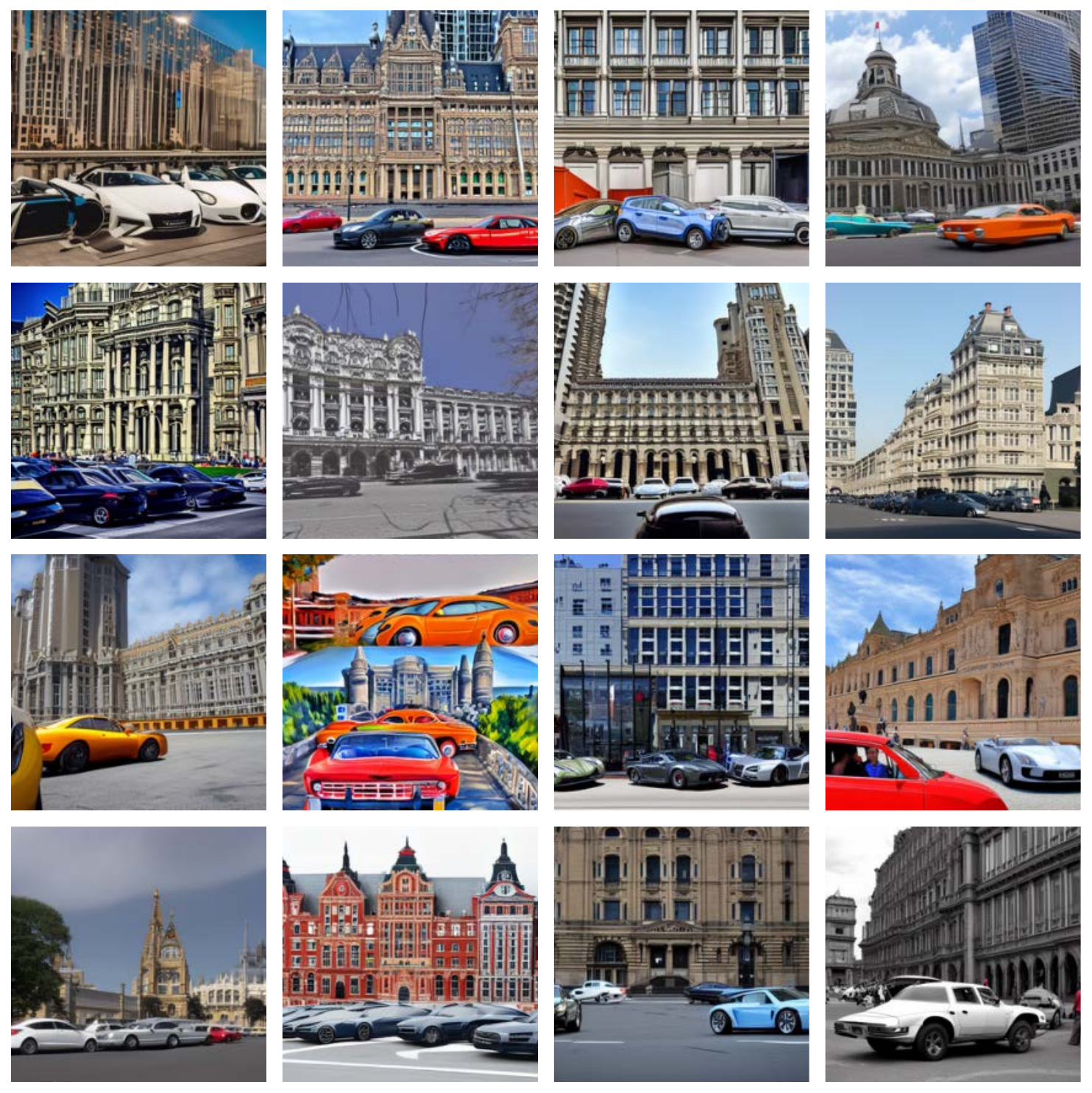}
        \caption{CADS guidance}
    \end{subfigure}
    \begin{subfigure}[ht]{0.25\textwidth}
        \includegraphics[width=\textwidth]{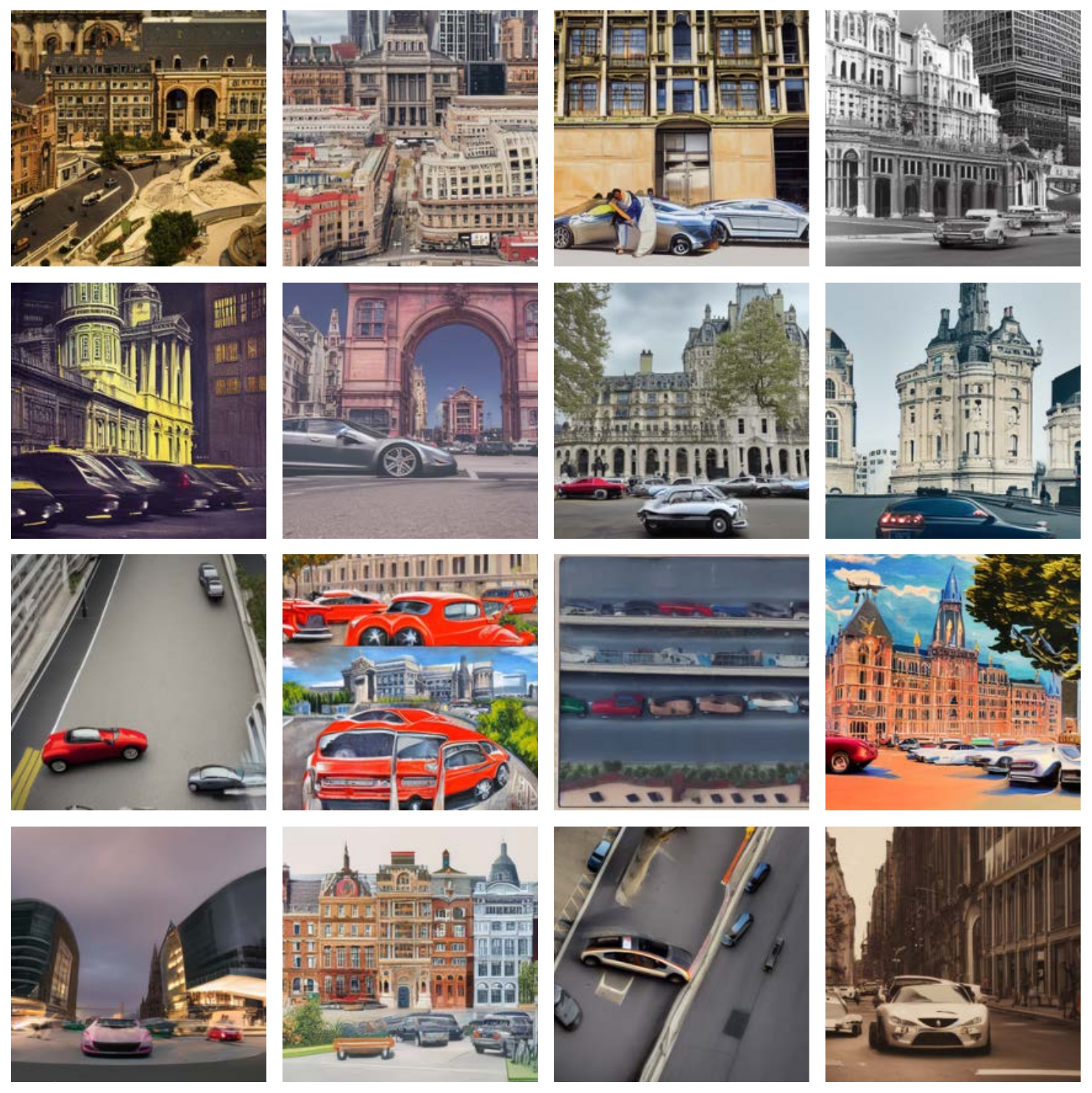}
        \caption{Interval guidance}
    \end{subfigure}
    \begin{subfigure}[ht]{0.25\textwidth}
        \includegraphics[width=\textwidth]{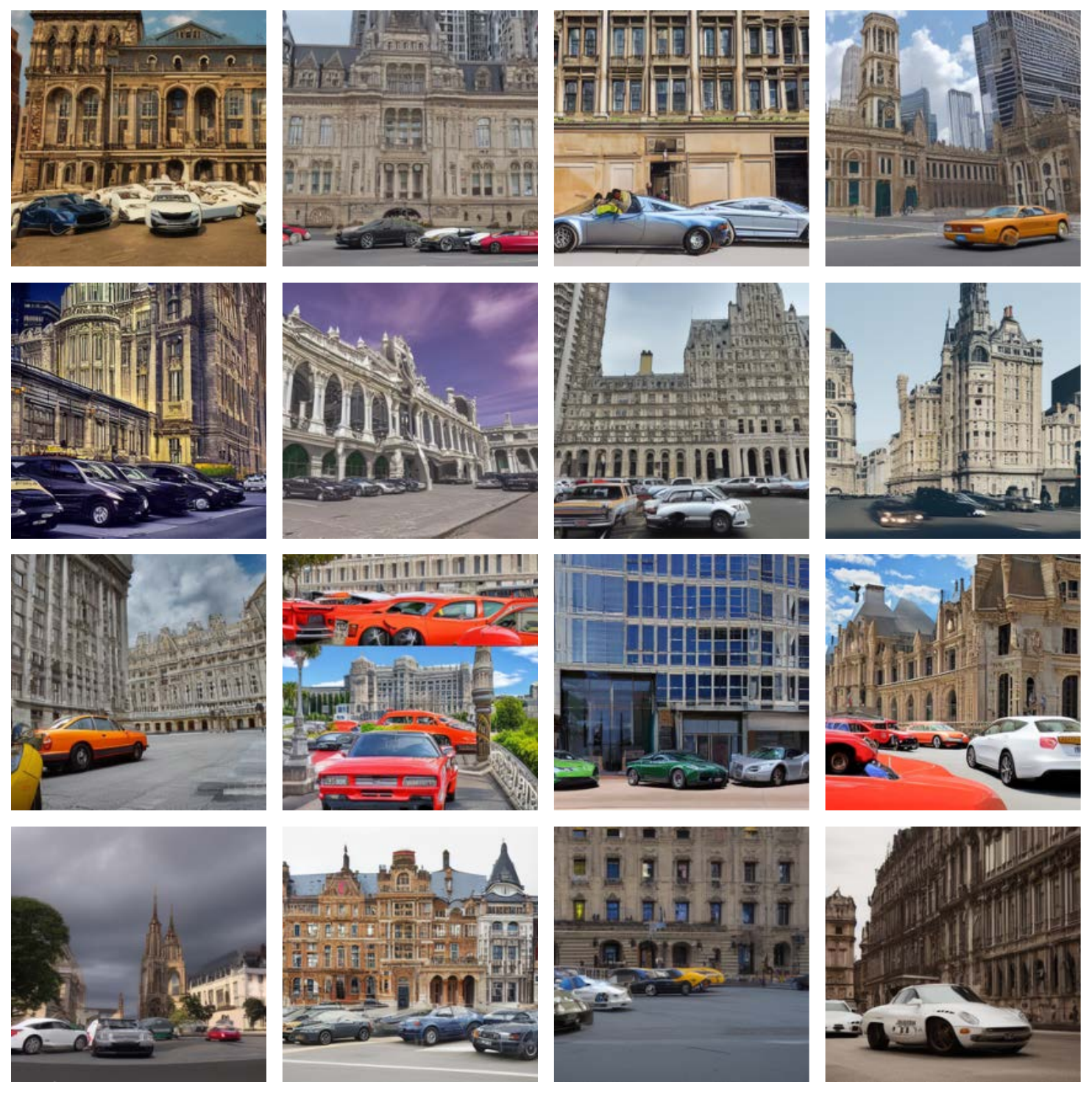}
        \caption{APG guidance}
    \end{subfigure}
    \caption{\textbf{Qualitative visuals for the prompt \texttt{``Cars and a grand building''} using the LDMv1.5 model with different sampling settings.}}
    \label{fig:header_visuals_6}
\end{figure}

\begin{figure}[ht]
    \centering
    \begin{subfigure}[ht]{0.25\textwidth}
        \includegraphics[width=\textwidth]{Imgs/visual_non_human/qualitative_data_7_2.pdf}
        \caption{Dataset}
    \end{subfigure}
    \begin{subfigure}[ht]{0.25\textwidth}
        \includegraphics[width=\textwidth]{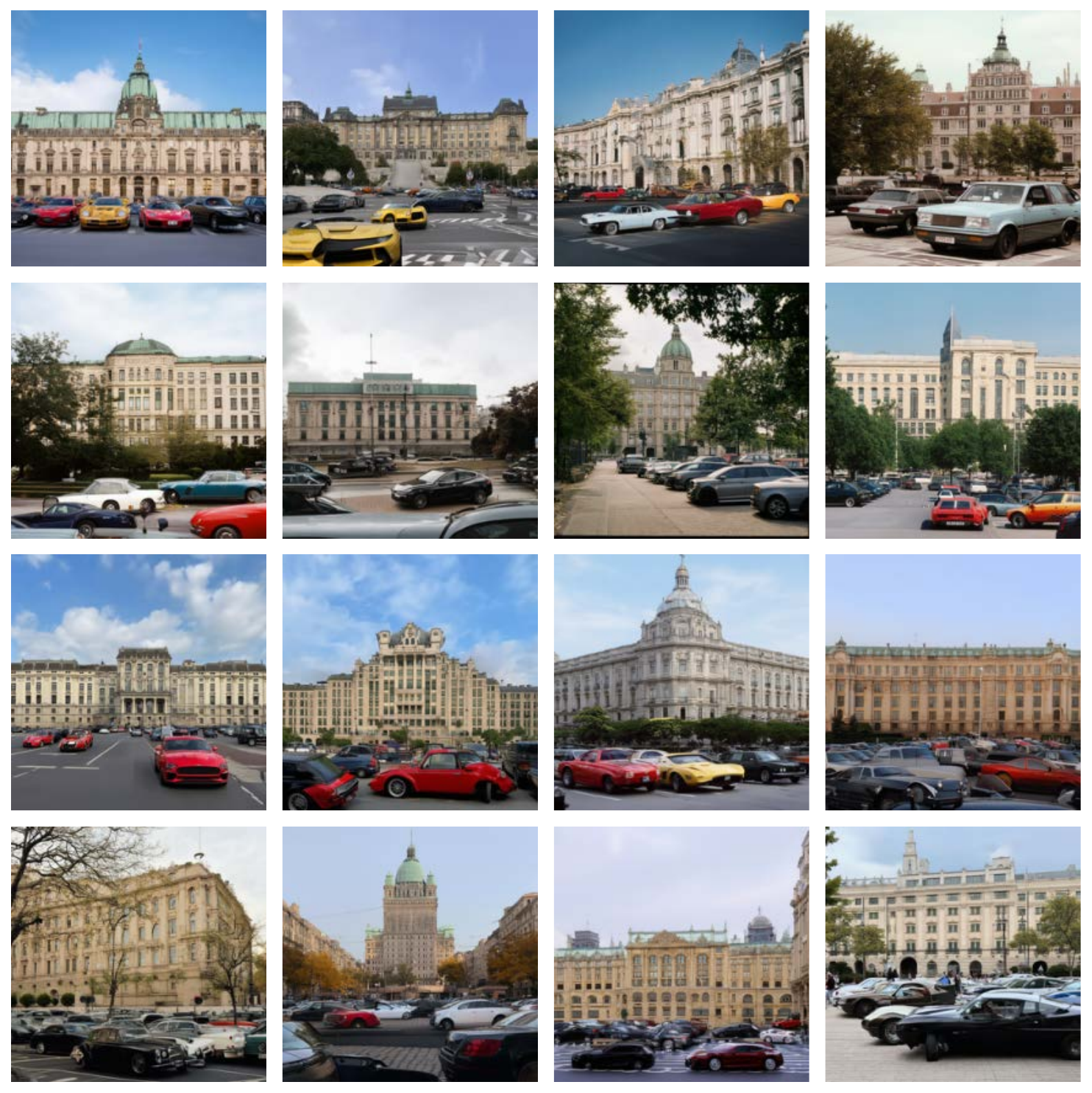}
        \caption{Vanilla guidance}
    \end{subfigure}
    \begin{subfigure}[ht]{0.25\textwidth}
        \includegraphics[width=\textwidth]{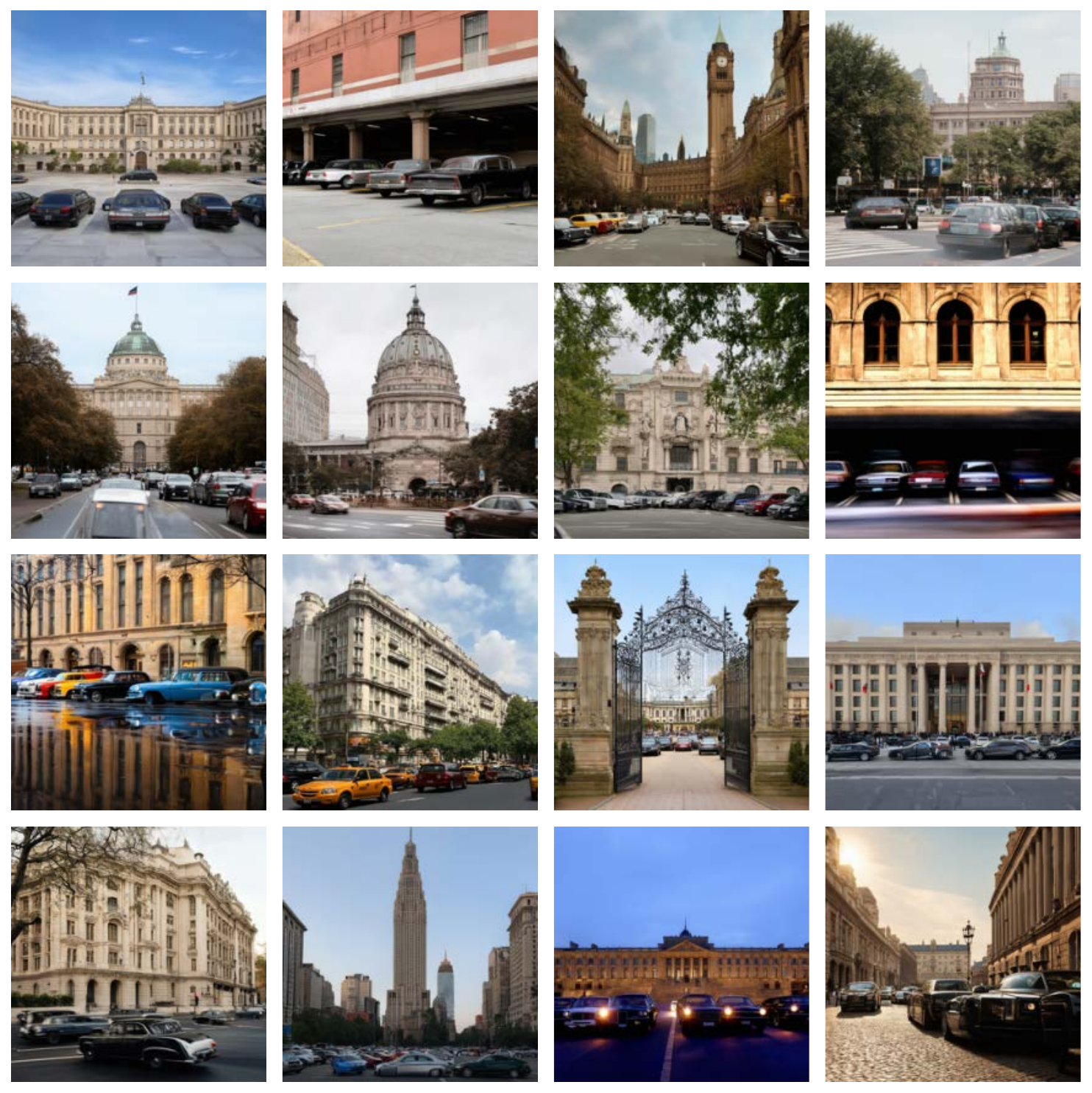}
        \caption{Prompt expansion}
    \end{subfigure} \\
    \begin{subfigure}[ht]{0.25\textwidth}
        \includegraphics[width=\textwidth]{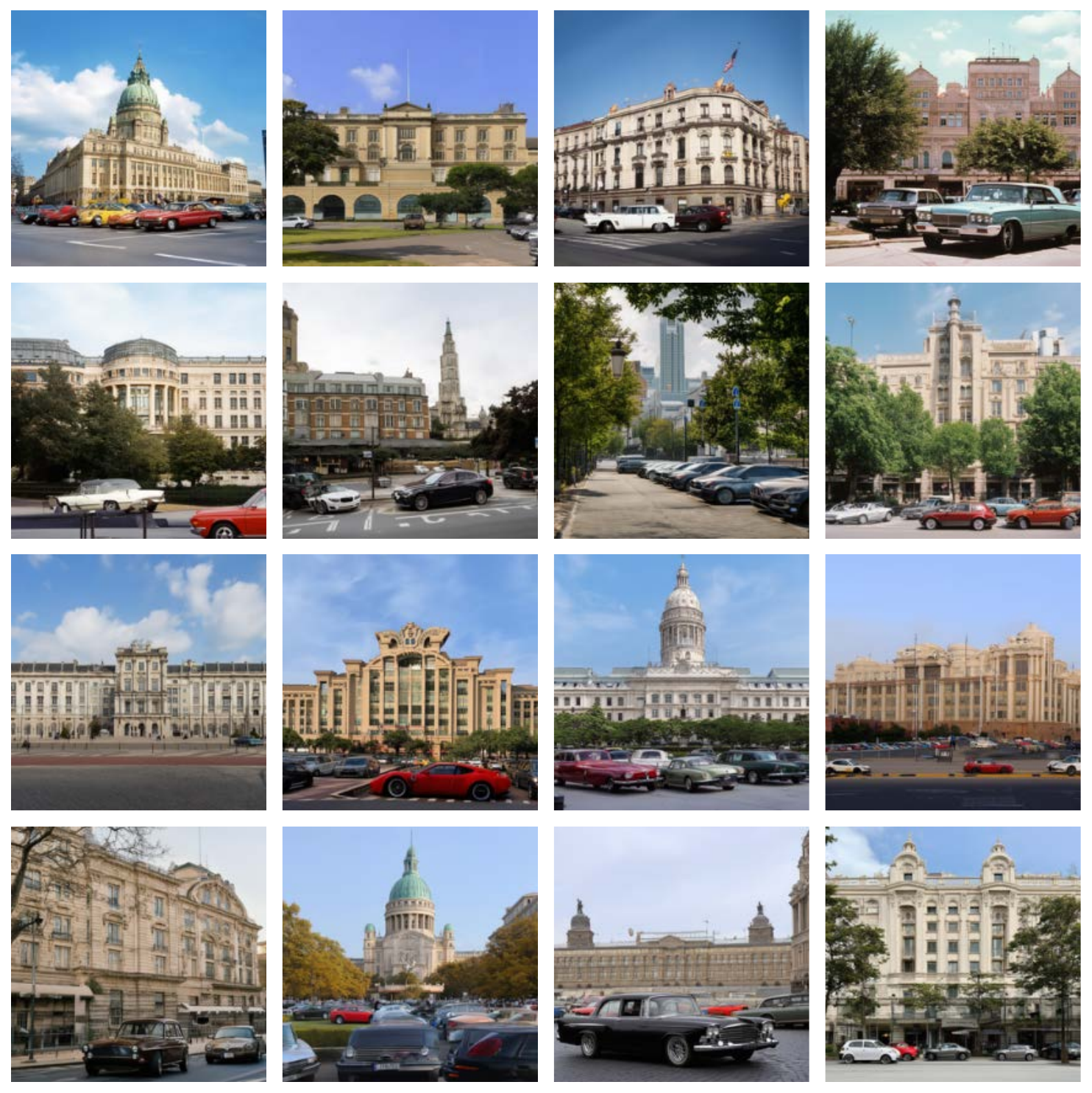}
        \caption{CADS guidance}
    \end{subfigure}
    \begin{subfigure}[ht]{0.25\textwidth}
        \includegraphics[width=\textwidth]{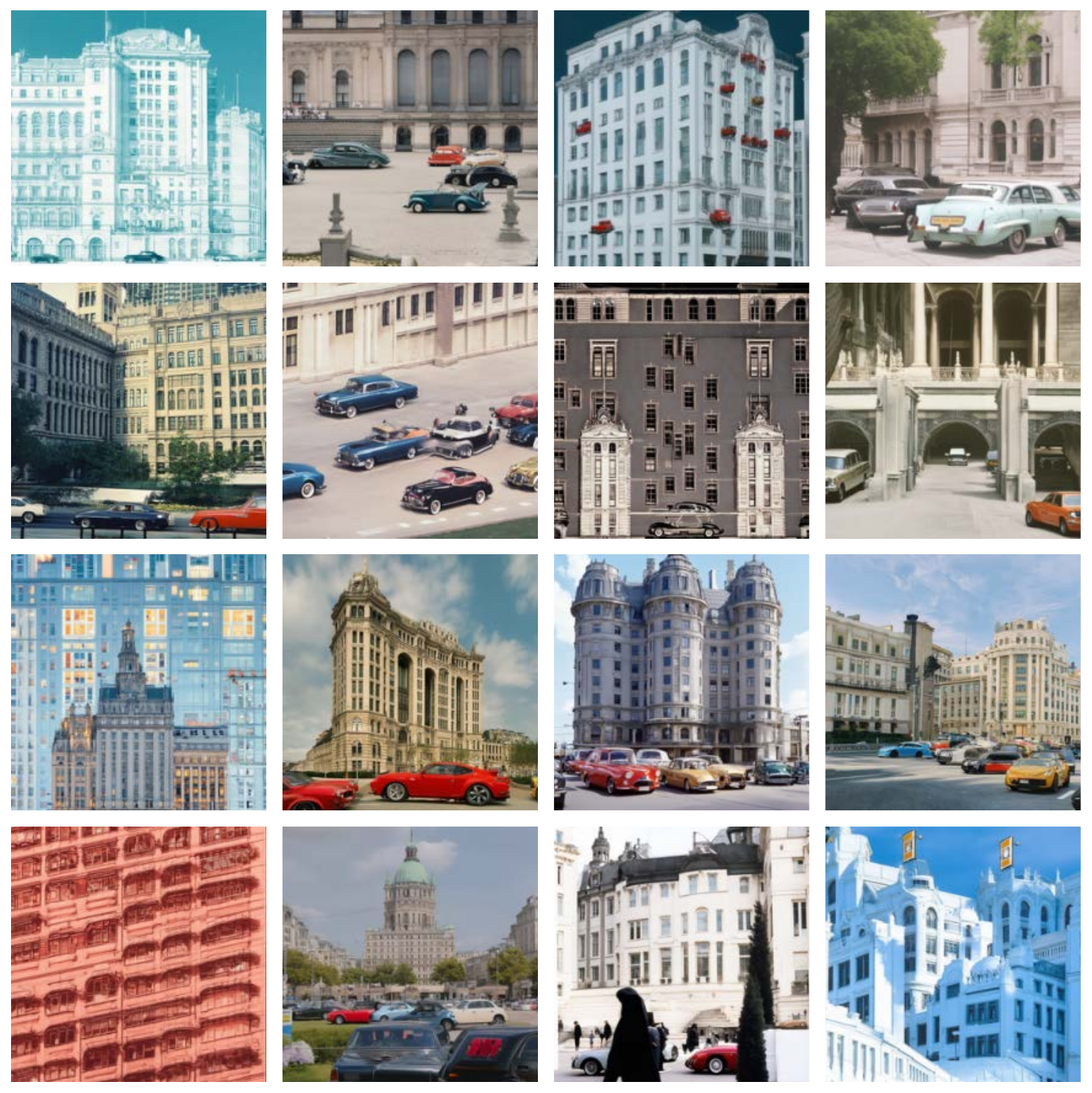}
        \caption{Interval guidance}
    \end{subfigure}
    \begin{subfigure}[ht]{0.25\textwidth}
        \includegraphics[width=\textwidth]{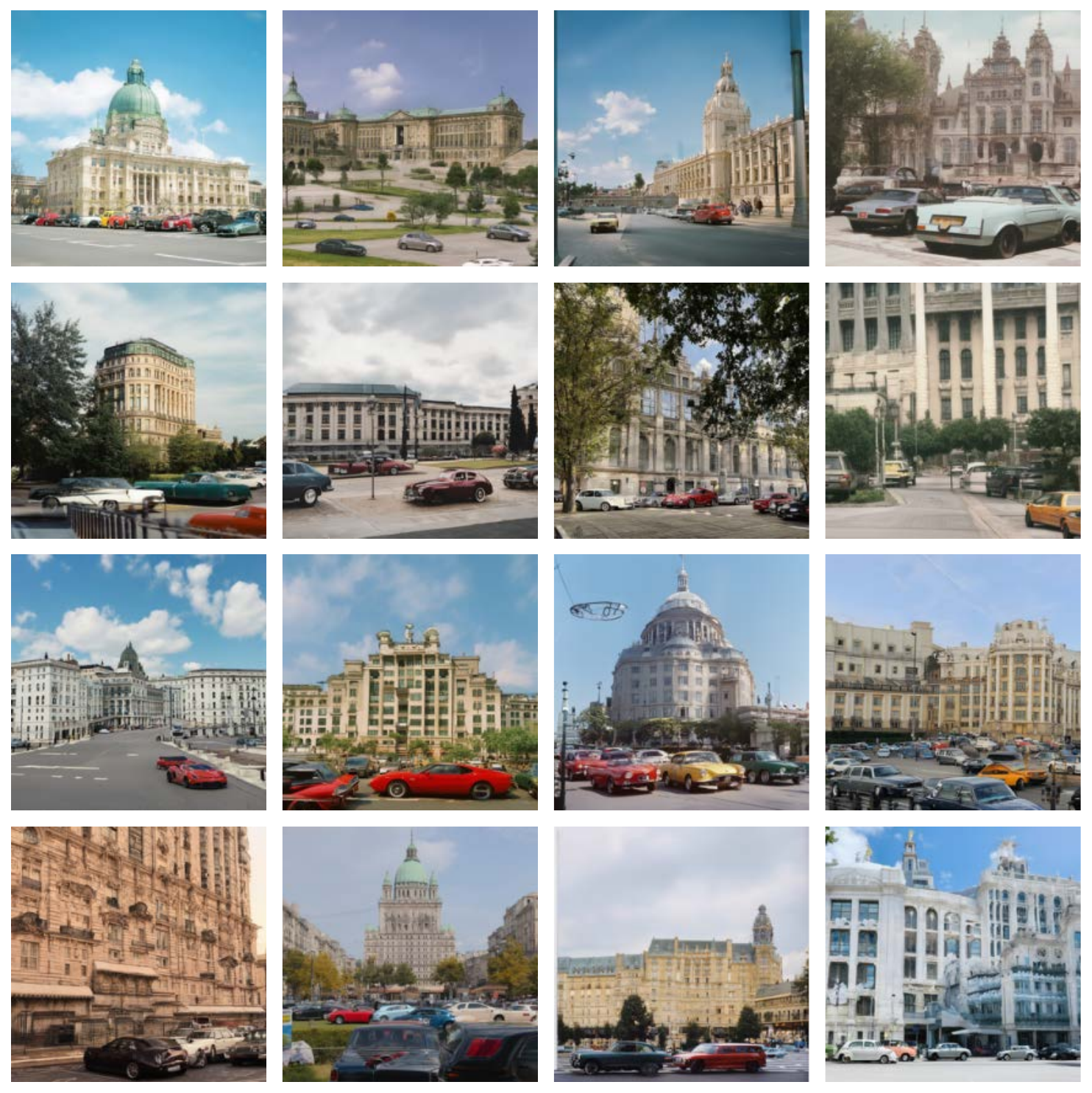}
        \caption{APG guidance}
    \end{subfigure}
    \caption{\textbf{Qualitative visuals for the prompt \texttt{``Cars and a grand building''} using the LDMv3.5M model with different sampling settings.}}
    \label{fig:header_visuals_7}
\end{figure}

\begin{figure}[ht]
    \centering
    \begin{subfigure}[ht]{0.25\textwidth}
        \includegraphics[width=\textwidth]{Imgs/visual_non_human/qualitative_data_7_2.pdf}
        \caption{Dataset}
    \end{subfigure}
    \begin{subfigure}[ht]{0.25\textwidth}
        \includegraphics[width=\textwidth]{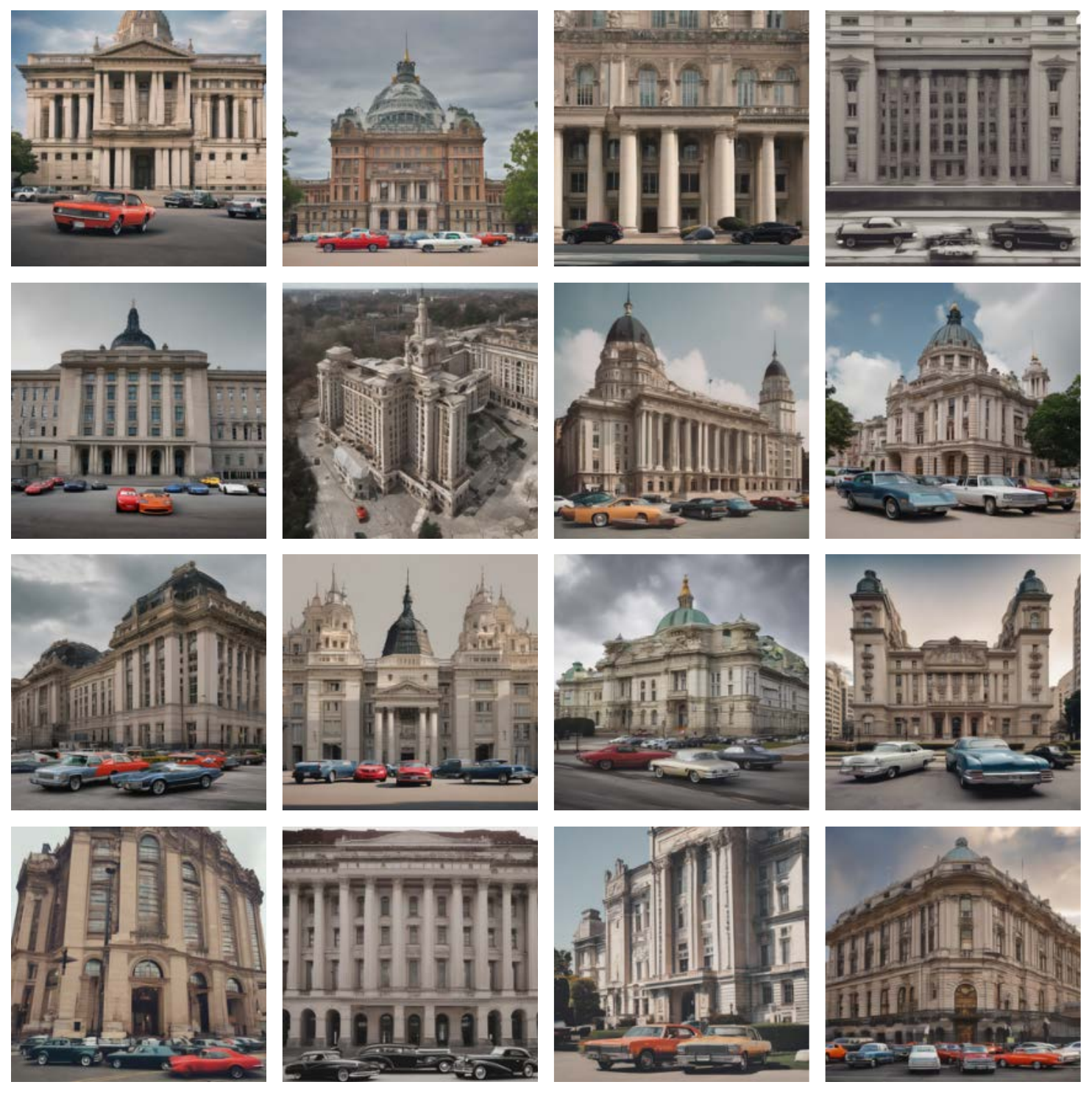}
        \caption{Vanilla guidance}
    \end{subfigure}
    \begin{subfigure}[ht]{0.25\textwidth}
        \includegraphics[width=\textwidth]{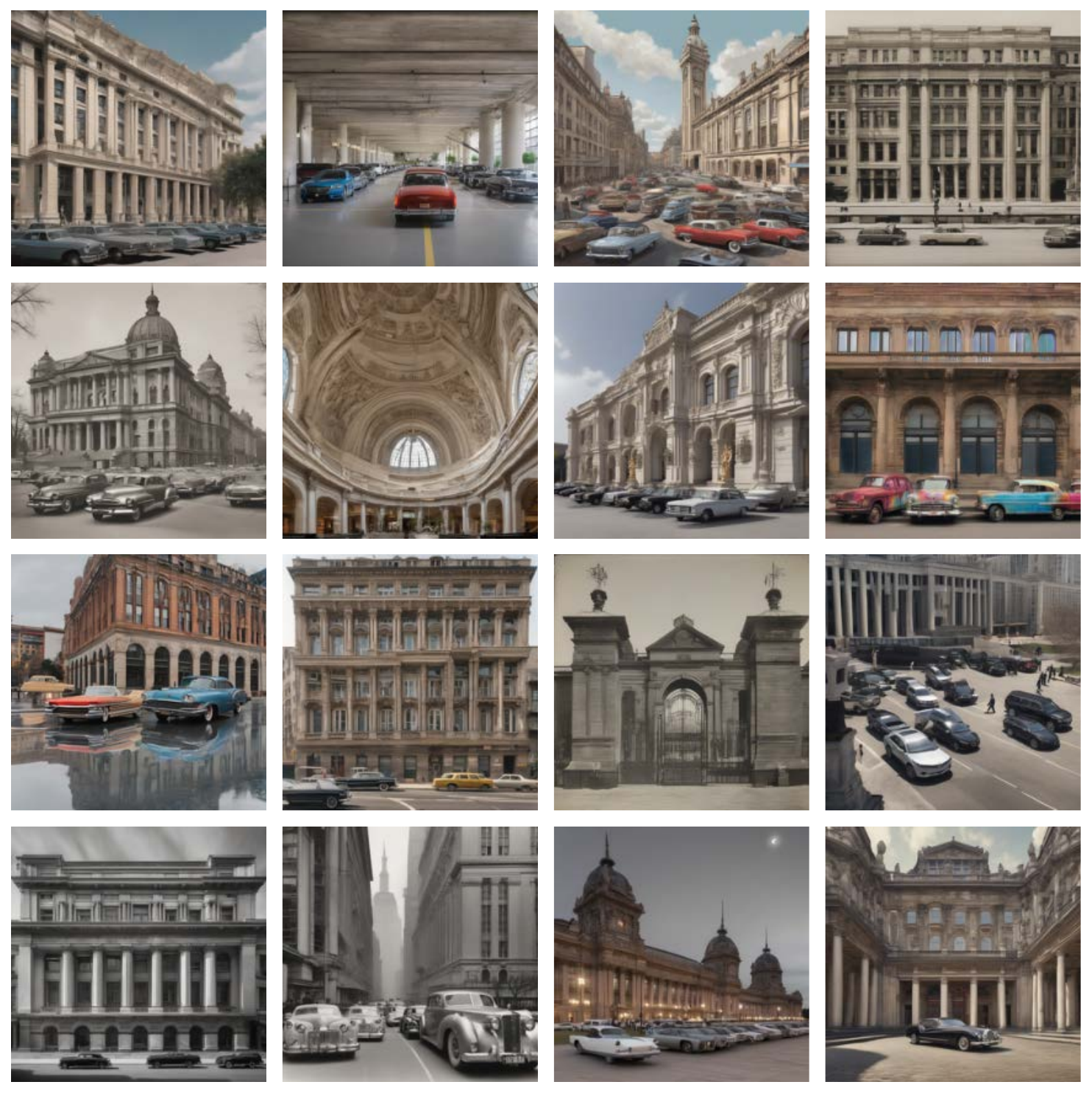}
        \caption{Prompt expansion}
    \end{subfigure} \\
    \begin{subfigure}[ht]{0.25\textwidth}
        \includegraphics[width=\textwidth]{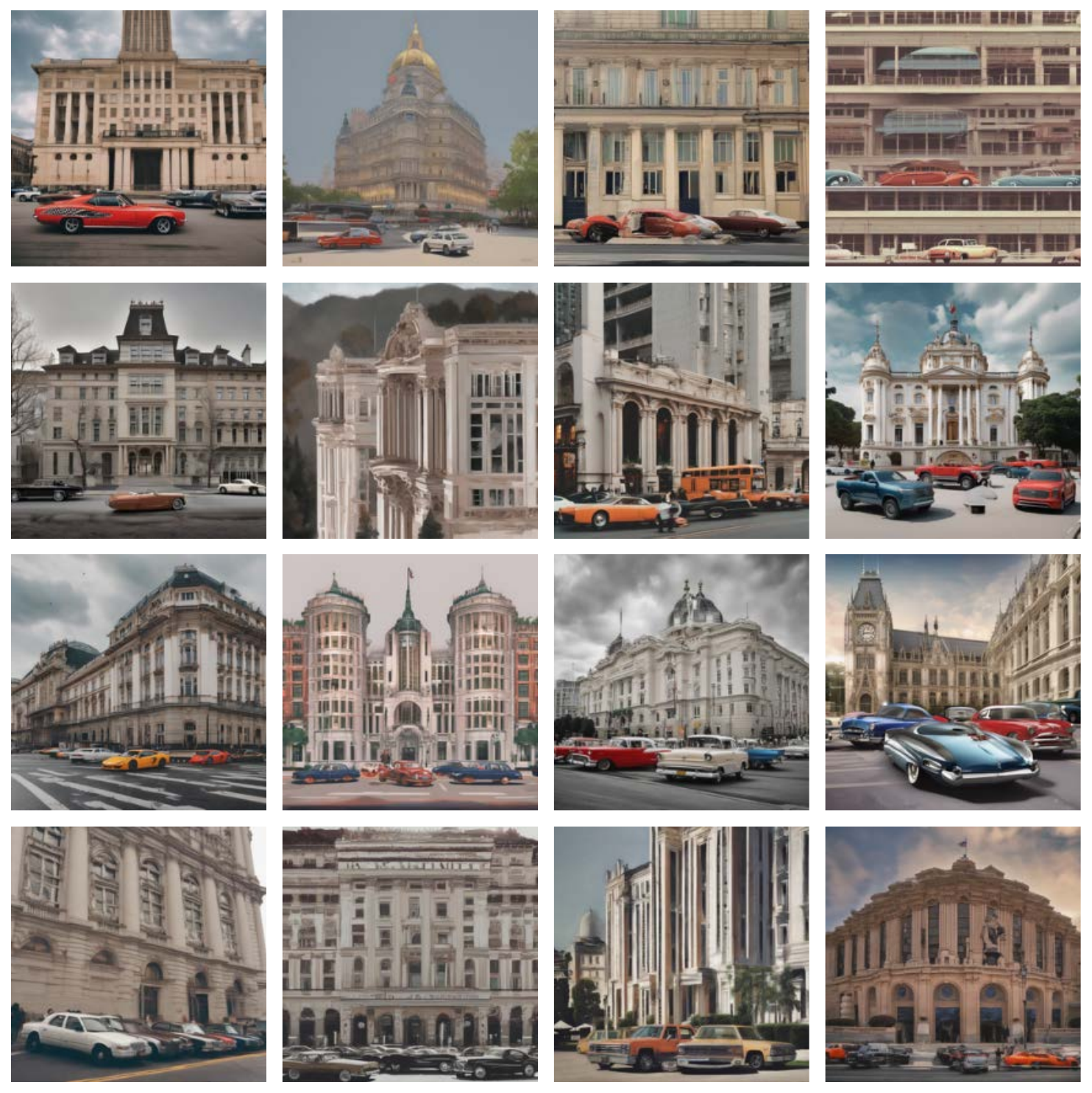}
        \caption{CADS guidance}
    \end{subfigure}
    \begin{subfigure}[ht]{0.25\textwidth}
        \includegraphics[width=\textwidth]{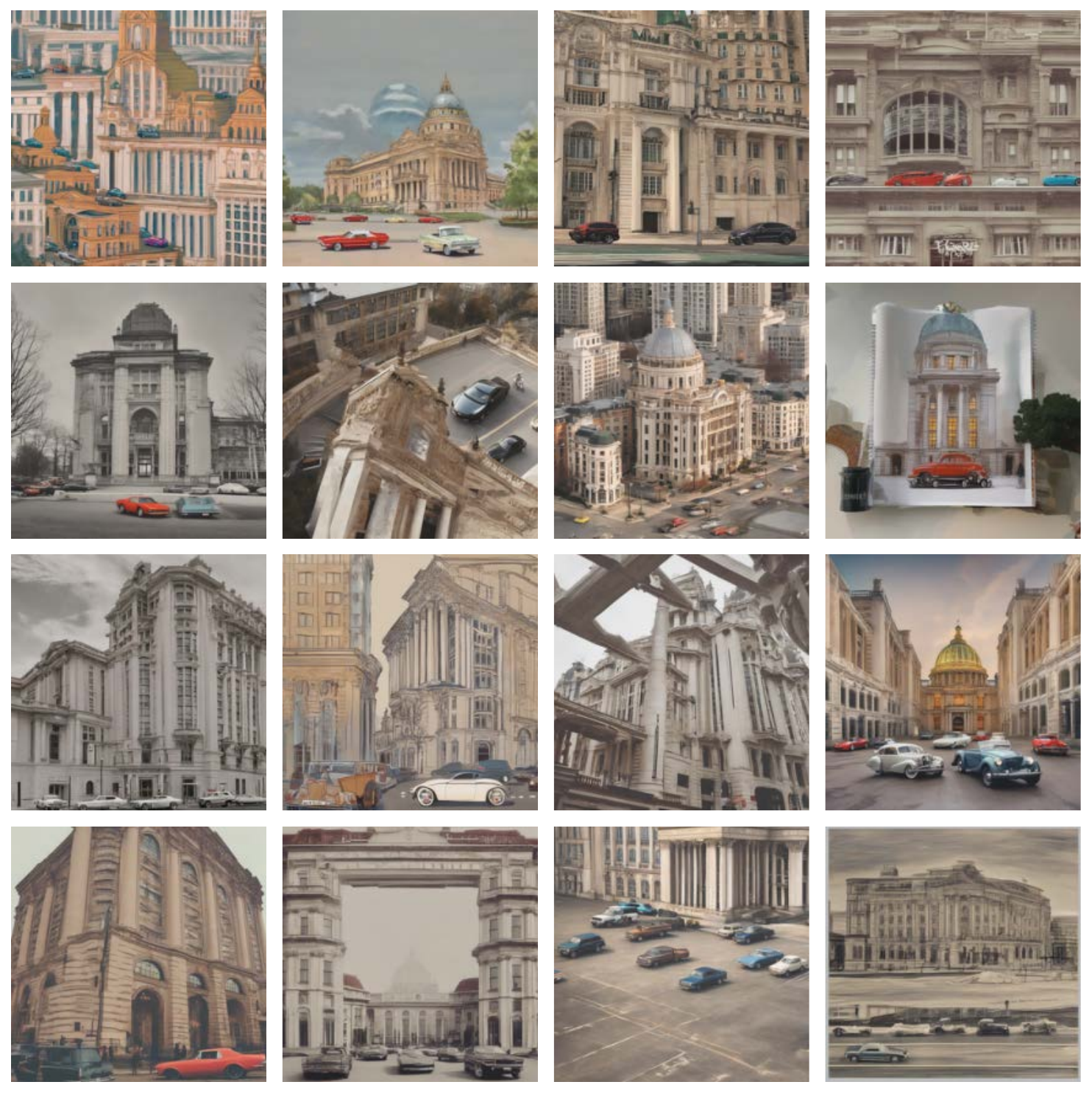}
        \caption{Interval guidance}
    \end{subfigure}
    \begin{subfigure}[ht]{0.25\textwidth}
        \includegraphics[width=\textwidth]{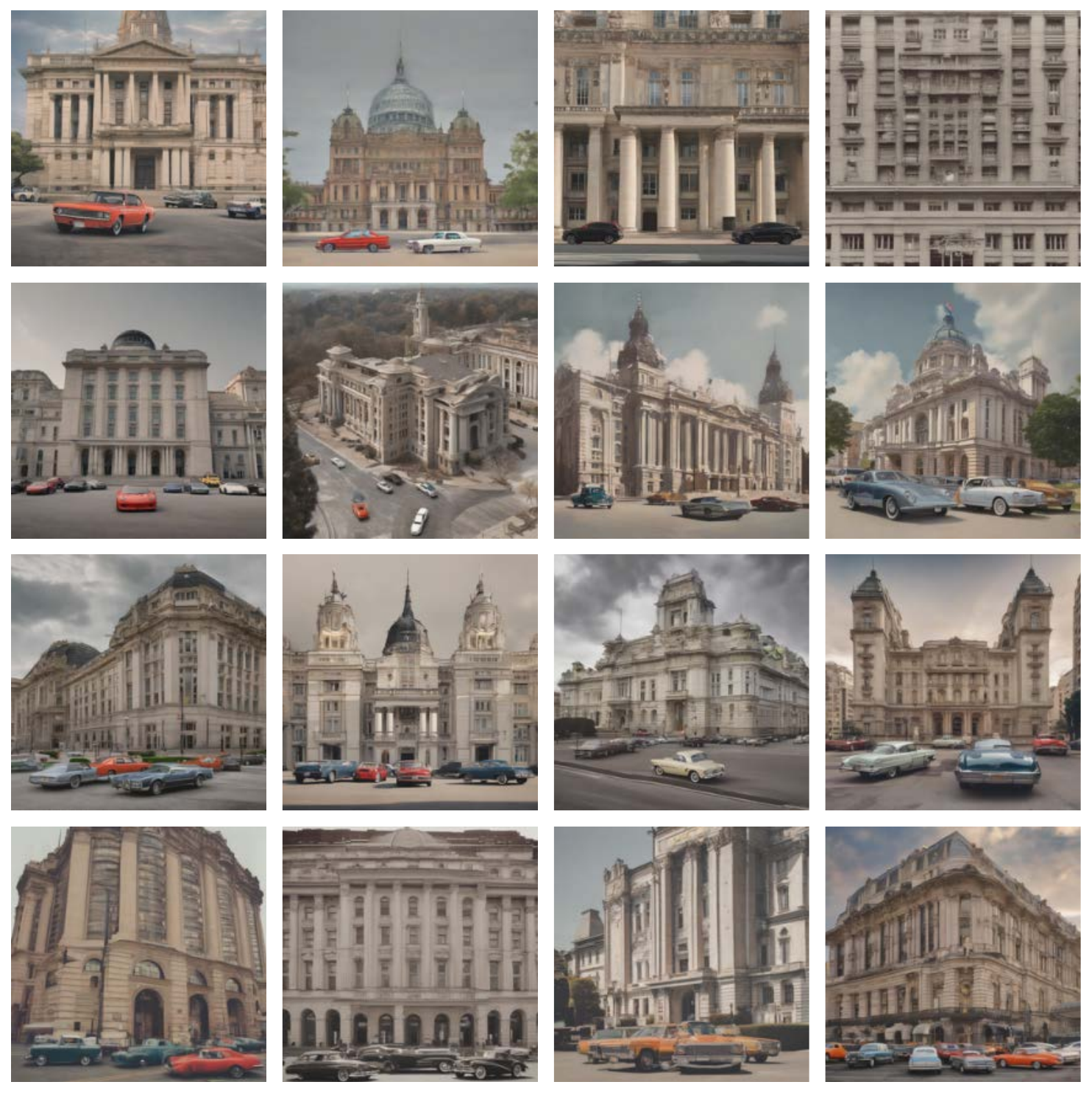}
        \caption{APG guidance}
    \end{subfigure}
    \caption{\textbf{Qualitative visuals for the prompt \texttt{``Cars and a grand building''} using the LDM-XL model with different sampling settings.}}
    \label{fig:header_visuals_8}
\end{figure}

We also show more qualitative samples in Figures~\ref{fig:header_visuals_1} to~\ref{fig:header_visuals_8} using all sampling intervention methods, generated by all four T2I models (LDMv1.5, LDM-XL, LDMv3.5M, and LDMv3.5L), complementing the results in Figure~\ref{fig:header_visuals_pmt_complexity}. We observe qualitatively that prompt expansion and advanced guidance methods lead to more diverse images than the vanilla guidance.

\end{document}